\documentclass{article}
\usepackage{microtype,graphicx,booktabs,needspace,placeins,array}
\usepackage{xurl,alphalph}
\usepackage{hyperref}

\usepackage[preprint]{icml2027}
\usepackage{amsmath,amssymb,amsthm,mathtools}
\usepackage{bookmark}
\makeatletter
\let\icmloriginalsection\section
\newcounter{outlineheading}
\RenewDocumentCommand\section{s o m}{%
  \IfBooleanTF{#1}{\icmloriginalsection*{#3}\stepcounter{outlineheading}\pdfbookmark[0]{#3}{outline.star.\arabic{outlineheading}}}{%
    \IfNoValueTF{#2}{\icmloriginalsection{#3}}{\icmloriginalsection[#2]{#3}}%
    \bookmark[dest=\@currentHref,level=0]{\thesection. #3}}}
\makeatother
\newtheorem{theorem}{Theorem}
\newtheorem{proposition}[theorem]{Proposition}
\newtheorem{lemma}[theorem]{Lemma}
\newtheorem{corollary}[theorem]{Corollary}
\newcommand{\OPT}{\mathrm{OPT}}
\newcommand{\one}{\mathbf{1}}
\newcommand{\E}{\mathbb E}
\icmltitlerunning{From NP-Hardness to Exact Tractability Through Sample Abundance}
\ifdefined\pdfinfoomitdate\pdfinfoomitdate=1\fi
\ifdefined\pdftrailerid\pdftrailerid{}\fi
\begin{document}
\twocolumn[
\icmltitle{From NP-Hardness to Exact Tractability Through Sample Abundance}
\begin{icmlauthorlist}\icmlauthor{Mojtaba Soltanalian}{uic}\end{icmlauthorlist}
\icmlaffiliation{uic}{Department of Electrical and Computer Engineering, University of Illinois Chicago, Chicago, IL, USA}
\icmlcorrespondingauthor{Mojtaba Soltanalian}{msol@uic.edu}
\icmlkeywords{sample abundance, exact optimization, noisy parity, low-degree estimation, active sensing}
\vskip 0.3in]
\printAffiliationsAndNotice{}
\hypersetup{pdfauthor={Mojtaba Soltanalian},pdfsubject={Research preprint}}

\begin{abstract}
Additional observations can make exact optimization tractable without changing the objective or feasible set. We establish this phenomenon for noisy parity fitting, a worst-case NP-hard problem, with uniformly sampled rank-one sign tensor entries, independent fair factor signs and independent errors. For fixed interaction order $k\ge3$, $n$ signs per block, logarithmic mean degree and known channel biases, $O(n/(\log n)^{1/(k-2)})$ further readings of the same source suffice with high probability for exact optimization in $O(n\log^2 n)$ worst-case work. The leading coefficient is sharp for weak recovery by depth-$O(\log\log n)$ local estimators with independent reading marks. A common signed-field law determines the gain over the generating source, the number of tied optima and the dimension needed for later reweightings. The geometry of the remaining coordinates determines the sharp radius of local verification for high-scoring assignments with high probability.
\end{abstract}

\section{Introduction}\label{sec:intro}
Sample abundance asks how additional observations can change a problem's computational complexity regime, often sharply reducing the work needed to solve it. Earlier work develops this idea through one-bit phase retrieval and sensing \citep{eamaz2022,eamaz2024,eamaz2026intro}. Enough threshold comparisons can make costly structural constraints unnecessary for accurate reconstruction. Here we ask when additional observations make exact optimization tractable without changing the objective or feasible set.

Consider noisy products, each taking one sign from each of $k$ blocks of $n$ unknown signs. Choosing signs to maximize agreement is NP-hard in the worst case \citep{garey1976}; see reduction in Section~\ref{sec:model}. We allow additional readings from the same source, on fibers that vary one index at a time, but score all assignments on the original products. If three of five products involving a sign were recorded incorrectly, reversing that source sign improves agreement. For comparison, read the same addresses on an independent source. The readings have the same marginal law, but only retained readings share the signs generating the original data. The source must persist, although readings remain noisy.

Let $D$ denote the original record, $R$ the additional readings, and $V=\OPT(D)$ the maximum score (agreements minus disagreements). Since $D$ determines $V$, their conditional mutual information is $I(V;R\mid D)=0$; the same holds for a fixed tie-broken optimizer. Statistical sufficiency need not preserve computational tractability \citep{montanari2014sufficient}. Here the question concerns the exact fit to a given finite record: additional readings can reveal how to compute an answer that the original record already determines.

The cost of obtaining this information depends on how signs interact: inference from a triple uses two partners, whereas a four-way product requires three. The local inference threshold rises from order $n/\log n$ readings for triples to $n/\sqrt{\log n}$ for four-way products. Above it by a fixed factor, local inference aligns nearly all signs. Observable fields then identify groups of bounded size for exact optimization with high probability.

The remaining optimization decisions obey a different law (Figure~\ref{fig:dimension-law}). Each observed product casts a vote for keeping or flipping each of its source signs. We derive the gain over the source, the multiplicity of optima and the number of signs that must stay free under reweighting from these nearly balanced vote counts. In a model with three-sign clauses, reweighting can select four orientations differing by whole-block flips. When nearly balanced coordinates are isolated, any single domain (some signs fixed and the rest free) must leave all $3n$ signs free. Yet four domains suffice, with their free signs in bounded-size groups linked by observed products.

The same geometry determines how many signs must change together to expose a suboptimal high-scoring answer. We obtain a sharp verification radius governed by density and noise rather than interaction order. Testing individual signs can miss profitable coupled changes. The probabilistic local-to-global guarantee separates finding an optimum from certifying it on a realized input.

\section{The fixed objective and the two experiments}\label{sec:model}
For a sign assignment $x=(x_{i,j})$, write $T_e(x)=\prod_{i=1}^k x_{i,e_i}$ at address $e\in[n]^k$, where $[n]=\{1,\ldots,n\}$. Coordinate $v=(i,j)$ belongs to $e$ if $e_i=j$. Let the mask $H$ be a uniform set of $m=\lceil cn\log n\rceil$ addresses ($c>0$), independent of the fair source signs $X$ with tensor $T=T(X)$. Observe $Y_e=T_e\eta_e$ on $H$, with independent errors $\Pr(\eta_e=-1)=q\in(0,1/2)$ and mean $\mu=1-2q$. Each coordinate belongs to $d=m/n$ observations on average; $d$ is the mean degree. For $D=(H,Y)$, maximize
\begin{equation}
 F_Y(x)=\sum_{e\in H}Y_eT_e(x),\qquad
 \OPT(H,Y)=\max_x F_Y(x).\label{eq:opt}
\end{equation}
Write $F_Y$ also on complete tensors (thus $F_Y(T)=F_Y(X)$), and abbreviate $\OPT(H,Y)$ by $\OPT$. The misfit count is $(m-F_Y(x))/2$. Even-block reversals preserve the tensor; we call the equivalent assignments its representatives.

For triples, deciding whether the maximum is at least a given integer is NP-complete. For blocks $a,b,c$, each oriented graph edge $uv$ gives terms $a_u b_{uv}c_1-a_v b_{uv}c_1$ with a private middle sign. Maximization over that sign gives $|a_u-a_v|$, so the optimum is twice the maximum cut \citep{garey1976}. The triples are distinct and unused signs equalize the block sizes. For larger $k$, give all terms the same extra coordinates; their common sign can be absorbed into the first block. This proves worst-case hardness, whereas our results use random masks and noise.

\paragraph{Measurement access.}
A reading returns the entry at a chosen address times fresh noise with error probability $s\in(0,1/2)$ and bias $\rho=1-2s$. Retention reads $T$; replacement reads an independent copy. For triples, read the fibers through $(1,1,1)$. For source blocks $a,b,c$, their noiseless entries are
\begin{equation}
 \begin{gathered}A_i=a_ib_1c_1,\quad B_j=a_1b_jc_1,\quad C_l=a_1b_1c_l,\\ A_iB_jC_l=T_{ijl}.\end{gathered}\label{eq:fiber}
\end{equation}
At general order the product of the $k$ noiseless fiber entries at $e$ equals $G^{k-1}T_e$, where $G=T_{1,\ldots,1}$ is the center sign. Multiplying noisy readings by the target label thus cancels the source factors under retention. Replacement leaves relative signs. The count $Q$ charges every reading.

\paragraph{Later uses of the same fit.}
After acquisition, allow changes $\xi=(\xi_e)_{e\in H}$ within a per-coordinate budget $\Gamma$:
\begin{align}
 F_\xi(x)&=\sum_{e\in H}(Y_e+\xi_e)T_e(x),\nonumber\\
 \mathcal U_\Gamma(H)&=\bigl\{\xi:\max_v\sum_{e\ni v}|\xi_e|\le\Gamma\bigr\}.
 \label{eq:main-task-neighborhood}
\end{align}
The budget is relative to $Y$. A domain fixes some signs and leaves others free; it is \emph{common} if it contains an optimizer for every allowed $\xi$, even for edits chosen after earlier answers.

\section{How many readings start local inference?}\label{sec:algorithm}
Mark block-$i$ coordinates for reading independently with probability $p_i$. A local estimator at a root coordinate sees observations and readings within a prescribed depth; one step traverses a product to another coordinate. It may use addresses and source-independent randomness, but no advice from outside the neighborhood. Block overlap is the mean product of estimated and true signs in the reference-fiber representation; weak recovery means absolute overlap bounded away from zero.

\begin{theorem}[Acquisition at interaction order $k$]\label{thm:order-main}
Fix $k\ge3$, $c>0$ and known rational biases $\mu,\rho\in(0,1)$. Put
\[
 L=(\mu^2d)^{1/(k-2)},\qquad
 B_k=\frac{k(k-2)}{(k-1)^{(k-1)/(k-2)}}.
\]
For any fixed $\epsilon>0$, an algorithm using at most
\begin{equation}
 (B_k+\epsilon)\,\frac{n}{\rho^2 L}\label{eq:order-budget}
\end{equation}
noisy reference-fiber readings for all sufficiently large $n$ attains the original empirical optimum with probability $1-o(1)$. On the same event it constructs a common domain with bounded free components (connected by observed products) for all reweightings in a predeclared local budget $\Gamma_n=o(\log n)$. The algorithm uses $O((m+n)\log(n+2))$ word operations on every input.

Conversely, if $L\rho^2\sum_i p_i\le B_k$, every estimator restricted to depth $O(\log\log n)$ with independent marks has vanishing absolute overlap in each block in probability, even with the common fiber-center sign supplied.
\end{theorem}

For triples, $B_3=3/4$. Balanced marks above the threshold give overlap tending to one using local neighborhoods, including at even orders when the common fiber-center sign is unknown (Appendix~\ref{app:order-law}). Sharpness concerns local weak recovery, not the minimum cost of exact optimization.

\paragraph{Where the exponent and coefficient come from.}
The local neighborhood is approximated by a branching tree with mean $d$ child products per coordinate (Appendix~\ref{app:order-law}). Let $x_{i,t}$ be the expected squared posterior mean, or energy, of a block-$i$ sign given depth-$t$ observations. A marked reading contributes $p_i\rho^2$ at the root. On a fresh product, child posterior means multiply, giving energy $\mu^2\prod_{j\ne i}x_{j,t}$. Write
\[
 u_i=p_i\rho^2+d\mu^2\prod_{j\ne i}x_{j,t}.
\]
The combined energy lies between $u_i/(1+u_i)$ and $u_i$. The upper bound follows by successive conditioning. For the lower bound, the sum of individual posterior means has covariance $u_i$ with the root sign and second moment at most $u_i+u_i^2$.

At a given depth set $z_i=Lx_{i,t}$ and $a_i=L\rho^2p_i$. For bounded $z_i$ the two bounds yield the same limiting next-depth update $z_i'$:
\[
 z_i'=a_i+\prod_{j\ne i}z_j.
\]
With equal allocations $a_i=a$ and states $z_i=z$, this is $z'=a+z^{k-1}$. A fixed point disappears when $a$ exceeds the maximum of $z-z^{k-1}$, attained at $z_*=(k-1)^{-1/(k-2)}$. Above it the iterates grow until the unscaled energy is of constant order. One further round gives overlap tending to one. For unequal allocations, Maclaurin's inequality bounds the sum of the products by $k(\sum_i z_i/k)^{k-1}$. Thus $\sum_i z_i\le kz_*$ is invariant whenever $\sum_i a_i\le B_k$, including equality. A finite-degree refinement enlarges the region of failure for local recovery. At $k=3$ this sufficient obstruction extends to $3/4+9/(32L)-1/(128L^2)+O(L^{-3})$. Matching it at finite degree remains open (Appendix~\ref{app:order-law}).

The same normalized constant occurs in noiseless peeling and hypergraph activation \citep{mori2015peeling,morrison2021}. Here both noise rates enter the budget, and the converse covers every estimator of the allowed neighborhood. Sum-product propagation is classical \citep{kschischang2001factor}. Noisy XOR recovery \citep{ahn2019,kim2021xor}, tensor estimation \citep{lesieur2017spiked}, and message passing for empirical MAX-3LIN at much higher density \citep{watanabe2012mls} provide related inference mechanisms. Here a local estimate initiates exact empirical fitting at logarithmic degree.

\paragraph{What if the source is replaced?}
For triples, supply both arms with an optimal-value threshold making the two decisions nearly equally likely. At sufficiently large fixed density, retained readings at prescribed addresses allow polylogarithmic-degree polynomial estimators with arbitrarily small limiting squared error (Propositions~\ref{prop:bootstrap-degree} and~\ref{prop:erasure-degree}). Under independent replacement at the same addresses, polynomials of target-label degree at most $D_n=\lfloor n^3/(16\mathrm e^2m^2)\rfloor$ have risk at least $1-o(1)$, even with all replacement readings. Trees inspecting at most $D_n$ target labels succeed with probability at most $1/2+o(1)$ (Theorem~\ref{thm:main}).

\begin{figure*}[t]
\centering
\includegraphics[width=\textwidth]{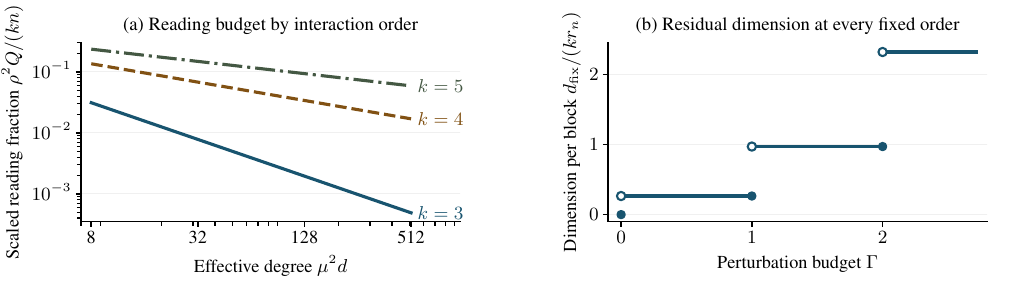}
\caption{\textbf{Interaction order governs acquisition, but not normalized residual freedom.} (a) Leading local-recovery thresholds for $Q$ readings, not finite success rates (Theorem~\ref{thm:order-main}). (b) Minimum free-coordinate count $d_{\rm fix}$ divided by $kr_n$, where $r_n=n^{1-\alpha}/\sqrt{\log n}$ and $\alpha=c(1-2\sqrt{q(1-q)})$; $q=.17$, $\alpha=.75$. Filled endpoints belong to the step: at $\Gamma=2$ the fields $-1,0,1$ can reverse preference, whereas magnitude two can only reach a tie.}
\label{fig:dimension-law}
\end{figure*}

\section{Exact optimization from alignment}
After choosing equivalent block signs to align with the source (the source gauge), overlap tending to one in each block gives score $\mu m-o_{\Pr}(m)$, where $o_{\Pr}(m)/m$ tends to zero in probability. Changing $o(n)$ signs affects $o(m)$ observations at maximum degree $O(\log n)$. Strict ascent flips one sign at a time until none improves the score; we call the resulting assignment stable. We ask how a high, stable score constrains the geometry of any better assignment.

For a set $S$ of source coordinates to flip, let $T^S$ be the resulting tensor. The signed sum over changed observations is (with $\one$ the indicator)
\[
 J(S)=\sum_{e\in H}\eta_e\,\one\{|e\cap S|\text{ is odd}\}.
\]
Then $F_Y(T^S)-F_Y(T)=-2J(S)$. A flip can improve on the source precisely when negative observations outnumber positive ones on the edges whose parity changes. Far from all representatives of the source, a positive fraction of the entire grid changes parity. An exponential bound for sampling without replacement excludes every such assignment from a sufficiently high score level, even after summing over all $2^{kn}$ possible factor assignments.

Within that level, join two coordinates when they share an observed product. A signed field at a coordinate is the sum of its incident labels times their predicted products. If a stable assignment differs from a source representative on a connected component $S$, summing its final fields on $S$ gives
\[
 \sum_{e\in H} |e\cap S|(-1)^{|e\cap S|}\eta_e\ge0.
\]
An edge meeting $S$ once contributes $-\eta_e$; edges meeting it twice or more are rarer when $|S|/n$ is small. Under an exponential tilt this imbalance pays for the choice of the component and of the observations connecting it. With high probability every such component has bounded size. At a changed coordinate $v$, the source field $D_v=\sum_{e\ni v}\eta_e$ is bounded above, since each pair belongs to at most two observations with high probability. This includes global maximizers. A high, stable reference thus reveals bounded components through observable fields. Fixed cores also occur in planted bisection \citep{coja2017bisection}. At even order, the solver tries both center signs and retains the higher score under~\eqref{eq:opt}.

Strict coordinate ascent takes at most $m$ improving flips, since each gains at least two score units. A queue stores the fields affected by each flip. Degree and component caps enforce the work bound on every input. If a cap is exceeded, the algorithm returns a feasible assignment. Section~\ref{sec:experiments} exhibits a suboptimal return that passes both caps.

\paragraph{Learning the channel for triples.}
For triples, a pilot sets an expected $O(n/\log n)$ reading budget without knowing either noise bias. A joint bound on errors and small source fields justifies two full-target updates followed by bounded-component repair at every fixed density. Put $h=1-2\sqrt{q(1-q)}$ and $\alpha=ch$. When $\alpha>1/2$, the second update suffices (Theorems~\ref{thm:geometric} and~\ref{thm:joint-localization}). The separate cascade has restrictive finite execution gates (Appendix~\ref{app:finite-gate}).

\section{Which decisions remain free?}\label{sec:local}\label{sec:common-reduction}\label{sec:exact-common-dimension}
Which signs must remain free under reweighting? In a coordinate domain (a Boolean cube), let $d_{\rm fix}(\Gamma)$ be the fewest free coordinates in such a domain containing an optimum for every allowed objective. Let $N_{\rm hit}(\Gamma)$ be the fewest complete tensors meeting the same requirement. At zero budget the minima are zero coordinates and one tensor. Appendix~\ref{app:residual-order} also counts distinct predictions on the observed addresses.

For bounded budgets, all optima align with the source outside $W=\{v:D_v\le L_0\}$ with high probability for a sufficiently large fixed $L_0$. At a positive-degree singleton of $W$, a negative field favors reversal and a zero field leaves both signs optimal. Reweighting with $\Gamma>0$ can force either sign exactly when $|D_v|<\Gamma$; equality permits only a tie. Negative fields thus contribute twice their magnitude to gain, zeros contribute one bit of multiplicity, and reversible signs contribute one free coordinate.

\begin{theorem}[One field law for gain, ties and necessary freedom]\label{thm:local}\label{thm:exact-common-dimension-main}
Fix $k\ge3$. For $0<\alpha<1$, write $r_n=n^{1-\alpha}/\sqrt{\log n}$ and $\gamma=\sqrt{q(1-q)}$. Put
\[
 \lambda=\tfrac12\log\frac{1-q}{q},\quad a=e^{-\lambda},\quad
 A=(4\pi c\gamma)^{-1/2}.
\]
If $G_n=\OPT-F_Y(T)$ and $\mathcal O_n$ is the set of optimizing complete tensors, then
\begin{align}
 G_n/r_n&\xrightarrow{\Pr}2kA\,a/(1-a)^2,\nonumber\\
 \log_2|\mathcal O_n|/r_n&\xrightarrow{\Pr}kA.\label{eq:main-gain-entropy}
\end{align}
For fixed $\Gamma>0$, the ratios $d_{\rm fix}(\Gamma)/r_n$ and $\log_2 N_{\rm hit}(\Gamma)/r_n$ have the common limit in probability
\begin{equation}
 kA\sum_{j\in\mathbb Z:\,|j|<\Gamma}e^{\lambda j}.
 \label{eq:dimension-constant-main}
\end{equation}
The dimension law is uniform over $0<\Gamma\le\overline\Gamma$ for every fixed finite $\overline\Gamma$. For $\alpha\ge1$, the source is uniquely optimal with high probability and bounded budgets require zero dimension with high probability.
\end{theorem}

For each fixed integer $j$, exponential tilting of the positive and negative incident counts gives
\[
 \Pr(D_v=j)\sim A e^{\lambda j}\,n^{-\alpha}/\sqrt{\log n}.
\]
All but $o_{\Pr}(r_n)$ vertices of $W$ lie farther than two steps from every other vertex of $W$. Summing their field masses gives the constants. Above $\alpha=1/2$, isolation gives $G_n=2\sum_v(-D_v)_+$ with high probability, where $z_+=\max(z,0)$. In Figure~\ref{fig:dimension-law}, zero fields determine the first dimension step. Only when the budget exceeds one can fields of magnitude one contribute.

Disjoint perturbations can prescribe any pattern on the separated reversible coordinates. Their optimizer sets are disjoint, so a common set of answers must accommodate every pattern. Retaining the interacting components and fixing every singleton whose preference can reach at most a tie gives the matching upper bound (Appendix~\ref{app:residual-order}).

\paragraph{Computing a reusable domain.}
Let $y$ be a reference with score at least $\mu m-o_{\Pr}(m)$, stable under single flips and let $J\ge2$ be an integer with $J>1/\alpha$. At every fixed order, free the coordinates whose observable fields satisfy the cutoff
\begin{equation}
 U=\left\{v:\sum_{e\ni v}Y_e T_e(y)\le4(J-2)+\lfloor\Gamma\rfloor\right\}.
 \label{eq:observable-domain}
\end{equation}
With probability $1-o(1)$ each component of $U$ has fewer than $J$ vertices, and every perturbed optimum has a representative equal to $y$ off $U$, even for deterministic $\Gamma=o(\log n)$. Undoing one sign in a disagreement component of size $b<J$ cannot improve the optimum. The other $b-1$ signs affect at most $2(b-1)$ incident observations, each by at most two units; the weight changes contribute at most $\Gamma$. This gives~\eqref{eq:observable-domain}.

An observation meets at most one free component. For each assignment of the free signs in a component, tabulate the score of observations meeting it. All other signs in those observations are fixed. The optimum is the sum of the table maxima and the score of observations meeting no free component. After linear preprocessing for fixed $k,J$, a change to $\ell$ specified coefficients costs $O(\ell 2^J)$ arithmetic operations to update the optimal value and the affected component choices. Writing out all $kn$ signs costs $O(kn)$. Related approaches use backdoor variables, compiled representations and block stability to identify tractable decisions \citep{williams2003backdoors,darwiche2002map,darwiche2004weighted,lang2019block}. Here the random field law also determines how many decisions must remain available.

\paragraph{How much can one observation change?}
Fix an observed product and optimize twice, once with its predicted sign fixed to $+1$ and once to $-1$. Half the absolute gap between the two scores is the smallest change in its coefficient that can bring them to a tie. Above $\alpha=1/2$, one optimum suffices to read off all these margins up to any fixed budget $B$. With high probability, at every observed address the clipped margin equals the smallest one-flip field at its endpoints, clipped at $B$. One pass accumulates the fields and a second scans the addresses. Throughout $0<\alpha<1$, the number of addresses whose preferred sign can be reversed at a fixed budget $\Gamma>0$ is asymptotic in probability to $2c\sqrt{q(1-q)}\,d_{\rm fix}(\Gamma)\log n$ (Proposition~\ref{prop:edge-response-order}). One free sign can control many margins.

Two reweightings chosen after seeing the target can force opposite signs on nearly every reversible coordinate. Their common coordinate domain then needs the full leading dimension in~\eqref{eq:dimension-constant-main} although two stored answers suffice. For a finite list in the isolated regime, join queries prescribing opposite signs on a coordinate. The minimum number of answers is the chromatic number of this conflict graph (Proposition~\ref{prop:order-query-graph}). For example, $(+,*)$ and $(*,-)$ share the answer $(+,-)$, while $(-,+)$ requires a second answer; a star permits either sign.

When the last corrections disappear, the counts themselves have a limit law. In the critical window $hm/n=\log n-\tfrac12\log\log n+t+o(1)$ with fixed offset $t$, the counts at any fixed finite set of fields $j$ converge to independent Poisson variables with means $\theta e^{\lambda j}$, where $\theta=kA e^{-t}$ with $A=(4\pi\gamma/h)^{-1/2}$. The source is optimal with limiting probability $\exp[-\theta a/(1-a)]$ and uniquely optimal with probability tending to $\exp[-\theta/(1-a)]$. For a growing budget $\Gamma_n$ with $b=\lceil\Gamma_n\rceil-1=O(\sqrt{\log n})$, the robustness window shifts by $(\lambda/h)nb$ observations to leading order, with a quadratic correction to the limiting probability (Proposition~\ref{prop:moving-budget}).

\section{Four orientations, all coordinates}\label{sec:clauses}
Four restricted optimizations can suffice even when a single common coordinate domain must leave all $3n$ signs free. The four even-block reversals of three factor blocks represent one parity tensor, but a clause can distinguish these orientations.

At each observed triple, a clause is satisfied unless its three signs match one forbidden pattern. Assign probability $b_3=q/4$ to each pattern whose product agrees with the source and $a_3=(1-q)/4$ to each of the other four. The objective counts satisfied clauses, with coefficient $1+\xi_e$ for a reweighted clause. No individual forbidden sign or pair of signs reveals anything about the source. Writing $L_e=(L_{e1},L_{e2},L_{e3})$ for that pattern, the statistic $-\prod_{i=1}^3 L_{ei}$ is a noisy parity observation with bias $\mu$. Conditional on that product, the four possible patterns are equally likely. Thus the parity statistic retains all the source information, while the observed pattern determines which assignments actually satisfy the clause \citep{feldman2018planted}.

A coordinate's signed change in clause satisfaction has small-field exponential rate $\alpha_3=c(\sqrt{a_3}-\sqrt{b_3})^2=\alpha/4$. At $0<\alpha_3<1$, gain over the best source orientation and log multiplicity both have order $n^{1-\alpha_3}/\sqrt{\log n}$ (Appendix~\ref{app:max3sat-transfer}). A coordinate is active in orientation $g$ if flipping it changes satisfaction of an incident clause there.

\begin{theorem}[Four orientations]\label{thm:four-main}
At $\alpha_3>1/2$ and fixed $\Gamma>0$, with probability $1-o(1)$ the smallest common coordinate domain for the clause objectives has $d_{\rm fix}=3n$ free coordinates. The smallest set of candidate answers has size $\sum_{g=1}^4 2^{R_g}$, where $R_g$ counts active coordinates with strictly reversible margins in orientation $g$. At $\alpha_3\ge1$, with high probability precisely four candidates suffice.
\end{theorem}
Use the parity readings to compute a reference. Apply the four even block reversals and form a domain around each reference using its \emph{clause} margins. With high probability these four domains have bounded-size components and contain, in their union, every optimizer at every allowed reweighting for $\Gamma=o(\log n)$. Comparing their four restricted optima gives the best clause score.

The four orientations are separated by $2n$ flips in factor space. A bounded positive local weight budget can favor any one of them by discounting the clauses it violates; the degree bound controls the local cost. In the singleton regime, any given physical coordinate is weak in at most one orientation with high probability. The other three orientations include both of its signs. Hence any cube meeting every forced answer leaves the coordinate free. Repeating this argument for all coordinates yields $3n$, even when comparing four scores solves the task.

At the critical density $h_3m/n=\log n-\tfrac12\log\log n+t+o(1)$, where $h_3=(\sqrt{a_3}-\sqrt{b_3})^2$, the four violation counts differ on the scale $\sqrt m$. A budget $\Gamma_n$ changes their gaps on the scale $n\Gamma_n$. When their ratio $n\Gamma_n/\sqrt m$ tends to a finite positive limit, Gaussian score gaps select the accessible orientations, each with an independent Poisson count of local ties. One orientation needs a vanishing fraction of the coordinates, two need fraction $2/3$, and three or four need them all asymptotically. The number of necessary answers stays bounded in probability (Appendix~\ref{app:sector-window}).

In a binary clause model, the forbidden pattern equals the source pair with probability $q<1/4$ and takes each other value with probability $(1-q)/3$. Here even individual literal frequencies carry source information. Appendix~\ref{app:max2sat} gives a target-only update and component search attaining the optimum with high probability in linear arithmetic work.

\section{What establishes that the fit is exact?}\label{sec:finite-certificates}
Checking a score takes one pass; verifying optimality requires excluding every better candidate. The radius criterion holds simultaneously for all high-scoring assignments, even those chosen after seeing the observations.

\begin{theorem}[Sharp radius for high-scoring assignments]\label{thm:radius-main}
For fixed $k\ge3$, $c,q$ there is a constant $\delta>0$ such that, with probability $1-o(1)$, every assignment with $F_Y(x)\ge\mu(1-\delta/3)m$ attains the global optimum if and only if no change of at most
\[
 R_*(\alpha)=\max\{1,\lfloor 1/\alpha\rfloor\}
\]
coordinates strictly improves it. Each smaller positive radius $R$ can miss a two-unit optimality gap above any fixed cutoff below $\mu m$, with probability $1-o(1)$ except at $R=1,\alpha=1/2$ (probability bounded away from zero).
\end{theorem}
First test single flips. For a high-scoring assignment stable under these flips, disagreement with a better optimum splits after block alignment into components of at most $R_*$ weak coordinates. Gains add, so one component must improve the score. Isolated paths show why the radius cannot be reduced. For the constructed path, only a flip of the entire path recovers the missing two score units.

At $\alpha=1/2$ the required radius is two. One family of oriented weak pairs alone gives a limiting inferior probability of at least $1-e^{-1/(4\pi)}$ that a one-flip test accepts some suboptimal high-scoring assignment. More generally, the number of specified oriented paths on $t$ vertices has a Poisson limit when $thm/n-\log n-(t/2-1)\log\log n$ converges (Theorem~\ref{thm:order-path-window}). Other obstructions can also occur. For triples the worst score deficit among passing one-flip-stable assignments has a separate two-type Poisson limit in the pair window (Appendix~\ref{app:pair-window}).

\paragraph{A finite test.}
For a realized candidate define $\eta_e(x)=Y_eT_e(x)$, $d_v(x)=\sum_{e\ni v}\eta_e(x)$, and let $w_{uv}$ count the positive signed observations containing both $u,v$. In an inclusion-minimal strictly improving flip set $S$, the following inequality holds for every $v\in S$:
\[
 \sum_{u\in S\setminus\{v\}}w_{uv}\ge
 \max\{0,\lfloor d_v(x)/2\rfloor+1\}.
\]
Indeed, removing $v$ from $S$ destroys its strict improvement. This forces a signed contribution from edges joining it to the other flips, and retaining just the positive edges bounds that contribution above. A proposed neighbor must itself have enough neighbors, all within the same cardinality budget. A two-hop integer test accounts for these obligations, prunes impossible sets, and enumerates what survives (Appendix~\ref{app:twohop-radius}). Successful exhaustion certifies a finite radius on that specific input. Upgrading it to a finite global certificate would also require an independently checkable exclusion of larger improving sets.

\paragraph{Global proof systems.}
On the same tripartite masks, the sum-of-squares (SOS) relaxation upper bound still treats all $m$ observations as satisfiable through degree of order $n/(\log^2 n\log\log n)$. The actual optimum is $(\mu+o(1))m$. This is a linear gap even when two signs in distinct blocks are fixed to choose a representative (Theorem~\ref{thm:sos-certification}). A syndrome check constrains errors along a set of products in which all coordinates occur evenly. Intersecting their individual convex hulls still allows a fractional low-error solution, since every nonzero check involves many observations. The resulting gap and size bound for parity linear-programming (LP) branch trees concern those proof systems (Appendices~\ref{app:sos-certification} and~\ref{app:syndrome-barrier}). An exact comparison of a proposed common domain against all competitors is available by mixed-integer optimization. A rational solution tree certifies an accepted finite instance, with no general polynomial bound on tree size.

\begin{figure*}[t]
\centering
\includegraphics[width=.95\textwidth]{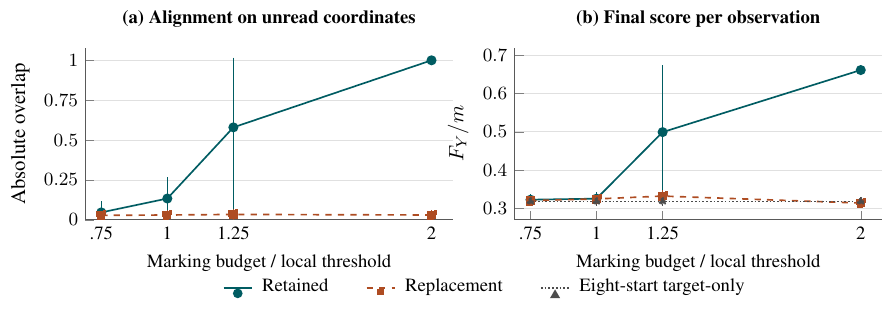}
\caption{\textbf{Returning to the source raises scores on the original data.} At higher tested budgets, retained readings align unread signs (a) and raise feasible scores (b), unlike independent replacement. Means and observed ranges on the same 12 order-four targets at $n=1{,}024$; the horizontal axis is the marking budget divided by the local threshold. Alignment is absolute block overlap on unread noncenter coordinates before score refinement. Scores are not certified optima. Budgets exclude 33 separately charged center readings; target-only uses eight starts.}
\label{fig:fourway}
\end{figure*}

\paragraph{What a positive certificate looks like.}
For tensors $x,y$, let $d(x,y)$ count the fewest differing signs between their representatives. Every observation whose product changes touches at least one flipped coordinate. Hence
\[
 F_\xi(y)-F_\xi(x)\ge F_Y(y)-F_Y(x)-2\Gamma d(x,y).
\]
If for each $x$ outside a proposed domain some $y$ inside makes the right side positive, the domain contains every perturbed optimum. On four constructed grids of orders $3,4,5,6$, this inequality certifies a two-answer domain for every real perturbation with $\Gamma=2$. The source loses by two unweighted score units, yet another admissible weighting makes it uniquely optimal. Every outside competitor was checked. Appendix~\ref{app:order-law} gives the grids and the strict margins.

\section{Finite comparisons}\label{sec:experiments}
On a small target we can enumerate every product tensor and check the quantities predicted by the theory directly. On larger targets we can measure local fields and compare feasible scores on the same original observations.

\paragraph{When do local margins explain the optimum?}
At $n=4$, all three isolated-field formulas can be checked by enumeration: gain and multiplicity on complete tensors and on the observed addresses.
\par\smallskip\noindent\begin{minipage}{\columnwidth}
\captionof{table}{Targets satisfying all three field formulas ($q=.2$, 32 per cell).}
\label{tab:field-check-main}
\centering
\begin{tabular}{lrrr}
\toprule
Interaction order & $\alpha=.4$ & $\alpha=.8$ & $\alpha=1.6$\\
\midrule
Three & 6/32 & 15/32 & 28/32\\
Four & 2/32 & 5/32 & 19/32\\
Five & 0/32 & 3/32 & 13/32\\
\bottomrule
\end{tabular}
\end{minipage}\par\smallskip
Two enumerations agreed on all 7,176,192 complete-tensor scores. On 31 of 288 targets every source field was strictly positive and the source still lost. Each individual flip lowers its score there. An improving simultaneous flip accounts for the difference (Appendix~\ref{app:edge-response-order}). 

\begin{table}[t]
\caption{Balanced readings for tripartite fitting. Entries in the last three columns count scores strictly above the generating source. Eight independent targets per size are reused across budgets; Cold uses one start. $Q/n$ is the observed mean.}
\label{tab:balanced-finite}
\centering
\begin{tabular}{r r r r r r}
\toprule
$n$ & Budget$^{\dagger}$ & $Q/n$ & Retained & Repl. & Cold\\
\midrule
4,096 & .6 & .106 & 0/8 & 0/8 & 0/8\\
      & .9 & .159 & 3/8 & 0/8 & 0/8\\
      & 1.2 & .211 & 8/8 & 0/8 & 0/8\\
16,384 & .6 & .088 & 0/8 & 0/8 & 0/8\\
       & .9 & .132 & 8/8 & 0/8 & 0/8\\
       & 1.2 & .176 & 8/8 & 0/8 & 0/8\\
\bottomrule
\end{tabular}\\
$^{\dagger}$Normalized expected budget $\mu^2d\rho^2\sum_i p_i$ (threshold $3/4$; unlike Figure~\ref{fig:fourway}, not divided by the threshold).
\end{table}

Enumeration also gives exact switching margins at all $7{,}680$ addresses. At budget two the source fields predict $4{,}634$ of the clipped margins correctly. The fields at an attained global optimum predict $4{,}677$ correctly, including every address on 63 of the 288 targets. At these sizes a single-bit test can therefore misstate the response even when the starting assignment is optimal.

\paragraph{Exact small-target comparisons.}
With $\lceil n/4\rceil$ one-fiber readings, retained and replacement arms attain 135/370 and 83/370 exact optima; a 32-start target-only solver attains 201/370. With $n$ readings on the 124 source-suboptimal middle-density targets, retained attains 50/124 versus 61/124 target-only. In a prospective reduced-reading study, retained attains 42/96 versus 38/96 replacement, an inconclusive difference. On its 64 source-suboptimal targets the counts are 20 and 23, respectively (Table~\ref{tab:finite-study}). The proved calibration cascade falls back to $3n$ readings at all tested sparse exact-reference sizes. Two additional updates improve its retained exactness from 73/536 to 200/536 (10/180 to 38/180 on source-suboptimal middle-density targets); the 32-start target-only solver attains 240/536. These schedules have different work allocations (Appendix~\ref{app:reuse-ablation}).

\paragraph{Reduced readings on large targets.}
Table~\ref{tab:balanced-finite} uses $c=2.5$, $q=.17$ and $s=.10$. Retained and replacement arms share addresses. All use 24 damped message rounds and eight refinement sweeps; the cold arm reads no source and starts with independent fair signs. At the middle budget all eight larger retained runs beat the source with $.132n$ readings on average ($.044$ per sign). On 48 fresh paired cases, direct initialization at the reading priors ties projection in 34 and improves it in 14; at the lowest shared budget the two beat the source on 15/16 and 3/16 targets. In a retrospective projected comparison at $n=4096,16384$, mean retained scores per observation were $.6590,.6602$, versus $.3365,.3151$ for replacement and $.3389,.3147$ for cold, with 16/16 wins over both controls at each size. These are feasible scores from loopy messages and polishing, not capped exact search. Controls match caps, not runtimes; projection adds message work. Global optima and the finite threshold remain unknown (Appendix~\ref{app:soft-experiment-study}).

\paragraph{Four-way products.}
Figure~\ref{fig:fourway} compares twelve targets at each of $n=256,1{,}024$ across four nested budgets ($m=\lceil4n\log n\rceil$, $\mu=.66$, $\rho=.76$). Both reading arms use the same marks, 33 extra center readings, twelve message rounds and six refinement sweeps. Target-only keeps the best of eight six-sweep starts.

At budget ratios $.75,1,1.25,2$, retained scores reach or exceed the source on $0,0,10,24$ of 24 targets; replacement on none. At ratio two, four exceed the source and twenty tie; all 24 beat both controls. Eight target-only starts cost more CPU time on every target ($.536$ seconds total versus $.332$ retained, excluding acquisition). The larger runs average $1{,}604.1$ readings for $4{,}096$ signs. No global optimum is available at these sizes.

On 24 four-way targets with exhaustive references ($n=5,6$), the generating source is suboptimal in 18 cases. At the four budget ratios, the retained arm attains $2,4,3,6$ exact optima and replacement attains $1,3,3,1$. The eight-start target-only solver attains five. Two enumerations agree on every optimal value and multiplicity. Target-only attained more optima at both intermediate budgets (Appendix~\ref{app:fourway-experiments}).

\paragraph{How often does verification change the answer?}
On a separate set of 32 large tripartite targets, all high-budget retained solutions from a one-fiber initializer outperform the eight-start cold solver. All are single-flip stable, yet nine admit strictly improving pairs. Repair accepts 15 moves in total, for 40 additional score units and no further readings. After repair, deterministic two-hop verification excludes every improving flip of at most 24 coordinates on 32/32 targets and at most 32 coordinates on all 16 larger targets. The available probability bound for exclusion beyond the verified radius is vacuous at these sizes. The global optima for these targets remain unknown.

\paragraph{Certifying future decisions.}
On $600$ prespecified domains ($n=8$, 150 targets, $3n$ readings per arm, two budgets), rational replay gave 22 proper-domain certificates, all at $\Gamma=1/2$ and none at $\Gamma=2$; 156 invalid, 127 unrestricted and 295 unresolved domains. Of the invalid domains, 22 had answered all 24 test queries correctly. Fixing the free signs missed five proper certificates (Appendix~\ref{app:component-certification-study}).

A four-way fixture passes every solver cap with zero readings and no free coordinates, yet scores 112 against optimum 144. On seven other constructed targets all 23 restricted answers match exhaustive optima, including three coupled-pair queries. On twelve sparse even-order targets the two fiber-center sign hypotheses yield different scores eight times, with each sign winning on some targets. A restricted optimum may still be beaten by an assignment excluded from the proposed domain.

\section{Discussion}
The original observations determine the optimum, but need not provide an efficient route to it. Returning to the source changes that route without changing the objective. Independent-source readings share the marginal law but lack the dependence needed for cancellation and local inference. Sufficient alignment followed by strict ascent identifies a bounded-component restriction with high probability. Maximizing the original score on these components can be essential even after exact recovery of the source.

At fixed density and noise, interaction order governs the scale of acquisition but not normalized residual freedom. That freedom reflects a family of future objectives, not just ties at the current optimum. Even a uniquely optimal source can require free signs under reweighting. A common domain prepared before the edits permits updates to just the affected component tables. The clause model shows why representation matters: a single common cube requires all $3n$ signs to remain free, whereas four separate domains preserve the decomposition into bounded components.

The verification radius determines how far a high-scoring assignment must be tested under the sampling law. A finite radius test alone cannot certify global optimality, and the relaxation gaps show why optimization and proof costs can differ. The finite comparisons establish score improvements, not practical reduced-reading exactness. Whether adaptive or nonlocal inference can reduce the reading budget, whether target-only exact optimization is tractable, and how to acquire with unknown channels at higher orders remain open. These results suggest a criterion for measurement design: the computational structure a measurement exposes, rather than only the uncertainty it removes. Measurement and computation can be traded against one another even when the objective and its exact answer are fixed.

\label{sec:main-end}
\section*{Impact Statement}
Additional observations can reduce the cost of optimizing a fixed objective when the source can be measured again. The worst-case hardness concerns arbitrary inputs, and the algorithmic guarantees concern noisy measurements from the stated model. Source drift or replacement can invalidate the gain from acquisition. In applications requiring a verifiable optimum, a feasible solution and its score alone do not establish global optimality.
\section*{AI Disclosure Statement}
Generic large language models helped with theorem formulations, proof exploration, coding, adversarial audits, and drafting of the manuscript. Seeded programs generated the synthetic records. Verification included separate derivations of the posterior bounds, independent field and path counts, direct enumeration of small instances, replay of stored witnesses against the original objectives, and exact arithmetic checks. The supplemental code and data identify the checks and their denominators. The largest tripartite targets have feasible solutions with finite-radius verification. The large four-way targets have feasible scores.
\bibliography{references,certification_barrier,references_certification,references_tree_size,references_residual,references_projection,references_clauses,references_seeded_projection,references_local_acquisition,references_basin}
\bibliographystyle{icml2027}
\clearpage
\appendix
\renewcommand{\thesection}{\AlphAlph{\value{section}}}
\onecolumn
\section*{Guide to the proofs and experiments}
Each entry gives the construction, its scope and the corresponding proof or experiment. The known-channel sum-product theorem is the primary sharp acquisition result; the separate triple calibration cascade estimates channel biases and uses fresh batches, with explicit finite execution gates in Appendix~\ref{app:finite-gate}. The projection and erasure schedules are alternative estimators and finite comparators, not additional stages of the primary theorem.

\paragraph{Notation used in the proofs.}
Unless a section specifies another model, $n$ is the block size, $k$ the interaction order, $H$ the observed mask, $m=|H|$, and $d=m/n$. A tuple $e$ is also viewed as the set of its $k$ endpoints $v=(i,e_i)$. The source signs are $X_v$, with tensor $T$, observed labels $Y_e=T_e\eta_e$, target and reading error probabilities $q,s$, and biases $\mu=1-2q$, $\rho=1-2s$. The unweighted score is $F=F_Y=F_0$, with maximum $\OPT$; $F_\xi$ and $\mathcal U_\Gamma(H)$ are given in~\eqref{eq:main-task-neighborhood}. The source-relative sign of an estimate $x_v$ is $x_vX_v$, and its source field is $D_v=\sum_{e\ni v}\eta_e$. An anchor is a prescribed factor sign used to choose one representative of a parity tensor. For triples, the source blocks are often written $a,b,c$, the noiseless reference fibers $A,B,C$ are those of~\eqref{eq:fiber}, and local uses of $k$ as a prefix length or coordinate index are stated separately. We use $\alpha=ch$, $h=1-2\sqrt{q(1-q)}$, and $\lambda=\tfrac12\log((1-q)/q)$ for the signed-field rate and tilt, unless locally redefined.

All asymptotic statements use $n\to\infty$ with the declared fixed parameters. Logarithms are natural unless a base is indicated. The notation $o_{\Pr}(a_n)$ means a quantity divided by $a_n$ tends to zero in probability; $O_{\Pr}(a_n)$ means that ratio is bounded in probability. We write $\E$ for expectation, $\one$ for the indicator, $(u)_j=u(u-1)\cdots(u-j+1)$ for the falling factorial, and $\operatorname{Bin}(N,p)$ and $\operatorname{Pois}(\theta)$ for binomial and Poisson laws. The abbreviation iid means independent and identically distributed. When a section uses a symbol in a different sense, its local convention takes precedence.
\begin{description}
\item[Acquisition.] Appendix~\ref{app:order-law} gives the order-dependent reading threshold, its local converse, and the passage from a high score to exact empirical optimization. Appendix~\ref{app:triple-calibration-reuse} states the triple calibration and reuse results. For triples, Appendix~\ref{app:balanced-acquisition} gives the sharp balanced construction, Appendix~\ref{app:geometric} handles unknown channel parameters, and Appendix~\ref{app:one-fiber} treats one-fiber projection. Appendices~\ref{app:seeded-projection} and~\ref{app:erasure} give the corresponding restricted thresholds. Appendix~\ref{app:lower} compares retained and replacement records.
\item[Residual decisions.] Appendix~\ref{app:residual-order} proves the fixed-order field, dimension and multiplicity laws, the critical window, and the characterization for finite query lists. Corollary~\ref{cor:marked-sensitivity} gives the joint dimension and sensitivity fluctuations in the critical window. Proposition~\ref{prop:moving-budget} gives the window for a growing reweighting budget. Appendices~\ref{app:local}, \ref{app:direct-mixed-moment}, and~\ref{app:recycling} prove localization, dependence control, and reuse of the target. Appendices~\ref{app:common-reduction}, \ref{app:dimension-exact}, and~\ref{app:residual-laws} give the common restriction, its minimum dimension, and the gain and multiplicity laws. The explicit tables for sparse edits are in Appendix~\ref{app:common-reduction}.
\item[Clause objectives.] Appendices~\ref{app:max2sat} and~\ref{app:max3sat-transfer} derive the two clause models from their forbidden-pattern laws. Appendix~\ref{app:sector-window} treats competing orientations, including the Gaussian and Poisson limits.
\item[Verification.] Theorem~\ref{thm:order-radius} and its proof give the sharp radius at every fixed interaction order. Appendices~\ref{app:score-basin} and~\ref{app:pair-window} give the sharp local verification radius and its critical window. Appendices~\ref{app:local-radius} and~\ref{app:twohop-radius} give the deterministic radius tests. Appendices~\ref{app:robust-certificates}, \ref{app:exact-common-certification}, and~\ref{app:common-search} treat finite common-domain proofs. Appendices~\ref{app:sos-certification} and~\ref{app:syndrome-barrier} give the global relaxation gaps.
\item[Experiments.] Appendix~\ref{app:edge-response-order} gives the exact fixed-order and field-count studies. The four-way matched comparisons, solver checks, and timing protocol are in Appendix~\ref{app:fourway-experiments}. The matched small-target projection study is in Appendix~\ref{app:one-fiber}, the large full-record trials in Appendix~\ref{app:soft-experiment-study}, and the frozen uniform-certificate study in Appendix~\ref{app:component-certification-study}. The underlying records accompany the reproducibility supplement, supplied as an ancillary archive with the article. Each finite result identifies whether its reference is an exhaustive optimum, a feasible score, or a restricted certificate.
\end{description}

\section{Acquisition at every fixed interaction order}
\label{app:interaction-order}\label{app:order-law}
Take $k$ blocks of $n$ vertices, with $k\ge3$ fixed. Draw $m=\lceil cn\log n\rceil$ distinct ordered tuples uniformly from $[n]^k$, and give each vertex $v$ an independent fair source sign $X_v$. A tuple $e$ reports $Y_e=\eta_e\prod_{v\in e}X_v$. The independent errors $\eta_e$ have mean $\mu\in(0,1)$. A noisy reading of a vertex reports its sign through an independent binary symmetric channel of bias $\rho\in(0,1)$. We give the implementation with physical reference fibers below. The score to be maximized is
\[
 F_\xi(z)=\sum_e(Y_e+\xi_e)\prod_{v\in e}z_v,\qquad
 \max_v\sum_{e\ni v}|\xi_e|\le\Gamma_n .
\]
Write $F=F_0$. Two assignments differing by constant block signs of product one represent the same products. A component of a set of vertices uses adjacency by a common observed tuple.
Put $d=m/n$, $r=k-1$, and
\[
 L=(\mu^2d)^{1/(k-2)},\qquad b=r^{-1/(r-1)},\qquad
 B_k=k(b-b^r)=\frac{k(k-2)}{(k-1)^{(k-1)/(k-2)}}.
\]
The coefficient per vertex, $B_k/k$, also occurs in noiseless hypergraph peeling and bootstrap percolation \citep{mori2015peeling,morrison2021}. In Theorems 2 and 3 of \citet{mori2015peeling}, specialize their parameter $r$ to $k$, with $N$ vertices, $\ell$ initially revealed vertices and $M=\nu N^{k-1}/\ell^{k-2}$ hyperedges. Their critical $\nu$ is $(k-2)^{k-2}/[k(k-1)^{k-1}]$. Writing the mean degree as $d=kM/N$ gives $(\ell/N)d^{1/(k-2)}=(k\nu)^{1/(k-2)}=B_k/k$ at the threshold. In their Theorem 1.1, an order-$k$ hypergraph of degree $D$ is thinned at probability $b D^{-1/(k-1)}$. Its mean degree after thinning is $d_{\rm thin}=bD^{(k-2)/(k-1)}$, and the critical initial active fraction is asymptotic to $(B_k/k)d_{\rm thin}^{-1/(k-2)}$. Here the recursion acts on the information in noisy channels. The conclusion concerns the maximizer of a noisy empirical score.

\begin{theorem}[Interaction order and acquisition]\label{thm:interaction-order}
Fix $c>0$, $k\ge3$, known rational biases $0<\mu,\rho<1$, and a deterministic budget $\Gamma_n=o(\log n)$, $\Gamma_n\ge0$. For each fixed $\epsilon>0$ an algorithm uses at most $(B_k+\epsilon)n/(\rho^2 L)$ readings for all sufficiently large $n$. Its degree and component caps are fixed from $c$ and $\mu$ before observing the labels. It returns a feasible unweighted answer on every input. With probability $1-o(1)$ the answer attains $\max F$, and the algorithm constructs one observable fixed-coordinate domain with bounded-size components containing a representative of every maximizer of every admissible $F_\xi$. Construction takes $O((m+n)\log(n+2))$ word operations on every input. On the success event each full-record reweighting query costs $O(m+n)$ arithmetic operations for fixed $k,c,\mu$; comparisons of supplied weights are exact.

For the converse, mark vertices independently with block probabilities $p_i$, independently of the target. Let $A=L\rho^2\sum_i p_i$. Suppose $A\le B_k$ at each sufficiently large $n$. In each block the absolute overlap of any sign estimates with the source tends to zero in probability if each output uses only its own radius-$O(\log\log n)$ record. Radius counts variable-to-variable steps. Vertex indices and randomness independent of the instance are allowed. The converse concerns local inference with these addresses. Above the threshold, balanced independent marks with $L\rho^2p_i=a>B_k/k$ yield overlap tending to one from neighborhoods of the same asymptotic depth, even for physical fibers with an unknown center sign. Global postprocessing and target-dependent choices of addresses lie outside the converse's scope.
\end{theorem}

\begin{proof} We first bound the information in the observations descending from a single factor.
On the comparison tree each variable has $\operatorname{Pois}(d)$ child factors, each with one child in every other block. Sources, channel errors, and marks are independent. For a type-$i$ root let $x_{i,t}$ be the expected square of its posterior mean from $t$ generations. Write $u_i=p_i\rho^2$. Then $x_{i,0}=u_i$. If $M_j$ denotes the posterior mean from child $j$, the channel comprising its sibling group and the observed tuple has posterior mean $M_f=\mu Y\prod_{j\ne i}M_j$ under a fair parent prior. Its second moment is $v=\mu^2\prod_{j\ne i}x_{j,t}$ by independence. The mark channel and the factor channels are conditionally independent given the parent sign $X$ and the observed child count. All are symmetric binary-input channels. With $s=u_i+dv$,
\begin{equation}\label{eq:arity-sandwich}
 \frac{s}{1+s}\le x_{i,t+1}\le s.
\end{equation}
For the upper bound, apply Lemma~\ref{lem:binary-energy-study} conditionally on the count and average. For the lower bound let $S_{\rm msg}$ be the sum of individual posterior means. A symmetric channel with second moment $w$ has conditional mean $wX$ and conditional second moment $w$, by symmetry and $\mathbb E[XM]=\mathbb E M^2$. Distinct channels have cross moment equal to the product of their second moments by conditional independence given $X$. Thus $\mathbb E[XS_{\rm msg}]=s$ and the second Poisson factorial moment gives $\mathbb E[S_{\rm msg}^2]=s+2u_i dv+(dv)^2\le s+s^2$. The sum is measurable in the full observations, so its inner product with the full posterior mean equals $\mathbb E[XS_{\rm msg}]$ by conditional expectation. Cauchy--Schwarz gives the lower bound, including $s=0$.

\paragraph{A threshold for local inference.}
Set $z_{i,t}=Lx_{i,t}$ and $a_i=Lu_i$. The upper recursion is
$z_{i,t+1}\le a_i+\prod_{j\ne i}z_{j,t}$. Let $S_t=\sum_i z_{i,t}$. Maclaurin's inequality for elementary symmetric means gives
\[
 S_{t+1}\le A+k(S_t/k)^r .
\]
Since $B_k+kb^r=kb$, the interval $[0,kb]$ contains $S_0=A$ and is invariant when $A\le B_k$. This includes equality at the actual finite-$n$ value of $A$.

For equal $a_i=a>B_k/k$, the scaled lower and upper bounds in \eqref{eq:arity-sandwich} converge at each fixed depth to $w_0=a$, $w_{t+1}=a+w_t^r$. The increasing sequence has no finite limit, because $\max_{w\ge0}(w-w^r)=b-b^r$. At some fixed depth $z_t\ge4$ for large $n$. While $z_t^r\le L$, the lower bound gives $z_{t+1}\ge z_t^r/2\ge2z_t$. Once $z_t^r\ge L$, the next energy is at least $1/2$. One further round gives an energy at least $1-O(1/d)$ since $d\mu^2 2^{-r}\to\infty$. A deterministic depth $C_k\log\log(n+3)$, with a sufficiently large fixed $C_k$, suffices. The energy is nondecreasing with depth. Choose a rational $a$ so that $B_k<ka<B_k+\epsilon$ with fixed slack. Compute $p_i=a/(\rho^2 L)$ up to dyadic error $O(n^{-3})$, truncated at one for small $n$, by rational bisection of the fixed-order root defining $L$. The needed precision and number of bisection steps are $O(\log n)$. Abort before reading if the stated cap would be exceeded. The mean total is $(ka+o(1))n/(\rho^2L)$ and diverges, so the cap changes the success probability by $o(1)$.

\paragraph{Reading the same graph at growing depth.}
For iid tuples, repetition has probability $O(m^2/n^k)=o(1)$, so any event of probability $1-o(1)$ transfers to the uniform mask. Poissonize the tuple count with mean $m$. Each address now has independent Poisson count of mean $d/n^{k-1}$. For one or two roots, the comparison trees have expected size at most $O_k(\sum_{j\le t+1}(1+rd)^j)=n^{o(1)}$. Stop exploration after $M=n^{1/20}$ variables or factors. Before stopping, incident tuples with fresh endpoints arrive with mean $d(1-O_k(M/n))$ per exposed vertex. Exposing tuples touching an old vertex costs $O_k(dM^2/n)$ in coupling probability. Coincidences among the new child indices cost $O_k(M^2/n)$. The stop probability is $o(1)$ by Markov's inequality on the comparison trees. Couple fresh indices, signs, marks and observations; the two approximating rooted records are independent. For a fixed tuple count use the same iid sequence: the Poisson count differs from $m$ by at most $n^{2/3}$ with probability $1-o(1)$. Conditional on the two counts, the neighborhood in the smaller graph has expected size $n^{o(1)}$ by counting simple paths. The extra tuples hit that neighborhood, up to the stop, with probability $O_k(n^{2/3}M/n)=o(1)$. These estimates are uniform over distinct specified roots.

Below the threshold, each tree output's expected correlation has absolute value at most $\sqrt{x_{i,t}}=O(L^{-1/2})$ by Cauchy--Schwarz. Condition on any independent shared random seed. The two-root coupling then makes the expected product of the source-signed outputs $O(1/L)+o(1)$ for distinct roots in a block. Summing the square of the block overlap gives $o(1)$. Above the threshold, applying sum--product and taking the sign of the posterior mean makes the expected error fraction $o(1)$ in each block, because $\mathbb E|M|\ge\mathbb E M^2=1-o(1)$. This conclusion uses a one-root coupling and Markov's inequality. It does not require independent coordinate errors.

\paragraph{From high empirical score to exact reduction.}
We prove a uniform fact about the original target. Fix any sufficiently small constant $\delta>0$, chosen below. With probability tending to one, every assignment whose score is within $o(m)$ of $\mu m$ or higher has a gauge representative with at most $\delta n$ errors relative to $X$. Indeed in source-relative coordinates $w_v=z_vX_v$ its normalized expected score is $\mu\prod_i\bar w_i$ in the iid model, where $\bar w_i$ is its block average. Hoeffding's inequality and a union bound over $2^{kn}$ relative assignments bound the uniform deviation by $(\log n)^{-1/4}$ with probability $1-o(1)$. A product within $o(1)$ of one has all magnitudes $1-o(1)$ and positive sign. Multiplication by the $k$ block-majority signs then preserves every tuple product. For the degree cap choose an integer $D>2+2c$. In the iid mask, $\E\exp(\deg(v))\le\exp((\mathrm e-1)d)$, so a union bound over $kn$ vertices puts the maximum below $D\log(n+2)$ with high probability. This event transfers to the simple mask. Hence the $o_{\Pr}(n)$ errors of the initializer change $F(X)$ by $o_{\Pr}(m)$, and $F(X)/m\to\mu$ in probability.

Starting there, repeatedly flip any coordinate that strictly improves $F$ until none does. Call the result $y$. It inherits the score lower bound. The same bound holds for every maximizer of every admissible $F_\xi$ on one event since $\sum_e|\xi_e|\le n\Gamma_n=o(m)$ and $X$ is feasible. Each such maximizer, and $y$, is one-flip stable for its own score. We now show why stability forces its remaining errors into a bounded-component set.

For any one of these representatives let $S$ be a connected component of its errors. Put $t=|S|$ and $j_e=|e\cap S|$. No tuple touching $S$ meets another error component. Summing the one-flip stability inequalities over $v\in S$ gives
\begin{equation}\label{eq:arity-stable}
 \sum_e j_e(-1)^{j_e}\eta_e\ge -t\Gamma_n .
\end{equation}
Set $h=1-\sqrt{1-\mu^2}$ and $\lambda=\operatorname{arctanh}\mu$. For a fixed $S$ with $t\le\delta n$, a fresh tuple has $\Pr(j=1)\ge(t/n)(1-t/n)$ and $\Pr(j\ge2)\le (t/n)^2/2$, by independence across blocks: for $t_i$ the number of vertices of $S$ in block $i$, the probability of exactly one hit is $\sum_i(t_i/n)\prod_{j\ne i}(1-t_j/n)\ge (t/n)(1-t/n)$, and a union bound on block pairs bounds the probability of at least two hits. Thus
\[
 Z_S=\mathbb E\exp(\lambda j(-1)^j\eta)
 \le 1-h(t/n)(1-t/n)+(e^{k\lambda}-1)(t/n)^2/2
 \le 1-ht/(2n)
\]
with $\delta$ sufficiently small. Under the product law tilted by these single-tuple factors, any tuple has probability at most $C n^{-s}$ of hitting $s$ specified compatible endpoints, since its density ratio is bounded by $e^{2k\lambda}$. Here and below $C$ can depend on fixed parameters.

To count connected $S$ under the tilt that depends on $S$, span its incidence graph using a rooted ordered tree with $t$ variable nodes and $l\le t-1$ distinct factor nodes. One can take a spanning tree of the induced incidence graph on $S$ and its connecting factors and delete all factor leaves. Each remaining factor joins at least two variables. For given $t,l$ there are at most $4^{t+l}$ plane tree shapes, $(kn)^t$ assignments of vertex indices, and $m^l$ assignments of observation indices. In the probability bound only distinct factors and compatible vertex indices are used. The incidences prescribe $t+l-1$ endpoints across the $l$ factors, giving at most $C^l n^{-(t+l-1)}$. Summing also over $l$ gives $n[C(1+d)]^t$ for the sum of these tilted connectivity bounds. Undo the tilt in \eqref{eq:arity-stable}. The expected number at size $t$ of connected sets satisfying the inequality is at most
\[
 n[C(1+d)]^t\exp(\lambda t\Gamma_n-hdt/2).
\]
The sum from any fixed integer $J\ge2$ with $J>2/(ch)$ through $\lfloor\delta n\rfloor$ tends to zero. All error components therefore have fewer than $J$ vertices. A fixed cross-block pair belongs to $n^{k-2}$ tuples and is hit with probability $n^{-2}$ by a uniform tuple. The maximum codegree is at most two with probability $1-o(1)$ since the expected number of vertex pairs in three distinct tuples is $O_k(n^2 m^3/n^6)=o(1)$. For an erroneous $v$, the vote of another flipped vertex changes its one-coordinate comparison by at most twice the codegree. Thus its field at the source, $D_v=\sum_{e\ni v}\eta_e$, satisfies $D_v\le L_0=\Gamma_n+4J$. This argument includes tied maximizers.

\paragraph{Constructing the domain from the observations.}
For any deterministic $B_n=o(\log n)$, set $W_{B_n}=\{v:D_v\le B_n\}$. Choose a fixed integer $K\ge2$ with $K>1/(ch)$. With high probability all components of $W_{B_n}$ have size less than $K$, and each vertex has fewer than $K$ distinct neighbors in $W_{B_n}$. Here are the counting details. For $K$ specified weak vertices tilt each iid observation by $\exp(-\lambda j\eta)$, with $j$ now its number of endpoints among those $K$. The single-tuple normalizer equals $1-hK/n+O_k(K^2/n^2)$. Requiring all fields at most $B_n$ costs at most $\exp(\lambda K B_n)$ on undoing the tilt. Count the same incidence trees on $K$ vertices, or on $K+1$ vertices with one distinguished unpenalized center. The latter covers every center with $K$ distinct weak neighbors. In either case the expected count is at most $n[C(1+d)]^{K+1}n^{-chK+o(1)}=o(1)$. Any connected set of size at least $K$ contains a connected subset of size $K$. We need this conclusion for only a fixed number of deterministic cutoffs.

Align $y$ and an arbitrary perturbed optimum to the source by block gauges. Their difference is contained in $W_{L_0}$. Each component of the difference has size less than $K$. At a differing vertex, one-flip stability of the perturbed optimum and the codegree bound give
\[
 d_v(y)=\sum_{e\ni v}Y_e\prod_{u\in e}y_u\le B=\Gamma_n+4K.
\]
Indeed changing $v$ alone reverses the sign of its field, while the other differing vertices in its component can change that field by at most $4K$. Use the observable domain that fixes $y$ outside $U=\{v:d_v(y)\le B\}$. Block gauges preserve $d_v(y)$, so undoing the alignment puts an appropriate representative of each optimum in this domain. Outside $W_{L_0}$ the aligned $y$ is correct, and at most $K$ of its distinct neighbors are incorrect. Codegree at most two gives $|D_v-d_v(y)|\le4K$. Hence $U\subset W_{\max(L_0,B+4K)}$, whose components also have fewer than $K$ vertices. Each tuple touches at most one free component and its table takes at most $O_k(2^K)$ operations to accumulate and enumerate. This proves common exactness on the same event for all admissible perturbations.

\paragraph{Physical fibers and all-input cost.}
With readings on the reference fibers, the source bit read at coordinate $(i,v)$ is $T_{i,v}=X_{i,v}\prod_{j\ne i}X_{j,1}$. Write $G=\prod_j X_{j,1}$. Conditional on the center signs the noncenter $T_{i,v}$ are independent fair bits. The $k$ center bits equal $G$, and $\prod_{v\in e}T_v=G^{k-1}\prod_{v\in e}X_v$. A typical one- or two-root neighborhood meets a center with probability $o(1)$ by the exploration bound. Giving the center signs free preserves the converse for the reference-fiber signs (absolute block overlap is unchanged by a block sign). For odd $k$, the target already gives noisy products of the $T_v$.

\emph{Local recovery of the center at even order.} Let $k\ge4$ be even and use balanced independent reading marks of probability $p=a/(\rho^2 L)$ above threshold (dyadic rounding does not affect the argument). In the comparison tree the number of variables in generation three is $(1+o_{\Pr}(1))((k-1)d)^3$: conditional on a generation of size $z$, the next size has law $(k-1)\operatorname{Pois}(dz)$, so this follows by three applications of Poisson concentration. Each generation-three variable with a child factor supplies one such factor, chosen by tree structure alone. These factors are vertex-disjoint, since their parents are distinct and their children are fresh. The number lost to childless vertices is negligible (the conditional expected lost fraction is $e^{-d}$). Thus their count $N_0$ is of order $d^3$ with probability tending to one. All these factors and their readings lie within four variable steps of the root. Ignore the negligible event that the neighborhood meets a center.

Write $Z_v=T_v\epsilon_v$ for a marked reading, with independent sensor errors of mean $\rho$. For each chosen factor let its vote be zero unless all $k$ endpoints are marked, and otherwise $W_e=Y_e\prod_{v\in e}Z_v$. By the fiber identity, $GW_e=\eta_e\prod_{v\in e}\epsilon_v$ on a fully marked factor. Conditional on the tree structure and the center sign, the votes $GW_e$ are independent, with mean $p^k\mu\rho^k$ and variance at most $p^k$. Hence the probability that their sum is nonpositive is at most $1/(N_0p^k\mu^2\rho^{2k})$ by Chebyshev's inequality. This tends to zero because $N_0p^k=\Theta(d^{3-k/(k-2)})$ diverges for every even $k\ge4$. Use the sign of the sum as a local estimate of $G$ (an arbitrary tie sign suffices). The tree coupling transfers this probability estimate to a typical root. At that root use the estimated sign to correct the labels in its depth-$O(\log\log n)$ sum--product computation. On the event of a correct center estimate, its output equals the output with the true center supplied. The union bound on the two error events does not require independence between the center estimate and the inference messages. The one-root expected error probability is $o(1)$, giving overlap tending to one in each block by Markov's inequality. This uses no additional addresses, global score selection or information outside the root neighborhood. The upper and lower local thresholds therefore refer to the same physical experiment.

For the capped global exact optimizer at even order, run the initializer and ascent under both hypotheses for $G$, multiplying labels by $G^{k-1}$ for each hypothesis. Map a fitted assignment back by multiplying its first block by the same sign. This preserves its attained score in the original coordinates. Keep the higher original-score stable reference and construct $U$ there. The correct branch has high score and the retained reference satisfies the same uniform stability argument, with no extra readings.

Compute the unsigned degrees from the mask and check the cap $D\log(n+2)$ before running the initializer. Factor messages are $u=\mu Y\prod M_j$ with $|u|\le\mu$. At a variable combine incoming messages and its unary message by the normalized difference of the two products $\prod(1+u)$ and $\prod(1-u)$. Use the fixed-precision implementation in the proof of Theorem~\ref{thm:balanced-acquisition}. For completeness, relative error $O(n^{-4})$ per multiplication or exclusion division gives relative error $O(\log n\,n^{-4})$ per product, using a normalized mantissa of $O(\log n)$ bits and a signed exponent. $1\pm u\ge1-\mu$ and $1\pm\rho>0$. The normalized difference is Lipschitz in relative product errors; its derivative in any incoming mean is bounded by $1/(1-\mu^2)$, by its half-log-likelihood representation. Summing the partial-derivative bounds over the capped degree gives a variable-update Lipschitz constant $O(\log n)$ in the maximum norm; the factor update has constant at most $k-1$. Clipping to $[-1,1]$ cannot enlarge the error. Over $O(\log\log n)$ rounds the amplification is at most $\exp(O((\log\log n)^2))=n^{o(1)}$, so the accumulated absolute error is $n^{-4+o(1)}$. The sign of an approximate posterior mean then loses at most twice this error in expected correlation on the comparison tree. Products computed once per vertex give all exclusions in linear work per round under the word model of Theorem~\ref{thm:balanced-acquisition}.

Strict integer-score ascent makes at most $m$ flips. Maintain signed fields and a queue of possibly improving vertices: a flip updates at most $kD\log(n+2)$ incident entries and queue items. For the component cap preselect any fixed integer $K\ge2$ with $chK>1$. A certified positive rational lower bound on $ch$ suffices to choose such a cap. Check that each component contains at most $K-1$ vertices before enumeration. On any cap failure return a feasible assignment, with its score, and report that the domain was not established. The capped work bound holds even on these inputs. Passing the caps gives a domain whose guarantee is probabilistic under the stated source law. An independent deterministic certificate of its global validity is a separate task.
\end{proof}

\paragraph{A computable obstruction at finite mean degree.}
The unary reading channel also limits the information gained from the factors. Keeping this effect gives a stronger finite-degree bound. Put $r=k-1$, $b=r^{-1/(r-1)}$, $w=b^r$ and $g(u)=1-e^{-u/L}$. For $b\le t\le kb$ write $s=(kb-t)/r$, $p=s^r$ and $v=ts^{r-1}$. Define
\begin{equation}\label{eq:arity-window}
 H_k(L)=\min_{b\le t\le kb}e^{p/L}\{kb-L[g(p)+r g(v)]\}.
\end{equation}
The local converse of Theorem~\ref{thm:interaction-order} holds whenever $A\le H_k(L)$ eventually, including equality. This bound allows unequal independent marking rates. Its expansion is
\begin{equation}
 H_k(L)=B_k+\frac{B_kw+kw^2/2}{L}+O_k(L^{-2}).
 \label{eq:arity-expansion}
\end{equation}
For triples one more term gives $H_3(L)=3/4+9/(32L)-1/(128L^2)+O(L^{-3})$, where $L=\mu^2d$. Equation~\eqref{eq:arity-window} gives the largest budget for which the upper recursion below preserves the simplex $\sum_i z_i\le kb$ under every relaxed mark allocation. In this range local recovery fails. The finite-degree threshold for recovery remains open.

\begin{proof}
Condition on the squared posterior magnitudes $u,v$ of two independent symmetric channels for a fair bit. Averaging over their relative signs gives combined energy $(u+v-2uv)/(1-uv)\le u+v-uv$. The difference has numerator $uv(1-u)(1-v)$, and a perfect channel gives the endpoint directly. Magnitudes are independent even without conditioning on the common bit, by symmetry. Iteration and Poisson averaging bound the scaled posterior energy by
\[
 z_i'\le a_i+(L-a_i)g(P_i),\qquad P_i=\prod_{j\ne i}z_j.
\]
We prove invariance of $\sum_i z_i\le kb$. Work at $L\ge kb$, as holds eventually. Since $g(u)\le u/L$ and Maclaurin's inequality gives $\sum_iP_i\le kw$ on this simplex, $H_k(L)\ge kb-kw=B_k$. The endpoint $t=kb$ gives $H_k(L)\le kb$. Thus the initial sum $A$ is allowed. The upper map is coordinatewise increasing, so it suffices to examine the boundary. At given $z$, maximize its sum over nonnegative $a_i$ of sum $A$. The coefficient of $a_i$ is $e^{-P_i/L}$. Its maximum is attained where $z_i$ is largest, also when several $P_i$ vanish. Put all $A$ there and call this coordinate $t$. This relaxation gives an upper bound for every admissible allocation.

At fixed $t$ replace any two remaining coordinates by their arithmetic mean. Their two terms $g(P_i)$ have equal weights and increase in sum by concavity. Every other $P_j$ weakly increases because it contains the product of the two coordinates. Its coefficient $L-a_j$ is nonnegative. Repeated averaging therefore maximizes the sum when the remaining coordinates are equal to $s$. The sum at this point is $Ae^{-p/L}+L[g(p)+rg(v)]$. Requiring it to be at most $kb$ for every $t$ gives precisely~\eqref{eq:arity-window}. The two-root argument above then gives vanishing absolute overlap.

For the expansion let $y=1/L$. The minimized function extends analytically to $y=0$, uniformly with its first two $t$-derivatives on the compact interval. Its first terms are
\[
 f_0(t)=kb-p-rv,\qquad f_1(t)=p f_0(t)+(p^2+rv^2)/2.
\]
Here $f_0'(t)=(r-1)(t-s)s^{r-2}$, so $f_0$ has its unique minimum at $b$ and $f_0''(b)=(r-1)(1+1/r)b^{r-2}>0$. Moreover $f_1'(b)=-B_k b^{r-1}<0$. Thus for small positive $y$ the minimum lies in the interior, at $b-yf_1'(b)/f_0''(b)+O_k(y^2)$. This follows by local strict convexity and the implicit function theorem, since compactness excludes the rest of the interval. Evaluating to first order yields~\eqref{eq:arity-expansion}. At $k=3$, $f_0''(b)=3/2$ and $f_1'(b)=-3/8$. The $y^2$-coefficient at the fixed point $t=b$ is $B_kw^2/2+kw^3/3=5/128$. Moving to the minimum subtracts $f_1'(b)^2/(2f_0''(b))=3/64$, giving $-1/128$.
\end{proof}

\paragraph{A finite check of the common domain.}
For $k=3,4,5,6$ fix three vertices per block, take the full grid and set the source signs to $+1$. On the fiber $e_1=2$, negate the first $(3^{k-1}+1)/2$ lexicographic observations. All other labels are positive. Let $D$ contain the source and its flip at vertex $(1,2)$. For every product class $x$ outside these two, enumerate both $y\in D$ and compute
\[
 \max_{y\in D}\{(F(y)-F(x))/2-2\,\mathrm{dist}(x,y)\},
\]
where the distance minimizes Hamming distance over block gauges. Every term whose product changes touches a differing coordinate, so a perturbation of per-vertex budget two changes the half-score difference by at most twice this distance. The minimum of the displayed quantity over outside competitors is respectively $1,7,25,79$ across $126,510,2046,8190$ competitors. These exact positive margins certify the domain simultaneously for all such real perturbations. The single flip uniquely wins the unweighted problem modulo gauge, by two over the source. Adding $2/3^{k-1}$ to each fiber coefficient instead makes the source uniquely optimal. Thus both answers are necessary. The supplement supplies two independent score calculations on each grid and checks the physical-fiber gauge identity on all $299{,}520$ source--tuple pairs with two vertices per block and the same four orders. These are constructed finite checks.

\subsection{Four-way measurements and empirical optima}\label{app:fourway-experiments}
We measured what the returned readings contributed at interaction order four.
Each target has four blocks of $n$ independent fair signs and
$m=\lceil4n\log n\rceil$ distinct, uniformly selected grid entries.
The target and reading biases are $\mu=0.66$ and $\rho=0.76$.
Let $d=m/n$, $L=(\mu^2d)^{1/2}$, and $a_*=2/(3\sqrt3)$.
At nominal multiplier $b$ each coordinate was marked independently
with probability $ba_*/(\rho^2 L)$. We reused the same uniform marks
across budgets by increasing the marking probability on the same random
ranks. Each arm also took 33 noisy readings of the common corner. Their
majority fixed the even-order orientation. The reported multiplier describes
the independent marks alone. Reading totals in the data include the 33
corner readings.

The two arms received the same target, addresses, sensor-error signs,
tie-breaking signs, and iteration budgets. One read the target's generating
source, and the other read an independent source. We ran twelve synchronous
sum--product rounds, then six sweeps of cyclic block-coordinate ascent.
Within each block we set the signs to their strict best responses and
kept ties. We measured sign overlap on unread noncorner vertices before
ascent. Each target also received eight independent target-only starts,
with six ascent sweeps per start. The comparison with this third arm
specifies update and restart counts. Its operations differ from the
message updates in the two acquisition arms.

\begin{table}[ht]
\centering\small
\caption{Matched four-way targets. Each row reuses the same twelve targets
at that size. Overlap is the mean absolute source correlation over unread
noncorner coordinates, first averaged over blocks. The next two columns
give mean feasible score divided by $m$ after ascent. The final column
counts retained answers strictly exceeding both the replacement and
eight-start target-only scores.}
\label{tab:fourway-empirical}
\begin{tabular}{rrccccc}
\toprule
&&\multicolumn{2}{c}{Unread overlap}&\multicolumn{2}{c}{Score/$m$}&\multicolumn{1}{c}{Paired}\\
$n$&$b$&Retained&Replacement&Retained&Replacement&wins\\
\midrule
256 & .75 & .063 & .051 & .357 & .353 & 3/12\\
256 & 1 & .119 & .055 & .360 & .360 & 4/12\\
256 & 1.25 & .480 & .061 & .464 & .363 & 9/12\\
256 & 2 & .999 & .080 & .654 & .346 & 12/12\\
1024 & .75 & .047 & .028 & .323 & .320 & 8/12\\
1024 & 1 & .134 & .031 & .325 & .325 & 6/12\\
1024 & 1.25 & .580 & .034 & .499 & .332 & 11/12\\
1024 & 2 & 1.000 & .031 & .661 & .314 & 12/12\\
\bottomrule
\end{tabular}
\end{table}

At multipliers $.75,1,1.25,2$, respectively, the retained answer attained
or exceeded the generating score on $0,0,10,24$ of the 24 targets. The
replacement answer did so on zero at each budget. At multiplier two the
retained answer strictly improved four generating scores and equaled twenty.
No exact reference optima are available at these two sizes. A verifier
decoded the stored bit strings independently, recomputed 600 target, cold,
and acquisition scores using Boolean parity, and checked the unread
overlaps and reading counts on all 192 arm records.

\paragraph{Exact scores on small targets.}
At $n=5$ and $6$, we applied the same procedure to twelve independent
targets per size. Here the 33 center readings alone exceed the number of factor signs. Fixing the first sign of each of the first three blocks
to one selects one representative per gauge orbit. A Walsh transform
evaluated every score in these coordinates. An independent calculation
enumerated the first three blocks. For fixed first-three-block signs let
$h_j$ be the field at the $j$th coordinate of the fourth block. The best
completion scores $\sum_j|h_j|$ and has $2^{|\{j:h_j=0\}|}$ ties.
The two calculations agreed on the optimum and its orbit multiplicity
on each of the 24 targets. They used 26,738,688 gauge-fixed assignment
scores and 442,368 three-block profiles across the two sizes.
Eighteen generating sources were strictly suboptimal. In order of increasing
multiplier, the retained answers attained the exact optimum on
$2,4,3,6$ of 24 targets. The corresponding counts were $1,3,3,1$
for replacement and $5,5,5,5$ for the eight-start comparator.
Thus the intermediate budget at $b=1.25$ gave a high score relative
to the finite controls on the larger graphs, while both acquisition arms
attained only three of 24 exact optima at the smaller sizes.

\paragraph{Exact checks of the proof.}
An independent rational calculation marginalized hidden child signs on
noisy parity stars of orders three through seven. It checked 400 likelihoods,
8,640 posterior means, and both bounds in \eqref{eq:arity-sandwich} on
1,080 two-factor experiments with fixed child counts and reading patterns.
Integer enumeration checked the elementary symmetric mean inequality and
the product formula for the odd-cut probability on 21,824 vectors each.
A separate calculation tested all 496 signed local contributions to
\eqref{eq:arity-stable} in these five orders.

\paragraph{Process-time comparison.}
On the 24 large targets at multiplier two, we timed one retained solution and target-only multistart on one CPU core after warming each routine. Both times exclude instance generation. The retained clock includes forming the marks, estimating the center sign, belief propagation, polishing and final scoring. The target-only clock includes sign generation, polishing and scoring. Each cold run completes at least eight starts and keeps running until its accumulated process time reaches the retained time on that target, including the full last start. Eight starts exceeded the time budget on every target. For $n=256$, the retained and target-only mean scores divided by $m$ were $.654223$ and $.354875$, with total CPU times $.058070$ and $.104323$ seconds across twelve targets. For $n=1{,}024$, the corresponding means were $.661172$ and $.317871$, with times $.273976$ and $.432135$ seconds. A separate parity calculation replayed the saved winners and the recorded start counts. Actual CPU time and the number of starts to exceed the budget depend on the machine. The physical acquisition time is excluded; each returned bit is charged separately.

\paragraph{Checking the component solver.}
The reproducibility supplement includes a reference implementation for any fixed interaction order. It uses double precision in belief propagation and exact arithmetic for the integer comparisons in the tests. On seven constructed targets of orders three through five it resolved 23 weight queries, including three queries on a coupled two-coordinate component. Full quotient enumeration evaluated 140,800 scores in these comparisons. Separate inputs exercised each of the reading, degree, and component refusals. On twelve sparse even-order targets the two center hypotheses produced different original scores eight times, with each hypothesis winning at least once. In an additional four-way counterexample with zero readings, the solver has no free coordinates and score 112, while its unrestricted maximum is 144. These checks exercise the implementation's restricted optimization and refusal rules. The asymptotic success probability requires the sampling model of Theorem~\ref{thm:interaction-order}.

\section{The residual laws at fixed interaction order}
\label{app:residual-order}
Consider the uniform order-$k$ target of Theorem~\ref{thm:interaction-order}, with fixed $k\ge3$, $q=(1-\mu)/2\in(0,1/2)$, $d=m/n$, and source $T$. Let $D_v=\sum_{e\ni v}\eta_e$ be the source field, $G=\max F-F(T)$ the empirical gain, and $\mathcal O,\mathcal O^H$ the optimal complete product tensors and their restrictions to the observed rows. Define
\[
 \gamma=\sqrt{q(1-q)},\quad h=1-2\gamma,\quad
 \lambda=\tfrac12\log((1-q)/q),\quad a=e^{-\lambda},\quad
 A(c)=(4\pi c\gamma)^{-1/2}.
\]
For $m=\lceil cn\log n\rceil$, put $\alpha=ch$ and $r_n=n^{1-\alpha}/\sqrt{\log n}$. Write $d_{\mathrm{fix}}$ for the fewest free factor coordinates in a single domain containing an optimizer for each allowed objective, and $N_{\mathrm{hit}}$ for the least number of complete product tensors meeting all the optimizer sets. Both minima may depend on the realized target and the allowed family. A positive budget $\Gamma$ permits every real coefficient change with per-vertex absolute load at most $\Gamma$.

\begin{theorem}[Residual laws at order $k$]\label{thm:order-residual}
For $0<\alpha<1$,
\begin{align*}
 G/r_n&\xrightarrow{\Pr}2k A(c)\,a/(1-a)^2,\\
 \log_2|\mathcal O|/r_n,\quad \log_2|\mathcal O^H|/r_n
 &\xrightarrow{\Pr}k A(c),\\
 d_{\mathrm{fix}}(\Gamma)/r_n,\quad
 \log_2 N_{\mathrm{hit}}(\Gamma)/r_n
 &\xrightarrow{\Pr}k A(c)\sum_{j\in\mathbb Z:\,|j|<\Gamma} e^{\lambda j}.
\end{align*}
The last two errors converge to zero uniformly on every fixed bounded budget interval $[0,\bar\Gamma]$. At $\Gamma=0$ the sum is empty and the common dimension and log hitting number are identically zero. If $\alpha>1/2$, the following identities hold with probability $1-o(1)$ simultaneously throughout that interval:
\begin{align}
 G&=2\sum_v(-D_v)_+,\quad
 \log_2|\mathcal O|=\#\{v:D_v=0\},\nonumber\\
 \log_2|\mathcal O^H|&=\#\{v:D_v=0,\ \deg(v)>0\},\nonumber\\
 d_{\mathrm{fix}}(\Gamma)=\log_2N_{\mathrm{hit}}(\Gamma)
 &=\#\{v: |D_v|<\Gamma,\ \deg(v)>0\}.\label{eq:order-singletons}
\end{align}
For $\alpha\ge1$, the source tensor is the unique optimum for every objective under the fixed budget envelope with probability $1-o(1)$.
\end{theorem}

\begin{proof}
We start with the joint fields under the fixed-size mask.
The following estimate also covers $d/\log n\to c>0$. For fixed $s$, any distinct specified vertices $v_1,\ldots,v_s$ and fixed integers $j_1,\ldots,j_s$,
\begin{equation}\label{eq:order-mass}
 \Pr\{D_{v_i}=j_i,\ i\le s\}
 =(1+o(1))\,(4\pi\gamma d)^{-s/2}
 e^{-s h d+\lambda\sum_i j_i}.
\end{equation}
All errors here are uniform over the vertex labels. To prove this, sample iid tuples and signs and condition on all tuple addresses being distinct. Tilt each row by $e^{-\lambda j\eta}$, where $j$ counts its endpoints in the specified set. The one-row normalizer is $Z=1-sh/n+O(n^{-2})$: a given named vertex is hit privately with probability $1/n+O(n^{-2})$ and the chance of a multiple hit is $O(n^{-2})$. Under the tilted law each private signed category has probability $\gamma/n+O(n^{-2})$. Multiple hits occur with probability $O(n^{-2})$. Couple each row to $2s$ independent Poisson counts of mean $\gamma/n$, one for each private signed category. Their probabilities of zero and of each single mark agree with the tilted row up to $O(n^{-2})$, and both laws assign $O(n^{-2})$ mass to all other outcomes. Coupling all $m$ rows therefore costs $O(m/n^2)$ in total variation. The resulting $s$ signed fields are independent differences of Poisson variables with equal means $\gamma d$. At a fixed integer $j$ the mass of one difference is asymptotic to $(4\pi\gamma d)^{-1/2}$ by Fourier inversion of $\exp[-2\gamma d(1-\cos t)]$ and Gaussian rescaling. Conditioning on distinct rows is harmless even at this local probability scale: under the tilted law each address has mass at most $C_s/n^k$, so the probability of a repetition is $O_s(m^2/n^k)=o(d^{-s/2})$. In the original law the conditioning denominator tends to one. Undoing the tilt on the fixed field event contributes $Z^m e^{\lambda\sum_i j_i}$, proving \eqref{eq:order-mass}.

We also need a bound uniform in the negative tail. For one specified vertex the single tilted row has equal $+1$ and $-1$ masses $u=\gamma/[n(1-h/n)]$. Integrating the absolute characteristic function $|1-2u(1-\cos t)|^m$ gives at most $C/\sqrt d$ for the tilted mass at any integer. Dropping the distinctness indicator in the numerator and dividing by its original probability bounds the true mass by $C e^{-hd+\lambda j}/\sqrt d$ for all integers $j$. In particular $\Pr(D_v\le-j)\le C e^{-hd}a^j/\sqrt d$ for $j\ge0$. For fixed density and $0<\alpha<1$, \eqref{eq:order-mass} with $s=1,2$ and the second moment method give $\#\{v:D_v=j\}/r_n\to kA(c)e^{\lambda j}$ in probability. For each single block the coefficient is $A(c)$.

\paragraph{Exceptional interacting vertices.}
Fix a cutoff $L$ large enough that, with probability $1-o(1)$, all perturbed optimizers can be aligned to the source off $W=\{v:D_v\le L\}$. The stability and counting proof of Theorem~\ref{thm:interaction-order} gives this event for the entire fixed envelope without any reading assumption. Its observable domain and an aligned original optimizer also lie in this restriction, in the sense that the domain's free coordinates and the optimizer's errors are contained in $W$. Increase $L$ past $\bar\Gamma$ if needed. The same proof applies for $d/\log n\to c>0$: take $J>2/(ch)$ and $K>1/(ch)$ with strict slack. The one-field bound gives $|W|=o_{\Pr}(n)$ and $W=\varnothing$ with high probability at fixed $\alpha\ge1$. In this regime $h<1$ implies $c>1$, so degree-zero coordinates disappear as well.

Let $B$ be the weak vertices at primal graph distance at most two from another weak vertex. Then
\[
 \mathbb E|B|=O_{k,L}(n(1+d)^2e^{-2hd}).
\]
Indeed, in the iid-row model a one-edge witness for an ordered pair has $O_k(m)$ expected choices before imposing weak fields. Two distinct witness rows with a common intermediate vertex contribute $O_k(m^2/n)$. Conditional on the prescribed endpoint coincidences or full witness addresses, their signed-field tilt factors are bounded by constants; the other rows still have the normalizer $1-2h/n+O(n^{-2})$. Multiplying by $e^{2\lambda L}$ bounds the weak-endpoint contribution. Conditioning the entire witness count on distinct rows costs a bounded factor. Thus $|B|=o_{\Pr}(r_n)$ for $0<\alpha<1$ and $B=\varnothing$ with high probability for $\alpha>1/2$. Vertices of degree zero have expected count at most $kn e^{-d}$ under sampling without replacement. Their contribution is $o(r_n)$ because $c>ch$.

\paragraph{Gain and multiplicity.}
Write $P_W$ for the number of adjacent pairs of weak vertices. The same one-edge estimate gives $P_W=o_{\Pr}(r_n)$. The maximum pair codegree is at most two with high probability. In a source-aligned flip set $S\subset W$, a row meeting $j$ flipped endpoints contributes $-2\eta\one\{j\text{ odd}\}$ to the change in score. Its deviation from $-2j\eta$ has magnitude $2(j-\one\{j\text{ odd}\})$, at most $4\binom{j}{2}$. Charge four units to each flipped pair in a row; each weak pair belongs to at most two rows. Summing and optimizing gives $|G-2\sum_v(-D_v)_+|\le8P_W$, since flipping all negative fields is an admissible choice in $W$. The uniform negative tail bounds the normalized expected contribution from $-D_v>M$ by $C\sum_{j>M}ja^j$. Truncation and the field counts give the gain limit.

The restricted score separates across components of $W$. A singleton supplies two choices exactly at zero field. All nonsingleton components combined involve at most $2P_W$ vertices. Full tensor classes have unique representatives agreeing with the source outside $W$ on the event $|W|<n$: a nonidentity gauge must change an entire block. Hence the full log multiplicity differs from the zero-field count by at most $2P_W$. For observed products, each nonisolated zero singleton has an incident row that distinguishes its two values independently of the other weak vertices. An isolated singleton is invisible. The same bound holds after deleting these isolated zeros. This proves the entropy conclusions and gives the exact identities whenever $B$ is empty.

\paragraph{Strict reversibility and the minimum over all domains.}
Let $S_\Gamma=\{v:\deg(v)>0,\ |D_v|<\Gamma\}$. On the weak domain $W$, fix all isolated coordinates and, among singletons, retain only the nonisolated coordinates in $S_\Gamma$. A singleton with $|D_v|\ge\Gamma$ can be fixed to an original optimal sign, even when a perturbation makes its field zero. Retain all nonsingleton components. This proves $d_{\mathrm{fix}}\le |S_\Gamma|+|B|$ for $\Gamma>0$. The same leading dimension is observable after acquisition: take an original optimum in the common domain, keep its nonsingleton free components, and keep a nonisolated singleton only when its nonnegative empirical field there is less than $\Gamma$. Outside $B$ all neighbors agree with the aligned source, so this singleton field is $|D_v|$. All reversible clean coordinates belong to the original common domain: their fields at the stable reference have magnitude $|D_v|\le\bar\Gamma$ and meet its cutoff. This trims the domain using one field scan.

Conversely each vertex of $S_\Gamma\setminus B$ has a private incident row. The chosen rows form a matching since the selected vertices are at distance greater than two from the other weak vertices. Relative to the source, a change on this row can strictly force either sign: at a nonzero field use zero change or an oppositely directed change of magnitude in $(|D_v|,\Gamma)$; at zero use small changes of either sign. Each imposed change on the signed field is realized by multiplying the desired change by the source product on the row. Matching makes all combinations admissible. Each altered row contains no second weak coordinate. Distinct forced sign patterns give opposite products on a chosen row for all corresponding optimizers. Consequently
\[
 |S_\Gamma|-|B|
 \le\log_2 N_{\mathrm{hit}}(\Gamma)
 \le d_{\mathrm{fix}}(\Gamma)
 \le |S_\Gamma|+|B|.
\]
For a sign-preserving family of the same leading size, discard vertices of degree at most $\bar\Gamma$ (expected count $O(n^{1-c}(1+d)^{\lceil\bar\Gamma\rceil})=o(r_n)$ by the one-vertex degree law). Every other selected vertex has both positive and negative contributions in the original optimal gauge. Amplifying an originally unfavorable term by a magnitude strictly between $|D_v|$ and $\Gamma$ reverses its preference without reversing the coefficient. At a tie use arbitrarily small changes on a private row. The complete real perturbation family supplies the exact singleton identity even when an original coefficient must change sign. Since only finitely many integer field values lie in the fixed budget envelope, all count errors are uniform in $\Gamma$, including approaches to an integer from above. At budget zero one original optimum always suffices. This proves the theorem.
\end{proof}

\begin{corollary}[The last corrections at order $k$]\label{cor:order-critical}
Assume $hd=\log n-\tfrac12\log\log n+t+o(1)$ for fixed real $t$. Let $P_j$ be independent Poisson variables of means $\theta e^{\lambda j}$, $j\in\mathbb Z$, where $\theta=kA(1/h)e^{-t}$. Jointly, for every fixed finite collection of budgets,
\begin{align*}
 G&\ \Rightarrow\ 2\sum_{j\ge1}jP_{-j},\qquad
 \log_2|\mathcal O|,\ \log_2|\mathcal O^H|\ \Rightarrow\ P_0,\\
 d_{\mathrm{fix}}(\Gamma),\ \log_2 N_{\mathrm{hit}}(\Gamma)
 &\ \Rightarrow\ \sum_{|j|<\Gamma}P_j .
\end{align*}
The source is optimal with limiting probability $\exp[-\theta a/(1-a)]$ and uniquely optimal with limiting probability $\exp[-\theta/(1-a)]$.
\end{corollary}
\begin{proof}
At every fixed list of fields, \eqref{eq:order-mass} gives the joint factorial moments of the field counts by summation over distinct labeled vertices. They converge to those of the stated independent Poisson variables, whose joint law is moment determinate. The path estimate gives $\mathbb E|B|=O(\log^3 n/n)$, and $kn e^{-d}=o(1)$ since $1/h>1$. The singleton identities therefore hold. The uniform negative tail makes the probability of a field below $-M$ at most $C a^M$ summed over the vertices, so finite-dimensional convergence passes to the gain. The source is optimal precisely in the absence of a negative field on this event. Uniqueness also excludes zeros.
\end{proof}

\subsection{When the permitted reweighting grows}
\label{app:moving-budget}
A source may be the unique empirical optimum and still require free signs for future objectives. How much more data makes the source optimal throughout a growing range of reweightings? The bounded-budget window does not extrapolate to this setting: on the scale of $\sqrt{\log n}$ the curvature of the signed-field distribution changes the limiting probability.

\begin{proposition}[The moving reweighting window]\label{prop:moving-budget}
Use the fixed-size mask at order $k\ge3$ and fixed $q\in(0,1/2)$. Assume $d/\log n\to c>1/(2h)$ and a deterministic budget $\Gamma_n\to\infty$ with $\Gamma_n=o(\log n)$. The largest integer strictly below $\Gamma_n$ is $b=\lceil\Gamma_n\rceil-1$. Define the tilt correction $\tau=\operatorname{arsinh}(b/(2\gamma d))$, the positive tilt $\beta=\lambda-\tau$ (for sufficiently large $n$), and
\begin{equation}\label{eq:moving-intensity}
 \Lambda_n=\frac{kn\exp\{-d+2\gamma d\cosh\tau+\beta b\}}
 {(1-e^{-\beta})\sqrt{4\pi\gamma d\cosh\tau}}.
\end{equation}
With probability tending to one, $d_{\rm fix}(\Gamma_n)=\log_2N_{\rm hit}(\Gamma_n)=\#\{v:|D_v|\le b,\ \deg(v)>0\}$. If $\Lambda_n\to L\in[0,\infty)$, both dimensions converge in distribution to $\operatorname{Pois}(L)$; if $\Lambda_n\to\infty$, both divided by $\Lambda_n$ converge to one in probability.

In particular, suppose $b/\sqrt{\log n}\to z\in[0,\infty)$ and
\begin{equation}\label{eq:moving-square-window}
 hd=\log n-\tfrac12\log\log n+\lambda b+t+o(1)
\end{equation}
for fixed $t\in\mathbb R$. Then the limiting mean is
\begin{equation}\label{eq:moving-square-mean}
 L=\frac{kA(1/h)}{1-a}\exp\left(-t-\frac{h z^2}{4\gamma}\right).
\end{equation}
Whenever $\Lambda_n\to L\in(0,\infty)$, the unperturbed source is uniquely optimal with probability tending to one, while the probability that it is optimal for \emph{every} allowed reweighting tends to $e^{-L}$. If $\Gamma_n$ is integer for all sufficiently large $n$, the probability of unique source optimality for every reweighting tends to $e^{-L/a}$. If $\Gamma_n$ is eventually noninteger, this probability instead tends to $e^{-L}$.
\end{proposition}

\begin{proof}
First localize all perturbed maximizers. The high-score and stability argument in Appendix~\ref{app:order-law} applies when $d/\log n\to c$: the uniform score deviation still tends to zero, $n\Gamma_n=o(m)$, and a fixed integer $J>2/(ch)$ gives strict slack in the connected-set bound. Thus, simultaneously for all permitted reweightings, every optimum has a representative agreeing with the source outside $W=\{v:D_v\le \Gamma_n+4J\}$. The same path count used in the proof of Theorem~\ref{thm:order-residual}, now with the varying cutoff $B=\Gamma_n+4J$, bounds the expected number of pairs in $W$ at distance at most two by $C n(1+d)^2\exp(-2hd+2\lambda B)=n^{1-2ch+o(1)}=o(1)$. Consequently all such coordinates are isolated from one another by more than two steps. The one-field tail bound gives $\E|W|\le Ckn e^{-hd+\lambda B}/\sqrt d=o(n)$, so with high probability $|W|<n$. Distinct assignments on this restriction represent distinct complete tensors because a nonidentity product-preserving block reversal changes an entire block. For nonisolated coordinates in the mask, the argument for strict reversibility in that theorem now applies to the full budget $\Gamma_n$. A matching of incident rows permits arbitrary independent sign preferences on the reversible coordinates. A field of magnitude exactly $\Gamma_n$ permits a tie but does not require a free sign. The expected number of degree-zero vertices is at most $kn e^{-d}$, which need not vanish for every $c$ in the proposition. We therefore first establish the law for the count $M=\#\{v:|D_v|\le b\}$, and account for degree-zero vertices below.

For fixed distinct vertices $v_1,\ldots,v_s$, tilt the iid-row law by $\exp(-\beta\sum_{i=1}^s D_{v_i})$. Here $\beta\to\lambda$, so the normalizer per row is $1+s(2\gamma\cosh\tau-1)/n+O_s(n^{-2})$, uniformly over the vertices. As in the proof of \eqref{eq:order-mass}, private signed categories under the tilt have probabilities $\gamma e^{\tau}/n+O_s(n^{-2})$ and $\gamma e^{-\tau}/n+O_s(n^{-2})$. Coupling to independent Poisson counts over the $m$ rows costs $O_s(m/n^2)=o(d^{-s/2})$. Conditioning on distinct addresses costs $O_s(m^2/n^k)=o(d^{-s/2})$ under the tilt; its original probability tends to one. The independent tilted field differences have mean $b$ and variance $2\gamma d\cosh\tau$. For each fixed integer offset $\ell$, Fourier inversion at $b-\ell$ gives mass asymptotic to $(4\pi\gamma d\cosh\tau)^{-1/2}$. Indeed the centered characteristic exponent, after rescaling by the standard deviation, tends to $-u^2/2$; the absolute characteristic function is $\exp[-2\gamma d\cosh\tau(1-\cos u)]$. The latter also bounds every lattice mass by $C/\sqrt d$. Both estimates hold for the joint tilted fields, including the coupling error, with bounds of order $d^{-s/2}$.

Undoing the tilt on $D_{v_i}=b-\ell_i$ yields, for fixed nonnegative offsets,
\[
 \Pr(D_{v_i}=b-\ell_i,\ i\le s)
 =(1+o(1))\left(\frac{\Lambda_n(1-e^{-\beta})}{kn}\right)^s
 e^{-\beta\sum_i\ell_i}.
\]
Uniformly in all offsets the probability is bounded by a fixed multiple of the right-hand expression without $1+o(1)$. Summing over $0\le\ell_i\le2b$ gives $(1+o(1))(\Lambda_n/(kn))^s$, since $b\to\infty$ and $\beta\to\lambda>0$. Summing over all distinct ordered vertices gives $\E(M)_s=(1+o(1))\Lambda_n^s$. This proves the Poisson law for positive finite $L$ by factorial moments, vanishing for $L=0$ by Markov's inequality, and concentration when $\Lambda_n$ diverges by the first two moments.

The actual minimum dimension and log hitting number both equal $M$ minus the number of degree-zero vertices on the separated localization event. In the finite-intensity case, \eqref{eq:moving-intensity} implies $c=1/h$ if $L>0$, and $c\ge1/h$ if $L=0$ (otherwise $\Lambda_n=n^{1-ch+o(1)}$ diverges). Hence $kn e^{-d}=o(1)$. In the diverging case the expected degree-zero count is $o(\Lambda_n)$ since $h<1$. In the window $c=1/h$ the unrestricted equality therefore holds with high probability. These observations yield all dimension limits.

Expanding the exponent in \eqref{eq:moving-intensity} gives $-hd+\lambda b-b^2/(4\gamma d)+O(b^4/d^3)$. Under \eqref{eq:moving-square-window}, $b=O(\sqrt{\log n})$, $d/\log n\to1/h$, so \eqref{eq:moving-square-mean} follows. If $\Lambda_n\to L>0$, the same uniform tilted mass bound gives $\Pr(\exists v:D_v\le0)=O(\Lambda_n e^{-\beta b})=o(1)$. As shown above, finite intensity also implies absence of degree-zero vertices with high probability. The isolated reduction then proves uniqueness of the unperturbed complete tensor. The source is an optimum for all reweightings exactly when no strictly reversible vertex occurs. Unique optimality for all reweightings instead requires every field to exceed $\Gamma_n$: equality at an integer budget can be attained by changing one incident coefficient. For an integer budget repeat the tail count with upper integer $b+1$, or include the offset $\ell=-1$ in the same geometric sum. Its mean tends to $e^{\lambda}L=L/a$, giving the uniqueness formula. Noninteger budgets give the same strict and inclusive integer cutoffs.
\end{proof}

\paragraph{Observation cost of robustness.}
In the square-root window~\eqref{eq:moving-square-window}, the leading displacement from the zero-field critical mean degree is $\lambda b/h$. Since $m=nd$, this corresponds to $(\lambda/h)nb$ additional original observations. The quadratic correction in~\eqref{eq:moving-square-mean} changes the limiting probability of robustness, so the fixed-budget law cannot simply be extrapolated to growing budgets.

\paragraph{A finite field diagnostic.}
We use $k=3$, $q=.17$, $b=\lfloor 2\sqrt{\log n}\rfloor$, and the integer mask size obtained by solving \eqref{eq:moving-intensity} for $\Lambda_n=2$ and rounding $m=nd$. At each of $n=512,2048,8192$, count fields on 96 independently drawn uniform masks. The mean counts are $1.646,1.917,1.969$ (respectively $19,13,18$ of 96 have zero count). Exact finite-mask marginal expectations, obtained by summing the binomial sign law over the hypergeometric degree law, are $1.882,1.905,1.920$. Dropping the moving-field correction and extrapolating the fixed-field mass instead gives $2.399,2.610,2.837$. These 288 masks check the field distribution, not global optimality or the separation event. The supplement gives the seeded code and every count, 36 exact rational checks of the finite marginal, and 15 comparisons of the saddlepoint mass to the independent Poisson-difference mass evaluated through a scaled Bessel function.

\paragraph{The observable cutoff and sparse updates.}
For a bounded envelope let $y$ be the high-scoring one-flip-stable reference of Theorem~\ref{thm:interaction-order}. Align it and any perturbed optimum to the source. Their errors lie in a fixed $W_L$, whose components have size less than any fixed integer $J\ge2$ with $J\alpha>1$ on an event of probability $1-o(1)$. Each connected component of their difference has at most $J-1$ vertices. At a differing vertex $v$, the other at most $J-2$ vertices contribute at most $4(J-2)$ to the comparison, by codegree at most two. The perturbed optimum is stable against flipping $v$. Since the unweighted field at $y$ is an integer, the observable restriction can therefore use
\[
 U=\{v:d_v(y)\le4(J-2)+\lfloor\bar\Gamma\rfloor\}.
\]
It contains a representative of every required optimum. Outside $W_L$, $y$ agrees with the aligned source and has fewer than $J$ neighbors in $W_L$ with high probability by the weak-neighbor bound. Thus $U$ is contained in another fixed-cutoff weak set and has the same component size cap. Given the original local score tables and their maxima, an $\ell$-row edit supplied as a list changes at most $\ell$ component tables, plus a scalar contribution for rows with no free vertex. Updating all entries of those tables and their cached maxima costs $O_{k}(\ell\,2^J)$ arithmetic operations. Reading or printing a full assignment has its own output cost.

\subsection{What a finite list of objectives can force}
\label{app:order-query-list}
For a nonempty finite family $\mathcal Q$ of objective changes, write $N_{\mathrm{hit}}(\mathcal Q)$ and $d_{\mathrm{fix}}(\mathcal Q)$ for the minima when only those queries will be asked. The domain and the list of candidate answers may use the whole query list in advance.

\begin{corollary}[Two queries suffice for coordinate dimension]\label{cor:order-two-queries}
At any fixed $\Gamma>0$ and $0<\alpha<1$, one can select two admissible queries on the realized target for which, with probability $1-o(1)$, $N_{\mathrm{hit}}(\mathcal Q)=2$ and
\[
 \frac{d_{\mathrm{fix}}(\mathcal Q)}{r_n}
 \xrightarrow{\Pr} k A(c)\sum_{|j|<\Gamma}e^{\lambda j}.
\]
For $\alpha>1/2$ their common dimension equals the full-family dimension in \eqref{eq:order-singletons} with high probability, allowing one answer when that dimension is zero.
\end{corollary}
\begin{proof}
Take the two opposite patterns on all $R=|S_\Gamma\setminus B|$ separated reversible vertices in the matching construction. If $R>0$ their optimizer sets are disjoint and two answers suffice. Any two canonical representatives of the respective optima differ at all $R$ vertices. They both agree with the source outside $W$. Representatives in different whole-block gauges differ at least $n-2|W|$ times, which exceeds $|W|$ with high probability. Hence any coordinate cube hitting both objectives has at least $R$ free coordinates even after optimizing over gauges. The upper bound for the full query family finishes the proof. The selections depend on the target; two arbitrary queries supplied in advance need not have this property.
\end{proof}

\begin{proposition}[Answer sets for finite query lists]\label{prop:order-query-graph}
Suppose $|W|<n/3$, every optimum of every query has a representative equal to the source off $W$, and no observed row meets two vertices of $W$. For each query and each vertex in $W$, record its strictly preferred sign in this representative, using a wildcard at a tie. Form a graph on the query list: a pair is adjacent if its strict signs disagree at some vertex. Then
\[
 N_{\mathrm{hit}}(\mathcal Q)=\chi(\mathcal G),\qquad
 d_{\mathrm{fix}}(\mathcal Q)
 =\#\{v:\text{both strict signs occur across the list at }v\}.
\]
On the fixed-density target with $0<\alpha<1$, for every fixed $\Gamma>0$, with high probability there are $(kA(c)+o_{\Pr}(1))r_n$ separated, nonisolated zero-field vertices available for query construction. On that event, every finite simple graph with at most that many edges and at least one vertex can be realized by an indexed list of admissible queries. The resulting hitting number is its chromatic number and the common coordinate dimension is its edge count. Identical queries for isolated graph vertices may be repeated without changing either minimum.
\end{proposition}
\begin{proof}
On the displayed hypotheses the canonical optimizer set of a query is a Boolean subcube: each wildcard is free and each strict sign is fixed. Complete products have unique canonical representatives. A group of queries has a common answer exactly when the partial sign assignments are consistent, which holds if and only if no pair conflicts. Partitioning the query list into groups covered by individual answers is therefore graph coloring. This is the Helly property of Boolean subcubes, familiar in the study of subcube intersection graphs \citep{johnsonmarkstrom2013}. The distinct step here is to obtain these subcubes as exact optimizer sets of a common noisy target.

At any coordinate without conflicting strict signs fix one sign compatible with every query. This gives the upper bound on dimension. For the lower bound take any successful cube. A cube of dimension at least $|W|$ already satisfies the lower bound. Otherwise all its chosen witnesses must use the same gauge relative to their canonical representatives, since witnesses using different gauges differ at least $n-2|W|>|W|$ times. Every conflicting coordinate must then be free.

For realization let each graph edge use one distinct nonisolated zero-field vertex at distance greater than two from every other weak vertex. Their available number follows from \eqref{eq:order-mass} and the bound on $B$. Orient each graph edge arbitrarily. Choose private incident rows as in the matching proof. In the query for a graph vertex, change the field of each incident graph edge's coordinate by $+\varepsilon$ for its head and $-\varepsilon$ for its tail, leaving all other selected fields zero, for a rational $0<\varepsilon<\min(1,\Gamma)$. The private rows form a matching, so even a high-degree query obeys every local budget. All changes preserve coefficient signs. The selected singletons share no row with any other weak coordinate. The restricted objective therefore separates additively, forcing their prescribed strict signs in every optimizer in the aligned gauge, regardless of ties in the remaining components. The other weak components have identical objective tables in all queries and can be fixed to one common maximizing pattern. This also proves realizability when $\alpha\le1/2$ and those other components can have several interacting vertices. On the selected coordinates, two queries conflict exactly on their mutual graph edge. The coloring and gauge arguments above give both equalities, including edgeless graphs. This is a worst-case construction of queries for a realized typical target; it gives no hardness claim for optimizing one typical query.
\end{proof}

\paragraph{Finite checks.}
On three free bits, direct searches for the smallest hitting set and the least-dimensional cube agreed with Proposition~\ref{prop:order-query-graph} on all $3{,}303$ nonempty lists of up to three distinct partial assignments. The graph construction gave the correct hitting number on all $1{,}099$ labeled simple graphs of orders one through five, checked by independent coloring and set-cover searches. An order-four target with four vertices per block has two distinguished fields, $0$ and $-2$. Enumeration of its $8{,}192$ product classes certifies the corresponding four-answer cube for every real perturbation of per-vertex budget $5/2$: each of the $8{,}188$ outside competitors has a cube representative whose original score lead, after subtracting five times their minimum gauge Hamming distance, is at least $39$. The proof of the reset inequality is the same as in Appendix~\ref{app:interaction-order}. At budget $2$, two opposite queries require just one free coordinate because the second field can only tie. At budget $5/2$, the same directions require both coordinates, yet two answers cover the pair of queries and four are necessary for the full family. All its forcing queries and all gauge representatives of the two-query comparisons are included in the exact check.

\section{Verification radius at fixed interaction order}
\label{app:order-radius}
A further reading can change the computation needed to find an answer. Once an answer is given, an observer can check its score and search for improving flips without any readings. The number of coordinates needed in that search has a different dependence on interaction order. Write $d=m/n$, $\gamma=\sqrt{q(1-q)}$, $h=1-2\gamma$, $\lambda=\frac12\log((1-q)/q)$, and $\alpha=ch$. The fields $D_v=\sum_{e\ni v}\eta_e$ are expressed at the source. In this section we consider the original, unweighted objective. A radius-$R$ stable assignment admits no strictly improving flip of at most $R$ coordinates.

\begin{theorem}[Sharp radius at fixed order]\label{thm:order-radius}
For fixed $k\ge3$, $m=\lceil cn\log n\rceil$, and fixed $0<q<1/2$, put
\[
 R_*=\max\{1,\lfloor1/\alpha\rfloor\}.
\]
There is $\epsilon_0>0$ such that, with probability $1-o(1)$, every radius-$R_*$ stable assignment with $F(x)\ge(\mu-\epsilon_0)m$ attains the global maximum. For any positive fixed $R<R_*$ and any $\epsilon>0$, there exists a radius-$R$ stable assignment scoring at least $(\mu-\epsilon)m$ and exactly two units below the maximum, with probability $1-o(1)$ except at $R=1,\alpha=1/2$, where its probability is bounded away from zero. Hence $R_*$ is the least positive fixed radius giving the stated uniform implication with high probability.
\end{theorem}

\begin{proof}[Proof of sufficiency]
The high-score argument in Appendix~\ref{app:interaction-order} uses the distance only to put a stable support in a ball of radius $\delta n$ for some fixed small $\delta>0$. A fixed score screen does so as well. Indeed if the relative block means $w_i$ have $\prod_iw_i>1-2\delta/k$, they have an even number of negative signs and $|w_i|>1-2\delta/k$ in every block. After a block gauge the total distance to the source is then less than $\delta n$. The uniform normalized score deviation from $\mu\prod_iw_i$ is $o(1)$ with high probability by Hoeffding and the union bound there. Take $\epsilon_0=\mu\delta/k$. Every passing assignment has such a representative, and the global optima pass with high probability because the source does. The connected stable-support estimate and the codegree bound in that appendix give a fixed $L_0$ for which every passing one-flip-stable representative is supported, relative to the source, in $W_{L_0}=\{v:D_v\le L_0\}$. For any fixed $J\ge2$ with $J\alpha>1$, all components of $W_{L_0}$ have fewer than $J$ vertices with high probability. Set $J=R_*+1$. The difference between aligned representatives of a passing stable assignment and any optimum lies within these components. Gains add over components, so any strict total gain has a strict summand on at most $R_*$ coordinates. Block gauge changes preserve the gain of a flip at any specified coordinates.
\end{proof}

\paragraph{Paths carrying an undetectable improvement.}
Use just the first two blocks for the path vertices. An oriented path of $t\ge2$ distinct vertices begins in the first block and alternates between them. Each consecutive pair must occur in exactly one observed row, with positive source-relative sign. No other observed row may meet two path vertices. Require fields $0$ at the first vertex, $1$ at the last, and $2$ at every internal vertex. We call these paths eligible. For $L\ge2$ a path is isolated at $L$ if it constitutes a component of $W_L$ in the full $k$-way observation graph. The numerical constant below counts the indicated orientation and block pair.

\begin{theorem}[Window for isolated path obstructions]\label{thm:order-path-window}
Fix $k\ge3$, $t\ge2$, $u\in\mathbb R$, and a constant $L\ge2$. If
\[
 t h d=\log n+(t/2-1)\log\log n+u+o(1),
\]
the number of eligible paths isolated at $L$ converges in law to a Poisson variable of mean
\[
 \Lambda_t e^{-u},\qquad
 \Lambda_t=(1-q)^{t-1}(4\pi\gamma)^{-t/2}
                  e^{-\lambda}(th)^{1-t/2}.
\]
Consequently, for every $\epsilon>0$ the limiting inferior probability that some radius-$(t-1)$ stable assignment scores at least $(\mu-\epsilon)m$ and exactly two below the optimum is at least $1-\exp(-\Lambda_t e^{-u})$. The Poisson law counts this specified family of paths. Other configurations can also cause verification failure.
\end{theorem}

\begin{proof}
Generate independent uniform $k$-tuples and independent noise signs. The probability of a repeated full address is $O(m^2/n^k)=o(1)$, so probability and distribution conclusions transfer to the uniform simple mask. Every eligible path, before isolation, depends only on the first two indices of each row and the noise sign. The other indices can change the path's neighbors.

Fix a bounded number $v$ of named vertices, in any blocks. For one independent row let $j$ be its number of named endpoints. The tilt by $e^{-\lambda j\eta}$ has normalizer $1-vh/n+O(n^{-2})$. With a bounded number $f$ of row indices forced, omit them from the tilt and condition on their endpoint contributions $a_i$. A private increment of either sign at a specified vertex has tilted probability $\gamma/n+O(n^{-2})$. The total probability of any row with two named endpoints is $O(d/n)$ over all unforced rows. The counts of the $2v$ private signed increments differ in total variation by $O(d/n)$ from independent Poisson variables of means $\gamma d$: one may couple a row's rare category indicators to independent Poisson counts at cost $O(n^{-2})$ per row, then change each total mean by $O(d/n)$. This additive error is $o(d^{-v/2})$. For two independent Poisson counts of mean $\gamma d$ the mass of their difference at any fixed integer is $(1+o(1))(4\pi\gamma d)^{-1/2}$, by Fourier inversion with characteristic function $\exp[-2\gamma d(1-\cos s)]$. Undoing the tilt, the joint mass at named fields $b_i$, also requiring no extra shared row, is
\begin{equation}\label{eq:order-path-fields}
 (1+o(1))e^{-vhd+\lambda\sum_i(b_i-a_i)}(4\pi\gamma d)^{-v/2}.
\end{equation}
Omitting the no-extra-shared-row requirement gives the same leading mass by the same tilted bound. For upper bounds, requiring every field to be at most fixed $L$ costs $O(e^{-vhd})$ conditional on the bounded forced records, uniformly over their labels and signs.

Let $X$ count eligible paths. There are $(1+o(1))n^t$ ordered choices of core vertices and $(m)_{t-1}$ ordered distinct row indices. The probability for each forced row to hit the designated first-two-block pair with positive sign is $(1-q)/n^2$. Such a row cannot hit another core vertex. Uniqueness of the specified observations means each realized path is counted once. In \eqref{eq:order-path-fields}, $\sum_i(b_i-a_i)=-1$. Thus in the iid model at every $d\asymp\log n$,
\begin{equation}\label{eq:order-path-mean}
 \mathbb E X\sim n (1-q)^{t-1} d^{t-1}
         e^{-thd-\lambda}(4\pi\gamma d)^{-t/2}.
\end{equation}
The stated centering gives the mean in the theorem.

For a fixed factorial moment, tuples of core-disjoint paths have disjoint forced rows, since a row has just one index in each of these two blocks. Equation~\eqref{eq:order-path-fields} on the union shows that their contribution factors to leading order. For the upper estimate one may drop the prohibition of extra rows across the union. For the lower estimate one may forbid all of them. The two leading masses agree. To bound overlapping tuples, group paths into connected clusters under core intersection. A cluster using $V$ core vertices and $E$ distinct required rows contributes at most
\[
 O\bigl(n^V(d/n)^E e^{-Vhd}\bigr)
\]
for each of the finitely many identification patterns. This follows by choosing the vertices and row indices, forcing their designated pairs, and applying the joint weak-field bound; it applies multiplicatively in the counts of $V,E$ over clusters. In a consistent cluster, the union of the required core edges is connected, hence $E\ge V-1$. If the cluster contains two distinct eligible paths then $V>t$: on a fixed set of $t$ vertices the entire induced edge record is a single positive path, and the fields $0,1$ determine its orientation. In the window such a cluster costs at most $n^{1-V/t+o(1)}=o(1)$. Singleton clusters cost at most $n^{o(1)}$ in this estimate. All factorial moments now converge to those of the asserted Poisson law.

A failure of isolation requires a further weak neighbor. Given one path pattern, such a neighbor either lies on a forced row or joins it through a new row. For the first case the choice of one further endpoint and the additional incidence probability contribute $O(1)$. For the second they contribute $O(d)$, from $O(nm)$ choices of a vertex and a row and probability $O(n^{-2})$ of hitting it and a named path vertex. In both cases tilt over the $t+1$ distinct fields. Compared with \eqref{eq:order-path-mean} the expected nonisolated count is at most an $O((1+d)d^{t/2}e^{-hd})$ fraction of the eligible mean. This fraction vanishes. In particular the two counts coincide with probability $1-o(1)$ in the window.

It remains to pass from an isolated path to the global objective. The stable-support and uniform weak-neighbor bounds used above hold also when $d/\log n$ tends to a positive constant, by taking a strictly smaller constant in their exponent bounds. Choose $L_0\ge2$ so a global maximizer aligns to the source outside $W_{L_0}$. With high probability every vertex has at most a fixed $K$ distinct neighbors in $W_{L_0}$ and the maximum pair codegree is at most two. Let $L^\prime=\max\{L,L_0,4K+4(t-2),2\}$, where $L$ is the fixed cutoff in the statement. In the window $\E X=O(1)$, and the preceding estimate at $L^\prime$ gives $o(1)$ expected eligible paths that fail isolation there. With probability $1-o(1)$ every eligible path is therefore isolated at both cutoffs. We use $L^\prime$ in the construction. The exact source-relative half-score cost of flipping path subset $B$ equals
\[
 \sum_{v\in B}D_v-2\sum_{i=1}^{t-1}
       \mathbf1\{v_i,v_{i+1}\in B\}.
\]
Indeed, each row touches at most two selected vertices, with double hits exactly the positive path rows. A proper interval costs zero, one or two according as it contains the first endpoint, contains the last, or contains neither. Costs add over disjoint maximal intervals. The full path costs $-1$. Since the path is isolated from all other possible flipped vertices of the maximizer, its unique optimal choice flips the entire path. Undo those flips at the maximizer to obtain $x$ with gap two.

Suppose a strict improvement at $x$ has at most $R=t-1$ vertices and take an inclusion-minimal such set $S$. At any of its vertices $v$, deleting $v$ loses the improvement, so $d_v(x)<2\sum_{w\in S\setminus\{v\}}\mathrm{codeg}(v,w)\le4(R-1)$. This follows by writing the added half-cost as the sum of the signed incident terms after the other flips. An incident term changes sign only if it contains another selected vertex. Outside $W_{L^\prime}$ the current field exceeds $L^\prime-4K$, since its sign agrees with the source and at most $K$ weak neighbors of codegree at most two can change incident terms. Hence $S\subset W_{L^\prime}$. Gains split across its weak components. Away from the path they are nonpositive by global optimality of the original configuration. On the path its proper subsets cannot improve. This contradiction gives the claimed stability. Finally $F(x)\ge F(X_{\rm source})-2$ and $F(X_{\rm source})/m\to\mu$. The high-probability structural events may be intersected with the Poisson occurrence event without assuming independence.
\end{proof}

\paragraph{Necessity in Theorem~\ref{thm:order-radius}.}
At fixed $c$ the eligible expectation \eqref{eq:order-path-mean} is asymptotic to a positive constant times $n^{1-t\alpha}(\log n)^{t/2-1}$. For $t\alpha<1$, or $t\alpha=1$ and $t\ge3$, this diverges. The core-disjoint pairs contribute $(1+o(1))(\mathbb EX)^2$. For two distinct overlapping eligible paths with $r\ge1$ shared core vertices and $f$ shared forced rows, their common core edges form a forest, so $f\le r-1$. If $r=t$ their entire vertex sets coincide, which precludes distinct eligible paths. The overlap contribution divided by $(\mathbb EX)^2$ is at most $n^{-r+f+\alpha r+o(1)}=o(1)$ for $1\le r<t$ when $t\alpha\le1$. The diagonal contributes $1/\mathbb EX$. Concentration, together with the preceding estimate for nonisolated paths, gives an isolated path with high probability. At $t=2$, $\alpha=1/2$, the path-window theorem applies with $u=0$ and gives an occurrence probability tending to $1-e^{-1/(4\pi)}$ for the specified orientation, since $(1-q)e^{-\lambda}=\gamma$ and $\Lambda_2=1/(4\pi)$ at every fixed noise level. This also lower bounds the limiting inferior failure probability. This completes the lower bounds.

\paragraph{Finite window and radius checks.}
For $k=4,6$, $n=512,2048$, and $t=2,3,4$, we generated 32 independent uniform simple targets in each cell, with $q=.17$ and $u=\log\Lambda_t$. The limiting eligible-path mean is then one per target. At $n=2048$ the eligible totals over 32 targets were respectively $(36,26,22)$ for order four and $(29,31,19)$ for order six, in increasing $t$. Across all 384 targets we counted 301 eligible paths. Only one was isolated even at cutoff $L=2$, which is smaller than the cutoff used to deduce a global score gap. These data measure the pre-isolation counts; they do not establish the limiting frequency of failed verification. Independent direct enumeration on 40 additional small masks gave the same counts in 120 order-and-length checks.

We also constructed actual finite assignments whose first improving radius can be verified exactly. Take $t$ anchor vertices per block and all $t^k$ positive anchor-only rows. Add the $t-1$ positive rows along an alternating path of fresh vertices, filling each unused index with the first anchor of its block. Add one negative row touching only the first path vertex and otherwise using first anchors. The all-positive assignment scores $t^k+t-2$. If an anchor block has mixed signs, the anchor score falls by at least $2t^{k-1}>2t$, beyond the possible gain on the $t$ other rows. If all anchor signs are blockwise constant, a negative product has the same exclusion. The remaining representatives fix all anchors positive. The path subset cost gives a unique improving product class on the active coordinates, at score $t^k+t$. Its closest representative flips $t$ vertices, since a nontrivial block gauge flips all $t$ anchors in at least one block. Padding the blocks with unused coordinates preserves the minimum improving distance and the gap. Different settings of the unused signs can give different complete tensors at the same score. Exhaustive Walsh enumeration agreed on the optimum and minimum improving distance for 11 choices covering orders three to six and path lengths two to five, with 497,600 active gauge-class scores in total. These constructions establish finite instances of the obstruction separately from its random occurrence probability.

\section{Sensitivity to a single observation}
\label{app:edge-response-order}

A coefficient change can be followed through the \emph{original} empirical
optimum. For an observed address $e\in H$, define
\[
 M_e^s=\max_{x:T_e(x)=s}F_Y(x)\quad(s=\pm1),\qquad
 b_e=\tfrac12|M_e^+-M_e^-|.
\]
Under the distribution proportional to $e^{F_Y(x)}$, the two constrained scores
are log max-marginals for the binary product variable, up to a shared
additive constant (see the definition in \citealp[Eq.~(5)]{wainwright2004tree}).
Both constraints are feasible, since changing a single factor sign in
$e$ reverses its product. When only $Y_e$ changes, by $t$, the best score
is $\max(M_e^++t,M_e^--t)$. Thus $b_e$ is the distance from zero to
the coefficient change at which the two product classes tie. For any
$\Gamma>0$, each sign can be made strictly better than the other by some
$|t|\le\Gamma$ exactly when $b_e<\Gamma$. The identity concerns the
prediction at one observed address. Different addresses may require
different changes, and the maximizing tensors within either class may tie.

\begin{proposition}[Single-observation margins]
\label{prop:edge-response-order}
Fix $k\ge3$, $c>0$, $0<q<1/2$ and $0\le B<\infty$.\par\smallskip
\noindent\emph{(i) Individual observations.} If
$\alpha>1/2$, with probability $1-o(1)$, simultaneously for all
$e\in H$,
\[
 \min(b_e,B)=\min\bigl(B,\min_{v\in e}|D_v|\bigr).
\]
For any original optimizer $w$, define its observable one-flip field
$H_v(w)=\sum_{f\ni v}Y_fT_f(w)$. On the same event,
simultaneously for all such $w$ and all $e\in H$,
\[
 \min(b_e,B)=\min\bigl(B,\min_{v\in e}H_v(w)\bigr).
\]
After finding one optimizer, all clipped global margins can therefore
be computed in $O(km+kn)$ operations by accumulating the fields and
scanning the addresses. Equality at $B$ identifies a lower bound on the
unrestricted margin.\par\smallskip
\noindent\emph{(ii) Number of sensitive observations.} For every $0<\alpha<1$, let $S_\Gamma=\#\{e\in H:b_e<\Gamma\}$ and
$s=\sqrt{q(1-q)}$. For each $0<\Gamma\le B$,
\begin{equation}\label{eq:sensitive-observation-count}
 \frac{S_\Gamma}{r_n\log n}
 \xrightarrow{\Pr}
 \frac{2cs\,k}{\sqrt{4\pi c s}}
       \sum_{j\in\mathbb Z:\,|j|<\Gamma} e^{\lambda j}.
\end{equation}
The convergence holds jointly over this bounded budget interval, which
contains only finitely many distinct strict integer cutoffs.
In the critical window of Corollary~\ref{cor:order-critical}, jointly
for these same budgets, the limit is instead
\[
 S_\Gamma/\log n\ \Rightarrow\ (2s/h)\sum_{|j|<\Gamma}P_j ,
\]
with the Poisson field counts of that corollary.
\end{proposition}

\begin{proof}
Fix $B>0$, the case $B=0$ being immediate. Use the common localization
event for budget $B$ and the weak-set estimates from the proof of
Theorem~\ref{thm:order-residual}. First assume $\alpha>1/2$. In source-relative signs, all these optimization problems
attain their maxima on a single cube fixing the complement of
$W=\{v:D_v\le L\}$ to $+1$ for a fixed $L>B$. Enlarging the
localization cutoff to exceed $B$ preserves the event and isolation
holds for the enlarged fixed cutoff. Every edge then contains at most
one vertex of $W$. On the cube the objective is a constant plus
$\sum_{v\in W}D_v x_v$. If $e$ misses $W$, changing its coefficient
adds the linear function $tT_e$ to the optimum. If $e$ meets $W$ at
$v$, the optimum as a function of $t$ is a constant plus
$|D_v+tT_e|$, where $T_e$ here is the source product. Throughout
$[-B,B]$ these functions equal the global envelope of the two
constrained scores. An interior change of slope therefore occurs
exactly at the indicated distance $|D_v|$ from zero. If there is no
interior change of slope, $b_e\ge B$. Every endpoint outside $W$ has
$D_v>L>B$, proving the clipped formula, including its endpoints.

On any instance, flipping just $v\in e$ at a global optimizer $w$
changes the observed product at $e$ and loses $2H_v(w)$.
Therefore $b_e\le\min_{v\in e}H_v(w)$. At an address with $b_e<B$
the preceding argument gives a weak singleton $v$ with $|D_v|=b_e$.
Its neighbors outside $W$ agree with the source in an aligned
representative of $w$, so $H_v(w)=|D_v|$ by optimality on that
singleton. The fields $H_v(w)$ are invariant under block gauge
changes. This proves the observable equality when $b_e<B$.
When $b_e\ge B$ the deterministic inequality already gives it.

On the same event each sensitive observed address belongs to exactly
one vertex with $|D_v|<\Gamma$. Consequently
\[
 S_\Gamma=\sum_{v}\deg_H(v)\,\mathbf 1_{\{|D_v|<\Gamma\}}.
\]
For the full range $0<\alpha<1$, let $I$ denote all vertices in
nonsingleton components of $W$. Theorem~\ref{thm:order-residual}
gives $|I|=o_{\Pr}(r_n)$. The maximum degree is $O(\log n)$ with
high probability by the factorial-moment bound below and a union
bound. There are therefore $o_{\Pr}(r_n\log n)$ rows incident to
$I$. Every remaining row either misses $W$ or meets it in a single
singleton, whose contribution to the restricted optimum separates
from every other component. The same envelope argument proves the
clipped formula on every such row. Each omitted row can change the
difference between the sensitivity count and the degree sum by at most
$k$. Thus the degree-sum identity has additive error
$o_{\Pr}(r_n\log n)$ uniformly on the stated budget interval.

For completeness, condition on one fixed signed field $D_v=j$.
Write $d_v=\deg_H(v)$ and $\ell=m/n$. Uniformly for
$d_v\le C\log n$, the one-star fixed-mask degree probabilities differ
by a multiplicative $1+o(1)$ from those of a Poisson variable of mean
$\ell$. Indeed the degree is hypergeometric with population $n^k$,
star size $n^{k-1}$ and draw size $m$. The logarithmic errors from the
falling factorials and binomial approximation are
$O((\log n)^2/n^{k-1}+m^2/n^k+(\log n)^2/m+
 (\log n)/n+m/n^2)=o(1)$. Conditional on the degree the sign law is
exactly binomial. The factorial-moment tail
$\Pr(d_v\ge t)\le(e\ell/t)^t$ makes
$\E[d_v\mathbf 1_{\{d_v>C\log n\}}]$ smaller than any prescribed
inverse power of $n$ by choosing $C$ large.

In the Poisson model, tilting by $e^{-\lambda D_v}$ leaves the
conditional law at $D_v=j$ unchanged. Under the tilt the two sign
counts are independent Poisson variables of common mean $s\ell$,
so their sum has mean $2s\ell$ and their difference has point mass
$\Theta(\ell^{-1/2})$ at $j$. Exponential concentration of the sum
then gives, for fixed $j$,
\[
 \E\!\left[\left.|d_v-2s\ell|\,\right|\,D_v=j\right]=o(\ell).
\]
To see this explicitly, restrict the sum to $C\log n$ and bound its
relative deviation by $\varepsilon$ plus $O(1)$ times the probability
of a deviation exceeding $\varepsilon\ell$. The latter under the
conditional tilted law is at most $O(\sqrt\ell)e^{-\Omega_\varepsilon(\ell)}$.
Take $\varepsilon\downarrow0$ after $n\to\infty$. The multiplicative
comparison and degree tail transfer the estimate to the fixed-size mask
because $\Pr(D_v=j)=\Theta(n^{-\alpha}/\sqrt{\log n})$.
Summed over vertices with $D_v=j$, the error in replacing $d_v$ by
$2s\ell$ is $o_{\Pr}(r_n\log n)$ by Markov's inequality.
The per-block count of the field $j$, divided by $r_n$, tends to
$e^{\lambda j}/\sqrt{4\pi cs}$. Summing over the $k$ blocks and the
finitely many allowed fields proves~\eqref{eq:sensitive-observation-count}.
In the critical window the same hypergeometric comparison and tilting
argument apply with $\ell/\log n\to 1/h$ and
$\Pr(D_v=j)=\Theta(1/n)$ for fixed $j$. The expected
degree-replacement error summed over all vertices at such a field is
now $o(\log n)$. With high probability the weak set has no interacting
pairs by Corollary~\ref{cor:order-critical}. The exact degree-sum
identity then gives the joint claimed limit by the Poisson field law.
\end{proof}

The factor $2cs\log n$ in \eqref{eq:sensitive-observation-count}
is the typical degree conditional on a bounded field. A reversible
coordinate changes the preferred product of all its incident
observations when it changes sign. Counting those observations using
the unconditional mean degree $c\log n$ would give a different
leading constant.

\begin{corollary}[Fluctuations in sensitivity at the last corrections]\label{cor:marked-sensitivity}
The critical-window limit can be refined to describe how many observations each remaining decision affects. Let $d=m/n$ vary as in Corollary~\ref{cor:order-critical}, put $v_n=2\gamma d$, and let $Z_{j,l}$, $l\ge1$, be independent standard normal variables, independent of the Poisson counts $P_j$ in that corollary. Jointly for any fixed finite collection of positive budgets,
\[
 \left(d_{\rm fix}(\Gamma),\frac{S_\Gamma-v_n d_{\rm fix}(\Gamma)}{\sqrt{v_n}}\right)
 \Rightarrow\left(\sum_{|j|<\Gamma}P_j,\sum_{|j|<\Gamma}\sum_{l=1}^{P_j}Z_{j,l}\right).
\]
For every fixed positive $\Gamma$ and integer $r\ge0$, conditional on $d_{\rm fix}(\Gamma)=r$ at finite $n$, the normalized centered sensitivity in the display converges to $N(0,r)$ (a point mass at zero if $r=0$). The same normal marks are used for nested budgets.
\end{corollary}

The variation within a fixed number of reversible coordinates is on the smaller scale $\sqrt{\log n}$, whereas their random number produces variation on the scale $\log n$.
\begin{proof}
Tilt the signs at any fixed list of distinct vertices as in the proof of~\eqref{eq:order-mass}. Under the independent Poisson comparison a named vertex has independent sign counts $X_+,X_-$ of mean $v_n/2$. For fixed $j$ and real $u$, the conditional moment generating function of $(X_++X_--v_n)/\sqrt{v_n}$ given $X_+-X_-=j$ is
\[
 \exp\{v_n(e^{u/\sqrt{v_n}}-1)-u\sqrt{v_n}\}\,
 \frac{p_j(v_ne^{u/\sqrt{v_n}})}{p_j(v_n)},
\]
where $p_j(v)$ is the probability of difference $j$ for two independent Poisson variables of mean $v/2$. The local limit $p_j(v)\sim(2\pi v)^{-1/2}$ makes the ratio tend to one, and the exponential tends to $e^{u^2/2}$. For a fixed list of $s$ vertices the independent comparison therefore gives independent normal marks conditional on their fields. The total variation coupling and distinct-address conditioning errors in the proof of~\eqref{eq:order-mass} are $o(d^{-s/2})$, smaller than the tilted probability of the specified fields. Consequently the same conditional convergence holds in the uniform fixed-size mask (it is enough to transfer bounded continuous functions, not unbounded moment generating functions). Undoing the tilt leaves the conditional law unchanged.

For finitely many field values and disjoint intervals with finite endpoints, the mixed factorial moments of the counts with marks in those intervals now converge to the independent Poisson factorial moments, with each intensity multiplied by its standard normal interval probability. These moments determine their joint law. Tightness of the total counts and of their marks follows from the unmarked Poisson convergence and the conditional one-vertex convergence; this yields joint convergence of the finite-field mark sums by truncation. Finally, with high probability the critical weak set is isolated and no vertex has degree zero, so the exact singleton and degree-sum identities above identify the dimension and sensitivity with these counts and degree sums. The integer-valued dimension permits conditioning on the interval $(r-1/2,r+1/2)$, whose boundary has zero limiting probability. Its limiting mass is positive for each $r\ge0$, so dividing the joint convergence of restricted bounded continuous integrals by this mass proves the conditional assertion. No independence of degrees in the original mask is required.\end{proof}

For two positive budgets $\Gamma_1\le\Gamma_2$, write $M_i=\sum_{|j|<\Gamma_i}P_j$ and $W_i=\sum_{|j|<\Gamma_i}\sum_{l=1}^{P_j} Z_{j,l}$. Conditional on all the field counts, the Gaussian increments over disjoint field ranges are independent. In particular, $(W_1,W_2)$ has conditional covariance $M_1$ and conditional variances $M_1,M_2$. If $M_2>M_1$, the conditional law of $(W_2-W_1)/\sqrt{M_2-M_1}$ is standard normal. This describes how the sensitivity changes when a budget crosses one or more integer field magnitudes; the increments come from newly reversible coordinates, rather than independent changes in already reversible coordinates.

A finite diagnostic samples 96 uniform masks at each of $n=512,2048,8192$ for $k=3$, $q=.17$ and window offset zero. The records include every vertex with field $-1,0,1$ (723 in total). For $n=8192$ the counts are 30, 64 and 137, and the empirical means of the standardized degrees are $-.101,-.094,-.094$, respectively. Their second moments are $.796,1.335,1.064$. Direct numerical summation of the one-star hypergeometric and binomial law on its full admissible support gives means $-.099,-.103,-.099$ and second moments $1.009,1.011,1.009$. These are local diagnostics of the normal marks, not checks of global optimality or of the isolation event. The recurrence was checked against 158 exact rational ratios on small stars. Independent 60-digit summation reproduced all nine conditional moment pairs with maximum absolute difference below $5\times 10^{-15}$. All seeds and counts, including zero counts, are in the accompanying program and records.

\subsection{Exhaustive objectives and larger signed stars}

For the exhaustive study we fixed four signs per block, $q=1/5$,
and used the same three normalized densities at each order. The
sample sizes $m$ were 12, 23 and 45 in increasing order of density.
Each of the nine cells contains 32 independent targets. Fixing one
sign in each of the first $k-1$ blocks leaves $2^{4k-(k-1)}$
distinct complete tensors. Direct multiplication of the factors and
an integer Walsh transform agreed on all $7{,}176{,}192$ scores. We
counted all maximizing tensors and deduplicated their restrictions to
the observed addresses. Independently, enumeration of the first
$k-1$ blocks followed by exact maximization over each last-block
sign reproduced all 288 global optimum scores and multiplicities and
all $7{,}680$ constrained-product margins. For every observed address
we also directly checked the optimum after each coefficient
increment $t\in\{-2,-1,0,1,2\}$, totaling $38{,}400$ checks.

\begin{table}[htbp]\centering
\caption{Exhaustive comparison with signed-degree predictions at $n=4$.
The gain and two count columns give the number of targets, out of 32,
on which each prediction was exact. The next column requires all three.
The last column counts observed addresses with exact clipped margin at
$B=2$. Addresses within a target are dependent.}
\label{tab:edge-residual-orders}
\begin{tabular}{ccrrrrr}\toprule
$k$&$\alpha$&Gain&Tensor count&Observed count&All three&Address margins\\\midrule
3&0.4&15&13&14&6&164/384\\
3&0.8&21&20&20&15&468/736\\
3&1.6&29&30&30&28&1206/1440\\
4&0.4&9&6&8&2&155/384\\
4&0.8&10&11&11&5&364/736\\
4&1.6&22&26&26&19&1053/1440\\
5&0.4&9&2&8&0&146/384\\
5&0.8&7&10&10&3&263/736\\
5&1.6&17&22&22&13&815/1440\\\bottomrule
\end{tabular}
\end{table}

The predictions used in Table~\ref{tab:edge-residual-orders} were
$2\sum_v(-D_v)_+$ for the gain, $2^{\#\{D_v=0\}}$ for the tensor
count and $2^{\#\{v:D_v=0,\,\deg_H(v)>0\}}$ for the observed count.
At four signs per block, 56 of 288 targets had the correct clipped
margin at every address. A subsequent check used the fields of the
first globally maximizing witness in the enumeration. All $7{,}680$
endpoint bounds held; the clipped observable margins were exact on
$4{,}677$ addresses, including all addresses on 63 of the 288 targets.
In 31 of the 288 targets the source lost
despite every signed degree being strictly positive. The breakdown
was 1 of 96 for triples, 14 of 96 for order four and 16 of 96 for
order five. Each of these sources strictly beat every single-coordinate
change. A test restricted to such changes would miss the improvement.

The larger field study used $\alpha=0.8$ and $q=1/5$, with 12
independent targets at each of the three orders for each
$n\in\{512,4096\}$. It recorded $663{,}552$ signed degrees on 72
targets. A subsequent calculation on these same records counted the
union of stars with $|D_v|<2$. At $n=512$ the sum of their degrees
was $3{,}451$ and their union contained $3{,}434$ addresses.
At $n=4096$ the corresponding counts were $4{,}915$ and $4{,}907$.
The finite-mask expected degree sums over those two panels were
$3{,}012.6$ and $5{,}325.8$, obtained by mixing binomial signs over
the hypergeometric degrees. The leading formula gave $3{,}128.3$
and $5{,}475.1$. Only the 288 small targets have exact optimum
response measurements. The released records include all target-level
counts, generation seeds, the small masks and noise signs, and checks
by each of the three exact computations.

\section{Calibration and reuse for triples}\label{app:triple-calibration-reuse}

For triples a second construction gives the number of residual errors after each full-data update. Fixed lower bounds on the channel biases suffice for one schedule. A second schedule learns the calibration it needs from a pilot.
\begin{theorem}[Exact fitting with or without channel bounds]\label{thm:geometric}\label{thm:geometric-known}
For triples, fixed $c,q,s$, and declared positive rational lower bounds on $\mu,\rho$, a capped algorithm uses $O(n/\log n)$ readings on every input. Without $c,q,s$, a second algorithm uses $O_{c,q,s}(n/\log n)$ expected readings, with a cap of $7n$ on its reading count $Q_n$ and tail $\Pr(Q_n>Cn/\log n)\le n^{-A}$ for some $C$ and each fixed $A>0$ for sufficiently large $n$. Both return a feasible assignment in $O((m+n)\log(n+2))$ word operations and attain~\eqref{eq:opt} with probability $1-o(1)$.
\end{theorem}

The original observations serve twice: they carry information about the source and define the objective ultimately maximized. Treating the fit as independent of the data used to initialize it would give an incorrect proof. For triples, separate target batches produce a preliminary estimate $z^0$; then reuse the entire target in simultaneous best responses $z^1,z^2$, keeping the previous sign at a tie. Let $D_v=\sum_{e\ni v}\eta_e$ be the signed source degree. If at most $b_v$ incident observations have the wrong product at their other endpoints, the computed field in source coordinates differs from $D_v$ by at most $2b_v$. This deterministic comparison reduces the dependence issue to a joint tail bound.

\begin{theorem}[Joint localization]\label{thm:joint-localization}
Fix a rational final-batch fraction $0<\tau<1$ and let $E$ be the error set of $z^0$. Put $h=1-2\sqrt{q(1-q)}$ and $\alpha=ch$. On intersecting with a pre-final calibration event whose failure probability is smaller than any fixed inverse power of $n$, the probability that fixed triples $P$ occur, fixed vertices $A$ lie in $E$, and vertices $B$ disjoint from $A$ have $D_v\le L_n$, is at most
\begin{equation}
 (m/n^3)^{|P|}n^{-\tau\alpha |A|-\alpha |B|+o(1)}
 \label{eq:main-joint-rate}
\end{equation}
uniformly over choices with those cardinalities, for $0\le L_n=o(\log n)$.
\end{theorem}
Prescribing the edges as well as the vertices in~\eqref{eq:main-joint-rate} bounds the number of errors incident to any coordinate. After one full-data update the errors of $z^1$ are confined to vertices of bounded signed degree. Repair around $z^1$ attains the global optimum at every fixed density. If $\alpha>1/2$, these weak vertices are isolated and $z^2$ already attains it (Appendices~\ref{app:direct-mixed-moment} and~\ref{app:recycling}).

\section{Additional Method and Comparison Details}
\label{app:main-details-study}

\subsection{Further Related Work}
Fiber sampling and adaptive tensor completion recover low-rank arrays from
selected entries \citep{caiafa2010,krishnamurthy2013}; side information
supports iterative recovery in tensor models \citep{montanari2014tensor}.
\citet{kim2021xor} give efficient noisy-XOR source recovery after a linear
number of degree-one initialization queries, without knowing the noise
parameters. Exact source-recovery thresholds are established for noisy
hypergraph observations \citep{ahn2019}. Algorithms for planted constraint satisfaction problems (CSPs) use
fresh-sample majority refinement, spectral methods, and sum-of-squares
\citep{basu2025,guruswami2023,basu2026}.

Recent preprints sharpen inference from noisy parity observations.
\citet{schmidhuber2026kikuchi} give tradeoffs between sample size and runtime for normalized
Kikuchi methods, with independent-sample refinement for exact source recovery
and pointwise-sound refutation certificates.
\citet{babakhani2026hce} use even covers for detection. For weak recovery
they aggregate products of clause labels into an $n\times n$ matrix.
Our acquisition model adds measurements of the retained source and asks for
the exact maximizer of the original noisy objective, including a regime in
which that source is suboptimal. The distinction concerns both access and
the required answer; neither source recovery nor a refutation bound alone
supplies that empirical maximizer.

Learning-augmented graph algorithms use predictions to improve approximation
guarantees \citep{cohen2024,aamand2025}, while noisy parity learning concerns
reconstruction from corrupted examples \citep{blum2003}. Worst-case
approximation of linear equations \citep{hastad2001} and average-case
refutation hypotheses \citep{feige2002} concern different computational
questions and access models. The activation rule used here is classical
hypergraph bootstrap percolation \citep{morrison2021}. We track the noisy
signs carried by the active set and compose acquisition with exact empirical
optimization. The replacement comparison uses even covers and low-degree
estimation \citep{hsieh2023,schramm2022,kunisky2019}.

For the acquisition schedule below, let $\mu_*,\rho_*$ be positive lower bounds on the two channel biases, $P$ the number of observations allocated to propagation, and $k$ the number of entries read per reference fiber (here the interaction order is three). Write $\mathcal M$ for one simultaneous majority update using the entire original target, retaining each old sign at a zero field.
\begin{algorithm}[t]
\caption{Acquisition, Reuse, and Exact Empirical Fitting (default repair)}
\label{alg:cascade-overview}
\begin{algorithmic}[1]
\STATE Use declared channel bounds, or estimate them with the charged pilot.
\STATE Compute the integer batch sizes and $k$ in Appendix~\ref{app:geometric}.
\IF{the bounds are nonpositive, $P=0$, or $k=n$}
\STATE Read all three fibers; return the full-fiber update and capped repair.
\ENDIF
\STATE Read the first $k$ entries of each reference fiber.
\STATE Propagate signs into unknown entries using fresh target batches.
\STATE Fill remaining unknowns fairly; calibrate on fresh batches.
\STATE Obtain $z^0$ by updating on the reserved final batch.
\STATE Compute $z^1=\mathcal M(z^0)$ and $z^2=\mathcal M(z^1)$ on the full target.
\STATE Retain the highest original-data score among $z^0,z^1,z^2$.
\STATE For the all-density guarantee, also repair $z^1$ by Algorithm~\ref{alg:repair}.
\STATE Return the highest feasible score and its assignment.
\end{algorithmic}
\end{algorithm}

\subsection{Intermediate Acquisition Schedules}
The displayed algorithm uses the explicit coefficient
\begin{equation}\label{eq:cascade-seed}
 K=\left\lceil\frac{5}{(\mu_*\rho_*)^4}\right\rceil,
 \qquad k=\min\{n,\lceil Kn^2/P\rceil\},
\end{equation}
with $P\sim m/4$; set $k=n$ when $P=0$. At declared bounds $\mu_*=.66$ and
$\rho_*=.80$, this gives $K=65$. The integer schedule is in
Appendix~\ref{app:geometric}; Algorithm~\ref{alg:cascade-overview} gives
its sequence of operations. Learning conservative channel bounds can increase
the finite seed coefficient. The expected and high-probability acquisition
bounds remain distinct from the all-input cap.

\subsection{Preserve the unknown entries}
An entry carries either a sign or an explicit unknown marker.
Read a predetermined prefix of each reference fiber. In a fresh batch of
original observations, a triple is usable when exactly one endpoint was
unknown at the start of that batch. Choose one usable triple uniformly
at each unknown vertex. Its label times the two known signs estimates
the missing sign. Retain the old known entries, keep an unknown marker
when no triple is available, and commit all new signs simultaneously.

This operation separates the extent of knowledge from its accuracy.
If the known fraction is \(p\), a vertex sees about
\(\lambda p^2\) usable observations in a batch with mean degree \(\lambda\).
If the known signs have conditional error bias at least \(r\), a new
sign has bias near \((1-2q)r^2\). The proof tracks both quantities.
It does not assume that stored errors are independent.

The distinction changes the acquisition scale. In the two-batch construction of Appendix~\ref{sec:sublinear-acquisition}, filling the unknown entries with fair signs after one initial extension gives full-block bias of order
\(\log n(k/n)^2\). Majority amplification of that diluted bias starts at
the scale \(k\gg n/(\log n)^{3/4}\). Preserving unknowns instead gives
the support-growth scale
\[
 p_{j+1}\asymp \min\{1,\log n\,p_j^2\}.
\] Here $p_j$ is the fraction known after round $j$. From \(p_0\asymp(\log n)^{-a}\),
its exponent is \(1-2^j(1-a)\). Every fixed \(a<1\) therefore reaches
constant support after finitely many steps, with a still positive
conditional bias.

For each fixed rational $0<a<1$, a fixed number of these rounds gives
an $O(n/(\log n)^a)$ reading guarantee without channel parameters.
The exact integer schedule and theorem appear in
Appendix~\ref{app:erasure} (Theorem~\ref{thm:erasure-fixed}).

\begin{table}[tb]
\centering
\caption{Additional readings for exact optimization under the stated law.
Constants can depend on fixed channel parameters and supplied bounds.
All methods retain the original objective. Full-fiber fallback costs \(3n\)
readings, plus the pilot when bounds are learned.}
\label{tab:acquisition}
\setlength{\tabcolsep}{3pt}
\begin{tabular}{@{}>{\raggedright\arraybackslash}p{.33\columnwidth}>{\raggedright\arraybackslash}p{.25\columnwidth}>{\raggedright\arraybackslash}p{.36\columnwidth}@{}}
\toprule
Schedule & Read order & Budget type\\
\midrule
Declared channel bounds & \(n/\log n\) & Deterministic\\
Learned channel bounds & \(n/\log n\) & Mean and high probability\\
Fixed \(0<a<1\) & \(n/(\log n)^a\) & Deterministic\\
Full fibers & \(3n\) & Deterministic\\
\bottomrule
\end{tabular}
\end{table}

For example, \(a=3/4\) uses two propagation batches and one final update;
\(a=7/8\) uses three propagation batches and one final update.
The cascade schedule reserves roughly one quarter of the target for
propagation, one quarter for calibration, and one half for final refinement.
Each of the first two quarters is divided into decreasing geometric batches.
A support bound with \(D=P/n\) is
\[
 p_{j+1}\ge\min\{(9/10)D2^{-j-1}p_j^2,1/8\}.
\]
The normalization \(y_j=(9/10)Dp_j/2^{j+2}\) gives repeated squaring
until support reaches one eighth. With \eqref{eq:cascade-seed}, the carried
bias remains at least \(D^{-\beta}\) for some fixed \(\beta<1/2\).

\begin{figure}[tb]
\centering
\includegraphics[width=.56\linewidth]{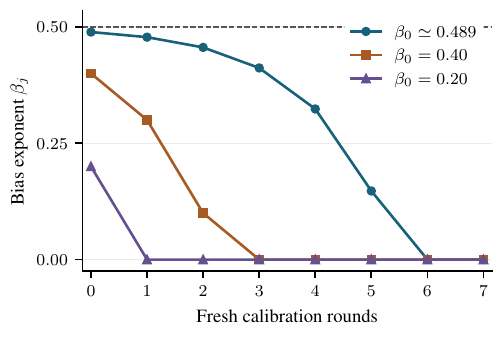}
\caption{Repeated calibration admits weaker initial alignment. If a block's
bias (mean source-relative sign) is at least a constant times \((\log n)^{-\beta_j}\), each fresh
logarithmic-degree batch gives the exponent bound
\(\beta_{j+1}=\max\{2\beta_j-1/2,0\}\). The displayed sequences are
analytic lower-bound exponents, not finite experimental accuracies.
Exponent zero means constant positive alignment; one further existing batch makes it tend to one as $n$ grows.}
\label{fig:calibration-cascade}
\end{figure}

The one-update crossover to constant alignment is the bias scale
\((\log n)^{-1/4}\). A cascade can start from
any bias at least a constant times \((\log n)^{-\beta}\), with
\(\beta<1/2\): a fresh logarithmic-degree batch changes its exponent by
\[
 \beta\longmapsto\max\{2\beta-1/2,0\}.
\]
Finitely many updates reach constant alignment; the next existing update
makes it tend to one (Figure~\ref{fig:calibration-cascade}). The remaining
stages preserve this accuracy. Appendix~\ref{app:geometric} proves the
schedule, its uniform conditioning, and the parameter-learning wrapper.

When the coefficient requests a whole fiber, the algorithm uses the simpler
full-fiber procedure directly. This finite branch agrees exactly with that
comparator when given the same readings. The branch condition \(k<n\)
only makes reduced acquisition possible; it does not assert that propagation
or repair will succeed at that size.

\FloatBarrier

\subsection{Capped Component Repair}
\begin{algorithm}[htb]
\caption{Exact repair on the original target}
\label{alg:repair}
\begin{algorithmic}[1]
\STATE Receive a reference tensor \(R\) and any feasible incumbents.
\STATE Compute \(D^R_v=\sum_{e\ni v}Y_eR_e\). Score \(R\) and its simultaneous negative-degree flip.
\FOR{\(u=0,\ldots,\max(12,\lceil\log_2(n+1)\rceil)\)}
\STATE Stop if the next scan exceeds the remaining work budget.
\STATE Find the components of \(W_u=\{v:D^R_v\le u\}\).
\STATE Mark those of size at most \(\max(12,\lceil\log_2(n+1)\rceil)\).
\STATE Stop if enumerating the whole marked stage exceeds the budget.
\STATE Enumerate marked components, fixing other signs to \(R\).
\STATE Combine their improvements; retain the highest feasible score seen.
\ENDFOR
\STATE Return the retained assignment and its score on \(H\).
\end{algorithmic}
\end{algorithm}

The budget is \(\lceil64(m+3n)\log(n+2)\rceil\) declared scan and
table units (Appendix~\ref{app:erasure-experiments}). Every stage starts from
the same reference \(R\): the reduced default uses the tensor of $z^1$,
whereas the full-fiber fallback uses the tensor of its one-update $z^0$.
Components above the size cap stay fixed.
The routine computes a feasible score even when it stops early.
\FloatBarrier

\subsection{A Balanced Threshold at Every Density}
A threshold at the mean of the generating-source score can be unbalanced for the empirical optimum, because empirical correction raises that score. Centering at the mean of the optimum instead gives a balanced decision at every fixed density.

\begin{proposition}[Fluctuations of the exact optimum]
\label{prop:clt}
For every fixed \(c>0\),
\[
\operatorname{Var}(G_n)=o(m),\qquad
\frac{\OPT-\E\OPT}{2\sqrt{mq(1-q)}}\ \Rightarrow\ N(0,1).
\]
\end{proposition}

The source score is a sum of independent signs. The gain is dependent on that sum, but its smaller variance suffices for the limit. The nearest point of \(m+2\mathbb Z\) to \(\E\OPT\), with ties rounded down, is an asymptotically balanced threshold. An independent training sample can estimate this center; its generation and solution costs are additional (Appendix~\ref{app:fluct}).

\subsection{Observable Noise Products}
The products of noise signs that the data reveal have an exact algebraic characterization. At fixed addresses, write measurements as \(X_i=\xi_i(-1)^{(B_0\theta)_i}\), where \(\theta\) is uniform over binary vectors and independent of a binary noise vector \(\xi\in\{-1,1\}^{r}\), where \(r\) is the number of measurement rows. The noise coordinates may have any joint distribution. Let \(\mathcal D_D\) contain the sets of at most \(D\) rows of \(B_0\) whose binary sum is zero, including the empty set.

\begin{proposition}[Which noise products are observable]
\label{prop:dual}
For any square-integrable \(f(\xi)\), the optimal degree-\(D\) polynomial risk equals
\[
\inf_{(c_S)}\E\left(f(\xi)-\sum_{S\in\mathcal D_D}c_S\prod_{i\in S}\xi_i\right)^2.
\]
\end{proposition}
Averaging over \(\theta\) erases all other monomials and improves squared error by conditional Jensen. Every surviving noise product is an observable product of measurements. This proves the equality.

When the three corresponding retained fiber entries have been measured, their rows and a target row sum to zero. Their product reveals a target error multiplied by sensor errors. The subset algorithm also uses relations among the original target observations during expansion and refinement. Replacement fibers use separate latent coordinates and do not create these target--source dependencies. For fixed reading addresses, the raw auxiliary record alone is independent of the decision conditional on \(H\): the optimum is invariant under a change of latent gauge and depends only on \(H,\eta\). Its computational use arises when it is combined with the original labels.

\subsection{Source Recovery Without the Target}

If the target observations are unavailable, even an adaptive reader needs \(\Omega_s(n\log n)\) noisy triple queries to recover the full tensor with a fixed positive success probability. In fact, with at most \(Q\) queries,
\begin{equation}
\Pr(\widehat T=T)\le
\left(1+\exp\!\left[-\frac{Q}{n}
\log\frac{1-s}{s}\right]\right)^{-n}.
\label{eq:source-only}
\end{equation}
Appendix~\ref{app:source-only} proves this for every adaptive strategy by revealing two factor blocks and bounding the remaining bit posteriors. It is a direct noisy-query argument \citep{feige1994,gu2026}. Fiber repetition attains the matching \(O_s(n\log n)\) order.

The retained-source algorithms use the target itself to infer and correct unmeasured factors. They give acquisition upper bounds for exact empirical optimization. A matching lower bound for arbitrary polynomial-time algorithms with both the target and source access remains open.

\subsection{Exact solutions on a fixed sparse grid}
\begin{figure*}[t]
\centering
\includegraphics[width=\textwidth]{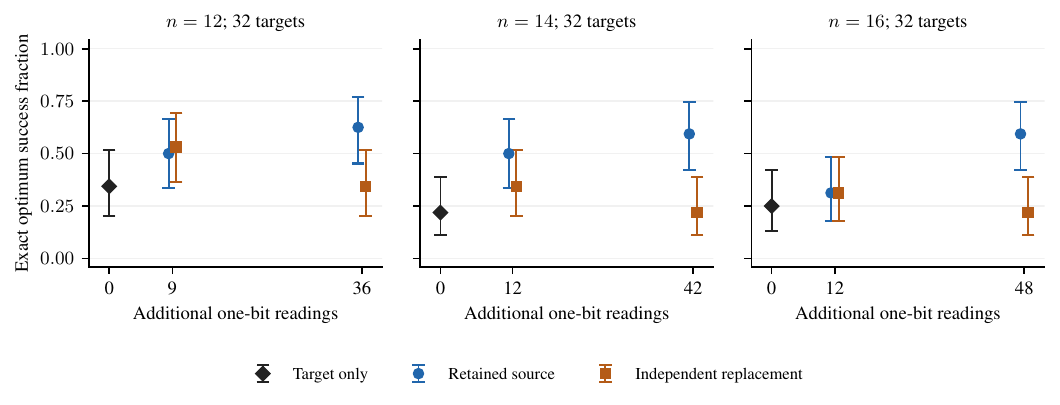}
\caption{Exact empirical optimization on the prospectively fixed 96-target grid.
Each panel contains 32 targets and shows the actual reading counts. Bars are
descriptive 95\% Wilson intervals; symbols at the same reading count are offset
slightly for visibility. All arms share the 32-start polishing cap; source arms
also perform the stated propagation scans. The reduced retained--replacement
comparison remains inconclusive.}
\label{fig:finite-study}
\end{figure*}

Table~\ref{tab:finite-study} and Figure~\ref{fig:finite-study} compare reduced and full fiber readings with target-only search on the same 96 targets.
\begin{table}[t]
\centering\small
\caption{Exact empirical successes on 96 targets. ``Suboptimal'' counts
successes among the 64 targets whose source is suboptimal. Reading counts
correspond to $n=12,14,16$. No target-only portfolio is included.}
\label{tab:finite-study}
\setlength{\tabcolsep}{4pt}
\begin{tabular}{lrrl}
\toprule
Method & Exact & Suboptimal & Readings\\
\midrule
Target only & 26 & 12 & 0\\
Reduced, retained & 42 & 20 & 9,12,12\\
Reduced, replacement & 38 & 23 & 9,12,12\\
Full, retained & 58 & 26 & 36,42,48\\
Full, replacement & 25 & 11 & 36,42,48\\
\bottomrule
\end{tabular}
\end{table}
\FloatBarrier

\section{Optimality on the frozen sample}\label{app:opt}
Throughout, $0<q<1/2$ is fixed, $N=n^3$, and $m\le N$. Each $m$-tuple of distinct locations is equally likely. Write $T_{ijk}=a_i b_j c_k$ for the latent tensor. Two representations differing by a simultaneous flip of two entire vectors define the same tensor. If a competitor has source-relative sign vectors $u,v,w$ in the three blocks, the proportion of the complete grid on which it disagrees with $T$ is
\[
 d=(1-\bar u\bar v\bar w)/2,
 \qquad \bar u=n^{-1}\textstyle\sum_i u_i,\quad\text{and similarly for }\bar v,\bar w.
\]
Suppose the sampled locations include $J$ disagreements. The competitor ties or beats the latent score exactly when the sum of the $J$ noise signs there is nonpositive. The exponential moment at $s=\tfrac12\log((1-q)/q)>0$ gives the conditional upper bound $[2\sqrt{q(1-q)}]^J$. Averaging over the mask gives
\[
 \Pr\{F(\mathrm{competitor})\ge F(T)\}\le (1-hd)^m\le e^{-mhd},\qquad h=1-2\sqrt{q(1-q)} .
\]
Indeed, this moment is the mean product of $m$ distinct members of a population containing a fraction $d$ of entries $1-h$ and a fraction $1-d$ of ones. Maclaurin's inequality bounds that mean by the $m$th power of the population mean. The same bound is immediate for independent sampling with replacement.

When $\bar u\bar v\bar w>0$, flip two blocks as needed to make all means positive. Let $t$ be the total number of minus signs in this representative. Then $d$ is at least the largest of the three blockwise fractions of minus signs, hence $d\ge t/(3n)$. At most $\binom{3n}{t}$ representatives have this $t$. For nonpositive product, $d\ge 1/2$ and there are at most $2^{3n}$ representatives. The probability that another tensor ties or wins is therefore at most
\begin{equation}
 E_{n,m}:=2^{3n}e^{-mh/2}
 +\big(1+e^{-mh/(3n)}\big)^{3n}-1.\label{eq:optbound}
\end{equation}
This argument is uniform over the hidden signs. For any $A>0$, $m=\lceil C n\log n\rceil$ with sufficiently large $C=C(q,A)$ gives $E_{n,m}=O(n^{-A})$. Gauge copies are excluded from the union. On the complementary event the unique maximizing tensor has score $m-2B$, where $B$ counts the corrupted target labels.

\section{Exact correction at every fixed logarithmic density}\label{app:local}
Use natural logarithms unless a base is displayed. Let $N=n^3$, $m=\lceil cn\log n\rceil$, $p=m/N$, $h=1-2\sqrt{q(1-q)}$, $\alpha=ch$, and $\lambda=\tfrac12\log((1-q)/q)$. Work in the true gauge, so an observed edge $e$ has sign $\eta_e$. The localization argument below is a property of the random objective and uses no acquired estimates. Appendices~\ref{sec:linear-acquisition}, \ref{app:sublinear}, and \ref{app:bootstrap} use it to justify their local searches.

\begin{lemma}[Localization of empirical corrections]
\label{lem:localization}
For each fixed \(c>0\) and \(q\in(0,1/2)\), there is a constant \(L_0\)
such that, with probability \(1-o(1)\), some maximizing assignment differs
from the source on components of size less than \(L_0\). Every changed
variable has true signed degree at most \(4L_0\).
\end{lemma}

\paragraph{Flip costs and localization.}
Write $S$ for a subset of the $3n$ vertices. Reversing those factor signs changes the planted score by $-2J(S)$, where $J(S)=\sum_{e\in H}\eta_e\one\{|S\cap e|\text{ is odd}\}$. Join two vertices in the interaction graph if they share an observed triple. No triple meets two different components of $S$ in that graph. Thus $J(S)$ is additive over components, and every component of an optimal flip set has $J\le0$. Select gauge representatives as in Appendix~\ref{app:opt}. For any fixed $\delta\in(0,1/2)$, with probability $1-o(1)$ all maximizers have a representative of size less than $\delta n$. Indeed, every remaining tensor has disagreement fraction at least $\min(\delta/3,1/2)$, and $2^{3n}\exp[-mh\min(\delta/3,1/2)]\to0$.

We next bound the size of each component by a constant. For $|S|=t\le\delta n$, the fraction $d$ of grid points with odd intersection is at least $(t/n)(1-t/n)$. To see this, write $u_1,u_2,u_3$ for the fractions flipped in the three modes, and expand $(1-\prod_i(1-2u_i))/2$. The negative term $2\sum_{i<j}u_i u_j$ is at most $(\sum_i u_i)^2$.

For any $s$ prescribed distinct grid edges $E$, the sampling law gives $\Pr(E\subseteq H)\le p^s$. Conditional on their inclusion, the $m-s$ remaining entries are a uniform sample of the other $N-s$. An unforced odd edge has exponential noise moment $1-h$ at $-\lambda$ and the remaining edges have moment one. The forced-edge factors are at most one. Maclaurin's inequality for sampling without replacement yields
\[
 \E\big[\one\{E\subseteq H\} e^{-\lambda J(S)}\big]
 \le p^s\exp[-hmd+2s].
\]
In fact at least $Nd-s$ odd locations remain, and $(m-s)(Nd-s)/(N-s)\ge md-2s$ when $Nd\ge s$; if $Nd<s$ the displayed upper bound is already trivial.

For a connected set $S$ of $t\ge2$ vertices, any spanning tree of its variable--triple incidence graph can be pruned of triple leaves. The remaining tree includes the $t$ variables and $s$ distinct triple nodes, each with one or two variable children when rooted at a variable. Hence $s\le t-1$. Restrict the incidence graph to variables in $S$, allowing the third endpoint of a triple to lie outside $S$. For fixed $s$ there are at most $4^{t+s}$ rooted plane tree shapes. The root has $3n$ possible labels. At each triple node at most $n^2$ choices complete the parent variable to a grid triple, and at most six choices assign its child or ordered children. Retain only valid trees with distinct edge labels; overcounting their number is harmless. Summing the preceding moment bound over witnesses, and absorbing the sum over $s$ in a constant $C_1^t$, gives
\begin{equation}
 \Pr\{\exists\ \text{connected }S,\ |S|=t,\ J(S)\le0\}
 \le 3n(C_1\log n)^t n^{-\alpha(1-\delta)t}.\label{eq:clusterbound}
\end{equation}
Choose a constant integer $L$ with $L\alpha(1-\delta)>2$. Geometric summation over $L\le t\le\delta n$ gives $o(1)$. Together with the large-distance bound, this implies that every optimal representative chosen above consists of components of fewer than $L$ vertices.

\paragraph{Which source coordinates can change.}
With high probability each pair of distinct vertices shares at most two observed triples: a union bound for three shared triples gives $3n^2\binom n3p^3=O(\log^3(n)/n)$. Fix the preceding optimal representative, one component $S_0$, and $v\in S_0$. Optimality against undoing the flip at $v$ requires the sum of the final signed observations at $v$ to be nonnegative. In the original gauge that sum is $-D_v+2\sum_e\eta_e$, where the correction sum includes incident edges having an odd number of other flipped endpoints. All such endpoints belong to $S_0$, so the number of correction terms is at most $2(|S_0|-1)$. It follows that $D_v\le4(|S_0|-1)\le4L$. For $\tau_n=\lceil\log\log(n+3)\rceil$ we eventually have $\tau_n\ge4L$. Thus $W=\{v:D_v\le\tau_n\}$ contains the flip set of an optimal representative with high probability. Given the source tensor, the solver computes $D_v$ by summing $Y_e T_e$ over incident observations. An estimated tensor instead yields the field at that estimate. The two fields agree everywhere when the estimate recovers the source tensor.

\paragraph{The induced components in $W$.}
Fix $t$ independently of $n$. For a specified $t$-vertex set $S$ and forced edge set $E$ of size $s\le t$, the event $S\subseteq W$ implies $\sum_{v\in S}D_v\le t\tau_n$. Each edge meeting $S$ in $j$ endpoints has moment $w_j=\E e^{-j\lambda\eta}$ in the exponential bound for this sum. We have $w_0=w_2=1$, $w_1=1-h$, and $w_3$ is finite. There are $tn^2+O_t(n)$ grid triples with exactly one endpoint in $S$, $O_t(n)$ with two, and $O_t(1)$ with three. Removing the forced edges changes the average moment by $O_{t,q}(1/N)$. The mean among the unforced locations is consequently $1-ht/n+O_{t,q}(n^{-2})$. The forced factors are bounded by a constant depending on $t,q$. Maclaurin's inequality and $e^{\lambda t\tau_n}=n^{o(1)}$ give, uniformly in all such labeled sets,
\begin{equation}
 \Pr\{S\subseteq W,\ E\subseteq H\}\le p^s n^{-\alpha t+o(1)}.\label{eq:weak}
\end{equation}
For $t=1$ this proves $\E|W|\le n^{1-\alpha+o(1)}$ (constant factors absorbed). For $t\ge2$, sum~\eqref{eq:weak} over the same tree witnesses as above. The expected number of connected weak sets of size $t$ is at most $n^{1-\alpha t+o(1)}$. A connected graph with at least $t$ vertices contains a connected subset of size exactly $t$. Taking any fixed integer $t>1/\alpha$ therefore bounds all components of $W$ by a constant with probability tending to one. When $\alpha>1/2$ the choice $t=2$ excludes every interacting pair in $W$; when $\alpha>1$, $W$ is empty with high probability.

\paragraph{Exact optimization and costs.}
Freeze the recovered source signs outside $W$. Every observed clause touching $W$ is assigned to exactly one component; all its weak variables are in that component. Enumerating the $2^k$ local sign choices on a component of size $k$ therefore yields its exact optimum with the outside signs fixed. Adding the local optimal values gives the full restricted optimum. On the event proved above this is the original optimum. One may take the better of this feasible solution and the feasible parallel-flip candidate. If any $k>b_n=\max(12,\lceil\log_2(n+1)\rceil)$, simply return the latter. Computing degrees, components by graph traversal, and per-component edge lists costs $O(m+n)$ operations. Enumeration by a direct scan costs $O((m+n)2^{b_n})$ in the worst case and $O_{c,q}(m+n)$ with high probability, in addition to acquisition. A Walsh transform of each component's table of signed parity weights gives the same local values. Its $O(k2^k)$ work per component obeys the same total bound since the component sizes sum to at most $3n$. An argmax yields actual factors; the experimental code records the score. Integer counters use $O(\log(n+m+r))$ bits. In the singleton regime, choosing all vertices with negative $D_v$ gives the exact formula in Theorem~\ref{thm:local}. For $\alpha>1$ every optimal representative chosen above is empty. At $\alpha=1$ a sharper one-vertex bound suffices because $W$ still consists only of isolated vertices. Let $d$ be the random unsigned degree of a specified vertex and $b=2\sqrt{q(1-q)}=1-h$. Conditional on $d$, the binomial tail for $D_v\le0$ is at most $C_q b^d/\sqrt{d+1}$: bound its first term by the central binomial coefficient and sum the rest geometrically with ratio at most $q/(1-q)$. For any $z\in(0,1)$ the hypergeometric moment satisfies $\E z^d\le[1-(1-z)/n]^m$. For $d\ge\kappa\log n$ use this bound at $z=b$ and the denominator $\sqrt{\kappa\log n}$. For $d<\kappa\log n$, bound $b^d$ by $n^\kappa(b/e)^d$ with $\kappa=cb(1-e^{-1})/2$. Together the two pieces give
\[
 \Pr(D_v\le0)=O_{c,q}(n^{-\alpha}/\sqrt{\log n}).
\]
A union bound at $\alpha=1$ excludes every nonpositive vertex with high probability. Each nonempty subset of isolated weak vertices then decreases the score, proving uniqueness also at equality.

\paragraph{How far the source score falls short.}
For $0<\alpha<1$ the upper exponent follows from $\E|W|\le n^{1-\alpha+o(1)}$, Markov's inequality, and the maximum unsigned vertex degree being $O_c(\log n)$ with high probability (apply a hypergeometric Chernoff bound to each degree). A representative supported on $W$ can improve the score by at most twice the number of incident edges, at most $O_c(|W|\log n)$.

For the matching lower bound take the Bernoulli mask with inclusion probability $p=m/N$ and consider only first-mode vertices, which have disjoint incident triple sets. The event $D_v<0$ has probability $n^{-\alpha+o(1)}$. For the upper bound use $\E e^{-\lambda D_v}=(1-ph)^{n^2}$. For the lower bound take $\ell=\lfloor n^2p\sqrt{q(1-q)}\rfloor$ positive and $\ell+1$ negative observations in its $n^2$ possible positions. The multinomial probability is
\[
 \frac{(n^2)_{2\ell+1}}{\ell!(\ell+1)!}
 [p(1-q)]^{\ell}(pq)^{\ell+1}(1-p)^{n^2-2\ell-1}.
\]
Expanding the falling factorial and applying Stirling gives log probability $-n^2ph+o(\log n)=-\alpha\log n+o(\log n)$. Thus, for every fixed $\epsilon>0$, at least $n^{1-\alpha-\epsilon}$ such vertices are negative with probability $1-\exp[-n^{1-\alpha+o(1)}]$ (for positive exponents, by binomial concentration). Conditioning on exactly $m$ sampled triples preserves this high-probability claim: the conditioning event has probability $\Theta(m^{-1/2})$ by Stirling. Flipping all these first-mode signs increases the score by at least two per vertex, proving the lower exponent for the stated fixed-size mask. This argument also shows directly that the planted tensor is strictly suboptimal with probability tending to one.

\paragraph{The balanced decision.}
Let $K_m=m-2\lfloor qm\rfloor$, $B=\sum_e\one\{\eta_e=-1\}$ and $U=2\one\{\OPT\ge K_m\}-1$. The planted score is a sum of $m$ independent signs with standard deviation of order $\sqrt{n\log n}$. For $\alpha<1/2$ the positive improvement dominates this fluctuation in probability, so $\Pr(U=1)\to1$. For $1/2<\alpha<1$ the improvement is $o_{\Pr}(\sqrt m)$. The binomial central limit theorem and the absence of a point mass at zero in its limiting Gaussian imply $\Pr(U\ne V)=o(1)$ for $V=2\one\{B\le\lfloor qm\rfloor\}-1$. For $\alpha\ge1$ this follows from uniqueness. At $\alpha=1/2$, let $B_W$ count adjacent pairs in $W$. The connected-pair estimate gives $\E B_W\le n^{o(1)}$, and at most $2B_W$ weak vertices belong to nonsingleton components. With high probability their total incident degree is at most $O(B_W\log n)$. The gain on the remaining components is bounded by $2\sum_v(-D_v)_+$. Since $\E e^{-\lambda D_v}\le(1-h/n)^m\le n^{-\alpha}$, summing the geometric tail $\Pr(D_v\le-j)\le n^{-\alpha}e^{-\lambda j}$ over $j\ge1$ bounds the expectation of this last gain by $O(\sqrt n)$. Both contributions are $o_{\Pr}(\sqrt m)$, proving balance at equality as well. The polynomial projection and common-degree arguments in Appendices~\ref{app:lower} and~\ref{app:upper} transfer with $\Pr(U\ne V)=o(1)$ in place of the large-density error bound. The repeated-fiber polynomial upper bound estimates the decision from the raw remeasured score. Appendix~\ref{app:sublinear-degree} gives a polynomial upper bound at sufficiently large density using the smaller acquisition budget. The exact all-density value comes from the component search.

\subsection{Fluctuations and independent calibration}\label{app:fluct}
Start with $m$ independent draws of a triple and a noise sign. Treat repeated draws as separate observations in the objective. With probability $1-O(m^2/n^3)=1-o(1)$ all triples are distinct; conditional on this event their unordered set is exactly the mask in the theorem. For $m$ or $m-1$ draws the localization proof above thus gives, with probability tending to one, a maximizing flip set of size $o(n)$: choose an optimal representative with minimum cardinality, breaking ties lexicographically. This rule is invariant to permuting the observations. For example, on the events of the proof the chosen cardinality is no larger than $|W|$, and $\E|W|\le n^{1-\alpha+o(1)}$ in the distinct model. The maximum unsigned degree is $O_c(\log n)$ with high probability, so the fraction of observations incident to the chosen set is $o_{\Pr}(1)$. Statements about these bounded fractions also hold in expectation under the independent sampling law.

The following finite-sample inequality makes the stability argument explicit.
It applies before any asymptotic or distinctness conditioning.

\begin{lemma}[Variance controlled by an optimizer's support]\label{lem:support-variance}
Let $Z_i=(e_i,\eta_i)$, $1\le i\le m$, be independent and identically
 distributed draws. For a finite family of vertex sets define
\[
 G_m=\max_S\sum_{i=1}^m g(S,Z_i),\qquad
 |g(S,Z_i)|\le M,\qquad g(S,Z_i)=0\ \text{if }e_i\cap S=\varnothing.
\]
Choose maximizers $S_m$ and $S_{m-1}$ by a fixed rule invariant under
permutation of the observations. Set
\[
 a_m=\E\!\left[\frac1m\sum_{i=1}^m
             \one\{e_i\cap S_m\ne\varnothing\}\right],\qquad
 b_m=\Pr\{e_m\cap S_{m-1}\ne\varnothing\},
\]
where $S_{m-1}$ uses only the first $m-1$ draws. Then
\[
 \operatorname{Var}(G_m)\le M^2m(a_m+b_m).
 \label{eq:support-var}
\]
For a uniform tripartite grid edge independent of $S_{m-1}$,
\[
 b_m=\E\!\left[1-\prod_{j=1}^3
       \left(1-\frac{|S_{m-1}\cap V_j|}{n}\right)\right]
 \le \E\min\{1,|S_{m-1}|/n\}.
\]
\end{lemma}
\begin{proof}
Delete observation $i$ and let $G_{-i}$ be the resulting maximum.
The increment $\Delta_i=G_m-G_{-i}$ lies in $[-M,M]$.
If $\Delta_i>0$, the deleted edge meets $S_m$: otherwise the value
at $S_m$ would be unchanged, contradicting the decrease of the maximum.
If $\Delta_i<0$, the deleted edge meets the selected optimizer $S_{-i}$
of the remaining observations, by the reverse argument. Exchangeability
therefore gives
\[
 \E\Delta_i^2\le M^2(a_m+b_m).
\]
Conditional on the other observations, $G_{-i}$ is a constant.
The conditional mean minimizes squared error, so
$\E\operatorname{Var}(G_m\mid Z_{-i})\le\E\Delta_i^2$.
Variance tensorization, equivalently the conditional-variance form of
Efron--Stein \citep{steele1986}, yields
$\operatorname{Var}(G_m)\le\sum_i\E\operatorname{Var}(G_m\mid Z_{-i})$.
For the final formula, a uniform edge misses the fixed set in all three
blocks with the displayed product probability; a union bound gives the
last inequality.
\end{proof}

Apply the lemma with
$g(S,(e,\eta))=-2\eta\one\{|e\cap S|\text{ odd}\}$ and $M=2$.
The localization argument gives $a_m=o(1)$ and $b_m=o(1)$ under independent
sampling. For the first statement, on its high-probability events
$|S_m|\le |W|\le n^{1-\alpha/2}$ and the maximum degree is $O_c(\log n)$,
so the incident fraction tends to zero; its boundedness handles the
exceptional event. The same support bound handles $b_m$ for $m-1$ draws.
Thus $\operatorname{Var}(G_m)\le4m(a_m+b_m)=o(m)$.
Conditioning on all addresses being distinct preserves this order:
conditional variance is at most the unconditional variance divided by
that event's probability, which tends to one. This proves the variance
claim for the fixed-size mask used in the paper. The lemma does not
require a unique optimizer or independence between the optimizer and its data.

The centered planted score is a sum of $m$ independent signs with variance $4mq(1-q)$. Chebyshev and the binomial central limit theorem prove Proposition~\ref{prop:clt} without requiring independence of the planted score and $G_n$. Let $K$ be the nearest integer of parity $m$ to $\E\OPT$ (break ties downwards). The true decision at $K$ and the sign of the centered planted score differ with probability $o(1)$, by the continuity of the limiting Gaussian. The independent-source lower bound in Appendix~\ref{app:lower} transfers by this $L^2$ agreement and contractivity of orthogonal projection: the projection of the centered planted-score decision has vanishing norm on the permitted polynomial space, and replacing it by the optimum decision changes that norm by $o(1)$. The same-record upper bound in Appendix~\ref{app:sublinear-degree} retains its sufficiently-large-density assumption.

Independent training instances provide a practical threshold. Independently generate an odd $k=\Theta(\log n)$ training instances from the same known generator, using fresh sources and auxiliary reads for those instances alone. Compute their scores with the complete two-batch procedure of Theorem~\ref{thm:sublinear}, and supply their median as one common threshold to both test experiments. Generating and solving these training instances costs polynomial time and $3kk_n=O(kn/\sqrt{\log n})$ auxiliary bits, paid separately from test acquisition. For every fixed $\epsilon>0$, the probabilities that a computed training score is below $\E\OPT-\epsilon\sqrt m$ or above $\E\OPT+\epsilon\sqrt m$ are each bounded below $1/2$ by a positive constant for large $n$, by the Gaussian limit and the $o(1)$ optimization error. Binomial concentration puts the median within these two limits with probability tending to one. Its independent deviation from the population center is thus $o_{\Pr}(\sqrt m)$. The decision agreement and degree conclusions carry over, including if the independent training information is disclosed to both sides.

\subsection{Testing the source alignment}\label{app:align}
Given the original locations, greedily retain a triple and discard every triple sharing one of its vertices. If the maximum unsigned degree is $\Delta$, each retained triple deletes at most $3\Delta$ triples, so the resulting matching has size at least $m/(3\Delta)$. Under the fixed-size sparse law $\Delta=O_c(\log n)$ with high probability. Hence any desired $k=O(\log n)$ checks are available with high probability. Conditional on the mask and an address-only choice of matching, use separate independent reads for the three reference signs for each triple, even when reference addresses coincide. Write $\widehat P_e$ for the product of the original label and these three sensor readings. In the coupled model the true response cancels by~\eqref{eq:fiber}. What remains is one target and three sensor errors. These are disjoint independent noise variables across checks. In the independent model there is also the multiplier $T_e T'_e$, which consists of fresh independent factors at each disjoint triple; these multipliers are independent uniform signs. A threshold at $b/2$ for the sample average with $b=(1-2q)(1-2s)^3$ gives error at most $\exp(-kb^2/8)$ in each direction, conditional on the matching. The comparison tests these two specified access laws. A persistent partial corruption could pass it without permitting exact reconstruction.

\section{Linear acquisition followed by local optimization}
\label{sec:linear-acquisition}

The logarithmic repetition factor is unnecessary when the original sample
can also be used to improve the initial estimate. Read each of the $3n$ reference-fiber entries $A,B,C$ of~\eqref{eq:fiber} once, with independent sensor errors of probability
$s<1/2$. Use separate visits at the common center. Write the resulting
factor arrays as $A^0,B^0,C^0$. Compute all three updated arrays
\emph{simultaneously}, with $\operatorname{sgn}_a$ denoting the sign function with value $a$ at zero:
\[
 A^1_i=\operatorname{sgn}_{A^0_i}\!\sum_{(i,j,k)\in H}Y_{ijk}B^0_jC^0_k,
\]
with the analogous formulas for $B^1_j,C^1_k$. Every field uses the initial
arrays. A zero field retains the previous sign. The proof also permits any fixed tie rule. The update takes one pass through
the original sample.

\begin{theorem}[Exact optimization after a linear number of new readings]
\label{thm:linear-acquisition}
Fix $c>0$ and $q,s\in(0,1/2)$, and let $m=\lceil cn\log n\rceil$.
There is an algorithm that uses exactly $3n$ additional noisy binary
readings and returns a feasible assignment and its score on every input.
Under the coupled-source law, that score equals $\OPT(H,Y)$ with probability
$1-o(1)$. The algorithm does not require $c,q$, or $s$, and uses
$O((m+n)\log(n+2))$ word operations on every input.

The algorithm first applies the synchronous update above. It then searches
the connected components of
\[
 W_\tau=\left\{v:\sum_{e\ni v}Y_eT^1_e\le\tau\right\}
\]
for successive integer cutoffs
$\tau=0,1,\ldots,\max(12,\lceil\log_2(n+1)\rceil)$.
Each completed stage optimizes exactly on the components that it enumerates,
with all other signs held fixed to the same updated reference $T^1$. Oversized components may be skipped. The algorithm retains the best feasible assignment found and
stops before a prescribed budget proportional to $(m+n)\log(n+2)$ is
exceeded. Components larger than $\max(12,\lceil\log_2(n+1)\rceil)$ need
not be enumerated.

If $c,q,s$ are supplied, a suitable constant cutoff gives
$O_{c,q,s}(m+n)$ work with probability $1-o(1)$, with a polynomial cap
on all other inputs.
\end{theorem}

The theorem concerns the optimum on the original noisy sample. The updated
assignment need not recover the source, and the source need not maximize
the objective. Local optimization removes the remaining errors in either
case. At every fixed positive logarithmic density, the additional record
contains only a constant number of bits per factor.

\subsection{Proof of Theorem~\ref{thm:linear-acquisition}}

Use the true reference-fiber gauge $A_iB_jC_k=T_{ijk}$ and write
$A^0_i=A_i\epsilon^A_i$, and similarly for the other two blocks.
The $3n$ sensor errors are independent of the target sample and of one
another. Put $\mu=1-2q$, $\rho=1-2s$, and choose any fixed
$\rho_0\in(0,\rho)$. Hoeffding's inequality gives
\[
 \Pr(\mathcal C^c)\le
 3\exp[-n(\rho-\rho_0)^2/2],
 \qquad
 \mathcal C=\{\bar\epsilon^A,\bar\epsilon^B,\bar\epsilon^C\ge\rho_0\}.
\]
Define
\[
 \lambda=\operatorname{arctanh}(\mu\rho_0^2),\qquad
 h_0=1-\sqrt{1-\mu^2\rho_0^4},\qquad \beta=ch_0>0.
\]
All constants below may depend on $c,q,s,\rho_0$.

In true coordinates, the field at a first-block vertex is
\[
 M_i=\sum_{(i,j,k)\in H}\eta_{ijk}\epsilon^B_j\epsilon^C_k.
\]
The other two fields have the same form. Let $E=\{v:M_v\le0\}$.
Every error in the updated assignment belongs to $E$, whatever tie rule
is used. Also let $D_v=\sum_{e\ni v}\eta_e$ be the true signed degree.

We first prove a joint bound that keeps the dependence between the
fields, degrees, and sampled graph. Fix a set $S$ of $t$ vertices, where
$t$ is independent of $n$. For every $v\in S$, choose either the event
$M_v\le0$ or the event $D_v\le L_n$, with $0\le L_n=o(\log n)$.
Let $Q$ be the sum of the chosen field or degree at these vertices.
The chosen events imply $Q\le tL_n$.

Conditional on any sensor-error arrays in $\mathcal C$, a grid triple
meeting $S$ at only a vertex with a field constraint has noise moment
$\cosh\lambda-\mu\sigma_e\sinh\lambda$, with
$\sigma_e$ the product of the two corresponding seed errors.
Averaging over the two free coordinates gives
\[
 \cosh\lambda-\mu\bar\epsilon^{\,b}\bar\epsilon^{\,c}\sinh\lambda
 \le \cosh\lambda-\mu\rho_0^2\sinh\lambda=1-h_0.
\]
For a degree constraint the corresponding average is
$\cosh\lambda-\mu\sinh\lambda\le1-h_0$.
There are only $O_t(n)$ grid triples meeting two or more vertices of
$S$, and their moments are at most $e^{3\lambda}$. Excluding these
triples from each single-vertex average therefore shows that the
population mean of the moment weights is at most
\[
 1-\frac{h_0t}{n}+O_{t,q,s}(n^{-2}).
\]
This bound is uniform over the error arrays in $\mathcal C$.

Now prescribe any fixed number $j$ of distinct target triples $F$.
Write $p=m/n^3$. Their inclusion probability is at most $p^j$.
Conditional on inclusion, the remaining triples are sampled without
replacement from the other locations. Removing $j$ bounded weights
changes the population average by $O_j(n^{-3})$. The forced noise
moments contribute only a constant. Maclaurin's inequality and
exponential Markov thus give
\[
 \Pr\!\left(
 F\subseteq H,\ \bigcap_{v\in S}\{\text{chosen constraint at }v\}
 \,\middle|\,\epsilon\right)
 \le C_{t,j,q,s}\,p^j
 \exp\!\left[\lambda tL_n-\frac{mh_0t}{n}+O_{t,j,q,s}(m/n^2)\right]
 =p^j n^{-\beta t+o(1)}.
 \tag{LA}
\]
The $o(1)$ is uniform in the labels, forced triples, and permitted seed
arrays. Summing over the $2^t$ choices of constraint proves the same
bound for $S\subseteq E\cup\{D\le L_n\}$.

Two consequences of (LA) are needed.

\emph{Bounded connected components.}
A connected $t$-vertex set has a rooted incidence-tree witness with at
most $t-1$ distinct triple nodes. There are at most $C^t$ plane shapes,
$3n$ root labels, and at most $6n^2$ extensions at each triple node.
Sum (LA) over these witnesses. For every fixed $t$,
\[
 \Pr\{\text{a connected $t$-set lies in }
            E\cup\{D\le L_n\}\}
 \le n^{1-\beta t+o(1)}.
\]
Choose a fixed integer $k>1/\beta$. With probability $1-o(1)$,
every component of this union has fewer than $k$ vertices.

\emph{A bounded number of erroneous neighbors.}
Fix a root vertex and $k$ of its distinct neighbors that belong to $E$.
Their adjacency has a witness consisting of at most $k$ distinct triples,
each containing the root and one or two designated neighbors. There
are $3n$ root labels and at most $C^k(n^2)^j$ such witnesses with $j$
triples. Apply (LA) only to the $k$ neighbors; the root is unconstrained.
The union bound is again $n^{1-\beta k+o(1)}$.
Thus, after increasing $k$ if needed, every vertex has fewer than $k$
neighbors in $E$, with probability $1-o(1)$.

The original sparse-localization proof supplies a fixed $L_0$ such that
some maximizing assignment differs from the true source only on
$\{D\le L_0\}$, with probability $1-o(1)$. This follows from Lemma~\ref{lem:localization} in Appendix~\ref{app:local},
with the field cutoff enlarged by a fixed factor if necessary. Work on the intersection of all these high-probability
events. The flip set $R$ from the updated assignment to that optimizer
is contained in $E\cup\{D\le L_0\}$. Every component of $R$ therefore
has fewer than $k$ vertices. For any $v\in R$, optimality against undoing
its flip, and pair-codegree at most two, give
\[
 D^1_v:=\sum_{e\ni v}Y_eT^1_e\le4(k-1).
\]
Consequently the fixed cutoff $\tau_*=4k$ contains the whole support
$R$.

It remains to show that the \emph{observable} set $W_{\tau_*}$ is small
enough to search. If $v\notin E$, its updated sign equals its true sign.
Only triples containing another updated error can then change its signed
degree. There are fewer than $k$ such error neighbors and at most two
triples per pair. Hence
\[
 D^1_v\ge D_v-4k,\qquad v\notin E,
\]
and therefore
\[
 W_{\tau_*}\subseteq E\cup\{D\le\tau_*+4k\}.
\]
The component bound already proved applies to this fixed threshold.
Every component of $W_{\tau_*}$ has bounded size. Enumerating those
components with all outside signs fixed finds the original optimum.

For the adaptive algorithm, all cutoffs before $\tau_*$ give nested
subsets of $W_{\tau_*}$. The total work through stage $\tau_*$ is
$O_{c,q,s}(m+n)$, so for all sufficiently large $n$ these stages fit
within any fixed positive multiple of $(m+n)\log(n+2)$.
The component cap and the final cutoff also exceed the relevant fixed
constants. The exact incumbent is therefore found before stopping.
Later feasible candidates cannot lower the retained score or exceed
the global optimum. The budget, implemented before each scan or local
enumeration, gives the stated bound on every input.

All scores and Walsh-transform entries are integers of magnitude at
most $m$. Counters and addresses use $O(\log(m+n))$ bits. Thus the
word-operation claim has at most an additional $O(\log^2(m+n+2))$ factor
under the conservative random-access arithmetic-bit accounting of
Appendix~\ref{app:erasure-experiments}. The acquisition has exactly $3n$ binary
outputs and requires $O(n\log n)$ address bits in an explicit address
list, or $O(n)$ word operations for the predetermined fiber schedule.

\paragraph{Direct recovery at sufficiently large density.}
The same calculation for a single field gives the explicit bound
\[
 \Pr\{T^1\ne T\}\le
 3e^{-n(\rho-\rho_0)^2/2}+3n e^{-mh_0/n}.
\]
Thus if $ch_0>1$, one synchronous update already recovers the source
with high probability. If $ch(q)>1$, a sufficiently large \emph{constant}
odd number of repeated reads per fiber makes $ch_0>1$, after replacing
$\rho$ by the majority-vote bias. The local stage is what allows the
linear-read theorem to hold at every $c>0$.

\section{Composition of calibration and empirical correction}
\label{app:direct-mixed-moment}

The final estimation errors and the weak true degrees are dependent.
Their joint bound can nevertheless retain both exponents. The final
sample is fresh conditional on the past, and the past degree contribution
has its own exponential moment. Integrating these two moments in that
order retains both deviation rates. The population-moment comparison uses
classical concentration tools for sampling without replacement
\citep{hoeffding1963}; we keep the prescribed-location factors throughout
the conditional calculation.

Write $\mathcal I(x)=1-\sqrt{1-x^2}$ for $0\le x\le1$. Consider a uniform partition of the target into predetermined capacities
$m_1,\ldots,m_k,t$, with $m=\sum_jm_j+t$. The last capacity $t$ is
reserved for final refinement. We call the batch membership a color; the number of colors may grow with $n$.
For a deterministic $b_0$ (possibly depending on $n$), let $\mathcal G$ be a pre-final measurable event on which the three full
calibration arrays have mean error sign at least $b_0\in[0,1]$. Capacities and coloring are independent of labels; the arrays may use a data-dependent number of source readings.
All prior observations, sensor readings, and private choices may be
included in the history. Define
\[
 \theta=\operatorname{arctanh}(\mu),\qquad
 \lambda=\operatorname{arctanh}(\mu b_0^2),\qquad
 w=e^{3\theta},\qquad 0<\mu<1.
\]
Fix deterministic disjoint vertex sets $A,B$ of sizes $a,b$, and put $s=a+b$.
As before, $M_v$ is the final field in source coordinates and $D_v$ is
the signed degree on the complete target.

For each color fix deterministic prescribed locations $F_j$, all mutually disjoint,
with $f_j=|F_j|$; write $f_f$ for the final color and
$f=\sum_jf_j+f_f$. Their exact inclusion probability is
\[
 \pi_F=\frac{(t)_{f_f}\prod_j(m_j)_{f_j}}{(n^3)_f}.
\]
If a prescribed count exceeds its capacity, the event is impossible.

\begin{lemma}[Direct mixed field and degree bound]
\label{lem:direct-mixed-moment}
Suppose $m\le n^3/2$. For every $L\ge0$,
\begin{align*}
 &\Pr\{F,\mathcal G,\ M_v\le0\ (v\in A),\ D_v\le L\ (v\in B)\}\\
 &\quad\le\pi_F w^f\exp\left\{
 \theta bL-\frac{ta}{n}\mathcal I(\mu b_0^2)-\frac{mb}{n}\mathcal I(\mu)
       +\mathcal R\right\},\\
 &\mathcal R=\frac{fs}{n}
       +\frac{4ws(s-1)m}{n^2}
       +\frac{2w(m^2+mf)}{n^3}.
\end{align*}
The bound is uniform in the arrays, histories, vertex labels, prescribed
locations, and number of colors. The quadratic term counts pairs among the $s$ selected vertices.
\end{lemma}

\begin{proof}
Condition first on all prescribed-color inclusions and keep their
probability $\pi_F$ outside the calculation. Put $R=m-t$. For the analysis,
let $\mathcal H_{\rm pre}$ contain the complete pre-final history and the latent source.
The calibration arrays and $\mathcal G$ are then $\mathcal H_{\rm pre}$-measurable.
Write $M_A=\sum_{v\in A}M_v$ and split the selected true degrees as
$D_B=D_B^{\mathrm{past}}+D_B^{\mathrm{final}}$, summing over $v\in B$.
Markov's inequality bounds the probability conditional on $F$ by
\[
 e^{\theta bL}\E\bigl[
 \mathbf1_{\mathcal G}e^{-\lambda M_A-\theta D_B}\mid F\bigr].
\]
The order of integration is explicit in the identity
\begin{align*}
 &\E\bigl[\mathbf1_{\mathcal G}e^{-\lambda M_A-\theta D_B}\mid F\bigr]\\
 &\quad=\E\Bigl[\mathbf1_{\mathcal G}
 e^{-\theta D_B^{\mathrm{past}}}
 \E\bigl[e^{-\lambda M_A-\theta D_B^{\mathrm{final}}}
       \mid\mathcal H_{\rm pre},F\bigr]\,\Bigm|\,F\Bigr].
\end{align*}
Conditional on $\mathcal H_{\rm pre},F$, the $t-f_f$ free final locations form a
uniform sample from the remaining $n^3-R-f_f$ locations. We first bound the
inner expectation uniformly over histories in $\mathcal G$.

For a final edge meeting only one selected vertex, its noise-averaged
weight is $\cosh\lambda-\mu x\sinh\lambda$ at an $A$ vertex and
$\cosh\theta-\mu\sinh\theta$ at a $B$ vertex. Here $x$ is the product
of the two opposite calibration-error signs. Its full-grid average at
an $A$ vertex is at most $1-\mathcal I(\mu b_0^2)$ on $\mathcal G$; at a $B$
vertex it is $1-\mathcal I(\mu)$. At most $\binom{s}{2}n$ locations meet two
or more selected vertices. The discrepancy of their combined weight
from the sum of the individual contributions is at most
$w+3e^\theta+4\le8w$. The complete population mean is consequently at
most
\[
 1-\frac{a \mathcal I(\mu b_0^2)+b \mathcal I(\mu)}{n}
       +\frac{8w\binom{s}{2}}{n^2}.
\]
All weights are positive and at most $w$. Removing $D\le m\le n^3/2$
locations increases their mean by at most $2wD/n^3$. Apply Maclaurin's
inequality to the free final locations; each forced final location
contributes at most $w$. Uniformly on $\mathcal G$, their conditional
moment is therefore at most
\begin{align*}
 w^{f_f}\exp\biggl\{&-\frac tn[a \mathcal I(\mu b_0^2)+b \mathcal I(\mu)]
                  +\frac{f_fs}{n}\\
 &+\frac{8w\binom{s}{2}t}{n^2}
                  +\frac{2wt(R+f_f)}{n^3}\biggr\}.
\end{align*}

Multiply this uniform bound by
$\mathbf1_{\mathcal G}e^{-\theta\sum_B D_v^{\rm past}}$ and integrate
over the past. The multiplier is nonnegative, so the indicator may now
be dropped. The remaining arrays no longer enter the calculation.
Conditional on $F$, the unforced past union is a uniform
$(R-f+f_f)$-subset of the $n^3-f$ unforced locations. Its true-degree
moment is at most
\begin{align*}
 w^{f-f_f}\exp\biggl\{&-\frac{Rb}{n}\mathcal I(\mu)
                 +\frac{(f-f_f)b}{n}\\
 &+\frac{8w\binom{b}{2}R}{n^2}
                 +\frac{2wRf}{n^3}\biggr\},
\end{align*}
by the same population calculation. Combine the two bounds, use
$t(R+f_f)+Rf\le m^2+mf$, and restore $\pi_F$.
\end{proof}

Let $m=\lceil cn\log n\rceil$, let $t/m\to\tau\in(0,1)$, and suppose
$b_0=b_n\to1$ on $\mathcal G$. For fixed $a,b,f$ and
$L_n=o(\log n)$, the lemma gives
\[
 \Pr\{F,\mathcal G,M_A\le0,D_B\le L_n\}
 \le\pi_F n^{-\tau\alpha a-\alpha b+o(1)},
 \qquad\alpha=c\mathcal I(\mu).
\]
For every fixed $s$-vertex set, summing its $2^s$ assignments to the two
kinds of constraint bounds membership in
$E\cup\{D\le L_n\}$ by $\pi_F n^{-\tau\alpha s+o(1)}$.
The sum of $\pi_F$ over all colors of $f$ fixed connecting edges is
exactly $(m)_f/(n^3)_f\le(m/n^3)^f$. Thus a growing number of colors
introduces no extra witness factor. The existing incidence-tree and
root-star counts now give $n^{1-\tau\alpha J+o(1)}$ for fixed witnesses
of $J$ exceptional vertices. Any fixed $J>1/(\tau\alpha)$ suffices
for the subsequent component argument. The calibration exception is
excluded once globally. At the default final fraction $\tau=1/2$, the
component exponent is $\alpha/2$.

\section{Using the target to reduce the acquisition budget}
\label{sec:sublinear-acquisition}

Theorem~\ref{thm:geometric} gives a smaller asymptotic reading budget.
This simpler construction provides a finite comparator and an amplification ingredient.

The original constraints can improve a noisy calibration. They can also
extend a calibration that covers only a small fraction of the variables.
Let
\[
 k_n=\min\left\{n,\left\lceil\frac{n}{\sqrt{\log(n+2)}}\right\rceil\right\}.
\]
Choose the first $k_n$ indices in each factor block and read their reference
fiber entries once. The budget is exactly $3k_n$ noisy binary readings.
Separate visits at the shared center have independent errors.

Randomly split the original sample into $H_1,H_2$ of sizes
$m_1=\lfloor m/2\rfloor$ and $m_2=m-m_1$. For an unmeasured variable,
use $H_1$ to select uniformly one observed triple whose other two variables
were measured. Multiplying its label by the two measured signs gives an
estimate of the remaining factor. If there is no such triple, use an
independent fair sign. Retain the measured signs on the seed variables.
One pass with a separate reservoir for each unmeasured variable performs
these selections. Each eligible triple belongs to only one reservoir.

Next, update all three factor blocks simultaneously using $H_2$. For
example, from the expanded arrays $A^0,B^0,C^0$, set
\[
 A^1_i=\operatorname{sgn}_{A^0_i}\!\sum_{(i,j,k)\in H_2}Y_{ijk}B^0_jC^0_k.
\]
Here $\operatorname{sgn}_a(0)=a$, so a zero field retains the previous
sign. Use the analogous formulas for the other blocks, always with the expanded
arrays on the right. Finally, apply the budgeted component search of
Theorem~\ref{thm:linear-acquisition} to the \emph{full} original sample,
starting from $T^1=A^1\otimes B^1\otimes C^1$.

\begin{theorem}[Exact optimization from sublinear source measurements]
\label{thm:sublinear}
Fix \(c>0\) and \(q,s\in(0,1/2)\), and let \(m=\lceil cn\log n\rceil\). The complete two-batch acquisition, expansion, update, and repair procedure just defined uses exactly \(3k_n=O(n/\sqrt{\log n})\) additional one-bit readings. It returns a feasible assignment on every input and attains \(\OPT(H,Y)\) with probability \(1-o(1)\) under the retained-source law.

Its total work is \(O((m+n)\log(n+2))\) word operations on every input. It does not use \(c,q\), or \(s\). The guarantee holds at every integer threshold supplied before or after the readings.
\end{theorem}

This appendix proves Theorem~\ref{thm:sublinear}. It also proves the logarithmic-degree upper bound for the same acquisition record at sufficiently large fixed density.

The local search in this theorem is necessary at low density: the measured
source itself can be a suboptimal fit. At sufficiently large fixed density,
the second update already recovers the source, and scoring that assignment
gives the exact optimum. The probability includes the independent random
split, reservoir choices, and fair signs used by the algorithm.

\subsection{Proof of Theorem~\ref{thm:sublinear}}
\label{app:sublinear-acquisition}\label{app:sublinear}

Write $N=n^3$, $p_j=m_j/N$, $\mu=1-2q$, and $\rho=1-2s$.
All calculations use the true reference-fiber gauge. Fix
$\rho_0\in(0,\rho)$. The mean of the $k_n$ sensor-error signs in each
seed block is at least $\rho_0$, except on an event of probability at most
\[
 3\exp[-k_n(\rho-\rho_0)^2/2]. \tag{S1}
\]

\paragraph{The first batch gives a positive bias at almost every variable.}
Analyze $H_1$ first under independent Bernoulli sampling at probability
$p_1$. An unmeasured variable has exactly $k_n^2$ eligible triples, and
the probability that at least one is observed is
\[
 a_n=1-(1-p_1)^{k_n^2}\longrightarrow1-e^{-c/2}.
\]
Conditional on the seed errors, the error sign of its expanded estimate
has mean
\[
 a_n\mu\bar\epsilon^{,b}_{\rm seed}
           \bar\epsilon^{,c}_{\rm seed}.
 \tag{S2}
\]
Uniform reservoir selection is used here: conditional on a nonempty
Bernoulli sample, the selected eligible grid address is uniform among all
$k_n^2$ possibilities. If the reservoir is empty, its fair sign contributes
zero to the mean.

The eligible triple sets are disjoint across \emph{all} unmeasured
variables, including variables in different blocks. Every eligible triple
has exactly one unmeasured endpoint. Thus, conditional on the seed errors,
the expanded error signs at unmeasured variables are mutually independent
under this Bernoulli law. Their means in (S2) are bounded below by a fixed
positive constant on the event (S1). Hoeffding's inequality, and
$k_n/n\to0$, show that all three full expanded blocks have mean at least
some fixed $b_0>0$, with failure probability
$\exp[-\Omega_{c,q,s}(k_n)]+\exp[-\Omega_{c,q,s}(n)]$.
For example, any fixed
\[
 0<b_0<\tfrac12(1-e^{-c/2})\mu\rho_0^2
\]
works for all sufficiently large $n$.

The event $|H_1|=m_1$ has probability $\Theta(m_1^{-1/2})$ under the
Bernoulli law. Conditioning therefore preserves this conclusion, with a
failure probability smaller than every fixed inverse power of $n$.
Conditional on that event, $H_1$ is precisely the required uniform
size-$m_1$ mask. Let $\mathcal C$ be the event that the three expanded
error means are at least $b_0$.

\paragraph{Bounds that retain the dependence on both batches.}
Conditional on the latent source, the whole first batch, its labels, the sensor readings, and the reservoir and fair-filling coins, the expanded error arrays are fixed.
The second batch is a uniform size-$m_2$ sample of the complement of
$H_1$, and its noise signs are fresh. Define
\[
 h_0=1-\sqrt{1-\mu^2b_0^4},\qquad
 \beta=\tfrac c2 h_0>0,\qquad
 \alpha=c\bigl(1-2\sqrt{q(1-q)}\bigr)>0.
\]
For a variable $v$, let $M_v$ be its second-batch field after multiplication
by the true factor sign. Put $E=\{v:M_v\le0\}$, so every actual error
after the second update belongs to $E$. Let
$D_v=\sum_{e\in H:e\ni v}\eta_e$ be its signed degree on the full target.

\begin{lemma}[Fields and weak degrees with prescribed locations]
\label{lem:joint-forced}
Fix disjoint vertex sets $A,B$ of sizes $a,b$, and disjoint sets of distinct
triples $F_1,F_2$ of sizes $r_1,r_2$, with all sizes fixed independently of $n$.
For every nonnegative $L_n=o(\log n)$,
\[
 \Pr\!\left(F_1\subseteq H_1,F_2\subseteq H_2,\mathcal C,
        M_v\le0\ (v\in A),\ D_v\le L_n\ (v\in B)\right)
 \le p_1^{r_1}p_2^{r_2}
       n^{-\beta a-\alpha b+o(1)}.
\]
The bound is uniform in the vertex and triple labels. For a fixed $t$-vertex
set $S$, membership in $E\cup\{D\le L_n\}$ under the same prescribed-location
event has probability at most $p_1^{r_1}p_2^{r_2}n^{-\gamma t+o(1)}$
on $\mathcal C$, where $\gamma=\min(\alpha,\beta)$.
\end{lemma}
\begin{proof}
Apply Lemma~\ref{lem:direct-mixed-moment} with two colors, final capacity
$m_2$, and calibration lower bound $b_0$. Its finite remainder is $o(1)$
at logarithmic density; the factor $e^{\theta bL_n}$ is $n^{o(1)}$.
The exact color probability is
$(m_1)_{r_1}(m_2)_{r_2}/(n^3)_{r_1+r_2}$, equal to
$p_1^{r_1}p_2^{r_2}(1+o(1))$ when the fixed capacities are nonzero.
Summing the $2^t$ possible assignments of constraints proves the union bound.
The proof keeps the past-degree exponential until after the conditional
final-field estimate, so it accommodates dependence between calibration,
true degrees, and final estimation errors.
\end{proof}

\paragraph{Localization and exact optimization.}
Sum Lemma~\ref{lem:joint-forced} over the rooted incidence-tree witnesses used in
Theorem~\ref{thm:linear-acquisition}. Sum also over the two possible
colors of each witness edge. Since $p_1+p_2=m/N$, the colored inclusion
weights have the same total as in the one-batch count. The probability of
a connected $t$-vertex set in $E\cup\{D\le L_n\}$ is at most
$n^{1-\gamma t+o(1)}$, on $\mathcal C$. The analogous root-star count
shows that no vertex has $t$ distinct neighbors in $E$, with the same
bound. Fix an integer $J>1/\gamma$. Consequently these components and
the number of erroneous neighbors of every vertex are bounded by $J$,
with probability $1-o(1)$. The exceptional event $\mathcal C^c$ is
excluded once, globally, rather than once per witness.

The original localization argument supplies an optimizer whose difference
from the true source is supported on $\{D\le L_0\}$ for a fixed $L_0$.
Also, the maximum pair-codegree is at most two with high probability.
The flip set from the updated assignment to that optimizer is therefore
contained in $E\cup\{D\le L_0\}$ and has bounded components. One-flip
optimality shows that every vertex of this difference set has updated
signed degree at most $4(J-1)$. If a vertex is outside $E$, its updated
degree differs from its true degree by at most $4J$, because there are
fewer than $J$ erroneous neighbors and at most two shared triples per
pair. Hence the observable weak set at cutoff $\tau_*=4J$ satisfies
\[
 W_{\tau_*}\subseteq E\cup\{D\le8J\}.
\]
It contains the difference set and has bounded components. Exact search
in its components therefore attains the global optimum on the original
sample. The budgeted sweep reaches this fixed stage for all sufficiently
large $n$ on the stated events, and retaining the best feasible score
preserves the result at every later stage.

The two initialization passes take $O(m+n)$ word operations; the budgeted
search has the stated $O((m+n)\log(n+2))$ bound. Reservoir sampling uses
only target addresses and private random coins. Scores are exact integer
sums. The operation bound uses the usual random-access model with
uniform random-word draws. For an implementation using fair bits, cap
each rejection sampler at $8\lceil\log_2(n+m+2)\rceil$ attempts and
return the all-positive feasible assignment if that cap is reached.
Each attempt accepts with probability at least $1/2$, and there are
$O(m+n)$ such draws, so the added failure probability is $o(1)$.
The added $O(m\log(n+m+2)+n)$ word work fits the stated bound because
$m\le n^3$. Across all erasure-propagation or majority stages, the target partition and
all reservoirs together require $O(m+n)$ random-integer draws: each
observation belongs to one batch and at most one erasure reservoir.
Initializing the arrays costs $O(n)$ per round. The supplied NumPy research
implementation uses random-word sampling; the explicit rejection cap
describes a separate fair-bit implementation of this mathematical algorithm.
The cascade uses the global attempt budget in
Appendix~\ref{app:global-fair-bit} to preserve its sharper error exponent.
With random-access storage, charging individual random bits and a conservative
quadratic bit cost for exact word arithmetic gives
$O((m+n)\log(n+2)\log^2(n+m+2))$ arithmetic-bit work.
Each capped sampling attempt uses $O(\log(n+m+2))$ fair bits.
Appendix~\ref{app:erasure-experiments} specifies the cost model; the theorem's
sharper displayed bound counts word operations.

\subsection{A low-degree estimator using this same record}\label{app:sublinear-degree}
For any fixed $A>0$, choose $c$ large enough that $\beta>A+1$ and the
source-optimality error is $O(n^{-A})$. The preceding pilot failure is
smaller than every inverse power. Applying Lemma~\ref{lem:joint-forced} with one field constraint and taking a union over vertices shows that the second update recovers the entire source with
failure $O(n^{-A})$.

Conditional on the mask, split, and reservoir coins, each expanded
coordinate is a polynomial of degree at most three in target and sensor
signs. Each second-stage field is a sum of products of degree at most
seven. A sign function of a sum of $d$ binary products has a multilinear
representation of degree at most $d$ in those products. If every
second-batch degree is at most $\Delta$, the updated coordinates have
degree at most $7\Delta+3$ when a zero field retains the previous sign.
Indeed, the tie indicator has degree at most $7\Delta$ and the previous
coordinate has degree at most three. Their original-sample score therefore
has degree at most $21\Delta+10$.

A hypergeometric tail bound makes $\Delta\le C_\Delta\log n$ hold with
probability tending to one, for a sufficiently large fixed
$C_\Delta=C_\Delta(c)$. Set the estimator to zero on the other masks.
Compose the normalized updated score with a fixed polynomial approximating
the sign function in Gaussian mean square, as in
Appendix~\ref{app:upper}. Choose $A$ larger than its degree plus two to
control its bounded-polynomial growth on reconstruction failures. The
result has total degree $O(\log n)$ and the claimed risk. For a finite fair-bit implementation, couple the private choices to the
ideal uniform draws and set the polynomial estimator to zero whenever a
rejection-attempt cap is exhausted, before composing the score polynomial.
That event depends only on the mask and private coins, not on any observed
sign. Its contribution to squared risk is its $o(1)$ probability; it does
not need the arbitrarily large failure exponent $A$ used to control the
polynomial's growth. On its complement the ideal choices and their error
bound apply. Averaging the coefficients over the private split, reservoir,
and tie coins then yields a polynomial in just the observed signs;
conditional Jensen cannot increase the risk or its degree. The independent-source argument allows arbitrary
coefficients depending on the mask and replacement transcript, and so
continues to apply to the same $3k_n$ acquisition schedule.

\section{Iterated amplification uses fewer source readings}
\label{app:bootstrap}

Theorem~\ref{thm:geometric} gives a smaller asymptotic reading budget.
This simpler construction provides a finite comparator and an amplification ingredient.

The first target batch need not produce a calibration with constant bias.
A sequence of fresh batches can amplify a smaller bias before the final
localization step. This reduces the number of source readings while keeping
the original objective fixed. Fresh-sample majority refinement is an
established ingredient in planted recovery \citep{basu2025}. The result here
accounts for the initial noisy fiber measurements, the dependence created
within each update, and the final exact optimization of the noisy sample.

\begin{theorem}[Acquisition reduced by repeated refinement]\label{thm:bootstrap}
For every fixed rational \(a\in(1/2,3/4)\), fixed \(c>0\), and fixed
\(q,s\in(0,1/2)\), the displayed algorithm uses exactly \(3k_{n,a}\)
additional bits and attains \(\OPT(H,Y)\) with probability \(1-o(1)\).
It returns a feasible assignment on every input and uses
\(O_a((m+n)\log(n+2))\) word operations. It requires \(a\), but not
\(c,q,s\). For \(a=2/3\), four batches suffice and the reading budget is
\(O(n/(\log n)^{2/3})\).
\end{theorem}

\paragraph{Schedule for Theorem~\ref{thm:bootstrap}.}
Fix a rational $a\in(1/2,3/4)$, $c>0$, and $q,s\in(0,1/2)$.
Set
\[
 \ell_n=\lceil\log_2(n+2)\rceil,\qquad
 d_n=\lceil\ell_n^a\rceil,\qquad k_n=\lceil n/d_n\rceil.
\]
Under the coupled-source law with $m=\lceil cn\log n\rceil$, exactly
$3k_n=O(n/(\log n)^a)$ additional noisy binary readings suffice to return
an exact maximizer of the original empirical objective with probability
$1-o(1)$. The algorithm does not require $c,q$, or $s$. For fixed $a$, it
uses $O_a((m+n)\log(n+2))$ word operations and returns a feasible assignment
on every input.

Let $r$ be the smallest nonnegative integer satisfying
$2^r(3-4a)>1$, and put $B=r+2$. Partition the original observations
uniformly into $B$ batches whose sizes differ by at most one. Use the
first batch to expand the $k_n$ measured signs in each block by the
reservoir rule in Appendix~\ref{app:sublinear}. Apply $r$ simultaneous
majority updates, each using the next batch and the preceding arrays.
Apply one final simultaneous update using the last batch. At every zero
field retain the preceding sign. Run the same budgeted component search
on the full original sample, starting from the final arrays.

In particular, $a=2/3$ uses four batches: expansion, two amplification
updates, and a final update before exact local optimization. Its acquisition
budget is $O(n/(\log n)^{2/3})$. Any fixed exponent below $3/4$ is allowed
by choosing a rational $a$ sufficiently close to $3/4$. The endpoint
$3/4$, an optimal acquisition rate, and a useful finite-size onset are not
consequences of this theorem. More batches reduce the observations available
to each update, so this improvement in acquisition order does not itself
predict better performance at the sizes in the experiments.

The schedule has an exact integer implementation. If $a=u/v$ in lowest
terms, $d_n$ is the smallest positive integer $d$ with
$d^v\ge\ell_n^u$. Also $\ell_n$ is the binary length of $n+1$.
Thus no real-number oracle is required to choose the measurement addresses.
Batch sizes, zero fields, empty reservoirs, and the component-search budget
are handled by the stated rules even at sizes for which the asymptotic
guarantee gives no information.

\subsection{A fresh batch amplifies a weak calibration}

The next lemma gives both a finite conditional bound and the asymptotic
amplification rule. Its concentration statement concerns a whole block of
updated signs. It therefore does not assume that different updated
coordinates are independent.

\begin{lemma}[Conditional amplification]
\label{lem:amplification}
For analysis, condition on the latent source, arbitrary previously inspected triples and their labels, and the private coins already used, with
at most $R<n^2$ distinct locations removed from the grid. The current arrays
must be measurable with respect to this revealed past, the sensor readings,
and independent private coins used to construct those arrays. Suppose that their three mean error
signs in the true gauge are at least $b>0$. Draw a uniform batch of $t$
distinct locations, where $0\le t$ does not exceed the number remaining,
with fresh independent target errors of mean
$\mu=1-2q$. Perform one simultaneous update, retaining the previous sign
at a zero field. Let $Z$ be the mean updated error sign in any one block.
Write
\[
 \lambda_-=\frac{t(n^2-R)}{n^3-R},\qquad
 \lambda_+=\frac{tn^2}{n^3-R},\qquad
 d_-=\lfloor\lambda_-/2\rfloor,\qquad
 d_+=\lceil2\lambda_+\rceil,
\]
and suppose $d_-\ge1$ and
$\delta_-:=\mu[b^2-R/(n^2-R)]\ge0$. Then
\begin{align}
 \E[Z\mid\text{past}]\ \ge\ &
 \frac18\min\{\delta_-\sqrt{d_-},1\}
 -\frac{2}{\sqrt{d_-}}
 -\frac{d_+^2}{n^2-R}
 -4e^{-\lambda_-/8}. \label{eq:finite-amplification}
\end{align}
For every $u>0$, without either of these extra numerical conditions,
\begin{equation}
 \Pr\{Z\le\E[Z\mid\text{past}]-u\mid\text{past}\}
 \le \exp\!\left[-\frac{n^2u^2}{136t}\right]
 \label{eq:amplification-concentration}
\end{equation}
when $t\ge1$; a batch of size zero is deterministic.

Consequently, if $R=O(n\log n)$, $t\sim\kappa n\log n$ for fixed
$\kappa>0$, and $b=b_n$ satisfies $b_n\sqrt{\log n}\to\infty$, there is
a constant $C=C(\kappa,q)>0$ such that, uniformly in the conditioned past,
\[
 \E[Z\mid\text{past}]
 \ge C\min\{b_n^2\sqrt{\log n},1\}
\]
for all sufficiently large $n$.
\end{lemma}

\begin{proof}
First consider $d$ independent signs of common mean $\delta\in[0,1]$.
Let $M_d(\delta)$ be the mean sign of their sum, assigning mean zero to a
tie. For odd $d=2h+1$, differentiation of the binomial tail gives
\[
 M'_{2h+1}(\delta)
 =(2h+1)\binom{2h}{h}4^{-h}(1-\delta^2)^h.
\]
For $h\ge1$, use $\binom{2h}{h}4^{-h}\ge1/(2\sqrt h)$ and
$(1-1/d)^h\ge1-h/d\ge1/2$. The derivative is therefore at least
$\sqrt d/4$ on $0\le\delta\le d^{-1/2}$; $d=1$ follows directly.
Integration and monotonicity give
$M_d(\delta)\ge\frac14\min\{\delta\sqrt d,1\}$ for odd $d$.
For even $d$, the binomial identity $M_d=M_{d-1}$ and the bound
$\Pr\{\text{tie}\}\le\sqrt{2/d}$ give, for either deterministic tie rule,
\begin{equation}
 \E\operatorname{sgn}_{\pm1}\!\left(\sum_{j=1}^d X_j\right)
 \ge\frac18\min\{\delta\sqrt d,1\}-\frac2{\sqrt d}.
 \label{eq:majority-bias}
\end{equation}

Fix a vertex. Among its remaining incident locations, the mean product
of the two current error signs is at least
$b^2-R/(n^2-R)$. Conditional on its batch degree $D=d$, its locations
are a uniform $d$-subset of that incident population, of size at least
$n^2-R$. Couple their ordered addresses to independent draws from that
population. The two address lists disagree with probability at most
$d(d-1)/(2(n^2-R))$. Fresh independent noise makes the independent
summands binary signs of mean at least $\delta_-$. Since a sign has
range two, the coupling costs at most $d(d-1)/(n^2-R)$ in its mean.

The degree is hypergeometric, with mean between $\lambda_-$ and
$\lambda_+$. The usual hypergeometric Chernoff bounds imply
\[
 \Pr\{D\notin[d_-,d_+]\}\le2e^{-\lambda_-/8}.
\]
On that interval, apply~\eqref{eq:majority-bias} and the coupling bound;
outside it use the lower bound $-1$. This proves
\eqref{eq:finite-amplification} for each vertex and hence for its block
average. The bound uses the total number of removed locations; it needs
no assumption about how the past mask is distributed among rows.

For concentration, expose an ordered sample of the $t$ distinct locations
and then its $t$ independent noise signs. Changing one location while
holding all other contributions fixed changes at most two coordinates
in a chosen updated block, so it changes $Z$ by at most $4/n$.
For two possible values of the next sampled location, couple the remaining
ordered samples by exchanging those two locations. At most two positions
differ, giving a Doob-martingale increment bounded by $8/n$.
Exposing one noise sign gives an increment bounded by $2/n$.
The sum of the squared increment bounds is at most $68t/n^2$.
Azuma's inequality \citep{azuma1967} therefore gives~\eqref{eq:amplification-concentration}.
In particular, no maximum-degree bound is needed for this concentration
argument: a changed observation enters only one field in each block.

Finally $\lambda_\pm\sim\kappa\log n$, and
$R/(n^2-R)=O(\log n/n)=o(b_n^2)$. The leading term in
\eqref{eq:finite-amplification} has order
$\min\{b_n^2\sqrt{\log n},1\}$, whereas the tie term has order
$(\log n)^{-1/2}$. Their ratio tends to infinity because
$b_n^2\log n\to\infty$. The other two error terms are smaller still.
This proves the last assertion.
\end{proof}

\subsection{From the amplified calibration to an exact optimizer}

\begin{proof}[Proof of Theorem~\ref{thm:bootstrap}]
Let the batches be $H_0,\ldots,H_{B-1}$, with sizes
$m_j=m/B+O(1)$, and put $p_j=m_j/n^3$.
The first-batch pilot calculation in Appendix~\ref{app:sublinear}
applies with the new seed size. In its Bernoulli version the probability
that an unmeasured vertex has an eligible triple is
\[
 1-(1-p_0)^{k_n^2}=
 \begin{cases}
 \Theta(1),&a=1/2,\\
 \Theta((\log n)^{1-2a}),&a>1/2.
 \end{cases}
\]
Conditional on the seed errors, the reservoirs at different unmeasured
vertices use disjoint triples. Hoeffding's inequality, followed by
conditioning the Bernoulli mask on $|H_0|=m_0$, shows that the expanded
block means are all at least
$C_0(\log n)^{-\nu_0}$, where $C_0>0$ and $\nu_0=2a-1<1/2$,
except with probability smaller than every fixed inverse power of $n$.
Here the unconditioned error bounds are
$\exp[-\Omega(k_n)]+\exp[-\Omega(n(\log n)^{-2\nu_0})]$;
the fixed-size conditioning costs only $O(\sqrt{m_0})$.

For the amplification stages put
\[
 \nu_j=\frac12-2^j\left(\frac32-2a\right).
\]
As long as $\nu_j\ge0$, a block bias bounded below by
$C_j(\log n)^{-\nu_j}$ satisfies the hypothesis of
Lemma~\ref{lem:amplification}. That lemma and its concentration bound
give, in the next stage, block bias at least
$C_{j+1}\min\{(\log n)^{-\nu_{j+1}},1\}$, simultaneously in all three
blocks, for some fixed $C_{j+1}>0$. The conditional failure probability
is $\exp[-n/(\log n)^{O_a(1)}]$, with constants also depending on
$c,q,s$. Our choice of $r$ makes $\nu_r<0$. Thus, before the final batch,
all three mean error signs exceed a fixed $b_*>0$ on an event
$\mathcal C$ whose failure probability is smaller than every inverse
power of $n$. This induction conditions on the latent source and only the batches and private choices already used.
The next batch remains uniform in their complement and its errors are
independent of that past. Formally, the uniformly colored target can be
generated by sequential sampling without replacement. The analysis reveals
the source, completed batch locations and labels, the sensor record, and the independent reservoir, filling or tie coins already used. It does not condition on unused locations or raw
random bits defining their partition. The current arrays depend only on
these revealed quantities, even though the complete target mask is supplied
to the algorithm.

We spell out why the dependence created in these rounds is compatible
with exact optimization. In the final batch, let $M_v$ denote the field
in true coordinates and $E=\{v:M_v\le0\}$. All actual final errors
belong to $E$. Define
\[
 h_*=1-\sqrt{1-\mu^2b_*^4},\qquad
 \beta=\frac cB h_*>0,\qquad \alpha=ch(q)>0.
\]
Prescribe finitely many distinct triples $F_j$ in each batch, with all
sets disjoint, and fix disjoint vertex sets $A,D$ of respective sizes
$u,w$. Their sizes are fixed as $n\to\infty$. Repeating the final-field
moment bound in Lemma~\ref{lem:joint-forced}, conditional on the whole
preceding history, gives
\[
 \Pr\{F_j\subseteq H_j\ \forall j,\ M_v\le0\ (v\in A),\mathcal C\}
 \le\left(\prod_jp_j^{|F_j|}\right)n^{-\beta u+o(1)}.
\]
Indeed, at $\lambda=\operatorname{arctanh}(\mu b_*^2)$, the full-grid
mean moment is at most $1-h_*u/n+O_u(n^{-2})$. Removing all earlier
batches changes it by $O(m/n^3)$, whose cost over the final batch is
$O(m^2/n^3)=o(1)$. The forced final locations have bounded moment factors.
Integrating over the previous history retains its prescribed-color
inclusion probability, bounded by a constant times the corresponding
product of $p_j$'s.

For the true signed degrees on the full original target, and any
$L_n=o(\log n)$, the corresponding bound is
\[
 \Pr\{F_j\subseteq H_j\ \forall j,\ D_v\le L_n\ (v\in D)\}
 \le\left(\prod_jp_j^{|F_j|}\right)n^{-\alpha w+o(1)}.
\]
The exact colored inclusion probability is
$\pi_F=\prod_j(m_j)_{|F_j|}/(n^3)_{\sum_j|F_j|}$.
Lemma~\ref{lem:direct-mixed-moment} bounds these constraints jointly by
$\pi_F n^{-\beta u-\alpha w+o(1)}$ on $\mathcal C$.
A union over the two constraint choices at each vertex gives
\[
 \Pr\{F_j\subseteq H_j\ \forall j,\ S\subseteq E\cup\{D\le L_n\},
              \mathcal C\}
 \le\left(\prod_jp_j^{|F_j|}\right)n^{-\gamma|S|+o(1)},
 \qquad\gamma=\min(\alpha,\beta)>0.
\]
All remainders are uniform in the fixed vertex and triple labels.

Summing over the finitely many batch colors of each incidence-tree
witness replaces their weights by $\sum_jp_j=m/n^3$. The component and
neighbor bounds, localization of a true optimum, and observable weak-set
argument in Appendix~\ref{app:sublinear} now apply without change.
The event $\mathcal C^c$ is removed once globally. Exact component search
therefore recovers an original-sample maximizer with probability $1-o(1)$.

There are $B=O_a(1)$ initialization and update passes. Their total cost
is $O_a(m+n)$; the prescribed component-search cap gives the claimed
worst-case word bound. No stage uses $c,q$, or $s$. All fields and scores
are integer sums, and the integer-power definition of $d_n$ makes the
acquisition schedule exact. The same fair-bit implementation convention
as in Appendix~\ref{app:sublinear} handles the private random choices.
\end{proof}

\subsection{A polynomial estimator on the smaller record}
\label{app:bootstrap-degree}

\begin{proposition}[Same-record polynomial comparison]
\label{prop:bootstrap-degree}
Fix a rational $a\in(1/2,3/4)$, $q,s\in(0,1/2)$, and $\varepsilon>0$.
For every sufficiently large fixed $c$, the retained record of
Theorem~\ref{thm:bootstrap} admits a polynomial estimator of the balanced
optimum-threshold decision with limiting mean squared error at most
$\varepsilon$ and total label degree
$O((\log n)^{r+1})$, where $r$ is its number of amplification rounds.
Its coefficients may depend on the mask and fixed model parameters.
With an independent replacement at exactly the same reading addresses,
the target-degree lower bound of Theorem~\ref{thm:main} still applies.
\end{proposition}

\begin{proof}
The strict inequality $\nu_r<0$ has a stronger consequence than constant
bias. Immediately before the last amplification update, each block bias
is at least $C(\log n)^{-\nu_{r-1}}$, with $\nu_{r-1}<1/4$.
A one-vertex exponential-moment bound for the next fresh batch gives error
probability at most
$\exp[-C'(\log n)^{1-4\nu_{r-1}}]=o(1)$, uniformly in the good past.
To obtain this bound, average the noise moment at
$\lambda=\operatorname{arctanh}(\mu b^2/2)$ over its remaining incident
population and use $R/(n^2-R)=o(b^2)$. Its full-grid mean is at most
$1-C''b^4/n$, and Maclaurin's inequality applies to the sampled batch.
Thus each block's expected error-sign mean tends to one.
Equation~\eqref{eq:amplification-concentration} shows that all three means
are at least $1/2$ before the final batch, except on an event smaller
than every inverse power of $n$.

We may consequently take $b_*=1/2$ in the final-field bound. Given any
fixed $A>0$, choose $c$ large enough that
\[
 \frac cB\left(1-\sqrt{1-\mu^2/16}\right)>A+2
\]
and the source-optimality bound~\eqref{eq:optbound} is $O(n^{-A})$.
A union bound then shows that the final updated assignment equals the
source with failure $O(n^{-A})$. No component search is needed for this
large-density comparison.

Condition on the mask, the split, the reservoirs, and all private fair
coins. Each expanded coordinate is a polynomial of degree at most three
in the target and acquired signs. On masks of maximum degree at most
$\Delta$, if the coordinates before an update have degree at most $d_j$,
then those after it have degree at most
\[
 d_{j+1}\le(2\Delta+1)d_j+\Delta.
\]
Indeed, a field is a sum of at most $\Delta$ binary products of degree
$2d_j+1$. Its sign with mean-zero ties, and its tie indicator, each have a
multilinear representation of degree at most $\Delta$ in those products.
Multiplying the tie indicator by the previous coordinate adds at most
$d_j$. After $r+1$ updates, the score on the original sample has degree
$O(\Delta^{r+1})$.

Truncate to masks with $\Delta\le C_\Delta\log n$ and assign estimator
zero otherwise; hypergeometric degree tails make the excluded probability
$o(1)$ for a suitable constant $C_\Delta$. Compose the normalized score
with a fixed polynomial approximating the sign of a standard Gaussian,
as in Appendix~\ref{app:upper}. If that polynomial has degree $k$, choose
$A>k+2$ above. Its squared growth on reconstruction failures is
$O(1+m^k)$, so their contribution vanishes. Gaussian convergence gives
the prescribed limiting risk, and the total degree is
$O((\log n)^{r+1})$. For a finite fair-bit sampler, couple its choices to ideal uniform draws
and set the estimator to zero on any rejection-cap exhaustion, before
applying the score polynomial. This event depends only on the mask and
private coins. Its squared-risk contribution is its $o(1)$ probability;
no arbitrarily large inverse-polynomial bound on this event is required.
On its complement the ideal-choice reconstruction estimate applies.
Averaging coefficients over the private algorithmic coins preserves the
degree and cannot increase squared error.
The independent-record projection bound allows arbitrary dependence of
coefficients on the replacement transcript, and therefore applies to
this same smaller acquisition schedule.
\end{proof}

For $a=2/3$, the degree in Proposition~\ref{prop:bootstrap-degree} is
$O((\log n)^3)$. The original two-batch schedule retains its sharper
$O(\log n)$ upper bound in Appendix~\ref{app:sublinear-degree}.

\section{Preserving unknown entries reduces the acquisition cost}
\label{app:erasure}

The initial readings need not be extended to a sign at every coordinate.
We retain an erasure at a coordinate that has no usable observation.
The active set then grows by the familiar hypergraph bootstrap rule: an
edge activates its last inactive endpoint \citep{morrison2021}. The contribution needed here is
to control the signs carried by that growing set, despite noisy readings
and observations, and then connect the resulting calibration to exact
optimization of the original empirical objective.

\begin{theorem}[A fixed number of propagation rounds]
\label{thm:erasure-fixed}
Fix a rational \(a=u/v\in(0,1)\). Let
\[
 \begin{gathered}
 \ell_n=\lceil\log_2(n+2)\rceil,\quad k_n=\lceil n/d_n\rceil,\\
 d_n=\min\{d\ge1:d^v\ge\ell_n^u\},
 \end{gathered}
\]
and let \(r\) be the smallest integer satisfying \(2^r(1-a)\ge1\).
Read \(k_n\) entries per fiber and divide the target uniformly into
\(r+1\) balanced batches. Use \(r\) batches for simultaneous propagation.
Fill remaining unknowns with independent fair signs, make one simultaneous
majority update on the last batch, and apply the full-target repair below.
For every fixed \(c>0\) and \(q,s\in(0,1/2)\), this uses exactly
\(3k_n=O(n/(\log n)^a)\) readings and attains the original optimum with
probability \(1-o(1)\). It uses no \(c,q,s\), has an all-input
\(O_a((m+n)\log(n+2))\) word cap, and always returns a feasible assignment.
\end{theorem}

\paragraph{One simultaneous extension.}
An active entry carries a sign; an inactive entry is zero. Keep every entry
that was active at the beginning of the batch. An observed triple is eligible
when exactly one endpoint was inactive at that time. At each inactive
vertex, choose one eligible triple uniformly, if any, and set its sign to
the observed label times the two old active signs. With no eligible triple,
retain the erasure. Commit all new entries together after processing the
batch. A triple belongs to at most one reservoir, so a single pass with
independent reservoir sampling implements this rule.

\begin{lemma}[Conditional extension of a partial calibration]
\label{lem:erasure-extension}
For analysis, condition on the latent source, a revealed history containing exactly $R$ distinct target locations with their labels, the sensor readings and the private choices already used. The three partial sign arrays are measurable with respect to these data.
In true coordinates, suppose every block has active fraction at least
$p>0$ and mean error sign among active entries at least $r>0$.
Put $\mu=1-2q$ and
\[
 \begin{gathered}
 N=n^3-R,\qquad M_-=n^2p^2-R,\\
 \delta=\mu\left(r^2-\frac R{M_-}\right),\qquad
 \theta=1-\exp\left(-\frac{tM_-}{N}\right),\\
 \bar A=p+(1-p)\theta.
 \end{gathered}
\]
Assume $M_->0$, $\delta>0$, and $1\le t\le N$. Take a uniform
$t$-subset of the remaining locations, with fresh independent target
errors, and perform one simultaneous extension. In block $i$, let $A_i'$
be the updated active fraction and $Z_i'$ its signed mass divided by $n$,
with erasures contributing zero. Conditional on the history,
\[
 \E Z_i'\ge\delta\E A_i',\qquad \E A_i'\ge\bar A.
\]
For every tolerance $u>0$,
\begin{align}
 \Pr\{Z_i'<\E Z_i'-u\mid\text{past}\}
 &\le e^{-n^2u^2/(144t)},\label{eq:erasure-sign-concentration}\\
 \Pr\{|A_i'-\E A_i'|>u\mid\text{past}\}
 &\le2e^{-n^2u^2/(32t)}.\label{eq:erasure-support-concentration}
\end{align}
On the complementary events, $A_i'\ge\bar A-u$ and
$Z_i'/A_i'\ge\delta-2u/p$, since activity increases and $A_i'\ge p$.
In particular, with conditional probability at least
\begin{equation}
 1-9\exp\left(-\frac{n^2\delta^2\bar A^2}{2304t}\right),
 \label{eq:erasure-success}
\end{equation}
every block has active fraction at least
$(1-\delta/4)\bar A\ge3\bar A/4$ and mean error sign among active
entries at least $\delta/2$. Active sets also increase deterministically.
For $t=0$ the arrays are unchanged.
\end{lemma}

\begin{proof}
Fix an inactive vertex. If the active sets in the other blocks are
$S_2,S_3$, its eligible grid contains $M_0=|S_2||S_3|$ locations before
removing the past. Their mean product of old error signs is at least $r^2$.
Deleting at most $R$ locations leaves at least $M_-$ and changes that mean
to at least $r^2-R/M_-$. Indeed, deleting $d$ signs from a population of
size $M_0$ and mean $z\ge r^2$ leaves mean at least
$z-d(1-z)/(M_0-d)\ge r^2-R/M_-$.

Let $M_v$ be the number of eligible locations still available. The
probability of a nonempty reservoir is
\[
 \theta_v=1-\frac{(N-M_v)_t}{(N)_t}
 \ge1-e^{-tM_v/N}\ge\theta.
\]
The ratio is zero when avoidance is impossible. Conditional on a nonempty
reservoir, its uniformly selected address is uniform over those $M_v$
locations. Its new error sign therefore has mean at least $\delta$.

Write $A_i'$ for the active fraction after extension in a specified block,
and $Z_i'$ for the sum of its error signs divided by $n$, with zero at
erasures. Old active entries have mean at least $r\ge\delta$.
Consequently, under the actual fixed-size conditional sampling law,
\begin{equation}
 \E Z_i'\ge\delta\E A_i',\qquad \E A_i'\ge\bar A.
 \label{eq:erasure-means}
\end{equation}
The means here do not assume independent stored signs.

For concentration, represent the batch by an ordered sample without
replacement and attach independent noise signs and independent continuous
priorities to its positions. Selecting the smallest priority in each
reservoir gives the same independent uniform choices conditional on the
mask. This is a representation of the sampler, not a real-arithmetic
requirement on the algorithm.

Expose locations, then noise signs, then priorities. Changing one sampled
location can affect only the old and new locations' reservoirs. It changes
$Z_i'$ by at most $4/n$ and $A_i'$ by at most $2/n$. Coupling two possible
values of the next sampled location by an exchange changes at most two
positions of a completed ordered sample. The corresponding martingale
increments are thus at most $8/n$ and $4/n$, respectively. A noise or
priority change affects at most one reservoir, giving a $2/n$ increment
for $Z_i'$ and none for $A_i'$. The squared increment sums are at most
$72t/n^2$ and $16t/n^2$. Azuma's inequality gives
\eqref{eq:erasure-sign-concentration}--\eqref{eq:erasure-support-concentration}.
Their complementary events imply
$Z_i'\ge\delta A_i'-(1+\delta)u\ge\delta A_i'-2u$,
which proves the ratio bound because $A_i'\ge p$.
Take $u=\delta\bar A/4$. On these events,
$A_i'\ge(1-\delta/4)\bar A$ and
\[
 Z_i'\ge\delta\E A_i'-u
       \ge\frac\delta2(\E A_i'+u)
       \ge\frac\delta2 A_i',
\]
because $0<\delta\le1$. Union over the three blocks proves
\eqref{eq:erasure-success}.
\end{proof}

The lemma uses the exact number of removed locations. If $R$ is only an
upper bound, replacing the denominator $N$ in $\theta$ by $n^3$ gives a
valid conservative statement. The one-unknown rule and simultaneous
commitment are essential to the stated calculation.

\subsection{Every fixed acquisition exponent below one}

\begin{proof}[Proof of Theorem~\ref{thm:erasure-fixed}]
Fix the rational $a\in(0,1)$, and let $r$ be the smallest integer with
$2^r(1-a)\ge1$. There are $r$ extension batches and one final refinement
batch, all of size $m/(r+1)+O(1)$. Write $L=\log n$,
$\mu=1-2q$, and $\rho=1-2s$.
The initial active fraction is $p_0=\Theta(L^{-a})$.
Sensor concentration makes its conditional error bias at least
$\rho/2$ in every block, except with probability at most
$3e^{-k\rho^2/8}$.

Set
\[
 \nu_j=1-2^j(1-a),\qquad r_0=\rho/2,\qquad
 r_{j+1}=\mu r_j^2/4.
\]
There are constants $C_j>0$, which we may take at most $1/2$, such that
after $j$ extensions every block has active fraction at least
$C_jL^{-\max(\nu_j,0)}$ and conditional bias at least $r_j$, with
superpolynomial success probability. To prove this by induction, first
use monotonicity of activity and the bound $R=O(nL)$ to obtain
\[
 \frac R{n^2p_0^2-R}=O(L^{2a+1}/n)=o(1).
\]
For fixed $j$ this loss is eventually at most $r_j^2/2$, so the lemma's
$\delta$ is at least $\mu r_j^2/2$. Each batch has
$t\sim[c/(r+1)]nL$, and its $\theta$ is bounded below by a positive
constant times $\min\{Lp^2,1\}$. Thus the next activity is at least a
constant times $\min\{Lp^2,1\}$, while its conditional bias is at least
$r_{j+1}$. This is the recurrence $\nu_{j+1}=2\nu_j-1$, with negative
exponents replaced by constant activity. The conditional failure in each
round is at most $\exp[-\Omega(n/L^{2a+1})]$, uniformly in the good past.

Our round choice gives $\nu_r\le0$, hence positive constant activity and
conditional bias. Equality suffices: it produces constant occupancy rather
than an asymptotically full active set. Fill residual erasures with
independent fair signs. Hoeffding's inequality then gives full-block
error-sign means at least a fixed $b_*>0$, except with
$\exp[-\Omega(n)]$ probability. Call the resulting calibration event
$\mathcal C$; its complement is smaller than every inverse power of $n$.

The analysis reveals the latent source, completed batch locations and labels, sensor readings, and reservoir or filling choices already used. It does not condition on unused target
locations or the raw coins defining their partition. Each stage ignores
future batch data, so its arrays are measurable with respect to this
filtration. The last batch is uniform in the complement of the revealed
extension batches, with fresh errors. The final-field argument of
Lemma~\ref{lem:joint-forced} applies with
\[
 \beta=\frac c{r+1}\left(1-\sqrt{1-\mu^2b_*^4}\right)>0,
 \qquad \alpha=ch(q)>0.
\]
It retains the probability of every fixed collection of prescribed colored
locations. The true-degree bound on the full target retains the same
factor. Lemma~\ref{lem:direct-mixed-moment} combines their constraints with exponent
$\gamma=\min(\alpha,\beta)>0$; summing the exact color probabilities replaces each
witness edge's weight by $m/n^3$. The component and neighbor bounds in
Appendix~\ref{app:sublinear} therefore apply. Intrinsic localization of an
empirical optimizer and the observable weak-set argument show that the
capped full-target search reaches an exact optimizer with probability
$1-o(1)$. This step resolves the empirical improvements as well as the
estimation errors.

Each extension processes its assigned observations once and initializes
$O(n)$ entries, for total cost $O(m+rn)$. The final update and the existing
repair cap give $O_a((m+n)\log(n+2))$ word operations on every input.
The seed size is computed by exact integer powers from the fixed rational
$a$, so neither model parameters nor a real-number oracle are needed.
The finite fair-bit convention of Appendix~\ref{app:sublinear} handles
uniform reservoir choices and returns a feasible incumbent on its rare
sampling-cap failure.
\end{proof}

\subsection{A limitation of local support propagation}
\label{app:erasure-converse}

A seed budget below $n/\log n$ cannot produce extensive activation
under this rule. The statement does not constrain algorithms allowed
to infer unknown signs by arbitrary computations.

\begin{proposition}[Too few seeds cannot spread]
\label{prop:erasure-converse}
Let \(S_0\) be a fixed set of \(s_0\) vertices, independent of a uniform
size-\(m\) target mask, with \(1\le s_0\le3n/2\).
Activate a new vertex only when an observed triple has both other endpoints
active. Allow arbitrary activation order and unlimited rounds.
For the resulting closure,
\[
 \Pr\{|\operatorname{cl}(S_0)|\ge2s_0\}
 \le\min\left\{1,
   \left(\frac{24e^2s_0m}{n^2}\right)^{s_0}\right\}.
\]
Thus \(s_0=o(n/\log n)\) cannot produce positive-density support with
high probability when \(m=\lceil cn\log n\rceil\).
With no seeds the closure is empty.
\end{proposition}

\begin{proof}[Proof of Proposition~\ref{prop:erasure-converse}]
Write $s=s_0$. If the closure reaches $2s$, let $T$ be the first $s$ newly activated
vertices. Each has an activating edge inside $S_0\cup T$. These $s$ edges
are distinct: after an edge activates its last inactive endpoint it has
no inactive endpoint left. There are at most $\binom{3n}{s}$ choices of
$T$, and at most $\binom{(2s)^3}{s}$ choices of its distinct witnesses.
Every specified $s$-edge set has inclusion probability at most
$(m/n^3)^s$. The union bound and $\binom{x}{s}\le(ex/s)^s$ give
\[
 \binom{3n}{s}\binom{(2s)^3}{s}\left(\frac m{n^3}\right)^s
 \le\left(\frac{24e^2sm}{n^2}\right)^s.
\]
If fewer than $s$ target or internal grid edges exist the event is empty.
For $s=0$ the closure is empty, and for $2s>3n$ the event is impossible.
The asymptotic conclusion follows because the base tends to zero.
\end{proof}

The geometric upper bound in Theorem~\ref{thm:geometric} matches this
order for predetermined seeding and local support propagation. Its additional
content is preserving enough noisy accuracy to optimize the empirical target.
A lower bound for arbitrary polynomial-time solvers with the same target and
source interface remains open.

\subsection{Polynomial comparison on the same acquisition records}

The erasure extensions introduce no majority gates. Conditional on the
mask and private choices, their supports and reservoirs depend only on
addresses, and every active sign is a product of recorded labels.
This yields a stronger polynomial-degree bound than repeated majority
amplification at a fixed acquisition exponent.

\begin{proposition}[Degree of erasure-assisted prediction]
\label{prop:erasure-degree}
Fix $q,s\in(0,1/2)$ and $\varepsilon>0$. For each fixed rational
$a\in(0,1)$ there is $c_0=c_0(a,q,s,\varepsilon)$ such that, for every
fixed $c\ge c_0$, the retained record of Theorem~\ref{thm:erasure-fixed}
admits a polynomial estimator of the balanced optimum-threshold decision
with limiting mean squared error at most $\varepsilon$ and total label
degree $O_a(\log n)$.
For fixed positive declared channel bounds, let $r_*$ be the fixed integer
in \eqref{eq:cascade-fixed-rounds}. There is also a sufficiently large
fixed density $c_1$ such that, for every fixed $c\ge c_1$, the record of
the declared-bound cascade in Theorem~\ref{thm:geometric} admits such an
estimator of degree
$O((\log\log n)(\log n)^{r_*+2})$.
The constants may depend on the declared bounds and fixed channel parameters.
Coefficients may depend on the mask and fixed model parameters.
The declared channel bounds are fixed before labels are observed. Under independent replacement at the same addresses, the target-degree
lower bound of Theorem~\ref{thm:main} applies to either record.
\end{proposition}

\begin{proof}
For the fixed schedule, take all occupancy constants at a fixed lower
density $c_0>0$ and consider $c\ge c_0$. Coverage increases with the
batch density, whereas the removal loss tends to zero for every fixed
$c$. The proof consequently supplies a lower calibration bias $b_*>0$
depending on $a,q,s,c_0$, but not on $c$. Its exceptional probability is
superpolynomially small. Increasing $c$ makes the final-field exponent
$c(1-\sqrt{1-\mu^2b_*^4})/(r+1)$ arbitrarily large. Together with the
source-optimality bound~\eqref{eq:optbound}, this implies that the final
assignment, before repair, equals the source and its score equals the
empirical optimum except with $O(n^{-A})$ probability, for any prescribed
fixed $A>0$ after choosing $c$ sufficiently large. For the declared-bound
cascade, the comparison estimator uses only the first $r_*+1$ calibration
batches and then the reserved final half. The cascade proof gives block
bias tending to one after these finitely many updates, with superpolynomial
success. Its coefficients may depend on the declared bounds. Ignored
calibration colors can be revealed analytically before the final half, so
the same final-field argument applies. The executable optimizer continues
through all batches that pass its deterministic size guard; the comparison
uses the same acquisition record and no additional readings.

Condition on the mask and private choices. Supports, reservoir choices,
and the geometric schedule's occupancy stopping rule are independent of the
sign labels. Initial measured entries have degree one. If active signs
have degree at most $d_j$, a new one has degree at most $2d_j+1$;
retained entries do not increase the bound. Hence after $j$ extensions
\[
 d_j\le2^{j+1}-1.
\]
Fair-filled signs have degree zero. On a mask of maximum degree at most
$\Delta$, one simultaneous majority update, with its stated tie rule,
changes degree $d$ to at most $(2\Delta+1)d+\Delta$ by the argument of
Proposition~\ref{prop:bootstrap-degree}. The final empirical score has
at most three times that degree plus one.

For fixed $r$, one majority therefore gives degree $O_a(\Delta)$.
For the declared-bound cascade comparison, $r_*+1$ calibration updates
and the final update give $O(2^{B_n}\Delta^{r_*+2})$.
Its exact schedule satisfies $2^{B_n}=O(\log\log n)$.
If all $B_n$ calibration updates are used instead, the resulting degree
is at most $\exp[O(\log\log n\,\log\log\log n)]=n^{o(1)}$;
the fixed-length comparison gives the sharper displayed bound.
Truncate to masks with
$\Delta\le C_\Delta\log n$; their complement has probability $o(1)$.
As in Proposition~\ref{prop:bootstrap-degree}, compose the normalized
score with a fixed polynomial approximating the sign of a Gaussian.
Choose its degree first and then choose $A$ and $c$ large enough to make
its squared contribution on reconstruction failures vanish. Gaussian
convergence supplies the limiting risk. Set the estimator to zero on
any finite sampler-cap failure before applying the score polynomial;
this event depends only on masks and private coins. Averaging those
coins does not increase degree or mean squared error. The resulting
degrees are the stated ones.

The replacement-record projection theorem allows arbitrary dependence
of coefficients on its independent transcript, so it applies at these
same acquisition addresses and budgets.
\end{proof}

\section{A finite law for coverage and carried errors}
\label{app:finite-propagation}

An active set can expand while its estimates become less reliable. The next
result gives their joint finite law. It also identifies which conditioning
makes the newly generated errors independent. Errors stored from earlier
rounds may have arbitrary dependence.

Fix the revealed past and the source for analysis. For an inactive vertex
$v$, let $\mathcal R_v$ be its remaining eligible triples: the other two
endpoints must already be active. Write $M_v=|\mathcal R_v|$ and let
$z_v$ be the mean product of those two old error signs over $\mathcal R_v$.
Empty reservoirs cannot be hit. All these reservoirs are disjoint, including
across blocks, because each eligible triple has exactly one inactive endpoint.
Let $N=n^3-R$ be the number of unexposed locations and take a uniform
$t$-subset, with fresh independent target errors of mean $\mu=1-2q$.
The nontrivial formulas below use $1\le t\le N$; when $t=0$, including
$N=0$, the arrays stay unchanged and no division by $N$ is needed.

\begin{proposition}[Conditional law of new errors]
\label{prop:finite-errors}
Choose one sampled location uniformly in each nonempty reservoir, independently
across reservoirs. Conditional on the past and on the set $V$ of hit reservoirs,
the selected locations are independent and uniform in their respective
$\mathcal R_v$. Consequently the new error signs $X_v$ are independent, with
\[
 \E(X_v\mid\text{past},V)=\mu z_v.
\]
If $J_i$ is the number of new incorrect signs in block $i$, then
\begin{equation}
 \E[x^{J_i}\mid\text{past},V]
 =\prod_{v\in V_i}(1-\pi_v+\pi_v x),
 \qquad \pi_v=\frac{1-\mu z_v}{2}.
 \label{eq:finite-error-pgf}
\end{equation}
The three $J_i$ are conditionally independent. With $h_i=|V_i|>0$,
\begin{equation}
 \Pr\!\left\{\sum_{v\in V_i}X_v<\mu\sum_{v\in V_i}z_v-u
       \,\middle|\,\text{past},V\right\}
 \le e^{-u^2/(2h_i)}\quad(u>0).
 \label{eq:finite-signed-tail}
\end{equation}
When $h_i=0$, the increment is zero deterministically.
\end{proposition}

\begin{proof}
Condition first on the number of sampled locations in every reservoir and
in their complement. Given these counts, the sampled subsets in the cells
are independent uniform subsets of the prescribed sizes: each tuple of such
subsets represents one equally likely global sample. Selecting a uniform
member of a uniform nonempty subset gives the uniform law on the full cell.
The resulting product law does not depend on the positive count values.
Mixing those counts conditional on $V$ therefore preserves the product law.
The selected locations are distinct and their noise signs are fresh, proving
\eqref{eq:finite-error-pgf}. Hoeffding's inequality for independent signs
gives \eqref{eq:finite-signed-tail}. The old signs have been fixed throughout.
\end{proof}

The old signed sum is retained exactly; only the new increment is random in
this conditional calculation. In particular, its concentration scale depends
on $h_i$, rather than the number of target observations in the batch. Equation
\eqref{eq:finite-error-pgf} is sharper still: if the seeds and target labels
are noiseless, it assigns zero probability to a new error. A recurrence that
loses a fixed fraction of accuracy on every noiseless step is a lower-bound
artifact, not an intrinsic noise-amplification law.

\paragraph{Coverage and depletion.}
The exact hit probability of reservoir $v$ is
\[
 \theta_v=1-\frac{\binom{N-M_v}{t}}{\binom Nt},
\]
where the numerator is zero when avoidance is impossible. Hit indicators
are negatively associated: uniform fixed-size sampling has this property,
and increasing functions on disjoint coordinate sets preserve it
\citep{joag1983}. Thus, for every real $\lambda$,
\[
 \E e^{\lambda |V_i|}
 \le\prod_{v\text{ inactive in }i}(1-\theta_v+\theta_v e^\lambda).
\]
This is a classical sampling fact. It controls coverage without assuming
that the hit indicators themselves are independent.

Let the two opposite active sets have sizes $a_j,a_k$, actual mean error
signs $r_j,r_k$, and put $M_0=a_ja_k$, $z=r_jr_k$. If $d_v$ old addresses
have been removed from $v$'s grid and their endpoint products sum to $u_v$,
then
\[
 z_v=\frac{M_0z-u_v}{M_0-d_v}
 \ge z-\frac{(1-z)d_v}{M_0-d_v}.
\]
For $z\ge0$, the exact mean update consequently satisfies
\begin{equation}
 \E Z_i'\ge p_i r_i+\mu z(\E A_i'-p_i)
                 -\frac{\mu(1-z)tR}{nN},
 \label{eq:retained-mass}
\end{equation}
where $A_i'$ is the new active fraction and $Z_i'$ its signed sum divided
by $n$. Indeed, $\theta_v\le tM_v/N$ and $\sum_vd_v\le R$.
The last term vanishes when $z=1$. Equation~\eqref{eq:retained-mass}
averages over hits; it must not be substituted as a conditional bound after
fixing an unusually depleted hit pattern.

\paragraph{Finite error bounds from observable counts.}
Suppose a preceding confidence event bounds the active wrong counts by
$e_j\le a_j/2$ and $e_k\le a_k/2$. For nonempty opposite active sets put
$r_j^-=1-2e_j/a_j$ and similarly for $k$. If either is empty there are no
eligible hits. For $M_0>d_v$ define
\[
 z_v^- =\max\!\left\{-1,\frac{M_0r_j^-r_k^--d_v}{M_0-d_v}\right\},
 \qquad
 \psi(z)=\begin{cases}\underline\mu z,&z\ge0,\\z,&z<0.\end{cases}
\]
Then $\pi_v\le(1-\psi(z_v^-))/2$ whenever
$0<\underline\mu\le\mu\le1$. The negative branch matters: multiplying a
negative lower bias by a lower channel bound would reverse the needed
comparison. If an opposite active set is empty, there are no eligible hits.

Substitute these observable upper probabilities in
\eqref{eq:finite-error-pgf}. For any $x>1$ and integer $u\ge0$, Markov's
inequality bounds the probability of more than $u$ new errors by that
product divided by $x^{u+1}$. Add $u$ to the previous error-count bound.
Allocate failure probabilities in advance across blocks and rounds. A
union bound over conditional good histories makes all resulting bounds
valid simultaneously, including at a bounded stopping time. If an error
bound exceeds half of its active count, stop this nonnegative-bias induction;
squaring two negative lower bounds would be invalid. These are confidence
bounds under the stated channel law, not certificates of a global optimum.

\paragraph{Several votes require stronger conditioning.}
For majority propagation, condition on the full reservoir count vector,
not just its nonzero pattern. If $d$ votes are selected uniformly in a
reservoir with $M$ eligible locations and $B_-$ negative old endpoint
products, their number $J$ of negative products is hypergeometric.
Conditional on $J$, the number of negative noisy votes is
\[
 W\ \overset{d}=\ \operatorname{Bin}(d-J,q)
                    +\operatorname{Bin}(J,1-q),
\]
with independent summands. The majority error is
$\Pr(W>d/2)+\tfrac12\Pr(W=d/2)$ for fair ties. Outputs at different
reservoirs are independent under the full-count conditioning, provided
vote counts and any trimming use only those counts and coins independent
across reservoirs, and within-reservoir subsets are uniform. Alternatively,
condition also on all selected vote counts. Arbitrary address-dependent
selection does not have this property. Mixing correlated counts conditional
only on hits need not preserve majority-output independence.

For $z=1-2B_-/M\ge0$, sampling-without-replacement moment domination gives
the simpler error bound $(1-\mu^2z^2)^{d/2}$ for $d>0$. Optimize
$[\cosh\lambda-\mu z\sinh\lambda]^d$ over $\lambda\ge0$; the endpoint
$\mu z=1$ has error zero directly. The exact finite convolution remains
useful when this upper bound is loose.

\subsection{An exact count recursion for a comparison experiment}
\label{app:finite-kernel}
There is a compact finite state description if target addresses are sampled
independently with replacement, with a fresh target error on every occurrence.
For one-vote propagation, let the state consist of active and wrong counts
$(a_i,e_i)_{i=1}^3$. Put
\[
 b_i=n-a_i,\quad w_i=a_ja_k/n^3,\quad
 w_0=1-\sum_i b_iw_i,\quad r_i=1-2e_i/a_i,
\]
with $r_i=0$ for an empty active set. There are $b_i$ eligible cells of
mass $w_i$ in block $i$. For a batch of $t$ observations, the probability
generating polynomial of the numbers $h_i$ of distinct cells hit is
\begin{equation}
 t!\,[z^t]e^{w_0z}
       \prod_{i=1}^3\bigl[1+x_i(e^{w_iz}-1)\bigr]^{b_i}.
 \label{eq:finite-occupancy-kernel}
\end{equation}
The coefficient counts ordered draws into unused and nonempty cells.
Conditional on $(h_i)$, new wrong counts are independent with laws
$\operatorname{Bin}(h_i,(1-\mu r_jr_k)/2)$, by the same reservoir argument.
The next state is $(a_i+h_i,e_i+J_i)_{i=1}^3$. Together with independent
seed errors $e_i\sim\operatorname{Bin}(k_i,s)$, this is an exact finite
Markov recursion for a fixed batch schedule.

For any event concerning the full algorithmic path on $m\le n^3$
observations, conditioning this comparison experiment on distinct addresses
changes its probability by at most
\[
 1-\frac{(n^3)_m}{(n^3)^m}\le\frac{m(m-1)}{2n^3}.
\]
That conditioning gives the original uniform-distinct-address law. The
bound is $O(\log^2n/n)$ at logarithmic density; it does not supply a
superpolynomial tail. The main proof instead works directly without
replacement. The exact small-state evaluator is provided as a research
tool, with exhaustive rational checks. Iterating expected states is not
the same as evaluating this distribution, and this evaluator is not part
of the optimizer's word-operation bound.

\section{Calibration from a logarithmic seed}
\label{app:geometric}

The one-update crossover to constant correlation occurs at input correlation
of order $(\log n)^{-1/4}$. Several fresh updates can amplify an initial
correlation bounded below by a fixed positive multiple of
$(\log n)^{-\beta}$, for any fixed $\beta<1/2$.
The distinction changes the sufficient seed constant. We first preserve
more of the accuracy supplied by erasure propagation, then use the
calibration sample in several stages.

\paragraph{Integer schedule.}
Let $\mu_*,\rho_*\in(0,1]$ be fixed rational lower bounds on
$\mu=1-2q$ and $\rho=1-2s$. Define
\[
 K=\left\lceil\frac{5}{(\mu_*\rho_*)^4}\right\rceil,\qquad
 \ell=\lceil\log_2(n+2)\rceil,\qquad
 u=\lceil\log_2(\ell+2)\rceil,
\]
\[
 B=\lceil\log_2(u+2)\rceil+4.
\]
Reserve $m_f=\lceil m/2\rceil$ observations for final refinement,
$C=\lceil\lfloor m/2\rfloor/2\rceil$ for calibration, and
$P=m-m_f-C$ for propagation. When $P>0$, read once the first
$k=\min\{n,\lceil Kn^2/P\rceil\}$ entries in each reference fiber.
The $P=0$ and $k=n$ branches use the stored full-fiber procedure on the
whole target, with these fresh readings.

Split each of $P$ and $C$ into $B$ predetermined geometric batches.
For a budget $Q$, their sizes are
\[
 t_j(Q)=\lfloor Q/2^{j+1}\rfloor\quad(0\le j<B-1),\qquad
 t_{B-1}(Q)=Q-\sum_{j<B-1}t_j(Q).
\]
Use the propagation batches for the simultaneous one-vote erasure
extension, stopping when every block has at least $\lceil n/8\rceil$
active entries. Fair-fill the remaining erasures. Apply a simultaneous
majority update for each calibration batch of size at least $32n$,
using an independent fair sign at a zero field. Skip smaller batches.
Then use the final batch and the original capped repair. The final
update keeps its stored tie convention.

The guard $t_j(C)\ge32n$ is a fixed finite stabilization rule. It prevents
very small calibration batches from replacing a useful reference with a
sparse update; it does not certify finite exactness. For every fixed
$c>0$, the guard eventually admits every calibration batch. All batch
capacities and this guard depend only on $(n,m)$. Skipped observations
remain in the empirical objective.

\begin{lemma}[Iterated majority can amplify a weaker calibration]
\label{lem:cascade-majority}
Fix $\mu\in(0,1]$, $0\le b_0<1/2$, and positive constants $c_b,c_1,c_2,C_R$.
Put $L=\log n$. For analysis condition on the latent source, the used private coins, and a history containing $R\le C_R nL$ distinct target locations with their labels and readings, and on the resulting full sign arrays with block error-sign means
at least $b\ge c_bL^{-b_0}$. Take a uniform fresh batch of $t$ remaining
locations, with independent target errors, where
\[
 c_1L/\log L\le\lambda:=t/n\le c_2L.
\]
After simultaneous majority with independent fair ties, all three block
means are at least
\begin{equation}
 \frac1{1024}\min\{\mu b^2\sqrt\lambda,1\}
 \label{eq:cascade-amplification}
\end{equation}
with conditional probability exceeding $1-n^{-A}$ for every fixed
$A>0$ and all sufficiently large $n$. The bound is uniform over the
stated histories, with constants and the large-$n$ cutoff depending only on the fixed parameters (including $C_R$ and $A$). If $b$ is bounded below by a positive constant, the
updated means exceed $1-\varepsilon$ for every fixed $\varepsilon>0$, with the same type of probability.
\end{lemma}

\begin{proof}
Fix a vertex. Its remaining opposite-pair grid has at least $n^2-R$
locations and mean error-sign product at least
$b^2-R/(n^2-R)\ge b^2/2$ for large $n$. Thus an iid vote from this grid,
including fresh target noise, has mean $\delta\ge\mu b^2/2$.

Let $M_d(\delta)$ be the mean of a majority of $d$ iid signs of mean
$\delta\ge0$, with a fair tie. For odd $d=2h+1$,
\[
 M_d'(\delta)=(2h+1)\binom{2h}{h}4^{-h}(1-\delta^2)^h.
\]
Integrate up to $\min\{\delta,d^{-1/2}\}$ and use
$\binom{2h}{h}4^{-h}\ge1/(2\sqrt h)$ for $h\ge1$.
The case $d=1$ is immediate. An even fair-tie majority has the same
mean as the preceding odd majority. These facts give the uniform bound
\begin{equation}
 M_d(\delta)\ge\tfrac18\min\{\delta\sqrt d,1\}
 \qquad(d\ge1).
 \label{eq:cascade-iid-majority}
\end{equation}

The actual degree $D_v$ is hypergeometric, with mean
$\lambda_v=(1+o(1))\lambda$ and variance at most $\lambda_v$.
Paley--Zygmund gives
$\Pr(D_v\ge\lambda_v/2)\ge1/8$ eventually. Conditional on
$D_v=d$, sampling the pair grid without replacement couples to iid
sampling with error in the majority expectation at most
$d(d-1)/(n^2-R)$. For $d\le L^2$ this is $O(L^4/n^2)$; the probability
of $D_v>L^2$ is $\exp[-\Omega(L^2\log L)]$.
The iid majority expectation is nonnegative at every degree, so these
bounds can be averaged without subtracting a fixed bad-degree probability.
On $D_v\ge\lambda_v/2$ we have $D_v\ge\lambda/4$ eventually.
Equation~\eqref{eq:cascade-iid-majority} therefore gives expected error
sign at least
\[
 \frac1{256}\min\{\mu b^2\sqrt\lambda,1\}
       -O(L^4/n^2)-e^{-\Omega(L^2\log L)}.
\]
The two remainders are negligible compared with the displayed main term.

For a block average, changing one sampled location affects at most two
output signs. The exchange coupling for sampling without replacement
therefore bounds a location martingale increment by $8/n$. A target-noise
change affects one output sign, with increment at most $2/n$; each tie
coin has the same bound. The squared increment sum is at most
$(68t+4n)/n^2$. Consequently a downward deviation $v$ has probability
at most
\[
 \exp\left[-\frac{n^2v^2}{144(t+n)}\right].
\]
Take $v$ to be half the eventual expected lower bound and union over
three blocks. Because $b\ge c_bL^{-b_0}$ and $b_0<1/2$, the exponent
exceeds every constant multiple of $\log n$. This proves
\eqref{eq:cascade-amplification}.

For a positive constant input bias, Proposition~\ref{prop:finite-majority-bound}
below gives expected output mean tending to one. The same concentration
bound makes it exceed every fixed $1-\varepsilon$ with a superpolynomially
small exception.
\end{proof}

\subsection{How close calibration comes to the source}
\label{app:near-perfect-calibration}

Once the calibration has positive constant correlation, a fresh majority
update brings its mean error sign close to one. The following finite bound
makes that step explicit. It conditions on the stored signs, so those
signs may have arbitrary dependence.

Use $\mathcal I(x)=1-\sqrt{1-x^2}$ as above. Fix a revealed history
with $R<n^3$ removed locations and put $N=n^3-R$. At a vertex $v$, let
$N_v$ be the number of remaining incident locations and $S_v$ the sum
of their two opposite error-sign products. A uniform fresh $t$-subset
of the $N$ remaining locations has independent target errors of mean
$\mu\in[0,1]$. For $N_v>0$, $S_v\ge0$, and $t>0$, set
\[
 \delta_v=\frac{\mu S_v}{N_v},\qquad
 B_v=\left[1-\frac{N_v}{N}\mathcal I(\delta_v)\right]^t.
\]
Set $B_v=1$ if $N_v=0$, $S_v<0$, or $t=0$.

\begin{proposition}[Finite conditional majority bound]
\label{prop:finite-majority-bound}
For any tie convention, the probability of an erroneous updated sign
at $v$, conditional on this history, is at most $B_v$. With independent
fair tie coins, write $r_i'$ for the updated mean error sign in block $i$.
For every $u>0$,
\[
 \Pr\!\left\{r_i'<1-\frac2n\sum_{v\in i}B_v-u
          \,\middle|\,\text{history}\right\}
 \le \exp\left[-\frac{n^2u^2}{144(t+n)}\right].
\]
If all three old block means are at least $b\ge0$ and $R<n^2b^2$, define
\[
 M_*=n^2-R,\qquad z_*=\frac{n^2b^2-R}{n^2-R},\qquad
 B_*=\left[1-\frac{M_*}{N}\mathcal I(\mu z_*)\right]^t.
\]
Then all three updated means are at least $1-2B_*-u$, except with
probability at most $3\exp[-n^2u^2/(144(t+n))]$.
\end{proposition}

\begin{proof}
For an exponential parameter $z\ge0$, the population mean of the
noise-averaged field weight is
\[
 1+\frac{N_v}{N}(\cosh z-1)
      -\frac{\mu S_v}{N}\sinh z.
\]
Maclaurin's inequality bounds its product over a uniform $t$-subset by
the $t$th power of this mean. An error requires a nonpositive field.
Minimizing at $z=\operatorname{arctanh}(\delta_v)$ gives $B_v$;
$\delta_v=1$ follows by a limit, or directly from the absence of negative
votes. The remaining cases use the trivial bound one.

The expected block mean is therefore at least $1-(2/n)\sum_v B_v$.
In the exchange martingale for a uniform sample without replacement,
a location change has increment at most $8/n$, a noise change at most
$2/n$, and a preassigned tie-coin change at most $2/n$. The squared
increment sum is at most $(68t+4n)/n^2$. Azuma's inequality gives the
stated downward deviation. Finally, removing $d_v\le R$ locations
from a pair grid of mean at least $b^2$ leaves
\[
 \frac{S_v}{N_v}\ge
 \frac{n^2b^2-d_v}{n^2-d_v}
 \ge\frac{n^2b^2-R}{n^2-R}=z_*.
\]
Together with $N_v\ge M_*$ and monotonicity of $\mathcal I$, this gives
$B_v\le B_*$. A union bound over the blocks completes the proof.
\end{proof}

For fixed $\mu>0$, $R=O(n\log n)$, $t=O(n\log n)$, a positive constant input bias, and
$t/n\ge c_1\log n/\log\log n$, the quantity $B_*$ tends to zero.
Taking $u=n^{-1/4}$ gives superpolynomial confidence. In particular,
the existing cascade reaches means at least $1-\epsilon_n$, where
\[
 \epsilon_n=2\exp\left[-\kappa\frac{\log n}{\log\log n}\right]
                   +n^{-1/4}\longrightarrow0
\]
for a fixed $\kappa>0$ depending on the fixed channel and density bounds.
To see this, first use the established $r_*$ amplification stages to
reach a positive constant mean. The next existing stage brings it above
$1/2$. All subsequent stages preserve that bound; apply the proposition
once more at the last stage. Every calibration density is eventually at
least $c_1\log n/\log\log n$ and passes the $32n$ guard. For example,
$\kappa=(c_1/2)\mathcal I(\mu/8)$ suffices after the input mean is at least
$1/2$. Summing the conditional exceptions over the predetermined batches
retains superpolynomial confidence. No additional algorithmic stage is
required. The finite proposition alone does not certify that a particular
run has reached the required input bias.

\begin{proof}[Proof of Theorem~\ref{thm:geometric}]
Write $L=\log n$ and $D=P/n\sim cL/4$. In the analysis reveal the latent source and the private choices already used, as well as the completed batch locations and labels and sensor readings. Do not reveal future batch locations or errors. This makes the source-relative arrays measurable at each stage; unused observations remain a uniform fresh sample. For sufficiently large $n$,
$k<n$ and $p_0=k/n=\Theta(L^{-1})$. At every extension the active
fraction is at least $p_0$. We will use the conditional extension
lemma with tolerance $v_n=n^{-1/4}$, rather than with a fixed fraction
of its mean. Its two concentration bounds imply
\[
 \frac{Z_i'}{A_i'}\ge\delta-\frac{2v_n}{p_0},
 \qquad
 \delta=\mu\left(r^2-\frac{R}{n^2p^2-R}\right),
\]
except with superpolynomially small conditional probability. This follows
from $\E Z_i'\ge\delta\E A_i'$ and
$|A_i'-\E A_i'|\le v_n$. It retains nearly all of the new vote's bias.
The depletion loss is $O(L^3/n)$ uniformly, since $R\le m=O(nL)$.

\paragraph{Growth of support.}
Set $a=9/10$. Uniformly in $j<B$, the integer propagation sizes satisfy
$t_j(P)/n\ge(1-o(1))D2^{-j-1}$; the last residual batch is no smaller.
Put $\nu=(t_j(P)/n)p_j^2$. The conditional expected active fraction in
each block is at least $1-e^{-(1-o(1))\nu}$, because the old support is
retained. For $\nu\le1/7$, this is at least
$(13/14-o(1))\nu$. For $\nu\ge1/7$, it is at least
$1-e^{-1/7}-o(1)>1/8$. Also
$\nu\ge\Omega((L\log L)^{-1})$ throughout the relevant rounds, so
$v_n=o(\nu)$. Thus, with the stated conditional probability,
\begin{equation}
 p_{j+1}\ge\min\{aD2^{-j-1}p_j^2,1/8\}.
 \label{eq:cascade-support}
\end{equation}
Normalize by $y_j=aDp_j/2^{j+2}$. Before the cap in
\eqref{eq:cascade-support} is reached, $y_{j+1}\ge y_j^2$, and
\[
 y_0\ge aK/4\ge\frac98(\mu_*\rho_*)^{-4}>1.
\]
Let $h=\log y_0$ and
$j_* =\lceil\log_2(\log D/h)\rceil$.
For large $n$, $1<y_0<D$ and $2^{j_*}\le2\log D/h$.
Failure to stop by $j_*$ would give $y_{j_*}\ge D$, contradicting
$p_{j_*}<1/8$. The chosen integer horizon satisfies $j_*<B$ eventually:
$h\ge\log(9/8)$, $u\sim\log L/\log2$, and hence
$\log D/h\le8u$ for large $n$.

\paragraph{Accuracy carried to the stopping time.}
Define the constants before the induction:
\[
 A=\log\frac1{\mu_*\rho_*},\qquad
 b_*=\log(9/8),\qquad \vartheta=e^{-b_*/16},
\]
\[
 r_0=\vartheta\rho_*,\qquad
 r_{j+1}=\vartheta\mu_*\,r_j^2.
\]
Seed concentration establishes $r_0$ with superpolynomial success. The
explicit solution satisfies
\[
 r_j=(\vartheta\mu_*)^{2^j-1}
             (\vartheta\rho_*)^{2^j}
       \ge e^{-(A+b_*/8)2^j}.
\]
Since $h\ge4A+b_*$, all $j\le j_*$ obey
\begin{equation}
 r_j\ge D^{-\zeta},\qquad
 \zeta=\frac{2(A+b_*/8)}{4A+b_*}
       =\frac12-\frac{b_*}{4(4A+b_*)}<\frac12.
 \label{eq:cascade-carried}
\end{equation}
These deterministic bounds make both $O(L^3/n)$ and $2v_n/p_0$
negligible relative to $r_j^2$. The conditional extension estimate
therefore verifies the recurrence used to define $r_j$, uniformly up to
$j_*$. Its conditional failure is at most
$\exp[-\Omega(\sqrt n/L)]$ per round. There is no circular reliance on
successful support growth to choose the horizon or accuracy bound.

At stopping, every block has signed mass at least $D^{-\zeta}/8$.
Fair filling leaves full-block means at least $D^{-\zeta}/16$, with a
superpolynomial exception. Reveal all skipped propagation colors for
analysis. The calibration batches remain fresh uniform subsets of the
remaining locations.

\paragraph{Amplification.}
For each fixed calibration index $j$, its density is $\Theta(L)$.
Lemma~\ref{lem:cascade-majority} sends a lower mean
$c_jL^{-a_j}$ to $c_{j+1}L^{-a_{j+1}}$, where
\[
 a_{j+1}=\max\{2a_j-1/2,0\}.
\]
Starting at $a_0=\zeta<1/2$, a positive constant mean is reached after
\begin{equation}
 r_* =\left\lceil\log_2\!\left(\frac1{1-2\zeta}\right)\right\rceil
     =\left\lceil\log_2\!\left(\frac{2(4A+b_*)}{b_*}\right)\right\rceil
 \label{eq:cascade-fixed-rounds}
\end{equation}
updates. Equivalently, $r_*$ is the least integer for which
$2^{r_*}\ge2(4A+b_*)/b_*$.
The next existing update gives means tending to one, with a
superpolynomially small exception. The integer $r_*$ is fixed for each
declared channel pair, whereas $B\to\infty$. Every calibration density
is eventually at least a positive constant times $L/\log L$ and passes
the $32n$ guard. Later stages preserve the $1/2$ bound; applying
Proposition~\ref{prop:finite-majority-bound} at the last stage gives
means at least $1-\epsilon_n$, with $\epsilon_n\to0$ as displayed above.
A union over the $B$ rounds retains the superpolynomial confidence.

\paragraph{Exact optimization and resources.}
After all skipped colors have been revealed, the final half is uniform
in the complement, with fresh errors. Lemma~\ref{lem:direct-mixed-moment}
therefore applies with $b_0=1-\epsilon_n$ and $t/m\to1/2$. For fixed
disjoint vertex sets $A,D$ and prescribed colored locations $F$, it gives
\[
 \Pr\{F,\mathcal G,M_A\le0,D^*_D\le L_n\}
 \le\pi_F n^{-\alpha|A|/2-\alpha|D|+o(1)}
\]
whenever $L_n=o(\log n)$. Summing exact inclusion probabilities over
colors costs at most $m/n^3$ per prescribed edge, even though the number
of colors grows. Exclude $\mathcal G^c$ once globally. Incidence-tree
and root-star witnesses of $J$ exceptional vertices then have total
probability at most $n^{1-\alpha J/2+o(1)}$. Choose a fixed $J>2/\alpha$.
The intrinsic localization and observable weak-set arguments in
Appendix~\ref{app:sublinear} now give exact repair of the full original
objective with probability $1-o(1)$. No independence between calibration
errors and profitable empirical changes has been used.

The seed count is $3k=O_{c,\mu_*,\rho_*}(n/L)$. All propagation and
calibration batches together process $O(m)$ observations and initialize
$O(Bn)$ entries. The existing repair cap therefore gives
$O((m+n)\log(n+2))$ word operations on every input. Exact fourth powers,
integer logarithmic schedules, the global fair-bit convention in
Appendix~\ref{app:global-fair-bit},
and the prior arithmetic-bit accounting suffice; the analytic constants
in the proof are not algorithmic inputs.
\end{proof}

\subsection{The final majority has a matching error exponent}
\label{sec:calibration-sharpness}

The exponent in the joint bound also describes the errors of the final
update itself. This supplies a comparison with the intrinsic correction
exponent without making a lower-bound claim about other algorithms.

\begin{proposition}[Sharp exponent for the final update]
\label{prop:calibration-sharpness}
Fix $c>0$, $q\in(0,1/2)$, and $\tau\in(0,1)$, and write
$\alpha=c(1-2\sqrt{q(1-q)})$.
Fix also $C_R>0$. Condition on a pre-final history with $R\le C_Rn\log n$ removed locations
and full error-sign arrays whose three block means are at least
$1-\varepsilon_n$, where the deterministic sequence
$\varepsilon_n\to0$. Let the final sample contain
$t\sim\tau c n\log n$ locations, conditionally uniform among the remaining locations, with fresh independent target errors. For every vertex $v$
and any tie convention, its final sign obeys
\begin{equation}
 \Pr\{\text{final sign at $v$ is wrong}\mid\text{history}\}
       =n^{-\tau\alpha+o(1)}.
 \label{eq:sharp-final-error}
\end{equation}
The exponent is uniform over these histories and vertices.
If the complement of the calibration event has superpolynomially small probability, the
expected number $X_n$ of wrong factor signs in the true reference-fiber
gauge satisfies
\[
 \E X_n=n^{1-\tau\alpha+o(1)}.
\]
\end{proposition}

\begin{proof}
The finite conditional majority bound gives the upper estimate in
\eqref{eq:sharp-final-error}, because the mean product on each remaining
pair grid is $1-o(1)$ and its size is $(1-o(1))n^2$.

For the reverse estimate, put $N=n^3-R$. At $v$ let $N_v$ be the number
of remaining incident locations and let $G_v$ count those whose two
old endpoint error signs have product $+1$. Uniformly in the stated
histories,
\[
 N_v=n^2-O(n\log n),\qquad
 G_v\ge(1-\varepsilon_n)n^2-O(n\log n).
\]
Set $\lambda=t/n$, $\ell=\lfloor\lambda\sqrt{q(1-q)}\rfloor$, and
$d=2\ell+1$. Consider the event that exactly $d$ final locations meet
$v$, all belong to those $G_v$ locations, and their target errors contain
$\ell$ positive signs and $\ell+1$ negative signs. Its field is $-1$,
so the update is wrong under every tie rule. The event has exact probability
\begin{equation}
 \frac{\binom{G_v}{d}\binom{N-N_v}{t-d}}{\binom Nt}
 \binom d\ell(1-q)^\ell q^{\ell+1}.
 \label{eq:sharp-final-witness}
\end{equation}
For sufficiently large $n$ all binomial coefficients are feasible.
Expanding the falling factorials incurs
$O(t^2/N+d^2/G_v)=o(1)$. Replacing $G_v/n^2$ by one costs $o(\log n)$,
uniformly. The logarithm of \eqref{eq:sharp-final-witness} is therefore
\[
 -\lambda+d\log\lambda-\log(\ell!)-\log((\ell+1)!)
 +\ell\log(1-q)+(\ell+1)\log q+o(\log n).
\]
Stirling's formula gives
$-\lambda(1-2\sqrt{q(1-q)})+o(\log n)
 =-\tau\alpha\log n+o(\log n)$.
This proves the lower estimate. Sum the uniform conditional bounds over
the $3n$ vertices. A superpolynomial exceptional probability contributes
less than every prescribed inverse power of $n$, yielding the expectation.
\end{proof}

The statement concerns factor errors after one specified update. It does
not assert concentration of $X_n$ or a computational lower bound. In
particular, exact repair can succeed when the expected number of those
errors diverges. Its task is to resolve their small connected components
together with the empirical improvements on the source.

\subsection{Allocating the final sample}
\label{app:allocation-tradeoff}

For any fixed rational $\tau\in(0,1)$, reserve
$t=\lceil\tau m\rceil$, put $Q=m-t$, and take
$C=\lceil Q/2\rceil$, $P=\lfloor Q/2\rfloor$. Keep the same $K$, $B$,
and update rules. Then $P/n\sim(1-\tau)c\log n/2$; the propagation
normalization, carried-bias exponent, and finite number of amplification
stages are unchanged. The sufficient reading count is
\[
 3k\sim\frac{6K}{(1-\tau)c}\frac n{\log n},
\]
while the final component exponent is $\tau\alpha$. The word bound
and exact-repair proof therefore hold for every such fixed allocation.
The default executable remains $\tau=1/2$. This statement permits
$\tau$ arbitrarily close to one as a fixed constant; it does not assert
a uniform result for $\tau=\tau_n\to1$.

For the learned-bound procedure, apply these integer capacities to $m-h$ after the charged pilot, retaining all $m$ observations in the objective. Since $h=o(m)$, the same exponents and reading order follow uniformly on the pilot confidence event. The unchanged default algorithm and the optional rational-allocation implementation are both included in the source package.

\subsection{A finite fair-bit implementation}
\label{app:global-fair-bit}
For the cascade, generate the uniform target partition by a single
Fisher--Yates permutation; the pilot, when present, uses its first $h$
positions. This requires at most $m$ uniform-integer requests, including the empty target.
Streaming one-vote reservoir selection makes at most one request per
propagation observation, hence at most $m$ further requests. Fair filling
uses at most $3n$ bits. Each enabled calibration stage draws $3n$ fair
bits, and its size is at least $32n$, so all such stages together use at
most $3m/32$ bits. Regard each fair bit as a request from a two-element
set. Thus the deterministic bound
\[
 M=3(m+n+1)
\]
exceeds the total number of requests. These counts apply to both the
declared- and learned-bound cascades, including every fixed rational
allocation. Sensor noise belongs to the measurement law and is not private
randomness generated by the optimizer.

Implement each uniform-integer request by rejection from enough fresh fair
bits, and cap the \emph{total} number of attempts at $4M$. If this cap
is exhausted, return the all-positive assignment and its actual target
score. Each attempt succeeds with conditional probability at least $1/2$.
Coupling its success indicators to independent fair bits gives
\[
 \Pr\{\text{cap exhausted}\}
 \le\Pr\{\operatorname{Bin}(4M,1/2)<M\}
 \le e^{-M/4}.
\]
The argument allows the requested ranges to depend on the preceding
history. The capped and ideal programs can be coupled to agree until
exhaustion. The probability of disagreement is therefore superpolynomially
small; the difference in their expected factor-error counts is at most
$3n e^{-M/4}$. In particular, the sharp expectation exponent is preserved.
For any fixed prescribed-edge and vertex counts, this error is also
negligible compared with the polynomial probabilities in the joint bound.

There are $O(m+n)$ attempts on every input, each using
$O(\log(n+m+2))$ fair bits. The resulting bit-generation work fits the
stated word bound when $m\le n^3$. Charging conservative quadratic cost
for exact word arithmetic retains
$O((m+n)\log(n+2)\log^2(n+m+2))$ arithmetic-bit work.
The supplied NumPy research programs use random-word sampling; the global
cap specifies the finite fair-bit realization of the same probabilistic
algorithm.

\paragraph{Learning the channel bounds.}
The pilot below estimates conservative channel bounds. We write $\mu_*,\rho_*$ for declared fixed bounds and $\underline\mu,\underline\rho$ for the lower confidence bounds computed from the pilot; $\mu,\rho$ remain the true channel biases. Only the seed count
depends on its estimates, and its target observations are included in the
removed-location count. On the pilot confidence event the estimates lie
in $[\mu/2,\mu]\times[\rho/2,\rho]$ eventually. Equations
\eqref{eq:cascade-carried} and \eqref{eq:cascade-fixed-rounds}, and all
concentration bounds, are uniform on this compact positive range.
Thus the same acquisition order holds in expectation and with
superpolynomially high probability; the all-input cap is still $7n$.
Every target label remains in the objective.

\paragraph{Polynomial comparison on the smaller acquisition record.}
For declared bounds, a comparison estimator can use only the first
$r=r_*+1$ calibration batches, ignore the later colors, and then use the
same final half. Its coefficients may depend on the fixed channel
bounds. The executable optimizer continues through every eligible batch;
the comparison needs no additional source readings. Conditional on a mask
of maximum degree $\Delta$, erasure propagation has degree at most
$2^{B+1}-1$, and a majority update sends degree $d$ to at most
$(2\Delta+1)d+\Delta$. The comparison score consequently has degree
\[
 O(2^B\Delta^{r+1})
   =O\bigl((\log\log n)(\log n)^{r_*+2}\bigr)
\]
on $\Delta=O(\log n)$. For sufficiently large fixed density, the final
source-recovery and source-optimality probabilities have any prescribed
fixed inverse-polynomial rate. Earlier calibration failures are
superpolynomial. The same fixed polynomial approximation to the
normalized score therefore gives the established balanced-decision
comparison at a degree polynomial in $\log n$. The replacement lower
bound and its charge for pilot-inspected labels are unchanged.

\subsection{Calibration by exact parity cancellation}
\label{app:pilot-cancellation-study}
Write $\mu=1-2q$ and $\rho=1-2s$. Let $H_p$ be a uniform $h$-subset of the target mask, selected independently of all labels, with $h\le m$. At every $e=(i,j,k)\in H_p$, take three fresh reference-fiber readings $R^A_e,R^B_e,R^C_e$. Separately take $h$ fresh readings $R^0_1,\ldots,R^0_h$ at the common fiber center. Every visit uses a fresh sensor error, including repeated addresses. Define
\[
 Z_e=Y_eR^A_eR^B_eR^C_e,\qquad
 \widehat z=h^{-1}\sum_{e\in H_p}Z_e,\qquad
 \widehat r=\left|h^{-1}\sum_{j=1}^hR^0_j\right|.
\]
For $h\ge1$ and $\epsilon>0$, put
\[
 z_-=\max\{0,\widehat z-\epsilon\},\quad
 \rho_+=\min\{1,\widehat r+\epsilon\},\quad
 \underline\mu=\min\{1,z_-/\rho_+^3\},\quad
 \underline\rho=\max\{0,\widehat r-\epsilon\}.
\]
If $\rho_+=0$, set $\underline\mu=0$. Then, with probability at least $1-4\exp(-h\epsilon^2/2)$,
\[
 \frac{\max\{0,\mu\rho^3-2\epsilon\}}{\min\{1,\rho+2\epsilon\}^3}\le\underline\mu\le\mu,
 \qquad
 \max\{0,\rho-2\epsilon\}\le\underline\rho\le\rho.
\tag{P1}
\]
The guarantee holds conditionally on the source and on every pilot address set.

\begin{proof}
The reference-fiber identity is exact even at coinciding coordinates:
$A_iB_jC_k=T_{ijk}$. If the fresh sensor errors for a pilot edge are
$\xi^A_e,\xi^B_e,\xi^C_e$, then
\[
 Z_e=(T_e\eta_e)(A_i\xi^A_e)(B_j\xi^B_e)(C_k\xi^C_e)
     =\eta_e\xi^A_e\xi^B_e\xi^C_e.
\]
The pilot edges are distinct, so their target errors are independent. All
sensor visits are independent by assumption. The $Z_e$ are therefore iid
signs of mean $\mu\rho^3$, independent of the source and the pilot addresses.
At the center, $R^0_j=T_{111}\xi^0_j$. Conditional on $T_{111}$, Hoeffding
bounds the deviation of the signed average from $T_{111}\rho$. The reverse
triangle inequality transfers that bound to $|\widehat r-\rho|$. The two
absolute deviations each exceed $\epsilon$ with probability at most
$2e^{-h\epsilon^2/2}$. On their complement (P1) follows.
\end{proof}

For an exact integer convention, set
$\ell=\lceil\log_2(n+2)\rceil$, $h=\min\{n,\lfloor m/16\rfloor,\ell^2\}$, let $d$ be the
largest positive integer satisfying $d^4\le\ell$, and take $\epsilon=1/(4d)$.
At logarithmic density $h=\ell^2$ eventually and
$h\epsilon^2=\Theta(\ell^{3/2})$. Thus the exceptional probability is
smaller than every fixed inverse power of $n$. Both lower bounds are
positive for all sufficiently large $n$ on the successful pilot event, and
converge to the positive constants $\mu$ and $\rho$. For example, when
$\epsilon\le\min(\rho/16,\mu\rho^3/16)$ the successful event implies
$\underline\mu\ge(448/729)\mu>\mu/2$ and
$\underline\rho\ge7\rho/8>\rho/2$.
The pilot uses exactly $4h\le4n$ readings, with
$4h=O(\log^2 n)=o(n/\log n)$ asymptotically. Its target mask occupies at
most $m/16$ observations.
If $h=0$, either lower bound is nonpositive, or the selected seed count is
$n$, read all fibers once with fresh sensor errors and use the full-fiber
linear procedure. Its readings are counted in addition to any pilot.

Only the inferred bounds, and not the pilot fiber signs themselves, are
used to initialize the propagation. Seed readings after the pilot use
fresh sensor errors. For the geometric branch, partition the remaining $m-h$ observations by the
integer schedule above: replace $m$ by $m-h$ only in its batch capacities,
and retain all $m$ observations in the final objective. Conditional on the
pilot history and the true source, all unrevealed target errors remain
independent, and unrevealed locations remain uniform in the complement
of the pilot mask. The removed-location count in every subsequent
conditional lemma includes the pilot edges.

Batch capacities and the original coloring must be fixed by $n,m$, not
by the estimated parameters, so their prescribed-edge inclusion factors
remain independent of target noise. Only the seed count is adapted to
(P1); any support stopping rule is handled by revealing all preassigned
skipped colors before all calibration batches and the final refinement.

\paragraph{Resource type.}
If a deterministic cascade algorithm succeeds with $k\le C(\mu_*,\rho_*,c)n/\log n$
seeds per mode given valid positive lower bounds $\mu_*\le\mu$ and
$\rho_*\le\rho$, use the pilot's lower confidence bounds for the two biases as these inputs and cap $k$ by $n$. Its acquisition count is $O_{c,q,s}(n/\log n)$ with probability
$1-n^{-A}$ for every fixed $A$, and in expectation. Indeed, on the pilot
event, eventually $\underline\mu\ge\mu/2$ and
$\underline\rho\ge\rho/2$, so the required seed constant is bounded;
on the complement, the cap is $3n+4h$ and its expected contribution is
$o(1)$. This conclusion is not a deterministic $O(n/\log n)$ bound on
arbitrary records. These statements use one constant $C(c,q,s)$ in the high-probability
reading bound, for every fixed inverse-polynomial tolerance. The compact
range $\underline\mu\in[\mu/2,\mu]$,
$\underline\rho\in[\rho/2,\rho]$ on the successful pilot makes the
cascade proof uniform in the random estimates. Conditional fresh seed
errors remain independent. Including $h=o(m)$ pilot edges in the revealed
history leaves all depletion and fixed-edge bounds unchanged, so the exact
optimization claim follows as well.

All pilot fractions and the fourth power defining $K$ are computed exactly
with rational integers of $O(\log(n+m))$ bits (up to fixed factors).
Pilot readings and their accumulation cost $O(h)$ word operations after the
common $O(m)$ target partition; exact integer preprocessing adds
$O(\log(n+m+2))$ word operations. Here $h\le n$.
The all-input reading cap is $4h+3k\le7n$; the word and conservative
arithmetic-bit bounds are those of the declared-bound algorithm.

\subsection{An observable test of the source coupling}
\label{app:pilot-distinguishability-study}
The same pilot permits a direct test of the two source laws, with an
observable condition on its addresses. Let $B_P$ be the binary incidence
matrix of the $h$ selected triples: one row per triple and one column per
latent coordinate, with three ones in each row. Compute its rank over
$\mathbb F_2$. If $\operatorname{rank}(B_P)<h$, report ``abstain.''
Otherwise declare retained source when $\widehat z>\epsilon$ and
independent replacement when $\widehat z\le\epsilon$.

\begin{proposition}[Coupling test with an address check]
\label{prop:pilot-coupling}
Fix any label-independent pilot address set $P$ of $h\ge1$ distinct
triples satisfying $\operatorname{rank}_{\mathbb F_2}(B_P)=h$.
Under the retained-source law the cancellation products $Z_e$ are iid
signs of mean $\mu\rho^3$. Under independent replacement they are iid
fair signs, even conditionally on all target errors and on the complete
independent replacement-source randomness.
Consequently, for $0<\epsilon\le\mu\rho^3/2$, each conditional error
probability of the displayed test is at most
\[
 \exp(-h\epsilon^2/2).
\]
These claims condition on the chosen pilot addresses. For the random
mask and a uniformly selected pilot of size $h=O(\log^2 n)$, the rank
check fails with probability $o(1)$.
\end{proposition}

\begin{proof}
The retained-source claim is the exact cancellation identity proved above
and requires no rank condition. Under replacement, write the uniform
latent target signs in binary coordinates as a uniform vector $x$ over
$\mathbb F_2$. The pilot source products are $B_Px$ in binary notation.
Full row rank makes this vector uniform on $\mathbb F_2^h$. Conditional
on all target errors and all replacement randomness, multiplying these
source products by the fixed error and replacement-reading signs preserves
that uniformity. Thus the $Z_e$ are independent fair signs. Hoeffding's
inequality gives replacement false-retention probability at most
$e^{-h\epsilon^2/2}$. Under retention it gives a false-replacement
probability at most $e^{-h(\mu\rho^3-\epsilon)^2/2}$, which is bounded
by the displayed expression.

The marginal law of a uniformly selected $h$-subset of the uniform target
mask is the uniform $h$-subset of $[n]^3$. Failure of full row rank is
exactly the presence of a nonempty even cover among those $h$ rows.
The existing pairing argument, now with $h$ in place of the full target
size $m$, gives for an independently sampled list
\[
 \Pr(\text{even cover of size }t)
 \le\left(\frac{eh\sqrt t}{n^{3/2}}\right)^t,
 \qquad 2\le t\le h.
\]
Only even $t$ contribute. Put $b=eh^{3/2}/n^{3/2}$. For sufficiently
large $n$, $b<1$ and the sum is at most $b^2/(1-b)=o(1)$.
The probability all sampled triples are distinct is at least
$1-\binom h2/n^3=1-o(1)$; conditioning on distinctness proves the
uniform-subset claim. When $h=1$, the row has rank one deterministically.
\end{proof}

Only coordinates incident to a pilot edge matter, so $B_P$ has at most
$3h$ nonzero columns. Compressing those coordinate names and binary
Gaussian elimination gives a rank check using $O(h^3)$ elementary field
operations, with $O(h\log h)$ address comparisons. For $h=O(\log^2n)$,
this is a polylogarithmic diagnostic cost and uses no extra readings.
It is an optional diagnostic, separate from the frozen optimization
algorithm.

The rank condition is necessary for the stated conditional uniformity.
For example, the four triples
\[
 (1,1,1),\quad(1,2,2),\quad(2,1,2),\quad(2,2,1)
\]
have incidence rank three. Their source products have product one. After
conditioning on errors and replacement readings, the four cancellation
products therefore have a fixed parity and cannot be independent fair
signs. The diagnostic abstains on this address set.

This is a test of the stated retention and independent-replacement laws.
It does not certify exact persistence against arbitrary drift: a replacement
still correlated with the target's source may also produce a positive pilot
mean. The target objective is unchanged, and the test does not compute its
optimum. The inequality $\epsilon\le\mu\rho^3/2$ is a condition on the
retained alternative's signal, not an observed certificate of that alternative;
for the prescribed $\epsilon_n\to0$ it eventually holds for every fixed
channel pair.

\subsection{Why the declared schedule has not reached the sparse finite regime}
\label{app:finite-gate}
The asymptotic regime in which the source is typically suboptimal is
$c\mathcal I(\mu)<1$, with $\mathcal I(x)=1-\sqrt{1-x^2}$.
For the declared schedule, the seed coefficient and the $32n$ calibration
guard impose a quantitative obstruction before its probability estimates are
used. The following statement concerns those execution rules.

\begin{proposition}[Finite gates of the declared cascade]
\label{prop:finite-gate}
Let $n\ge2$ and let $\mu,\rho\in(0,1)$ be the true channel biases.
Declare bounds $0<\mu_*\le\mu$ and $0<\rho_*\le\rho$, and set
$K_{\mathrm{decl}}=\lceil5/(\mu_*\rho_*)^4\rceil$ and
$m=\lceil cn\log n\rceil\le n^3$, with $c\mathcal I(\mu)<1$.
For a final fraction $\tau\in(0,1)$, write
$R=m-\lceil\tau m\rceil$.
The schedule computed from these declared bounds has a reduced seed exactly when
\[
 R\ge2\left\lceil\frac{K_{\mathrm{decl}}n^2}{n-1}\right\rceil;
\]
its precomputed calibration schedule has a batch satisfying the
capacity guard exactly when $R\ge128n-1$. Execution of that batch also
requires passing the reduced-seed gate; the full-fiber fallback executes
no calibration batches.
If both gates hold, then
\begin{equation}
 (1-\tau)\log n+\frac{(2-\tau)\mathcal I(\mu)}n
 >\frac{D(\rho)}2,
 \label{eq:finite-gate-bound}
\end{equation}
where
\[
D(\rho)=
\begin{cases}
256\left(1-\sqrt{1-\dfrac{\sqrt5}{8\rho^2}}\right),
 &\rho^2\ge 9\sqrt5/64,\\[4pt]
135/(8\rho^4),&\rho^2<9\sqrt5/64.
\end{cases}
\]
In particular, put
\[
 D_0=256\left(1-\sqrt{1-\sqrt5/8}\right)
       =38.70266654\ldots.
\]
The default $\tau=1/2$ requires $\log n+3/n>D_0$.
Even after choosing another final fraction, the two gates require
$\log n+2/n>D_0/2$.
\end{proposition}

\begin{proof}
The propagation capacity is $P=\lfloor R/2\rfloor$.
The inequality $\lceil K_{\mathrm{decl}}n^2/P\rceil\le n-1$ is equivalent to the
first gate. The calibration capacity is $C=\lceil R/2\rceil$.
Its first batch has $\lfloor C/2\rfloor$ observations, and each middle
batch is smaller. For the implemented number $B$ of batches,
$6\le B\le n+7$. The last batch has size
\[
 \left\lfloor\frac{C}{2^{B-1}}\right\rfloor
 +\operatorname{popcount}(C\bmod2^{B-1}).
\]
If $C<32n$, no batch can meet the guard. If $32n\le C<64n$,
the last batch is smaller than $2n+B-1\le9n<32n$, and the
first is also too small. Conversely $C\ge64n$ enables the first
batch. This proves the second gate.

Write $I=\mathcal I(\mu)$. The assumption $cI<1$ gives
$m<n\log n/I+1$. Combining this inequality with the two gates yields
\[
 (1-\tau)\log n+\frac{(2-\tau)I}{n}
 > I\max\{2K_{\mathrm{decl}},128\}.
\]
Valid bounds give $K_{\mathrm{decl}}\ge5/(\mu\rho)^4$. With $u=\mu^2$, the minimum
of the resulting right side is half of
\[
 \inf_{0<u\le1}
 \max\left\{\frac{20}{\rho^4u^2},256\right\}
 (1-\sqrt{1-u}).
\]
The function $(1-\sqrt{1-u})/u^2$ decreases to $u=8/9$ and then
increases. Equating the two coefficients gives
$u_0=\sqrt5/(8\rho^2)$. If $u_0\le8/9$ the minimum occurs at
$u_0$; otherwise it occurs at $8/9$. These cases give $D(\rho)$.
For example, the minimum $27/32$ of the first function follows by
writing $x=\sqrt{1-u}$ and using
\[
 \frac{32}{27}-(1-x)(1+x)^2
 =(x-1/3)^2(x+5/3).
\]
Finally $D(\rho)\ge D(1)=D_0$ and $I<1$ give the last claims.
\end{proof}

The exact seed gate uses the declared bounds; the relaxation in
\eqref{eq:finite-gate-bound} uses the true biases through
$K_{\mathrm{decl}}\ge5/(\mu\rho)^4$. Replacing the declared
coefficient by the smaller true-parameter coefficient need not preserve
the exact gate.

The default gate has the scale $\exp(D_0)=6.43\times10^{16}$
coordinates per block, even as sensor noise tends to zero. Optimizing the
allocation reduces this necessary scale to
$\exp(D_0/2)=2.54\times10^8$.
At sensor bias $\rho=.8$, the respective scales are
$5.47\times10^{27}$ and $7.40\times10^{13}$.
These scales minimize over target-channel parameters at the specified sensor bias. They are necessary execution gates for one calibration stage of the declared schedule, not sufficient thresholds for success or lower bounds for other algorithms. For instance, with $c=2.5$, $\mu_*=.66$, $\rho_*=.8$ and the default allocation, $K_{\mathrm{decl}}=65$ and the seed gate alone has the scale $n\gtrsim\exp(104)$. The constant $D_0$ is sharp for
the displayed relaxation: as $\rho\uparrow1$, rational valid channel bounds can approach
$\mu^2=\sqrt5/(8\rho^2)$ from above while keeping $K_{\mathrm{decl}}=64$.
The leading seed and calibration constraints then agree in the limit. The finite rounding terms remain explicit in
\eqref{eq:finite-gate-bound}.

This calculation explains why the existing dense test can execute the
reduced cascade while ordinary sparse tests fall back to full acquisition.
A useful sparse finite guarantee requires smaller sufficient coefficients,
a different schedule, or a different acquisition mechanism. The obstruction
places no lower bound on other algorithms or on the underlying problem.

\Needspace{8\baselineskip}
\section{Reusing the complete target}
\label{app:recycling}

Fresh batches establish the geometry of the estimation errors. Once each
vertex has only a bounded number of wrong neighbors, the complete target
can be used again: its new field differs from the true signed degree by
only a constant. This deterministic comparison restores the deviation
rate of the full sample. It also explains when majority updates alone
complete the empirical optimization.

For any factor assignment $z$, let $\mathcal M(z)$ denote a simultaneous
majority update using all $m$ original observations, retaining $z_v$ at a
zero field. Write $z^0$ for the estimate produced by the reserved final
batch of the cascade, or by the single update in full-fiber acquisition, and put
\[
 z^1=\mathcal M(z^0),\qquad z^2=\mathcal M(z^1).
\]
Every update uses the same frozen labels and requires no source readings.
The reduced-acquisition default uses $z^1$ as its fixed repair reference
and scores $z^2$ as its parallel-flip candidate. It retains $z^0$
as an additional feasible incumbent. The default full-fiber branch instead
returns its one-update-plus-repair procedure directly. The explicit
two-pass variant computes $z^1,z^2$ in both acquisition branches and performs
no component search. The probability and expectation statements below
refer to the defined iterates, not the default solver's selected output.

Use the true reference-fiber gauge. Let $\epsilon_v(z)\in\{-1,1\}$ be
its error sign, $E(z)=\{v:\epsilon_v(z)=-1\}$, and
$D_v=\sum_{e\ni v}\eta_e$ its true signed degree. Write
$W_L^*=\{v:D_v\le L\}$ for a fixed threshold $L$. Define
\[
 b_v(z)=\#\{e\in H:e\ni v,\ \prod_{u\in e\setminus\{v\}}
                    \epsilon_u(z)=-1\}.
\]
This counts the incident observations whose other two estimates have the
wrong product. It is used in the proof, not supplied to the algorithm.

\begin{lemma}[A deterministic comparison of reused fields]
\label{lem:recycle-field}
For every target, source assignment, and estimate $z$, its all-target field
in source coordinates obeys
\[
 \left|M_v(z)-D_v\right|\le2b_v(z),\qquad
 M_v(z)=\sum_{e\ni v}\eta_e
          \prod_{u\in e\setminus\{v\}}\epsilon_u(z).
\]
Consequently, if $2b_v(z)\le C$ for every vertex, then
\begin{equation}
 \{v:D_v<-C\}\ \subseteq E(\mathcal M(z))
                 \ \subseteq\ \{v:D_v\le C\}.
 \label{eq:recycle-sandwich}
\end{equation}
Both inclusions permit every convention at a zero field.
\end{lemma}
\begin{proof}
An incident observation contributes zero to $M_v(z)-D_v$ when the other
error signs have product $+1$, and contributes $-2\eta_e$ otherwise.
The triangle inequality proves the field bound. A field below zero
forces an error, while a field above zero forces the correct sign.
\end{proof}

Put $\alpha=c\mathcal I(1-2q)$, where
$\mathcal I(x)=1-\sqrt{1-x^2}$. The next theorem applies to every fixed
rational final fraction $\tau\in(0,1)$, including the default $1/2$.

\begin{theorem}[The full-sample exponent after reuse]
\label{thm:recycling}
Under the retained-source law, fix $c>0$, $q,s\in(0,1/2)$,
and a rational final-batch fraction $\tau\in(0,1)$.
Let $z^0$ be the cascade estimate, using either fixed positive declared
channel bounds $\mu_*\le1-2q$, $\rho_*\le1-2s$, or the charged
parameter-learning pilot. Constants and sufficiently large sample sizes
may depend on $c,q,s,\tau$ and, for the declared-bound procedure,
on $\mu_*,\rho_*$. Then:
\begin{enumerate}
\item For every fixed $A>0$, there is a finite constant $C_A$, depending
only on $A$ and the fixed quantities just specified, such that
\[
 \Pr\!\left\{\max_v 2b_v(z^0)>C_A\right\}\le n^{-A}
\]
for all sufficiently large $n$.
\item Uniformly over vertices,
\[
 \Pr\{v\in E(z^1)\}=\Theta\!\left(
                      \frac{n^{-\alpha}}{\sqrt{\log n}}\right),\qquad
 \E|E(z^1)|=\Theta\!\left(
                      \frac{n^{1-\alpha}}{\sqrt{\log n}}\right).
\]
For $0<\alpha<1$, there are constants $a_-,a_+>0$ such that
\[
 \Pr\!\left\{a_-\frac{n^{1-\alpha}}{\sqrt{\log n}}
       \le |E(z^1)|\le
       a_+\frac{n^{1-\alpha}}{\sqrt{\log n}}\right\}\longrightarrow1.
\]
For $\alpha\ge1$, $z^1$ equals the true reference-fiber factors with
probability $1-o(1)$.
\item Use $z^1$ as the reference in the capped component repair, and
retain $z^0$ as an additional feasible candidate. The result attains
$\OPT(H,Y)$ with probability $1-o(1)$ at every fixed $c>0$.
All acquisition bounds are unchanged, and total work remains
$O((m+n)\log(n+2))$ word operations on every input.
For every fixed integer $J>1/\alpha$, a suitable constant repair cutoff
has components of size less than $J$ with probability $1-o(1)$.
\end{enumerate}
\end{theorem}

The component conclusion concerns a suitable fixed cutoff reached by the
adaptive sweep. Later cutoffs may produce larger components. The algorithm
continues to enforce its stated work cap.

\subsection{Uniform control of the reused observations}

We first record how the existing joint estimate controls exceptional
probabilities strongly enough for the expectation claim. Put
$\beta=\tau\alpha>0$. For any fixed integer $k$, a root with $k$ distinct
neighbors in $E(z^0)$ has a witness of $j\le k$ distinct observed triples.
Each witness triple contains the root and one or two designated neighbors.
There are $3n$ root labels and at most $C_k(n^2)^j$ labeled witnesses
with $j$ triples; the finitely many choices of roles are included in $C_k$.
The prescribed-edge factor from Theorem~\ref{thm:joint-localization}
cancels the $n^{2j}$ choices, leaving
\begin{equation}
 \Pr\{\exists v:\#(N(v)\cap E(z^0))\ge k,\ \mathcal G_n\}
       \le n^{1-\beta k+o(1)}.
 \label{eq:recycle-root-star}
\end{equation}
A triple serving two designated neighbors is counted once. The root is
unconstrained, so the calculation does not assume that it is correct.
The complement of $\mathcal G_n$ is superpolynomially small.

Let $\Delta_2$ be the largest number of observed triples containing a
fixed pair of distinct vertices. Only pairs in different blocks matter.
For every fixed integer $r\ge1$,
\begin{equation}
 \Pr\{\Delta_2\ge r\}
 \le 3n^2\binom nr\left(\frac{m}{n^3}\right)^r
 =O_{c,r}\!\left(n^{2-r}(\log n)^r\right).
 \label{eq:recycle-codegree}
\end{equation}
This follows by prescribing the $r$ triples and using their exact
without-replacement inclusion bound.
Choose fixed integers $k,r$ with
$\beta k>A+3$ and $r>A+4$. Outside an event of probability at most
$n^{-A}$ for large $n$, every root has fewer than $k$ erroneous neighbors
and $\Delta_2<r$. Every observation counted by $b_v(z^0)$ contains at
least one such neighbor. Hence $b_v(z^0)<kr$, proving the first part with
$C_A=2kr$.

For the learned-bound schedule, the pilot count and batch capacities are
predetermined by $n,m$. On its superpolynomial-confidence event the
estimated bounds lie in the fixed compact rectangle
$[\mu/2,\mu]\times[\rho/2,\rho]$, where $\mu=1-2q$ and
$\rho=1-2s$. The calibration argument is uniform on this rectangle and
provides a deterministic $1-\varepsilon_n$ lower bound with
$\varepsilon_n\to0$. The finite mixed-moment lemma permits the resulting
data-dependent seed count. Its root-star calculation is unchanged;
the pilot occupies $o(m)$ prescribed earlier observations. Thus
\eqref{eq:recycle-root-star} holds with the same $\beta$ for this schedule.
The exponentially small exhaustion probability of the global fair-bit
sampler can likewise be included in the exceptional event.

\subsection{The number of errors after reuse}

Put $r_n=n^{1-\alpha}/\sqrt{\log n}$. We need a fixed-threshold
signed-degree estimate that retains its logarithmic factor:
\begin{equation}
 \Pr\{D_v\le L\}=\Theta_{c,q,L}\!\left(
                    \frac{n^{-\alpha}}{\sqrt{\log n}}\right)
 \quad\text{for every fixed real }L.
 \label{eq:recycle-degree-tail}
\end{equation}
It holds both for the uniform size-$m$ mask and for Bernoulli masking
with inclusion probability $p=m/n^3$.

For the upper bound, condition on the unsigned degree $d$, and write
$\theta=\operatorname{arctanh}(1-2q)>0$ and
$b=2\sqrt{q(1-q)}\in(0,1)$. Exponential tilting to $d$ fair signs gives
\begin{equation}
 \Pr\{D_v\le L\mid d\}
 =b^d\E_{1/2}[e^{\theta D_v}\mathbf1_{D_v\le L}]
 \le C_q e^{\theta L}\frac{b^d}{\sqrt{d+1}}.
 \label{eq:recycle-tilted-tail}
\end{equation}
Indeed, every fair-binomial mass is $O((d+1)^{-1/2})$, and its exponential
weights below $L$ form a convergent geometric series. The constant $C_q$
is independent of the real threshold $L$.
For $d\ge\kappa\log n$, use the denominator and the moment
$\E b^d\le[1-(1-b)/n]^m$ in the fixed-size model. For
$d<\kappa\log n$, use $b^d\le n^\kappa(b/e)^d$, with
$\kappa=cb(1-e^{-1})/2>0$. This contribution is
$O(n^{-\alpha-\kappa})$. The binomial degree under Bernoulli masking has
$\E z^d=[1-p(1-z)]^{n^2}\le e^{-(m/n)(1-z)}$, which gives the same
bounds. Thus the upper estimate in \eqref{eq:recycle-degree-tail} holds;
moreover its constant is at most a fixed multiple of $e^{\theta L}$.

For the lower bound, choose a fixed positive integer $r>\max\{0,-L\}$,
put $\lambda_n=m/n$, and write
$\nu_n=\lambda_n\sqrt{q(1-q)}$. For each integer
$\ell$ in $[\nu_n-\sqrt{\nu_n},\nu_n+\sqrt{\nu_n}]$, require
$\ell$ positive and $\ell+r$ negative observations incident to $v$.
This forces $D_v=-r<L$. In the fixed-size model its exact probability is
\[
 P_\ell=
 \frac{\binom{n^2}{2\ell+r}
             \binom{n^3-n^2}{m-2\ell-r}}{\binom{n^3}{m}}
 \binom{2\ell+r}{\ell}(1-q)^\ell q^{\ell+r}.
\]
Uniformly over this interval, expansion of the falling factorials gives
\begin{equation}
 P_\ell=(1+o(1))e^{-\lambda_n}
 \frac{\lambda_n^{2\ell+r}(1-q)^\ell q^{\ell+r}}
      {\ell!(\ell+r)!}.
 \label{eq:recycle-witness-prob}
\end{equation}
The errors in this multiplicative approximation are
$O(m^2/n^3+(\ell+r)^2/n^2+\lambda_n/n)=o(1)$.
The Bernoulli model gives the same expression by expanding its exact
multinomial probability. Stirling's bounds, uniformly for
$|\ell-\nu_n|\le\sqrt{\nu_n}$, give
\[
 \frac{\nu_n^{2\ell+r}}{\ell!(\ell+r)!}
            =\Theta_{q,r}(e^{2\nu_n}/\nu_n).
\]
Each event in \eqref{eq:recycle-witness-prob} therefore has probability
$\Theta_{c,q,r}(n^{-\alpha}/\log n)$.
The events are disjoint and there are $\Theta(\sqrt{\log n})$ of them.
Their sum proves the lower bound in \eqref{eq:recycle-degree-tail}.

Choose $A>\alpha+2$ in the first part of the theorem. The deterministic
sandwich \eqref{eq:recycle-sandwich} and
\eqref{eq:recycle-degree-tail}, with an exceptional probability at most
$n^{-A}$, give the matching marginal orders. Summing over $3n$ vertices
proves the expectation statement, even when that expectation tends to
zero. In particular, for $\alpha\ge1$ the expectation tends to zero,
which proves exact source recovery including the boundary.

For $0<\alpha<1$, use Bernoulli masking first. Within each block the
incident location sets are disjoint, so its indicators $\mathbf1_{D_v\le L}$
are independent for every fixed $L$. Their common probability has the
order in \eqref{eq:recycle-degree-tail}, and $r_n\to\infty$.
Chernoff's inequality puts their counts between positive constant
multiples of $r_n$ except with probability $\exp[-\Omega(r_n)]$.
Use $L<-C_A$ for the lower count and $L=C_A$ for the upper count, and
union over the three blocks. Conditioning on a total of $m$ sampled
locations costs only $O(\sqrt m)$ and recovers the specified uniform
mask. Since $\sqrt m\exp[-\Omega(r_n)]=o(1)$, the simultaneous count
bounds survive conditioning. Intersect with the sandwich event to prove
the claimed high-probability bounds for $|E(z^1)|$.

\begin{corollary}[The scale of the empirical correction]
\label{cor:correction-scale}
For $0<\alpha<1$ there are constants $g_-,g_+>0$ such that
\[
 \Pr\!\left\{g_-r_n\le\OPT(H,Y)-F_Y(T)\le g_+r_n\right\}
          \longrightarrow1,
 \qquad r_n=\frac{n^{1-\alpha}}{\sqrt{\log n}}.
\]
\end{corollary}
\begin{proof}
Flipping all negative-degree vertices in one block increases the source
score by at least twice their number, which is at least a positive
constant times $r_n$ with high probability by the preceding count bound.

For the upper bound, intrinsic localization gives an optimal flip set
$S\subseteq W_{4L_0}^*$ whose components have fewer than $L_0$ vertices,
for a fixed constant $L_0$, with probability $1-o(1)$. Work also on
$\Delta_2\le2$. Its gain has the deterministic bound
\begin{align*}
 -2J(S)
 &= -2\sum_{v\in S}D_v
       +4\sum_{e:\,|e\cap S|\ge2}\eta_e\\
 &\le2\sum_v(-D_v)_++4(L_0-1)|W_{4L_0}^*|.
\end{align*}
Every edge in the second sum contains a pair from one component; the
codegree bound gives at most $2\sum_{S_i}\binom{|S_i|}{2}
\le(L_0-1)|S|$ such edges. The weak-vertex count is $O(r_n)$ with high
probability.

It remains to control $\sum_v(-D_v)_+$. Equation
\eqref{eq:recycle-tilted-tail}, averaged over the degree as above, gives
uniformly for all integers $j\ge1$
\[
 \Pr\{(-D_v)_+\ge j\}
 \le C_{c,q}e^{-\theta j}\frac{n^{-\alpha}}{\sqrt{\log n}}.
\]
Hence, for $t=\theta/2$,
$\E e^{t(-D_v)_+}\le1+C'_{c,q}n^{-\alpha}/\sqrt{\log n}$.
In the Bernoulli model these variables are independent within each block,
so exponential Markov bounds each block sum by a fixed multiple of $r_n$
except with probability $\exp[-\Omega(r_n)]$. Union over the blocks and
condition on the total mask size as before. This proves the upper bound.
\end{proof}

\subsection{Exact repair with the intrinsic component exponent}

For a fixed threshold $L$, write $W_L^*=\{v:D_v\le L\}$.
The true-degree calculation in Appendix~\ref{app:local} gives, for fixed
vertex sets $S$ and prescribed triples $F$,
\[
 \Pr\{S\subseteq W_L^*,F\subseteq H\}
 \le(m/n^3)^{|F|}n^{-\alpha|S|+o(1)}.
\]
Consequently, for every fixed integer $J>1/\alpha$, all components of
$W_L^*$ have size less than $J$ with probability $1-o(1)$. A root-star
count also shows that every vertex has fewer than $k_*$ neighbors in
$W_L^*$, for any fixed integer $k_*>1/\alpha$. These facts hold for each
fixed $L$, with the same sufficient $J$ and $k_*$.

Choose a fixed sandwich threshold $C$ with probability $1-o(1)$.
Intrinsic localization supplies an optimizer whose difference from the
source is contained in $W_{4L_0}^*$ for a fixed $L_0$.
Set $u_*=4J$ and choose the constant
\[
 L=\max\{C,4L_0,u_*+4k_*\}.
\]
Take the component and root-star events for this predetermined $W_L^*$,
and $\Delta_2\le2$, simultaneously. The difference from $z^1$ to that
optimizer lies in $W_L^*$ and has components of fewer than $J$ vertices.
Optimality against undoing a single sign in such a component gives its
$z^1$-signed degree at most $4(J-1)$, exactly as in the original repair
proof. Thus the observable cutoff $u_*$ contains that difference.

At a vertex outside $E(z^1)$, the $z^1$-signed degree differs from
$D_v$ by at most $4k_*$: there are fewer than $k_*$ erroneous neighbors,
with at most two observations per pair. It follows that the observable
weak set at cutoff $u_*$ is contained in
$E(z^1)\cup W_{u_*+4k_*}^*\subseteq W_L^*$.
Its components therefore have fewer than $J$ vertices. Exact component
optimization finds the global optimum. The growing cutoff, component cap,
and work budget eventually permit this fixed successful stage. Earlier
stages are nested subsets and cost $O_{c,q,s}(m+n)$ altogether.
The additional full-target update and score evaluation take $O(m+n)$
word operations; all reading and all-input resource claims follow.

\subsection{Two passes suffice when the weak vertices are isolated}

\begin{corollary}[Exact optimization by two reused majority passes]
\label{cor:two-pass-exact}
Under the assumptions of Theorem~\ref{thm:recycling}, suppose
$\alpha>1/2$. Then
\[
 F_Y(z^2)=\OPT(H,Y)
\]
with probability $1-o(1)$. Thus the cascade followed by two simultaneous
full-target majority passes is an exact empirical optimizer with high
probability in this regime, without component enumeration. It retains
all acquisition bounds and the all-input work bound. When
$1/2<\alpha<1$, the generating source itself remains suboptimal by positive constant
multiples of $r_n$ with high probability (Corollary~\ref{cor:correction-scale}).
\end{corollary}

\begin{proof}
Choose the first-pass sandwich constant $C$. Fix an integer
$k_*>1/\alpha$, let $L_0$ be the intrinsic localization constant, and set
$L=\max\{C,4L_0,4k_*\}$. These are fixed before applying any
weak-set probability bound. With probability $1-o(1)$, every vertex has
fewer than $k_*$ neighbors in $W_L^*$, $\Delta_2\le2$, and $W_L^*$
contains no adjacent pair: the last assertion follows from the
connected-pair bound $n^{1-2\alpha+o(1)}$.

The first recycled error set is contained in $W_L^*$. A vertex in
$W_L^*$ has no erroneous neighbor, so its next field in source coordinates
is exactly $D_v$. A vertex outside $W_L^*$ has $D_v>L\ge4k_*$ and
its field differs from $D_v$ by at most $4k_*$, hence its next sign is
correct. Therefore $z^2$ flips precisely the negative-degree vertices,
with arbitrary choices at the zero-degree vertices, and all these
vertices lie in the independent set $W_L^*$.

Some global optimizer differs from the source only inside $W_L^*$ by
intrinsic localization. Every observed triple meets this set in at most
one vertex, so optimizing on it separates into independent sign choices.
The choices just made attain that restricted, and hence global, optimum.
\end{proof}

A simultaneous pass can decrease a feasible score on an arbitrary input.
The full algorithm therefore retains the best of its earlier and newly
scored candidates. The result above concerns the stated random target;
its proof supplies no sparse all-input optimality certificate. The
threshold $\alpha>1/2$ is a sufficient regime for two-pass exactness. It
is not a proved failure threshold for the procedure at smaller $\alpha$.

\section{One acquisition for adaptively reweighted objectives}
\label{app:task-transfer}

The observations used to construct an estimate need not be acquired again
when the scoring rule changes. We prove this for an entire neighborhood of
the recorded objective, allowing the new weights to depend on every previous
observation and computed answer. The new objective may have a different
maximizer. The guarantee concerns solving each objective, rather than
preserving one optimizer under perturbation, as in classical instance
stability \citep{bilu2012stable}.

For a deterministic nonnegative budget $\Gamma_n$, define
\begin{equation}
 \mathcal U_{\Gamma_n}(H)=\left\{\xi\in\mathbb R^H:
           \max_v\sum_{e\ni v}|\xi_e|\le\Gamma_n\right\},\qquad
 F_\xi(x)=\sum_{e\in H}(Y_e+\xi_e)\prod_{v\in e}x_v.
 \label{eq:task-neighborhood}
\end{equation}
The locations $H$ remain fixed. Coefficients may change sign or vanish.
Let $\mathcal M_\xi$ be the simultaneous majority update for $F_\xi$,
retaining the old sign at a zero field. Acquire $z^0$ from the original
unweighted target, once. Start each new objective from that same stored
estimate, rather than from the answer to the preceding objective, and define
\[
 z^1_\xi=\mathcal M_\xi(z^0),\qquad
 z^2_\xi=\mathcal M_\xi(z^1_\xi).
\]
The error influence of $z^0$ is controlled under the original law; no
independence is assumed between $\xi$ and that estimate.

\begin{theorem}[Uniform exact optimization after acquisition]
\label{thm:task-transfer}
Under the retained-source assumptions of Theorem~\ref{thm:recycling}, let $\Gamma_n\ge0$ be a deterministic sequence with
$\Gamma_n=o(\log n)$, and fix an integer $J>1/\alpha$.
There is one event with probability
$1-o(1)$ on which the following statements hold simultaneously for every
$\xi\in\mathcal U_{\Gamma_n}(H)$.
\begin{enumerate}
\item If $\alpha>1/2$, then
$F_\xi(z^2_\xi)=\max_x F_\xi(x)$.
\item At every fixed $\alpha>0$, weighted capped component repair about
$z^1_\xi$ attains $\max_x F_\xi(x)$. The cutoff $u_n=\lceil4J+2\Gamma_n\rceil$ has components of size less than $J$.
The existing growing cutoff and work cap eventually reach that stage.
\end{enumerate}
No additional source readings are used. Consequently, any sequence of such
objectives may be chosen after acquisition, including choices depending on
all earlier answers, without a probability union bound over that sequence.
For dyadic coefficients with a common denominator $2^b$, each two-pass query
uses $O(m+n)$ exact additions and comparisons; capped repair uses
$O((m+n)\log(n+2))$. The integers have
$O(b+\log(m+n)+\log(1+\Gamma_n))$ bits. The original acquisition costs are
paid once. Dyadic thresholds use the same scaling after matching the
denominators. Arbitrary rational thresholds are compared by cross multiplication,
with their encoded lengths and arithmetic costs charged separately.
\end{theorem}

The coefficient array is an explicitly supplied part of each query; forming
that array is an additional cost. Reading it and checking a supplied rational
local budget take $O(m+n)$ arithmetic operations. A sequence of queries pays
the sum of its per-query costs. The postprocessor does not need $\Gamma_n$:
its existing sweep eventually reaches the cutoff used in the proof.

Every fixed budget $\Gamma_n=\Gamma<\infty$ is included. For
$b=O(\log(n+m))$, the arithmetic counts are word bounds because
$\log(1+\Gamma_n)=O(\log\log n)$ eventually. In general the displayed bit length must be charged. The theorem
provides an asymptotic correctness event under the target law; it does not
supply an observable certificate of membership in that event. The incidence structure remains fixed. The theorem does not cover
arbitrary added interactions or perturbation budgets of order $\log n$.

\subsection{A deterministic principle for signed factor graphs}

For a finite hypergraph with real coefficients $a_e$, fix a nonnegative
scalar budget $\Gamma$ (using the same definition of $\mathcal U_\Gamma$) and an arbitrary
reference assignment $\sigma$. Write
\[
 D_v=\sum_{e\ni v}a_e\prod_{u\in e}\sigma_u,
 \qquad
 I_v(z)=\sum_{\substack{e\ni v:\prod_{u\in e\setminus\{v\}}
                         z_u\sigma_u=-1}}|a_e|.
\]
The field for the perturbed coefficients $a_e+\xi_e$, expressed relative
to $\sigma_v$, satisfies
\begin{equation}
 |M_v^\xi(z)-D_v|\le 2I_v(z)+\sum_{e\ni v}|\xi_e|.
 \label{eq:weighted-reuse-field}
\end{equation}
This holds for every arity, every estimate, and every dependence among
coefficients and estimates: subtract the reference field term by term.
Define
\[
 h_v(W)=\sum_{\substack{e\ni v:(e\setminus\{v\})\cap W\ne\varnothing}}
                   |a_e|.
\]

\begin{lemma}[Exactness through an isolated ambiguity set]
\label{lem:isolated-transfer}
Suppose $2I_v(z^0)\le C$ for every $v$. Let $W$ contain
$\{v:D_v\le C+\Gamma\}$, and suppose each edge meets $W$ at most once.
Assume $D_v>2h_v(W)+\Gamma$ for $v\notin W$.
For any $\xi\in\mathcal U_\Gamma(H)$ whose objective has an optimizer
agreeing with $\sigma$ outside $W$, the second simultaneous update from
$z^0$ is globally optimal for that objective.
\end{lemma}
\begin{proof}
Equation~\eqref{eq:weighted-reuse-field} puts every first-pass error in $W$.
At a vertex outside $W$, the second field differs from $D_v$ by at most
$2h_v(W)+\Gamma$ and is positive. At a vertex in $W$, every other endpoint
of each incident edge is correct, so the second field is its exact
perturbed reference field. The second pass therefore maximizes the objective
with the outside assignment fixed to $\sigma$. Edgewise isolation makes this
optimization separable. The assumed localized optimizer shows that its
value is the unrestricted optimum.
\end{proof}

The last hypothesis is a structural localization requirement, not a
conclusion inferred from small estimation error alone. We next derive it
simultaneously for the full perturbation family from the stated random law.

\subsection{Uniform localization from the original target law}

Work in the true source gauge. Put
$J(S)=\sum_{e\in H}\eta_e\mathbf1\{|e\cap S|\text{ is odd}\}$ and
$\lambda=\tfrac12\log((1-q)/q)>0$. The corresponding perturbed flip cost
$J_\xi(S)$ obeys, deterministically,
\begin{equation}
 |J_\xi(S)-J(S)|
 \le\sum_{e:\,|e\cap S|\text{ odd}}|\xi_e|
 \le\Gamma_n |S|.
 \label{eq:task-cut-bound}
\end{equation}
Thus any component of a perturbed optimal flip set must satisfy
$J(S)\le\Gamma_n|S|$. This implication holds for all admissible perturbations
on the same sample.

Fix $\delta\in(0,1/2)$. If a tensor has no source-gauge representative
supported on fewer than $\delta n$ vertices, its grid disagreement
fraction is at least $\delta/3$. Indeed, a smaller fraction forces the
product of the three factor correlations to exceed $1-2\delta/3$.
Changing two whole-block signs when necessary makes all three correlations
positive; each then exceeds $1-2\delta/3$, giving fewer than $\delta n$
flips in total. Maclaurin's inequality and exponential Markov therefore give
\[
 \Pr\{\exists S:\ d(S)\ge\delta/3,\ J(S)\le3\Gamma_n n\}
 \le 2^{3n}\exp\{3\lambda\Gamma_n n-m\mathcal I(1-2q)\delta/3\}
 =o(1),
\]
because $\Gamma_n n=o(n\log n)$. Every admissible perturbed objective consequently has an optimal gauge
representative of size less than $\delta n$, on one common event.

For a connected set with $t\le\delta n$ vertices, the incidence-tree
calculation in Appendix~\ref{app:local} gives the strengthened estimate
\begin{equation}
 \Pr\{\exists\text{ connected }S:\ |S|=t,\ J(S)\le\Gamma_n t\}
 \le 3n(C_1e^{\lambda\Gamma_n}\log n)^t
                          n^{-\alpha(1-\delta)t}.
 \label{eq:task-cluster-bound}
\end{equation}
The only change is the Markov factor $e^{\lambda\Gamma_n t}$; the prescribed
edge probabilities and tree count are unchanged. Put
$q_n=C_1e^{\lambda\Gamma_n}\log n\,n^{-\alpha(1-\delta)}$. Since
$\Gamma_n=o(\log n)$, $q_n=n^{-\alpha(1-\delta)+o(1)}$. Choose a
fixed integer $L_0$ with $L_0\alpha(1-\delta)>2$. Geometric summation
bounds the probability over $L_0\le t\le\delta n$ by
$3nq_n^{L_0}/(1-q_n)=o(1)$. This is uniform over the entire growing
range of component sizes.

Also impose the original event that every pair of vertices has codegree at
most two. If $v$ is flipped in an optimal component $S$ with $|S|<L_0$,
comparison with undoing that single flip gives
\begin{equation}
 D_v\le4(|S|-1)+\Gamma_n<4L_0+\Gamma_n.
 \label{eq:task-optimum-weak}
\end{equation}
There are at most $2(|S|-1)$ base observations with another flipped
endpoint; their field contribution changes by at most two each. The total
coefficient perturbation in that field is at most $\Gamma_n$.
Therefore all perturbed objectives have an optimal representative supported
in the same set $W^*_{4L_0+\Gamma_n}$. No perturbed optimizer was used in
constructing the probability event.

\subsection{Weak-set geometry at growing thresholds}

The common weak set now has a threshold of order $1+\Gamma_n$, so the
fixed-threshold estimates alone are insufficient. The following consequence
of the finite mixed-moment bound makes the required uniformity explicit.

\begin{lemma}[Weak sets below the logarithmic field scale]
\label{lem:sublog-weak-geometry}
Let $L_n\ge0$ be deterministic with $L_n=o(\log n)$. For fixed integers
$s,f\ge0$, every $s$-vertex set $S$ and every set $F$ of $f$ distinct triples
satisfy
\begin{equation}
 \Pr\{S\subseteq W^*_{L_n},F\subseteq H\}
 \le (m/n^3)^f n^{-\alpha s+o(1)},
 \label{eq:sublog-weak-moment}
\end{equation}
uniformly over their labels. Consequently, for every fixed integer
$J>1/\alpha$, all components of $W^*_{L_n}$ have fewer than $J$ vertices
with probability $1-o(1)$. For every fixed integer $k_*>1/\alpha$, each
vertex has fewer than $k_*$ neighbors in that set with probability $1-o(1)$.
If $\alpha>1/2$, no observed triple contains two vertices of $W^*_{L_n}$
with probability $1-o(1)$.
\end{lemma}
\begin{proof}
Apply Lemma~\ref{lem:direct-mixed-moment} with $A=\varnothing$, $B=S$,
one color of capacity $t=m$, deterministic all-positive calibration
arrays, and $\mathcal G$ the whole sample space.
With $\lambda=\operatorname{arctanh}(1-2q)$ and $w=e^{3\lambda}$, its
finite bound is
\[
 \Pr\{S\subseteq W^*_{L_n},F\subseteq H\}
 \le\frac{(m)_f}{(n^3)_f}w^f
 \exp\left\{\lambda sL_n-\frac{ms}{n}\mathcal I(1-2q)
 +\frac{fs}{n}+\frac{4ws(s-1)m}{n^2}
 +\frac{2w(m^2+mf)}{n^3}\right\}.
\]
For fixed $s,f$ the last three terms are $o(1)$. The threshold term is
$o(\log n)$, and $(m)_f/(n^3)_f\le(m/n^3)^f$, proving
\eqref{eq:sublog-weak-moment}. Its error terms do not depend on the
selected labels. Sum it over the incidence-tree witnesses for $J$ weak
vertices to obtain $n^{1-\alpha J+o(1)}$; a larger component contains a
connected $J$-vertex subset. For a root with $k_*$ weak neighbors, prescribe
the distinct triples joining those neighbors to the root. With $j\le k_*$
triples there are at most $C_{k_*}n(n^2)^j$ labeled witnesses, and their
inclusion factor cancels the $n^{2j}$ term. The resulting probability is
$n^{1-\alpha k_*+o(1)}$. The choice $J=2$ gives the last assertion.
\end{proof}

\subsection{Uniform postprocessing and its cost}

Theorem~\ref{thm:recycling} supplies a fixed $C$ such that
$\max_v 2b_v(z^0)\le C$ with probability $1-o(1)$.
Hence, simultaneously for every admissible $\xi$,
\[
 E(z^1_\xi)\subseteq W^*_{C+\Gamma_n}.
\]
Choose a fixed integer $k_*>1/\alpha$. For each predetermined $L_n=o(\log n)$,
Lemma~\ref{lem:sublog-weak-geometry} gives fewer than $k_*$ neighbors in
$W_{L_n}^*$ at every vertex with probability $1-o(1)$. Together with codegree at most
two, this bounds every first-pass error influence by $4k_*$ in field units.

When $\alpha>1/2$, take
$L_n=\max\{C+\Gamma_n,4L_0+\Gamma_n,4k_*+\Gamma_n\}$.
The growing-threshold lemma makes $W_{L_n}^*$ independent with probability
$1-o(1)$. For a vertex outside it, $D_v>L_n\ge4k_*+\Gamma_n$.
Lemma~\ref{lem:isolated-transfer} applies simultaneously to all $\xi$ and
proves the two-pass statement.

For arbitrary $\alpha>0$, fix an integer $J>1/\alpha$, set
$u_*=\lceil4J+2\Gamma_n\rceil$, and now choose
\[
 L_n=\max\{C+\Gamma_n,4L_0+\Gamma_n,u_*+4k_*+\Gamma_n\}.
\]
Here $L_n=o(\log n)$, so Lemma~\ref{lem:sublog-weak-geometry} implies
that every component of $W_{L_n}^*$ has
fewer than $J$ vertices. The difference between $z^1_\xi$ and a perturbed
optimizer lies in $W_{L_n}^*$. Optimality against undoing one flip, now measured
relative to $z^1_\xi$, bounds its signed weighted degree by
$4(J-1)+2\Gamma_n<u_*$. The extra $2\Gamma_n$ bounds the twice-counted
perturbed interaction weights. Thus the observable cutoff $u_*$ contains
the difference. Conversely, a vertex outside $E(z^1_\xi)$ has observed
signed degree differing from $D_v$ by at most $4k_*+\Gamma_n$, so the cutoff
set is contained in $W_{L_n}^*$. Its components have size less than $J$ and
can be optimized exactly. Every smaller cutoff is a subset, so the
increasing sweep reaches this successful stage within
$O_{c,q,s,J}((1+\Gamma_n)(m+n))$ work on the common event. Because $u_*=o(\log n)$ and $J$ is fixed, the cutoff and component
limits eventually admit this stage, and its total work is
$o((m+n)\log n)$. Thus the usual fixed-factor work budget also admits it
for all sufficiently large $n$. That budget enforces
$O((m+n)\log(n+2))$ on every input.

For common denominator $2^b$, multiply every coefficient by $2^b$.
A signed-degree cutoff \(u\) is applied to its integer numerator as \(\widetilde D_v\le 2^b u\); scaling the coefficients never leaves the cutoff unscaled. Updates, signed degrees, component tables, Walsh transforms and objective
comparisons then use integers only. Every table entry is a signed sum of
at most $m$ weights, bounded in magnitude by $m(1+\Gamma_n)2^b$.
This proves the stated bit length and arithmetic counts. If separately
encoded rational denominators are allowed instead, their common product
has at most the sum of their bit lengths; clearing denominators gives a
polynomial bit bound, but need not preserve logarithmic word size.

\subsection{Why local influence alone is insufficient}

A finite counterexample isolates the role of ambiguity geometry. Take four
vertices per block. Use all $3\times3\times4=36$ positive triples on
$A\setminus\{a_0\}$, $B\setminus\{b_0\}$, and $C$. Add
\[
 -(a_0,b_0,c_0),\quad +(a_0,b_1,c_1),\quad -(a_0,b_2,c_2),
 \quad +(a_1,b_0,c_1),\quad -(a_2,b_0,c_2).
\]
Start from the all-positive source, with no estimation error. The two weak
vertices $a_0,b_0$ both have signed degree $-1$; all others have degree at
least seven. A simultaneous pass reverses both weak signs, and the next
reverses them back. The strong signs remain positive throughout. Every
iterate has score $35$; reversing just one weak sign has score $37$ and
is globally optimal. To verify optimality, the positive core has value
$36$ or at most $18$. In the latter case the five additional terms raise
the total to at most $23$. In the former case a whole-block gauge puts the
core factors at $+1$, the private terms cancel, and the remaining objective
is $36-a_0b_0\le37$.

Thus even perfect source knowledge does not make synchronous updates an
exact optimizer on every signed hypergraph. The example is a deterministic
obstruction; it does not prove a random-model failure threshold at
$\alpha=1/2$.

\section{One observable reduction for all later objectives}
\label{app:common-reduction}

The weighted updates in Appendix~\ref{app:task-transfer} revisit the original
data for each objective. A single unweighted update instead suffices to
identify the variables that any admissible objective may need to change.
The resulting components, their objective tables, and an index of the
observations can all be constructed before the later coefficients are known.

Acquire $z^0$ once and compute the unweighted iterate
$r=z^1=\mathcal M_0(z^0)$. For an observed triple write
$T_e(r)=\prod_{v\in e}r_v$, and define the observable signed degrees
\[
 d_v(r)=\sum_{e\ni v}Y_eT_e(r).
\]
The prescribed local budget remains relative to the original coefficients:
$\xi\in\mathcal U_{\Gamma_n}(H)$ as defined in
\eqref{eq:task-neighborhood}. In a sequence of updates, each accumulated
coefficient array must belong to this same set.

\begin{theorem}[A common observable reduction]
\label{thm:common-reduction}
Under the assumptions of Theorem~\ref{thm:recycling}, fix an integer
$J\ge2$ with $J>1/\alpha$, and a deterministic sequence
$\Gamma_n\ge0$ with $\Gamma_n=o(\log n)$. Set
\begin{equation}
 u_n=4(J-2)+\lfloor\Gamma_n\rfloor,
 \qquad U_n=\{v:d_v(r)\le u_n\}.
 \label{eq:common-observable-set}
\end{equation}
There is one event with probability $1-o(1)$ on which:
\begin{enumerate}
\item Every global optimizer of every
$F_\xi$, $\xi\in\mathcal U_{\Gamma_n}(H)$, has a whole-block gauge
representative agreeing with $r$ outside $U_n$.
\item Every component of the interaction graph induced by $U_n$ has fewer
than $J$ vertices. In particular, when $\alpha>1/2$, one may take $J=2$;
then $U_n=\{v:d_v(r)\le\lfloor\Gamma_n\rfloor\}$ and its vertices
share no observed triple.
\item If $0<\alpha<1$, then
$|U_n|=n^{1-\alpha+o(1)}$ with high probability. If the budget
sequence is bounded, the sharper bound is
\[
 |U_n|=\Theta_{c,q,s,J,\sup_n\Gamma_n}
       \!\left(\frac{n^{1-\alpha}}{\sqrt{\log n}}\right).
\]
If $\alpha>1$, the set is empty with high probability, including for growing
sublogarithmic budgets. At $\alpha=1$ it is empty for every bounded budget;
for general sublogarithmic budgets the bound is $|U_n|\le n^{o(1)}$.
\end{enumerate}
Thus the same observable restriction $x_{U_n^c}=r_{U_n^c}$ preserves
the optimum of every allowed objective. Both the restriction and its component
partition are computed once, without further source readings. The coefficient
arrays may subsequently be chosen adaptively from all observations and answers.
\end{theorem}

The component bound $J$ and the budget envelope are inputs to this reduction.
The acquisition procedure can remain parameter-free. Choosing $J$ requires a
valid lower bound on $\alpha$, or a fixed choice whose stated density
condition holds; the theorem does not infer that bound from the observed
component sizes.

\subsection{The one-coordinate comparison}

The cutoff in \eqref{eq:common-observable-set} follows by retaining the
perturbation at the final assignment, rather than bounding it twice during
the comparison. The following deterministic identity also applies to general
signed hypergraphs.

\begin{lemma}[One-coordinate cutoff]
\label{lem:common-one-coordinate}
Let a finite hypergraph have base coefficients $a_e\in\mathbb R$, let $r$
be any sign assignment, and let $x$ maximize the objective with coefficients
$a_e+\xi_e$. Put $S=\{v:x_v\ne r_v\}$ and
$d_v^a(r)=\sum_{e\ni v}a_eT_e(r)$. For every $v\in S$,
\begin{equation}
 d_v^a(r)\le
 2\!\sum_{\substack{e\ni v:\ |(e\setminus\{v\})\cap S|\text{ odd}}}
                   a_eT_e(r)
       +\sum_{e\ni v}\xi_eT_e(x).
 \label{eq:common-flip-comparison}
\end{equation}
If $|a_e|=1$, every pair has codegree at most two, and the component
of $S$ containing $v$ has $b$ vertices, then
\[
 d_v^a(r)\le4(b-1)+\sum_{e\ni v}|\xi_e|.
\]
\end{lemma}
\begin{proof}
Undoing the sign of $x_v$ cannot improve an optimizer, so
$\sum_{e\ni v}(a_e+\xi_e)T_e(x)\ge0$. Since $v\in S$, the base part of
this sum is
\[
 -d_v^a(r)+
 2\!\sum_{\substack{e\ni v:\ |(e\setminus\{v\})\cap S|\text{ odd}}}
                 a_eT_e(r).
\]
This proves \eqref{eq:common-flip-comparison}. Every edge in the correction
sum contains another vertex from the same component of $S$. The codegree
bound gives at most $2(b-1)$ such edges. Their base contributions have
absolute value one, and the perturbation sum is bounded by its local
$\ell_1$ norm.
\end{proof}

This argument gives $4(J-2)+\Gamma_n$ when $b<J$. The base degrees are
integers, which gives the floor in \eqref{eq:common-observable-set}.

\subsection{Proof of the common probability event}

Write $D_v$ for the true signed degree in source coordinates and
$W_L^*=\{v:D_v\le L\}$. Theorem~\ref{thm:recycling} and its field comparison
give a fixed $C$ for which
\[
 E(r)\subseteq W_C^*
\]
with probability $1-o(1)$. Here $E(r)$ is measured in the true reference-fiber
gauge. Fix $\delta\in(0,1/2)$. The uniform localization argument in
Appendix~\ref{app:task-transfer} gives a fixed $L_0$ such that every
perturbed optimizer has a source-gauge representative supported in
$W_{4L_0+\Gamma_n}^*$. We record why the quantifier covers every optimizer.
For any optimal flip set $S$, comparison with the source gives
$J_\xi(S)\le0$, and \eqref{eq:task-cut-bound} implies
$J(S)\le\Gamma_n|S|\le3\Gamma_n n$. The macroscopic union bound in that
appendix therefore gives every optimal tensor a representative with
$|S|<\delta n$. For each interaction component $A$ of this representative,
additivity of the flip cost and comparison with undoing $A$ give
$J_\xi(A)\le0$, hence $J(A)\le\Gamma_n|A|$. The event in
\eqref{eq:task-cluster-bound} excludes every such component with
$L_0\le|A|\le\delta n$. Finally, the single-flip comparison
\eqref{eq:task-optimum-weak} puts every changed vertex in
$W_{4L_0+\Gamma_n}^*$. These events are constructed from the original
target and exclude the offending sets uniformly over the coefficient array.

Fix an integer $k_*>1/\alpha$ and set
\begin{equation}
 L_n=\max\{C,4L_0+\Gamma_n,u_n+4k_*\}.
 \label{eq:common-proof-threshold}
\end{equation}
It is deterministic and $o(\log n)$. By
Lemma~\ref{lem:sublog-weak-geometry}, every component of $W_{L_n}^*$ has
fewer than $J$ vertices and every vertex has fewer than $k_*$ neighbors
in that set, with probability $1-o(1)$. Work also on the event that every
pair has codegree at most two. All of these requirements concern the
single chosen threshold \eqref{eq:common-proof-threshold}; none depends
on a future query.

Let $x$ be any optimizer of any admissible objective, with the localized
source-gauge representative just described. Its difference
$S=\{v:x_v\ne r_v\}$ is contained in $W_{L_n}^*$. Every component of $S$
therefore has fewer than $J$ vertices. Lemma~\ref{lem:common-one-coordinate}
gives, for $v\in S$,
\[
 d_v(r)\le4(J-2)+\Gamma_n,
\]
and hence $S\subseteq U_n$. This establishes the common restriction for
all optimizers and all allowed coefficient arrays at once.

To control the observable set itself, a vertex in $E(r)$ already belongs to
$W_C^*$. At a vertex outside $E(r)$, the unweighted signed degree relative
to $r$ differs from $D_v$ only through incident triples with an erroneous
other endpoint. There are fewer than $k_*$ such neighboring vertices and
at most two triples per pair. Consequently
\[
 |d_v(r)-D_v|\le4k_*\qquad(v\notin E(r)).
\]
If such a vertex belongs to $U_n$, then $D_v\le u_n+4k_*$.
Thus $U_n\subseteq W_{L_n}^*$. Its components satisfy the same bound.
For $\alpha>1/2$, $J=2$ gives isolated vertices.

For completeness, the one-vertex case of
\eqref{eq:sublog-weak-moment} gives
$\E|W_{L_n}^*|\le n^{1-\alpha+o(1)}$. Markov's inequality shows that
for every fixed $\epsilon>0$,
\[
 \Pr\{|U_n|>n^{1-\alpha+\epsilon}\}=o(1).
\]
Equivalently, the $o(1)$ in the theorem can be chosen as a deterministic
sequence tending to zero: if the expectation exponent is bounded by
$1-\alpha+\delta_n$, with $\delta_n\ge0$ tending to zero, use the
threshold exponent $1-\alpha+2\delta_n+1/\sqrt{\log n}$.
This also proves the stated $n^{o(1)}$ bound at $\alpha=1$.
For $\alpha>1$, the expectation tends to zero, so $U_n$ is empty
with high probability.

When the budget is bounded, $L_n$ is bounded by a fixed constant. The
fixed-threshold count argument in Appendix~\ref{app:recycling} gives
$|W_{L_n}^*|=O(n^{1-\alpha}/\sqrt{\log n})$ with high probability for
$0<\alpha<1$. Specifically, use a fixed upper bound on $L_n$, apply
Chernoff's inequality within each block under Bernoulli masking, and
condition on exactly $m$ locations; the exponentially small count error
absorbs the $O(\sqrt m)$ conditioning factor.

For the matching lower bound, put $r_n=n^{1-\alpha}/\sqrt{\log n}$ and
let $Z_0$ be the nonisolated vertices with $D_v=0$.
Lemma~\ref{lem:zero-and-unit-counts} gives $|Z_0|=\Theta(r_n)$ with high
probability for every $0<\alpha<1$; its count is independent of this
reduction argument. Increase the fixed first-pass threshold $C$ to be
nonnegative. Let $B_C$ count adjacent pairs in $W_C^*$. Prescribing their
common triple and applying \eqref{eq:sublog-weak-moment} with $s=2,f=1$
gives
\[
 \E B_C\le3m\,n^{-2\alpha+o(1)}
          =n^{1-2\alpha+o(1)}=o(r_n).
\]
Thus $B_C=o_{\Pr}(r_n)$ by Markov's inequality. On the first-pass event,
any vertex of $Z_0$ with an erroneous neighbor is an endpoint of one of
these pairs. There are at most $2B_C$ such vertices. At each remaining
vertex, all other signs in its incident triples agree with the source,
so $d_v(r)=r_v\sigma_vD_v=0$, where $\sigma$ is the source reference.
Since $u_n\ge0$, it belongs to $U_n$. Hence $|U_n|\ge|Z_0|-2B_C=
\Omega(r_n)$ with high probability. This lower bound holds for every
allowed budget, and completes the fixed-budget $\Theta(r_n)$ estimate.
Together with the upper exponent, it also gives
$|U_n|=n^{1-\alpha+o(1)}$ for every allowed growing budget.

For $\alpha\ge1$ and bounded $\Gamma_n$, the cutoff $L_n$ is bounded, so the fixed-threshold marginal estimate gives
$\E|W_{L_n}^*|=O(n^{1-\alpha}/\sqrt{\log n})=o(1)$.
This proves the remaining size and emptiness claims and completes the
proof of Theorem~\ref{thm:common-reduction}.

The boundary qualification is substantive: at $\alpha=1$, a general
growing budget need not leave the reduction empty. The uniform tilted
tail bound gives emptiness whenever
$e^{\lambda\Gamma_n}/\sqrt{\log n}\to0$, where
$\lambda=\tfrac12\log((1-q)/q)$, because
$L_n\le\Gamma_n+O_{c,q,s,J}(1)$. Bounded budgets satisfy this condition.

\subsection{An exact finite representation of the restricted objectives}

The remaining construction is deterministic. It applies to every observed
input and every chosen $U$, whether or not the probability event holds.
Let $C_1,\ldots,C_h$ be the components of the interaction graph induced
by $U$. An edge meets at most one component, since any two of its endpoints
in $U$ are adjacent. Index the edges of $H$ from $1$ through $m$.

For each edge, store its component index (zero if it misses $U$), its
intersection mask in that component, and its fixed reference product
$t_e=T_e(r)$. Write a component configuration as a flip mask
$s\in\{0,1\}^{C_i}$. With current weights $w_e=Y_e+\xi_e$, define
\begin{align}
 B(w)&=\sum_{e:\,e\cap U=\varnothing} w_et_e,\nonumber\\
 A_i^w(s)&=\sum_{e:\,e\cap U\subseteq C_i,\ e\cap U\ne\varnothing}
       w_et_e(-1)^{\langle s,\one_{e\cap C_i}\rangle}.
 \label{eq:common-component-tables}
\end{align}
The inner product in the exponent is over $\mathbb F_2$. Direct substitution
gives the exact identity
\begin{equation}
 \max_{x:\,x_{U^c}=r_{U^c}}F_\xi(x)
   =B(w)+\sum_{i=1}^h\max_s A_i^w(s).
 \label{eq:common-table-optimum}
\end{equation}
A maximizing configuration in each table yields an actual feasible assignment.
Theorem~\ref{thm:common-reduction} identifies when the left side equals the
unrestricted optimum simultaneously for the entire objective family.

Graph traversal and edge indexing take $O(m+n)$ operations. The initial
unweighted table for a component is the Walsh transform of its histogram
of signed parity masks. It takes $O(|C_i|2^{|C_i|})$ operations, in addition
to the edge scan. Preprocessing therefore costs
\[
 O\!\left(m+n+\sum_i |C_i|2^{|C_i|}\right)
\]
and stores $O(m+n+\sum_i2^{|C_i|})$ words. When all components have fewer
than $J$ vertices these are $O_J(m+n)$; only the component-table portion
has size $O_J(|U|)$. The original edge index and reference assignment still
occupy $O(m+n)$ words.

For an all-input implementation, finish the graph traversal before allocating
tables. If a component has at least $J$ vertices, or exceeds the stored
cap $b_n=\max(12,\lceil\log_2(n+1)\rceil)$, return $r$ and its feasible
score with the reduction marked unavailable. Likewise stop if the declared
table and scan work would exceed
$\lceil64(m+3n)\log(n+2)\rceil$ units. The component sizes determine the
table cost before any exponential allocation; saturate the cost estimate at
the budget plus one. For each fixed $J$, the
successful preprocessing cost is eventually below this budget on the common
event. These guards give an all-input $O((m+n)\log(n+2))$ bound in
addition to the original acquisition guarantee. Passing them certifies
the finite restricted-table computation; it does not certify that the
restriction contains a global optimum on an arbitrary input.

\begin{proposition}[Exact updates from sparse coefficient changes]
\label{prop:incremental-common-reduction}
Suppose the common reduction has been constructed with component sizes
less than $J$. Fix a common dyadic denominator $2^b$ for the coefficient
stream. Supply each nonempty update as replacement values at $k\ge1$ distinct indexed
edges, and suppose the resulting coefficient array remains in
$\mathcal U_{\Gamma_n}(H)$. The restricted optimum value and a maintained
maximizing assignment can be updated in
\[
 O(k2^J)
\]
exact arithmetic operations, with $O(k)$ additional transient words.
A supplied exact rational local budget can also be checked in $O(k)$
arithmetic operations, charging its encoded comparison precision. On the common
event of Theorem~\ref{thm:common-reduction}, these are global optima for
every adaptive update in the stream. No observation outside the indexed
updates is revisited.
\end{proposition}
\begin{proof}
Maintain each coefficient, the scalar $B$, every component table and its
maximizer, and the sum in \eqref{eq:common-table-optimum}. A change
$\Delta$ to an edge missing $U$ changes only $B$, by $\Delta t_e$.
An edge meeting $C_i$ changes its table entry at $s$ by
\[
 \Delta t_e(-1)^{\langle s,\one_{e\cap C_i}\rangle}.
\]
There are $2^{|C_i|}<2^J$ entries. A parity array for all local masks,
formed during preprocessing, evaluates the sign by one indexed lookup
after intersecting the masks. Mark the components touched by the update
list. After applying the changes, scan each marked table once to obtain
its new maximum and a fixed tie-broken maximizer, and adjust the maintained
sum. At most $k$ components are marked. Updating the corresponding entries
of a stored full assignment costs at most $O(Jk)$ operations, already
covered by the displayed bound.

For budget verification, retain
$g_v=\sum_{e\ni v}|\xi_e|$. Replacing $\xi_e$ by $\xi'_e$ changes the
three incident values by $|\xi'_e|-|\xi_e|$. Check the affected vertices
after processing the proposed transaction, before committing table changes.
Only $O(k)$ vertices are affected. Boolean arrays and lists of touched
indices permit deduplication and resetting in $O(k)$ work without scanning
all vertices. Store the old values of the changed entries to reject an
invalid transaction in the same work bound. Distinct edge indices can
be checked by the same marking procedure. This proves both the update
and validation bounds.
\end{proof}

The stream may contain arbitrarily many admissible objectives: uniformity
of the original event avoids a further union bound. The accumulated
computational work is still charged for every update. Constructing its
replacement values is additional work, and a dense coefficient array has
input cost $\Omega(m)$. Indexed edge access is part of the input convention;
converting raw triples to those indices also has a cost. Returning the
optimum value or the list of modified component configurations uses the
maintained data. Writing a complete assignment costs $\Theta(n)$ output
operations. An empty update returns the stored optimum value in $O(1)$ work.

All tables and scalars use integer numerators after multiplication by $2^b$.
Their bit length is
$O(b+\log(m+n)+\log(1+\Gamma_n))$, since they are sums of at most $m$
current weighted terms. A scaled implementation tests a degree against
$2^b u_n$ in the same units. The operation bounds are logarithmic-word
bounds when $b=O(\log(m+n))$. Arbitrary-precision arithmetic must charge
this bit length. A stream that increases its common denominator must also
pay to rescale the stored tables; that operation is not included in a
sparse update at fixed precision. Rational budget and objective thresholds
are handled by exact cross multiplication with their encoding and arithmetic
costs. The mathematical family with arbitrary real coefficients is distinct
from this finite-input arithmetic guarantee.

\begin{figure}[t]
\centering
\includegraphics[width=\textwidth]{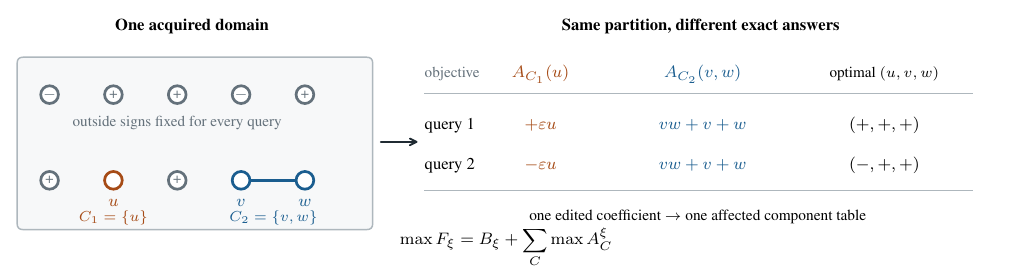}
\caption{A common exact domain with changing answers (schematic).
Gray signs remain fixed. Colored vertices form two components of the
free interaction graph. The illustrative component functions give
opposite optimal choices for $u$ when one local field changes from
$+\varepsilon$ to $-\varepsilon$; the other component and all outside
signs are unchanged. An indexed coefficient edit touches at most one
component. The probability theorem establishes when this decomposition
is globally exact; a finite certificate verifies that property for a
given instance.}
\label{fig:common-mechanism}
\end{figure}

\section{Local influence for general binary factor objectives}
\label{app:general-local-influence}

The one-coordinate comparison does not require parity factors. This
appendix separates its deterministic content from the localization and
component geometry supplied by the random XOR model.

Let $V$ be a finite set of $N$ binary variables and let $E$ be an indexed
family of nonempty factor scopes, each of size at most $a$. Repeated
scopes are permitted. For $x\in\{-1,1\}^V$, consider
\[
 F(x)=\sum_{e\in E}\phi_e(x_e),\qquad
 \Psi(x)=\sum_{e\in E}\psi_e(x_e).
\]
The base and perturbation factors are arbitrary real functions. Their
declared scopes remain fixed. The interaction graph joins distinct
variables belonging to a common declared scope, even if its base factor
is zero or happens to separate into smaller factors. This convention is
necessary because a later perturbation may couple those variables.

For $A\subseteq V$, write $x^A$ for $x$ with the signs in $A$ reversed;
write $x^v=x^{\{v\}}$ and $x^{uv}=x^{\{u,v\}}$ for distinct $u,v$.
For any function $h$, put $\Delta_vh(x)=h(x)-h(x^v)$. A difference is
zero if the function does not depend on that variable. Define
\begin{align}
 d_v^F(x)&=\tfrac12\Delta_vF(x),\nonumber\\
 \kappa_{vu}
 &=\tfrac12\sum_{e\supseteq\{v,u\}}
       \|\Delta_u\Delta_v\phi_e\|_\infty,
 \qquad \kappa=\max_{v\ne u}\kappa_{vu},
 \label{eq:general-influence-def}
\end{align}
where the maximum is zero when there is no pair of distinct variables.
The norm ranges over the finite factor domain. In expanded form, its
mixed difference is
\[
 \Delta_u\Delta_v\phi_e(x_e)
 =\phi_e(x_e)-\phi_e(x_e^v)-\phi_e(x_e^u)+\phi_e(x_e^{uv}).
\]
Let $\mathcal P$ be any family of perturbations satisfying the common
factorwise local envelope
\begin{equation}
 \gamma_v(\Psi):=\tfrac12\sum_{e\ni v}
                        \|\Delta_v\psi_e\|_\infty\le\Gamma
 \quad(v\in V,\ \Psi\in\mathcal P).
 \label{eq:general-local-budget}
\end{equation}
Thus $|d_v^\Psi(x)|\le\Gamma$ for every assignment. The family may be the
entire envelope or any specified subset. No probability law is assumed.

\begin{theorem}[A conditional common reduction for binary factors]
\label{thm:general-local-influence}
Fix a reference assignment $r$, an integer $B\ge1$, and $\Gamma\ge0$.
Suppose every $F+\Psi$, $\Psi\in\mathcal P$, has an optimizer $x^\Psi$
whose disagreement set
$S^\Psi=\{v:x_v^\Psi\ne r_v\}$ has interaction components of size at
most $B$. Define the observable set
\begin{equation}
 U=\{v:d_v^F(r)\le\kappa(B-1)+\Gamma\}.
 \label{eq:general-common-set}
\end{equation}
Then, simultaneously for every $\Psi\in\mathcal P$,
\begin{equation}
 \max_x(F+\Psi)(x)
   =\max_{x:\,x_{U^c}=r_{U^c}}(F+\Psi)(x).
 \label{eq:general-common-exact}
\end{equation}
If the observed components of $U$ have size at most $K$, the right side
separates into independent tables with at most $2^K$ entries each.
The localization bound $B$ and the observable component bound $K$ are
separate hypotheses. If the localization premise holds for every
optimizer, every optimizer agrees with $r$ outside $U$.
\end{theorem}

\begin{proof}
For distinct $u,v$ and every $z$,
\begin{align}
 |d_v^F(z)-d_v^F(z^u)|
 &=\tfrac12|\Delta_u\Delta_vF(z)|\nonumber\\
 &\le\tfrac12\sum_{e\supseteq\{v,u\}}
               \|\Delta_u\Delta_v\phi_e\|_\infty
 =\kappa_{vu}.
 \label{eq:general-coordinate-lipschitz}
\end{align}
Fix any allowed $\Psi$ and a localized optimizer $x=x^\Psi$, and write
$S=S^\Psi$. At a changed coordinate $v\in S$, optimality against a
single reversal gives $d_v^{F+\Psi}(x)\ge0$. Set $y=x^v$. Since
$d_v^F(y)=-d_v^F(x)$,
\begin{equation}
 d_v^F(y)\le d_v^\Psi(x)\le\Gamma.
 \label{eq:general-final-perturbation}
\end{equation}
The assignments $r$ and $y$ differ exactly on $S\setminus\{v\}$.
Telescope along any ordering of these coordinates and apply
\eqref{eq:general-coordinate-lipschitz} at each step. This yields
\begin{equation}
 d_v^F(r)\le\Gamma+\sum_{u\in S\setminus\{v\}}\kappa_{vu}.
 \label{eq:general-changed-coordinate}
\end{equation}
If $u$ is not in the component of $S$ containing $v$, then no factor
scope contains both variables, so $\kappa_{vu}=0$. There are at most
$B-1$ other variables in that component. Hence
$d_v^F(r)\le\Gamma+\kappa(B-1)$, proving $S\subseteq U$.
The restricted domain therefore contains an optimizer of this objective.
This argument applies to every allowed perturbation using the same $U$,
which proves \eqref{eq:general-common-exact}. If the premise covers every
optimizer, apply the same comparison to each one.

For the component statement, a scope meeting two different components of
$U$ would join them in the interaction graph, a contradiction. Fixing the
outside signs to $r$ consequently assigns every nonconstant remaining
factor to one component. Optimizing these components independently proves
the claimed table representation.
\end{proof}

The proof uses only single-coordinate optimality of the chosen global
optimizer, but this does not replace the global localization premise.
It establishes neither that premise nor a certificate for it from local
fields. It also makes the role of ties explicit: coordinates with equality
in \eqref{eq:general-common-set} must be retained.

The assignments in the theorem are actual binary assignments. In a model
with objective-preserving symmetries, a localization proof may first select
a suitable representative of each required optimizer. Existence of that
representative supplies the stated premise. No symmetry or gauge is
assumed for general factors.

\paragraph{The two structural conditions are independent.}
For $N\ge2$, consider the path objective
$F(x)=\sum_{i=1}^{N-1}x_ix_{i+1}$, zero perturbation, and the alternating
reference $r_i=(-1)^i$. An all-positive optimum differs from $r$ only on
one bipartition of the path, so its disagreement components are singletons:
the localization premise holds with $B=1$. Nevertheless every reference
field is negative, so $U=V$ and its observed component has size $N$.
Small optimal-change components do not imply small observable components.

Conversely, local fields alone do not justify freezing variables. For
$F(x,y)=3xy-x-y$, reference $(+1,+1)$, and $\Gamma=0$, both reference
fields are $2>0$. Choosing $B=1$ would make $U$ empty, but the unique
optimum is $(-1,-1)$: its score is $5$, compared with reference score $1$.
Its disagreement component has size two, violating the chosen localization
bound. The failure is in the structural premise, not the cutoff algebra.

\paragraph{Sharper nonuniform bounds.}
The bound in \eqref{eq:general-changed-coordinate} can be retained directly.
For each $v$, sum the largest $B-1$ values in
$\{\kappa_{vu}:u\ne v\}$, using all values if fewer are present.
Calling this sum $\ell_v(B)$, the set
$\{v:d_v^F(r)\le\ell_v(B)+\Gamma\}$ also preserves every allowed
optimum value. Vertex-specific supplied budgets $\Gamma_v$ can replace
$\Gamma$ in both the hypothesis and this cutoff. The uniform version is
sufficient for the parity model.

\subsection{Finite representation and costs}

Let $C$ range over the components of $U$, and write $x^{C,s}$ for the
assignment equal to a local configuration $s\in\{-1,1\}^C$ on $C$ and
to $r$ elsewhere. With $f_e^\Psi=\phi_e+\psi_e$, store
\begin{align*}
 b^\Psi&=\sum_{e:\,e\cap U=\varnothing}f_e^\Psi(r_e),\\
 A_C^\Psi(s)&=
   \sum_{e:\,\varnothing\ne e\cap U\subseteq C}
                         f_e^\Psi(x_e^{C,s}).
\end{align*}
The exact restricted optimum is
$b^\Psi+\sum_C\max_s A_C^\Psi(s)$ on every input. Its equality with
the unrestricted optimum uses the localization premise.

For an explicit finite-input implementation, index variables and factors
and supply their complete binary truth tables. Computing all the local
mixed-difference maxima takes
$O(\sum_e |e|^2 2^{|e|})$ arithmetic operations. Contributions belonging
to the same variable pair can be aggregated in
$O(N+\sum_e|e|^2)$ additional indexed operations: two stable counting-sort
passes on the pair's vertex indices suffice. This computes $\kappa$ without
an implicit dense $N\times N$ array. The fields, the interaction graph,
and its components have comparable or smaller cost. After verifying that
the observed component sizes are at most $K$, constructing the tables
costs at most
\[
 O\!\left(\sum_{e\in E}|e|2^K+N2^K\right)
\]
additional indexed operations and factor-table evaluations. At fixed
maximum arity $a$ and component cap $K$, these bounds are linear in the
explicit input size, including $N$. The input and any factor evaluations
are charged; a compact oracle representation does not automatically inherit
the truth-table bound. Component sizes are checked before exponential
allocation. If the declared cap fails, return a feasible reference score
and report that this reduction procedure did not finish.

The partition and the factor-to-component incidence map remain fixed as
objectives change. Adaptive choices are permitted whenever each accumulated
perturbation from $F$ remains in the allowed family. A nonempty transaction
replacing $k$ distinct indexed
factor tables can update the restricted optimum value and a maintained
assignment in
\begin{equation}
 O\!\left(k\{a2^a+(a+K)2^K\}\right)
 \label{eq:general-factor-update-cost}
\end{equation}
exact arithmetic and indexed operations. To verify this bound, read each
changed table and compute its perturbation table relative to the stored
base factor. For every incident vertex, recompute that factor's
half-oscillation in \eqref{eq:general-local-budget}, in $O(a2^a)$ work
per factor. Maintain the sum of these contributions at each vertex, so
only the $O(ak)$ touched vertices need budget validation. Check the
transaction before committing table changes, retaining its old values for
rollback. Touched-index lists avoid any full-array reset.

If a factor misses $U$, change its contribution to $b^\Psi$. Otherwise,
for each configuration of its one component, evaluate its old and new
table entries with the outside signs fixed to $r$ and update that
component value. Constructing the local factor index costs $O(a)$, giving
$O(a2^K)$ work per changed factor. Rescan each touched component once
for its maximum and tie-broken maximizer; there are at most $k$ such
components. At most $Kk$ entries of a maintained full assignment change.
This proves \eqref{eq:general-factor-update-cost}. Reading dense queries,
converting unindexed factor names, and writing complete assignments have
their respective input and output costs. An empty update returns the
stored optimum value in constant work.

Exact arithmetic requires an encoding convention. With a fixed common
dyadic denominator, scaled values, mixed differences, and table sums are
integers, allowing one additional factor of two for the field definitions.
If the base and current factor numerators have at most $L$ bits, stored sums have
$O(L+\log(|E|+1))$ bits. Supplied rational budgets are compared exactly
with their own encoded lengths charged. The local perturbation budget
does not bound additive constants in factor values, so numerator lengths
cannot be inferred from $\Gamma$ alone. Increasing a common denominator
requires rescaling stored values and is additional work. Arbitrary real
factors in the mathematical theorem carry no finite-input bit-complexity
claim.

\subsection{Parity specialization and interpretation}

For $\phi_e(x_e)=Y_e\prod_{v\in e}x_v$ with $|Y_e|=1$, every incident
pair contributes $2$ to $\kappa_{vu}$. Thus pair codegree at most two
gives $\kappa\le4$. For
$\psi_e(x_e)=\xi_e\prod_{v\in e}x_v$, its contribution to
$\gamma_v(\Psi)$ is $|\xi_e|$. The first field is precisely
$d_v^F(r)=\sum_{e\ni v}Y_eT_e(r)$.
With $B=J-1$, the general cutoff becomes $4(J-2)+\Gamma$; integer base
fields give $4(J-2)+\lfloor\Gamma\rfloor$. The common probability event
in Theorem~\ref{thm:common-reduction} proves both structural conditions,
with observed components smaller than $J$, uniformly over the perturbation
family.

Fixing values that extend to a global optimum is the established notion
of persistency or partial optimality \citep{shekhovtsov2016persistency}.
The theorem above is an elementary quantitative transfer from a uniform
localization premise to a common persistent assignment. It is not a
general algorithm for proving that premise. The model-specific content
here is that a small acquisition and the random target jointly supply
localization, computable fields, and the required observed component
geometry. Applying the criterion to another factor model requires proving
those properties there.

\section{Exponentially many incompatible exact answers}
\label{app:many-answers}

The common reduction need not return the same answer to every objective.
We quantify this at the sharp correction scale
\[
 r_n=\frac{n^{1-\alpha}}{\sqrt{\log n}}.
\]
Throughout this appendix, two factor assignments are identified when they
give the same complete tensor. An \emph{observed product vector} is the
restriction $(T_e(x))_{e\in H}$. These are different notions when a variable
has no incident observation. The distinction is necessary for uniqueness.

\subsection{Nonisolated vertices with zero and unit signed degree}

\begin{lemma}[Fixed signed-degree counts]
\label{lem:zero-and-unit-counts}
Suppose $0<\alpha<1$. In each of the three blocks, both of the following
sets have cardinality $\Theta_{c,q}(r_n)$ with probability $1-o(1)$:
\begin{align*}
 Z^*&=\{v:D_v=0,\ \deg_H(v)>0\},\\
 Q^*&=\{v:D_v=1,\ \text{some incident observation has }\eta_e=-1\}.
\end{align*}
The bounds hold jointly across the three blocks under the uniform
size-$m$ mask.
\end{lemma}
\begin{proof}
First use Bernoulli masking with $p=m/n^3$, and put
$\lambda_n=m/n$ and $\nu_n=\lambda_n\sqrt{q(1-q)}$.
For an integer $\ell$ with
$|\ell-\nu_n|\le\sqrt{\nu_n}$, the event that a specified vertex has
exactly $\ell$ positive and $\ell$ negative incident observations has
probability
\begin{align*}
 P_{\ell,0}
 &=\frac{(n^2)_{2\ell}}{(\ell!)^2}
       [p(1-q)]^\ell(pq)^\ell(1-p)^{n^2-2\ell}\\
 &=(1+o(1))e^{-\lambda_n}\frac{\nu_n^{2\ell}}{(\ell!)^2}.
\end{align*}
The multiplicative error is uniform on the displayed interval because
$\ell=O(\log n)$ and $p=O(\log n/n^2)$.
Uniform Stirling bounds give
\[
 \frac{\nu_n^{2\ell}}{(\ell!)^2}
       =\Theta_{c,q}(e^{2\nu_n}/\nu_n).
\]
As $\lambda_n-2\nu_n=\alpha\log n+o(1)$, each such event has
probability $\Theta_{c,q}(n^{-\alpha}/\log n)$. There are
$\Theta_{c,q}(\sqrt{\log n})$ disjoint choices of $\ell$, all positive
for sufficiently large $n$. Their union is contained in the nonisolated
zero-degree event. This gives the lower bound
$\Omega_{c,q}(n^{-\alpha}/\sqrt{\log n})$ for its probability. The
upper bound follows from \eqref{eq:recycle-degree-tail} at threshold zero.

For the unit-degree event, require $\ell+1$ positive and $\ell$ negative
observations on the same interval. Its probability is
\[
 P_{\ell,1}=(1+o(1))e^{-\lambda_n}
       \frac{[\lambda_n(1-q)]^{\ell+1}(\lambda_n q)^\ell}
            {(\ell+1)!\ell!}.
\]
Its ratio to $P_{\ell,0}$ is
$(1+o(1))\lambda_n(1-q)/(\ell+1)$, bounded above and below by positive
constants. Again there is at least one negative observation because
$\ell\ge1$. Summation gives the same lower bound, and
\eqref{eq:recycle-degree-tail} at threshold one gives the upper bound.

Within a block, the incident location sets of distinct vertices are
disjoint. Under Bernoulli masking and independent signs, the indicators
of each event are therefore independent. Their expected counts are
$\Theta_{c,q}(r_n)$, so Chernoff's inequality puts each count between
positive constant multiples of $r_n$, except with probability
$\exp[-\Omega_{c,q}(r_n)]$. Union over the six counts. The total number
of observations is binomial with mean $m$, and its probability of being
exactly $m$ is $\Theta(m^{-1/2})$. Conditioning on this event produces
the uniform size-$m$ mask and multiplies the failure bound by at most
$O(\sqrt m)$. For every fixed $\alpha<1$,
$\sqrt m\exp[-\Omega(r_n)]=o(1)$.
\end{proof}

\subsection{The common event and its observable coordinates}

Assume now that $1/2<\alpha<1$ and fix a budget $\Gamma>0$.
Use Theorem~\ref{thm:common-reduction} with $J=2$, write
$r=z^1$, and set
\[
 U=\{v:d_v(r)\le\lfloor\Gamma\rfloor\}.
\]
Its proof supplies a fixed intrinsic weak set $W_L^*$ containing
$U$, the errors of $r$, and a source-gauge representative of every
admissible optimizer. Increase the fixed $L$ to at least one when needed.
By Lemma~\ref{lem:sublog-weak-geometry}, on one event with probability
$1-o(1)$, every edge contains at most one vertex of $W_L^*$ and every
vertex has at most one neighbor in $W_L^*$. Here the choices
$J=k_*=2$ are valid precisely because $2>1/\alpha$.

It follows that choosing one incident edge for each of any collection of
distinct nonisolated vertices in $W_L^*$ gives pairwise disjoint edges.
Indeed, an edge cannot contain another selected weak vertex. If two
selected edges shared an outside vertex, that vertex would have two
distinct neighbors in $W_L^*$, contradicting the second property.
This matching property controls the local budget without a loss in the
number of coordinates.

Using the unperturbed component tables of the common reduction, compute
one original optimizer $w$ agreeing with $r$ outside $U$. Define observable
integers
\begin{equation}
 d_v^w=\sum_{e\ni v}Y_eT_e(w),\qquad
 Z=\{v\in U:d_v^w=0,\ \deg_H(v)>0\}.
 \label{eq:many-observable-zero}
\end{equation}
Every $d_v^w$ is nonnegative: otherwise a single flip improves $w$.
For an assignment agreeing with $w$ outside $U$, put $t_v=x_vw_v$.
Since each edge meets $U$ at most once, the restricted base objective is
\begin{equation}
 F_0(x)=C_0+\sum_{v\in U}d_v^w t_v.
 \label{eq:many-linear-restriction}
\end{equation}
For a perturbed objective it is
$C_\xi+\sum_{v\in U}(d_v^w+\delta_v)t_v$, where
\[
 \delta_v=\sum_{e\ni v}\xi_eT_e(w),\qquad |\delta_v|\le\Gamma.
\]
Every global optimizer has such a representative, by
Theorem~\ref{thm:common-reduction}; thus these elementary linear
comparisons describe all optimizing tensors, not just one chosen answer.

Let $\sigma_v$ denote the true factor sign in the reference-fiber gauge.
Every neighbor of a vertex in $W_L^*$ is correct in $w$.
Consequently $d_v^w=(w_v\sigma_v)D_v$ there. Optimality of $w$
makes this $|D_v|$ when $D_v\ne0$. Every true zero-degree vertex belongs
to $U$, since $d_v(r)=(r_v\sigma_v)D_v=0$. Thus $Z=Z^*$, with the
nonisolated restriction understood, and Lemma~\ref{lem:zero-and-unit-counts}
gives $|Z|=\Theta(r_n)$. This equality is used only on the common event;
the algorithm computes $Z$ from the displayed observable quantities.

\subsection{Different exact answers within any positive budget}

\begin{theorem}[Exponentially many disjoint optimizer sets]
\label{thm:many-answers}
Under the retained-source assumptions of Theorem~\ref{thm:recycling},
suppose $1/2<\alpha<1$ and fix $\Gamma>0$.
With probability $1-o(1)$, the one acquired common reduction gives an
observable integer $R=\Theta_{c,q,s,\Gamma}(r_n)$ and an explicitly
constructible family of $2^R$ coefficient arrays in
$\mathcal U_\Gamma(H)$ such that:
\begin{enumerate}
\item Every coefficient retains the sign of its original label.
\item Each objective has a unique optimizing observed product vector.
These vectors are different for every pair of objectives, so their sets
of globally optimizing complete tensors are pairwise disjoint.
\item Across the entire budget neighborhood, the total number of
complete tensors that can be optimal for any objective is at most
$2^{O_{c,q,s,\Gamma}(r_n)}$.
\end{enumerate}
If $c>1$, every objective in the constructed family has a unique globally
optimizing complete tensor with probability $1-o(1)$. Uniqueness of the
factor assignment is always interpreted modulo the four whole-block gauges.
The family has a common-denominator dyadic description of constant
precision for fixed $\Gamma$.
\end{theorem}
\begin{proof}
Work on the common event just established. Set $R=|Z|$, and choose one
incident edge $e_v$ for each $v\in Z$. These edges form a matching.
Choose a fixed dyadic number $0<\varepsilon<\min\{1,\Gamma\}$.
For every $\tau\in\{-1,+1\}^Z$, set
\begin{equation}
 \xi^{\tau}_{e_v}=\varepsilon\tau_vT_{e_v}(w)
       \quad(v\in Z),\qquad
 \xi^{\tau}_e=0\quad\text{otherwise}.
 \label{eq:many-tie-perturbation}
\end{equation}
Because the selected edges are disjoint, each vertex receives total
absolute perturbation at most $\varepsilon<\Gamma$. Since
$|\xi_e|<1$, each nonzero original coefficient retains its sign.
The restricted field at $v\in Z$ becomes $\varepsilon\tau_v$.
No other $U$ coordinate has an incident modified edge, so its field stays
$d_v^w$.

Every nonisolated coordinate in $U$ now has a nonzero field and a unique
preferred relative sign: $t_v=\tau_v$ for $v\in Z$, and $t_v=+1$
otherwise. Only isolated coordinates remain free. They do not affect any
observed product, which proves uniqueness on $H$. If two sign vectors
differ at $v$, their optimizing products on $e_v$ differ, so no complete
tensor can optimize both objectives. These are distinct optimization
problems with incompatible answers, even though this construction selects
among ties of the original objective.

For the upper bound, every global optimum of every admissible objective
has a representative fixed to $r$ on $U^c$. There are at most $2^{|U|}$
such assignments, hence at most that many complete tensors. The fixed-budget
bound $|U|=O(r_n)$ in Theorem~\ref{thm:common-reduction} proves the claim.

If $c>1$, the expected number of isolated variables is at most
\[
 3n\frac{\binom{n^3-n^2}{m}}{\binom{n^3}{m}}
 \le3n(1-1/n)^m=O(n^{1-c})=o(1).
\]
There are then no free coordinates in the restricted problem. Every global
optimizer has the same restricted representative, hence the same complete
tensor. The coefficients in \eqref{eq:many-tie-perturbation} use only the
fixed dyadic number $\varepsilon$ and observed signs.
\end{proof}

Full-tensor uniqueness cannot be asserted throughout $1/2<\alpha<1$:
these conditions allow $c<1$, when isolated variables persist. Changing
coefficients on $H$ cannot identify a sign absent from $H$.
Indeed, under Bernoulli masking their probability is
$(1-p)^{n^2}=\Theta(n^{-c})$; within-block concentration followed by the
same conditioning argument gives $\Theta(n^{1-c})$ isolated variables
per block with high probability when $c<1$.
The observed-product and disjoint-set statements require no additional
condition on $c$.

\subsection{Forcing reversals of strict original decisions}

The preceding family uses original ties. A local budget greater than one
also changes strict empirical decisions, while preserving every label sign.
The value one is the smallest nonzero signed degree of the original
unit-weight objective.

\begin{theorem}[Exponentially many forced strict reversals]
\label{thm:many-strict-answers}
Under the assumptions of Theorem~\ref{thm:many-answers}, suppose
$\Gamma>1$. With probability $1-o(1)$ there is an observable set
$Q\subseteq U$ of size $R=\Theta_{c,q,s,\Gamma}(r_n)$ and an explicit
family $(\xi^b)_{b\in\{0,1\}^Q}\subseteq\mathcal U_\Gamma(H)$ with
the uniqueness and disjointness conclusions of that theorem, such that
every optimizer $x^b$ satisfies
\begin{equation}
 \OPT(H,Y)-F_0(x^b)=2\sum_{v\in Q}b_v.
 \label{eq:many-original-loss}
\end{equation}
Every original coefficient retains its sign. Every nonzero bit vector
therefore forces an answer that is strictly suboptimal for the original
objective. If $c>1$, every constructed objective has a unique optimizing
complete tensor.
\end{theorem}
\begin{proof}
Use the same observable base optimizer $w$ and zero set $Z$, and let
\[
 Q=\{v\in U:d_v^w=1,\ \exists e\ni v\text{ with }Y_eT_e(w)=-1\}.
\]
For $\Gamma>1$, every true unit-degree vertex belongs to $U$: its
observed degree relative to $r$ is either $1$ or $-1$, and
$\lfloor\Gamma\rfloor\ge1$. At a true unit-degree vertex,
optimality forces $w_v\sigma_v=+1$, and its neighbors agree with the
source. Thus the vertices counted by $Q^*$ in
Lemma~\ref{lem:zero-and-unit-counts} belong to $Q$.
The lemma gives $|Q|=\Omega(r_n)$, while $Q\subseteq U$ gives
$|Q|=O(r_n)$.

For each $v\in Q$, choose an incident edge $e_v$ with
$Y_{e_v}T_{e_v}(w)=-1$. For each $z\in Z$, choose an incident edge
$f_z$. Since $Q$ and $Z$ are disjoint weak-vertex sets, all these chosen
edges form one matching. Choose fixed dyadic numbers
$1<a<\Gamma$ and $0<\varepsilon<\min\{1,\Gamma\}$.
For $b\in\{0,1\}^Q$, define
\begin{equation}
 \xi^b_{e_v}=a b_vY_{e_v}\quad(v\in Q),\qquad
 \xi^b_{f_z}=\varepsilon T_{f_z}(w)\quad(z\in Z),
 \label{eq:many-strict-perturbation}
\end{equation}
and set all remaining entries to zero. The local budget is at most
$\max\{a,\varepsilon\}<\Gamma$, because the chosen edges are
disjoint. Each selected unit-field coefficient is multiplied by
$1+a b_v>0$, and every zero-field perturbation has magnitude less than one,
so no coefficient changes sign.

The restricted field at $v\in Q$ is $1-a b_v$, whose sign is
$(-1)^{b_v}$. At each $z\in Z$ it is $\varepsilon>0$.
Every other nonisolated coordinate in $U$ keeps its positive integer field.
Thus the preferred relative sign is $-1$ exactly on the selected unit-field
vertices with $b_v=1$. The proof of Theorem~\ref{thm:many-answers} now gives
the uniqueness and disjointness statements. In the original objective,
each of these flips costs exactly two, and no edge meets two flipped
vertices. Equation~\eqref{eq:many-linear-restriction} therefore gives
\eqref{eq:many-original-loss} for every optimizing tensor. Changes at
isolated vertices have no score effect.
\end{proof}

\begin{proposition}[The unit-budget boundary]
\label{prop:unit-budget-boundary}
On the common event above, if $0\le\Gamma<1$, every optimizer of
every $F_\xi$ with $\xi\in\mathcal U_\Gamma(H)$ is also an optimizer
of the original objective. If $\Gamma=1$, every perturbed objective has
at least one original-optimal tensor among its optimizers, but it can also
have new tied optimizers that are not original-optimal. If $\Gamma>1$,
Theorem~\ref{thm:many-strict-answers} supplies exponentially many tasks
that force prescribed reversals of strict original choices.
\end{proposition}
\begin{proof}
In \eqref{eq:many-linear-restriction}, every positive $d_v^w$ is an
integer and hence at least one. The perturbation changes it by at most
$\Gamma$. For $\Gamma<1$ its sign stays positive; only original zero
fields can choose a different sign. For $\Gamma=1$ a positive field
can become zero but cannot become negative. Retain $t_v=+1$ at all
originally positive fields and optimize the originally zero fields. This
gives a common optimizer of both objectives. To see why the stronger
all-optimizers statement fails at equality, use the strict-reversal
construction with $a=1$ and one selected $b_v=1$. The corresponding
unit field becomes zero. Both signs optimize the perturbed objective,
and its negative choice loses two in the original objective. Such a
vertex exists with probability $1-o(1)$ by
Lemma~\ref{lem:zero-and-unit-counts}.
\end{proof}

\paragraph{Representation and work.}
The matching, base optimizer, sets $Z,Q$, and selected observation indices
are constructed once by scans of the common reduction and its stored
incidences. A member of either family is specified by its $R$ signs or bits
and the shared fixed coefficients. Producing all $2^R$ explicit objectives
or full answers still incurs their total description and output cost.
After installing the shared perturbations, changing one coordinate in either
family changes a single coefficient. The maintained component tables of
Proposition~\ref{prop:incremental-common-reduction} update the score and
an implicit optimizing assignment in constant arithmetic work here, since
all components are singletons. Reporting all $3n$ factor signs separately
costs $\Omega(n)$ per report. No further source readings are used, and no
finite-sample certificate of the common probability event is inferred from
the observed singleton geometry alone.

\section{The minimum dimension of a common exact restriction}
\label{app:dimension-optimality}

The isolation condition $\alpha>1/2$ makes every residual component a
singleton. Exponentially many incompatible answers require less: enough
residual coordinates must be independent of the other residual
coordinates. Almost all fixed-field vertices have this property at every
fixed $0<\alpha<1$. This gives a matching lower bound for any common
fixed-coordinate restriction.

\subsection{Most weak vertices have no nearby weak vertex}

Let $G_H$ be the interaction graph on the $3n$ factor coordinates: distinct
vertices are adjacent when they share an observed triple. Graph distances
are infinite between different connected components. Write
$W_L^*=\{v:D_v\le L\}$ for the intrinsic weak set in source coordinates.

\begin{lemma}[Few nearby pairs of weak vertices]
\label{lem:nearby-weak-pairs}
Fix $L\ge1$ and $0<\alpha<1$. Let $B_L$ count unordered pairs of
distinct vertices of $W_L^*$ whose distance in $G_H$ is at most two.
Under the uniform size-$m$ mask,
\[
 \E B_L\le n^{1-2\alpha+o(1)}=o(r_n),
 \qquad r_n=n^{1-\alpha}/\sqrt{\log n}.
\]
Consequently, removing every weak vertex at distance at most two from
another weak vertex removes $o(r_n)$ vertices with probability $1-o(1)$.
In each block, $\Theta(r_n)$ nonisolated zero-degree vertices remain.
The same conclusion holds for unit-degree vertices with a negative
incident observation.
\end{lemma}
\begin{proof}
Put $p=m/n^3$. For any two distinct prescribed vertices $u,v$ and any
$f\in\{1,2\}$ distinct prescribed triples $F$,
\eqref{eq:sublog-weak-moment} gives, uniformly over the choices of $u,v,F$,
\[
 \Pr\{u,v\in W_L^*,\,F\subseteq H\}
 \le p^f n^{-2\alpha+o(1)}.
\]
This estimate is already for the uniform size-$m$ mask; no conditioning
factor is introduced in the witness count below.

An adjacent weak pair has a witness consisting of one triple and two
of its endpoints. There are at most $3n^3$ such witnesses. A pair at
distance two has a witness consisting of a middle vertex, two distinct
triples incident to it, and one other endpoint from each triple.
If the two triples coincide, the pair was already counted as adjacent.
There are at most
$3n\cdot(n^2)^2\cdot4=12n^5$ witnesses with distinct triples; retaining
only choices with distinct designated endpoints can only reduce this
number. The triples may share further vertices, which is permitted in
the prescribed-location bound. Summing the displayed probabilities gives
\[
 \E B_L
 \le (3n^3p+12n^5p^2)n^{-2\alpha+o(1)}
 \le n^{1-2\alpha+o(1)}.
\]
The ratio to $r_n$ is $n^{-\alpha+o(1)}\sqrt{\log n}=o(1)$.
Markov's inequality therefore gives $B_L=o_{\Pr}(r_n)$. At most $2B_L$
vertices are endpoints of these pairs. Subtract this quantity from the
six $\Theta(r_n)$ counts in Lemma~\ref{lem:zero-and-unit-counts}.
\end{proof}

The absolute number of nearby weak pairs can grow when $\alpha\le1/2$.
Its ratio to the number of zero or unit fields still tends to zero.
The proof needs this relative estimate, not complete weak-set isolation.

\subsection{An observable collection of independent coordinates}

Fix $\Gamma>0$ and an integer $J\ge2$ with $J>1/\alpha$.
Use the common restriction of Theorem~\ref{thm:common-reduction}, and
write its reference as $r=z^1$ and its free set as $U$. Define
\[
 I(U)=\{v\in U:\operatorname{dist}_{G_H}(v,U\setminus\{v\})>2\}.
\]
Each $v\in I(U)$ is a singleton component of $G_H[U]$, and no two
vertices in $I(U)$ share a neighbor. In particular, choosing one
incident triple for each of a collection of nonisolated vertices in
$I(U)$ gives pairwise disjoint triples. Every chosen triple meets $U$
only at its selected vertex.

The set $I(U)$ can be found in $O(m+n)$ indexed operations. In a first
incidence scan, retain for each vertex its first distinct neighbor in
$U$ and a flag indicating whether it has another. A second scan marks
vertices of $U$ that have a neighbor in $U$, or have a neighbor whose
flag is set. The unmarked vertices form $I(U)$. Repeated incidences of
the same pair do not count as distinct neighbors.

Compute one base optimizer $w$ by maximizing the unweighted common
component tables. Define the observable sets
\begin{align}
 Z&=\{v\in I(U):d_v^w=0,\ \deg_H(v)>0\},\nonumber\\
 Q&=\{v\in I(U):d_v^w=1,\
       \exists e\ni v\text{ with }Y_eT_e(w)=-1\}.
 \label{eq:dimension-observable-coordinates}
\end{align}
On the common probability event, $w$ is globally optimal. Its agreement
with $r$ outside $U$, the inclusion $U\subseteq W_L^*$, and
$E(r)\subseteq W_L^*$ make $w$ agree with the source outside $W_L^*$,
for a fixed $L$ supplied by the proof of
Theorem~\ref{thm:common-reduction}. Increase this fixed $L$ to at least
one.

Every intrinsic zero-degree vertex left by
Lemma~\ref{lem:nearby-weak-pairs} has all its neighbors outside $W_L^*$.
Its degree relative to $r$ is zero, so it belongs to $U$ and then to
$I(U)$. Its degree relative to $w$ is also zero, so it belongs to $Z$.
Thus
\[
 |Z|=\Theta(r_n).
\]
The upper bound follows from $Z\subseteq U$ and the bounded-budget
common-reduction theorem.

For $\Gamma>1$, a remaining intrinsic unit-degree vertex belongs to $U$:
its degree relative to $r$ is either $1$ or $-1$, and the cutoff
$4(J-2)+\lfloor\Gamma\rfloor$ is at least one. All its neighbors are
correct in $w$. Single-coordinate optimality therefore forces its own
sign to agree with the source, so $d_v^w=1$ and its negative intrinsic
observation remains negative relative to $w$. It belongs to $Q$.
Consequently $|Q|=\Theta(r_n)$ when $\Gamma>1$.

For any assignment fixed to $w$ outside $U$, let $t_v=x_vw_v$.
Because the vertices of $Z\cup Q$ are singleton components, the
restricted objective separates into their individual linear terms and
the unchanged tables for all other components. Those other components
may interact internally and may have several maximizing configurations.

\subsection{Incompatible answers at every subcritical density}

\begin{theorem}[Answer diversity throughout the correction regime]
\label{thm:all-density-answers}
Under the retained-source assumptions, let $0<\alpha<1$ and fix
$\Gamma>0$ and an integer $J\ge2$ with $J>1/\alpha$.
With probability $1-o(1)$, the acquired common reduction
constructs a family of $2^R$ objectives in $\mathcal U_\Gamma(H)$,
where $R=\Theta_{c,q,s,J,\Gamma}(r_n)$, whose sets of globally optimizing
observed product vectors are pairwise disjoint. In particular, their
global optimizer sets of complete tensors are pairwise disjoint.
All coefficients retain their original signs.

If $\Gamma>1$, another such family can be indexed by
$b\in\{0,1\}^R$ so that every global optimizer $x^b$ obeys
\begin{equation}
 \OPT(H,Y)-F_0(x^b)=2\sum_{j=1}^R b_j.
 \label{eq:all-density-strict-loss}
\end{equation}
Neither conclusion requires all residual components to be singletons.
Uniqueness within each optimizer set is not asserted at
$\alpha\le1/2$.
\end{theorem}
\begin{proof}
Use the event above and put $R=|Z|$. Choose one incident edge $e_v$
for each $v\in Z$ and a fixed dyadic
$0<\varepsilon<\min\{1,\Gamma\}$. For each
$\tau\in\{-1,+1\}^Z$, change only the selected coefficients by
\[
 \xi^\tau_{e_v}=\varepsilon\tau_vT_{e_v}(w).
\]
The edges form a matching, so the local budget is at most
$\varepsilon$. Their coefficient signs remain unchanged. The restricted
field at $v$ changes from zero to $\varepsilon\tau_v$, and the
perturbation touches no other $U$ coordinate. Every global optimum has
a representative with the fixed outside signs. It must therefore have
$t_v=\tau_v$ at each selected vertex, while its other components
maximize their unchanged original tables.

On the selected edge, the optimal observed product is necessarily
$T_{e_v}(w)\tau_v$. Two distinct sign vectors prescribe different
products on at least one such edge. Thus no optimizing observed product
vector, and no complete tensor, can be shared by those two objectives.
The remaining component tables can retain ties; their choices do not
affect this disjointness. In this family every perturbed optimizer is
also original-optimal, since the selected original fields are zero
and all other component objectives are unchanged.

For $\Gamma>1$, instead use the set $Q$, choose a negative incident
edge $e_v$ for each $v\in Q$, and take a fixed dyadic $1<a<\Gamma$.
For $b\in\{0,1\}^Q$ set
\[
 \xi^b_{e_v}=a b_vY_{e_v},\qquad
 \xi^b_e=0\quad\text{on all other edges}.
\]
The matching again gives the required local budget. The coefficient
is $Y_{e_v}(1+a b_v)$, preserving its sign, and its field relative
to $w$ is $1-a b_v$. Every optimizer therefore has
$t_v=(-1)^{b_v}$ at these coordinates. All other components maximize
the unchanged base objective. The original loss is exactly two for
each reversed selected vertex, and the singleton components do not
share observations. This proves the loss formula for every global
optimizer. Selected-edge products again give pairwise disjointness.
\end{proof}

The two constructions have a shared description of size $O(m+n)$ plus
their $R$ task bits or signs. The common component tables and
\eqref{eq:dimension-observable-coordinates} are constructed in
$O_J(m+n)$ work after acquisition. Each selected variable is itself
a singleton component even when other components are larger. Changing
one family coordinate therefore changes one coefficient and two entries
of a single component table; the update is constant arithmetic work
at the fixed precision of the construction. Listing every task or
serializing every complete assignment pays its full input or output cost.

The universal unit-budget boundary in
Proposition~\ref{prop:unit-budget-boundary} still uses
$\alpha>1/2$. The theorem above extends the existence of the displayed
families, not that universal assertion about all perturbations when
larger residual components are present.

\subsection{Optimal dimension and minimum common candidate sets}

For a realized $(H,Y)$ and a budget $\Gamma\ge0$, call a fixed-coordinate
domain
\[
 \mathcal D(V_0,r_0)=
 \{x\in\{-1,+1\}^{3n}:x_{V_0^c}=(r_0)_{V_0^c}\}
\]
a common exact restriction if it contains a global optimizer of every
$F_\xi$ with $\xi\in\mathcal U_\Gamma(H)$. The reference and coordinate
set may depend on the realized data and acquired measurements but are
fixed before the future objective is specified. Let
\[
 d_{\mathrm{fix}}(H,Y;\Gamma)
 =\min_{\mathcal D(V_0,r_0)\ {\rm common\ exact}} |V_0|.
\]
No efficient algorithm is required for a candidate in this minimum.

Let $N_{\mathrm{hit}}(H,Y;\Gamma)$ be the smallest cardinality of a
query-independent set of complete tensors containing an optimizer of
every objective in the same family. This permits an arbitrary finite
candidate set, not only a coordinate subcube. Always
\[
 \log_2N_{\mathrm{hit}}(H,Y;\Gamma)
 \le d_{\mathrm{fix}}(H,Y;\Gamma),
\]
because a domain with $d$ free factor coordinates contains at most
$2^d$ distinct complete tensors.

\begin{theorem}[Optimal order of the common dimension]
\label{thm:optimal-common-dimension}
Under the retained-source assumptions, fix $\Gamma>0$. If
$0<\alpha<1$, then with probability $1-o(1)$,
\begin{equation}
 \log_2N_{\mathrm{hit}}(H,Y;\Gamma)
 =\Theta(r_n),
 \qquad
 d_{\mathrm{fix}}(H,Y;\Gamma)=\Theta(r_n).
 \label{eq:optimal-common-dimension}
\end{equation}
The common observable restriction attains this order in
$O_J(m+n)$ preprocessing after acquisition. For $\alpha\ge1$ and
bounded $\Gamma$, both quantities are zero with probability $1-o(1)$.
\end{theorem}
\begin{proof}
Theorem~\ref{thm:all-density-answers} supplies $2^R$ objectives with
pairwise disjoint global optimizer sets of complete tensors. A common
candidate set must contain a different member for each objective,
so $N_{\mathrm{hit}}\ge2^R$. Therefore
$d_{\mathrm{fix}}\ge\log_2N_{\mathrm{hit}}\ge R=\Omega(r_n)$.
The common domain supplied by Theorem~\ref{thm:common-reduction}
contains every required optimizer up to whole-block gauge and has
$|U|=O(r_n)$ free coordinates. It is admissible in the defining minimum,
so $d_{\mathrm{fix}}\le|U|=O(r_n)$. These inequalities prove both
claims. At $\alpha\ge1$ the bounded-budget common domain is a singleton
with high probability, giving $d_{\mathrm{fix}}=0$ and
$N_{\mathrm{hit}}=1$.
\end{proof}

The lower bound also applies to the number of binary choices in any
fixed, query-independent parametrization of a common candidate set:
a decoder with $d$ binary inputs has at most $2^d$ possible outputs.
It is not a lower bound on program length, total data-structure memory,
the number of source readings, or an algorithm allowed to construct
answers outside that fixed image after seeing the query.

\paragraph{Every positive bounded neighborhood.}
Fix a deterministic upper envelope $\overline\Gamma>0$ and use the
common event and the set $Z$ constructed for that envelope. The same
matching biases its zero fields with any
$0<\varepsilon<\min\{1,\Gamma\}$, for every
$0<\Gamma\le\overline\Gamma$. The upper domain for
$\overline\Gamma$ remains valid for every smaller budget. Hence, on
one event, the two bounds in
\eqref{eq:optimal-common-dimension} hold simultaneously for all these
positive budgets, with constants independent of the smaller budget.
This includes positive sequences tending to zero. Encoding the
correspondingly small perturbations may require increasing precision;
only the fixed-budget construction has the stated constant dyadic
precision. At $\Gamma=0$, one original optimum always suffices, so
$N_{\mathrm{hit}}=1$ and $d_{\mathrm{fix}}=0$ on every input.

\subsection{A sufficient budget for preserving original optima}

The earlier universal unit-budget boundary uses singleton residual
components. A smaller sufficient budget remains available when those
components interact.

\begin{proposition}[Component-size sufficient budget]
\label{prop:component-budget-stability}
Let $Y_e\in\{-1,1\}$ and suppose that, for a declared local budget
$\Gamma$, every global optimizer of every admissible $F_\xi$ has a
whole-block gauge representative in a common domain
$x_{U^c}=r_{U^c}$. Suppose every component of the induced interaction
graph on $U$ has at most $K$ vertices, where $K\ge1$.
If $\Gamma<1/K$, every optimizer of every admissible objective is an
optimizer of $F_0$. If $\Gamma=1/K$, each admissible objective has at
least one optimizer that also optimizes $F_0$.
These are sufficient bounds; no sharpness is asserted when $K>1$.
\end{proposition}
\begin{proof}
No observed triple meets two different components of $U$, so every
restricted objective is a sum of independent component tables and a
constant. Fix a component $C$ and a base-optimal assignment $p$ on it.
For any assignment $q$ that is not base-optimal, the original component
gap is at least two: each table is a sum of a fixed number of unit
signed terms. If $S\subseteq C$ is the set of coordinates where $p$
and $q$ differ, the perturbation contribution to their difference has
absolute value at most
\[
 2\sum_{e:\,T_e(p)\ne T_e(q)}|\xi_e|
 \le 2\sum_{v\in S}\sum_{e\ni v}|\xi_e|
 \le 2\Gamma |S|\le2\Gamma K.
\]
For $\Gamma K<1$, every base-optimal $p$ therefore has strictly larger
perturbed score than every non-base-optimal $q$. Every optimizing
component choice remains base-optimal. For $\Gamma K=1$, a base-optimal
choice maximizing the perturbed table weakly dominates every other
choice, so a common component optimum exists. Combining components
proves the assertions inside the domain. Its all-optimizer localization
and the gauge invariance of both objectives give the stated global
conclusions.
\end{proof}

For the common random-model domain one may take $K=J-1$. The resulting
$1/(J-1)$ bound is available at every fixed positive $\alpha$; when all
components are singletons it reduces to the unit-budget statement.
At the boundary only the existence of a common optimizer is claimed.

\subsection{An independent finite check with an interacting residual}

The supplied program \texttt{check\_dimension.py} imports no
acquisition or common-reduction implementation. It exhaustively evaluates
all $2^{16}=65{,}536$ gauge-fixed factor assignments of an explicit
$n=6$, $m=104$ distinct-triple instance using integer quarter units.
The observed restriction contains four coordinates: two singleton
components and one interacting two-vertex component. The latter has
three different optimizing observed patterns.

The program verifies the observable distance-two filter, a disjoint
selected-edge matching, the budget and coefficient-sign conditions,
and the global optimizer sets of two tie-selection tasks and four
combined tie/strict-reversal tasks. Each task retains three optimizing
observed patterns, while the task optimizer sets are pairwise disjoint.
Every strict-reversal task loses exactly two original units per reversal.
Independent enumeration of fixed-coordinate domains gives minimum
dimensions one and two for the two-task and four-task families,
respectively. These checks illustrate why the extension needs
disjointness rather than uniqueness. They test the finite mechanism
and the counting argument, not the asymptotic probability estimates.

\section{An exact law for the minimum common dimension}
\label{app:dimension-exact}

The order in Theorem~\ref{thm:optimal-common-dimension} can be sharpened
to a leading constant. A singleton coordinate needs to remain free
precisely when its empirical preference can be reversed strictly within
the local budget. Merely making a preference tie does not require a
second answer when the domain need only preserve one optimum.

Put
\[
 h=1-2\sqrt{q(1-q)},\quad \alpha=ch,\quad
 \lambda=\frac12\log\frac{1-q}{q},\quad
 r_n=\frac{n^{1-\alpha}}{\sqrt{\log n}},
\]
and, for $\Gamma>0$, define
\begin{equation}
 C(c,q,\Gamma)=
 \frac{3}{\sqrt{4\pi c\sqrt{q(1-q)}}}
       \sum_{\substack{j\in\mathbb Z\\ |j|<\Gamma}}e^{\lambda j}.
 \label{eq:exact-dimension-constant}
\end{equation}
Set $C(c,q,0)=0$. The strict inequality in this sum is part of the
statement, including when $\Gamma$ is an integer.

\subsection{The local limit and its fixed-size count}

\begin{lemma}[Signed-degree local limit]
\label{lem:exact-field-local-limit}
For every fixed integer $j$, under either the uniform size-$m$ mask or
Bernoulli masking with probability $p=m/n^3$, where
$m=\lceil cn\log n\rceil$,
\begin{equation}
 \Pr\{D_v=j\}
 =(1+o(1))
 \frac{n^{-\alpha}e^{\lambda j}}
      {\sqrt{4\pi c\sqrt{q(1-q)}\log n}}.
 \label{eq:exact-field-local-limit}
\end{equation}
The error is uniform over any fixed finite set of integers.
If $0<\alpha<1$, then in each factor block $B$, jointly over such a set,
\begin{equation}
 \frac{\#\{v\in B:D_v=j\}}{r_n}
 \xrightarrow{\Pr}
 \frac{e^{\lambda j}}{\sqrt{4\pi c\sqrt{q(1-q)}}}.
 \label{eq:exact-field-count}
\end{equation}
These count limits hold for the uniform size-$m$ mask.
\end{lemma}
\begin{proof}
Write $\ell_n=m/n$ and $d=\deg_H(v)$. We first justify a
\emph{multiplicative} Poisson approximation for the rare event in
\eqref{eq:exact-field-local-limit}. For the fixed-size mask,
\[
 \Pr\{d=k\}
 =\binom mk\frac{(n^2)_k(n^3-n^2)_{m-k}}{(n^3)_m}.
\]
For any fixed $A>0$, comparison of the falling factorials with their
powers, and then of $\mathrm{Bin}(m,1/n)$ with
$\mathrm{Pois}(\ell_n)$, gives, uniformly for $0\le k\le A\log n$,
\begin{equation}
 \Pr\{d=k\}
 =(1+o(1))e^{-\ell_n}\frac{\ell_n^k}{k!}.
 \label{eq:exact-degree-poisson}
\end{equation}
For the first comparison the logarithmic error is
$O(k^2/n^2+m^2/n^3)$; for the second it is
$O(k^2/m+k/n+m/n^2)$. Both tend to zero uniformly on this range.
The binomial degree $\mathrm{Bin}(n^2,p)$ under Bernoulli masking
satisfies the same approximation, with logarithmic error
$O(k^2/n^2+kp+n^2p^2)$.

For all three degree laws, the factorial moment bound
$\E(d)_t\le\ell_n^t$ gives
\[
 \Pr\{d\ge t\}\le \ell_n^t/t!\le(e\ell_n/t)^t.
\]
In the hypergeometric case this follows from
$\E(d)_t=(m)_t(n^2)_t/(n^3)_t\le(m/n)^t$.
Choose fixed $A$ large enough that the tail above $A\log n$ is
$o(n^{-\alpha}/\sqrt{\log n})$. Conditional on $d$, the noise signs
have the same independent law in all three models. Summing
\eqref{eq:exact-degree-poisson} times their conditional probabilities,
and using this tail bound, reduces the desired mass to that of
$P_+-P_-$, where $P_+$ and $P_-$ are independent Poisson variables
with means $\ell_n(1-q)$ and $\ell_nq$.

Exponential tilting by $e^{-\lambda(P_+-P_-)}$ makes both means equal
to $\nu_n=\ell_n\sqrt{q(1-q)}$, and its normalizing factor is
$e^{-h\ell_n}$. Under the tilted law, Fourier inversion gives
\[
 \Pr_{\rm tilt}\{P_+-P_-=j\}
 =\frac1{2\pi}\int_{-\pi}^{\pi}
       e^{-2\nu_n(1-\cos t)}e^{-ijt}\,dt
 \sim\frac1{\sqrt{4\pi\nu_n}}.
\]
To verify the last equivalence, substitute $u=t\sqrt{2\nu_n}$.
The integrand tends to $e^{-u^2/2}$ for every fixed $u$, and
$1-\cos t\ge2t^2/\pi^2$ on $[-\pi,\pi]$ gives an integrable
Gaussian majorant after this substitution. Dominated convergence
applies, uniformly for bounded $j$. Multiplication by
$e^{-h\ell_n+\lambda j}$ proves
\eqref{eq:exact-field-local-limit}, since
$\ell_n=c\log n+O(n^{-1})$.

For the count, first use Bernoulli masking. Within a given block the
incident location sets are disjoint, so the indicators of $D_v=j$ are
independent. The local limit gives their mean as
$(a_j+o(1))r_n$, where
$a_j=e^{\lambda j}/\sqrt{4\pi c\sqrt{q(1-q)}}$.
For each fixed $\varepsilon>0$, a deviation exceeding
$\varepsilon r_n$ has probability
$\exp[-\Omega_{\varepsilon,c,q,j}(r_n)]$ by the binomial Chernoff
bound. Union over the three blocks and the finite collection of $j$.
The total mask size has mean $m$, and the probability that it equals
$m$ is $\Theta(m^{-1/2})$. Conditioning therefore multiplies the
failure probability by at most $O(\sqrt m)$. Since $\alpha<1$,
$r_n/\log m\to\infty$, so the conditioned failure probability still
tends to zero. This proves \eqref{eq:exact-field-count}.
\end{proof}

\subsection{An observable restriction with the leading dimension}

Fix a deterministic budget envelope $\overline\Gamma>0$ and
$J\ge2$ with $J>1/\alpha$. Construct the common domain $U$ for
$\overline\Gamma$ and compute an original optimizer $w$ using its
unweighted component tables. On the common probability event this is
a global optimizer, and $w=r$ outside $U$.

For each $0\le\Gamma\le\overline\Gamma$, let $V_\Gamma$ contain
every vertex in a nonsingleton component of $G_H[U]$, together with
the nonisolated singleton vertices satisfying $d_v^w<\Gamma$.
Freeze all other coordinates to $w$. At $\Gamma=0$, simply take $V_0=\varnothing$ (one original optimum suffices).

\begin{proposition}[Trimming the common restriction]
\label{prop:exact-domain-trim}
On the common probability event, the domain
$x_{V_\Gamma^c}=w_{V_\Gamma^c}$ contains a global optimizer of every
$F_\xi$ with $\xi\in\mathcal U_\Gamma(H)$, simultaneously for
$0\le\Gamma\le\overline\Gamma$. The references, components, and
signed degrees are computed in $O_J(m+n)$ work after acquisition.
For a specified budget the additional trimming takes $O(n)$ work.
\end{proposition}
\begin{proof}
The original common domain preserves every required optimum, up to
whole-block gauge. Its objective separates across the components of
$G_H[U]$. At a singleton vertex put $t_v=x_vw_v$. Its term is
$(d_v^w+\delta_v)t_v$, with $d_v^w\ge0$ by optimality of $w$ and
$|\delta_v|\le\Gamma$ by the local budget. If $d_v^w\ge\Gamma$,
the choice $t_v=+1$ is always maximizing, including equality.
A vertex with no incident observation can also be fixed to $w$.
These choices do not affect another component, so they preserve at
least one global optimum of every objective. Retaining the other
components proves the claim. The work follows from the existing
component tables and one incidence scan.
\end{proof}

At an integer boundary a frozen coordinate can also have another
maximizing sign. The proposition preserves one optimum of each
objective; it need not retain every optimizer. This is the property
in the definitions of $d_{\mathrm{fix}}$ and $N_{\mathrm{hit}}$.

\subsection{Matching upper and lower constants}

\begin{theorem}[Exact common-dimension law]
\label{thm:exact-common-dimension}
Under the retained-source assumptions, fix $0<\alpha<1$,
$\Gamma>0$, and an integer $J\ge2$ with $J>1/\alpha$. Then
\begin{equation}
 \frac{d_{\mathrm{fix}}(H,Y;\Gamma)}{r_n}
 \xrightarrow{\Pr} C(c,q,\Gamma),\qquad
 \frac{\log_2N_{\mathrm{hit}}(H,Y;\Gamma)}{r_n}
 \xrightarrow{\Pr} C(c,q,\Gamma).
 \label{eq:exact-common-dimension}
\end{equation}
The observable restriction in
Proposition~\ref{prop:exact-domain-trim} attains this constant.
It also constructs
$2^{(C(c,q,\Gamma)+o_{\Pr}(1))r_n}$ objectives whose coefficients
retain their original signs and whose global optimizer sets are
pairwise disjoint.

For every fixed $\overline\Gamma>0$, both convergences are uniform
over $0<\Gamma\le\overline\Gamma$: the supremum of each absolute
normalized error tends to zero in probability. The same acquired
domain for $\overline\Gamma$ supplies all the trimmed restrictions.
\end{theorem}
\begin{proof}
The proof of Theorem~\ref{thm:common-reduction} supplies a fixed
$L\ge\overline\Gamma$ such that $U$ and the source-coordinate
errors of $r$ lie in $W_L^*=\{v:D_v\le L\}$. Since $w=r$ outside
$U$, its errors also lie in $W_L^*$. Let $B$ be the vertices of
$W_L^*$ within graph distance two of another member of $W_L^*$.
Lemma~\ref{lem:nearby-weak-pairs} gives $|B|=o_{\Pr}(r_n)$.

For $v\in W_L^*\setminus B$, all its neighbors lie outside
$W_L^*$ and are correct in $w$. Therefore
$d_v^w=(w_v\sigma_v)D_v=|D_v|$, with either choice of $w_v$ at
zero. Such a vertex is a singleton whenever it belongs to $U$.
Conversely, if $|D_v|<\Gamma\le\overline\Gamma$ and $v\notin B$,
then
$d_v(r)=(r_v\sigma_v)D_v\le|D_v|\le\lfloor\overline\Gamma\rfloor$,
so $v\in U$ and has distance greater than two from $U\setminus\{v\}$.

Define
\[
 S_\Gamma=\{v:\deg_H(v)>0,\ |D_v|<\Gamma\}.
\]
Every nonsingleton component of $U$ is contained in $B$, and the
singleton trimming agrees with $S_\Gamma$ outside $B$. Consequently,
simultaneously for all $0<\Gamma\le\overline\Gamma$,
\begin{equation}
 |V_\Gamma|\le |S_\Gamma|+|B|.
 \label{eq:exact-dimension-upper-count}
\end{equation}
Isolated vertices have probability
$(1+o(1))e^{-m/n}=n^{-c+o(1)}$. Their total count is
$o_{\Pr}(r_n)$ because $c>\alpha$.
Lemma~\ref{lem:exact-field-local-limit} thus gives
\begin{equation}
 |S_\Gamma|/r_n=C(c,q,\Gamma)+o_{\Pr}(1),
 \label{eq:exact-dimension-central-count}
\end{equation}
uniformly in the displayed budget interval: only finitely many integer
field values occur in these sets.

For the lower bound, use the observable distance-two filter $I(U)$
from Appendix~\ref{app:dimension-optimality}. Select its vertices
with $d_v^w<\Gamma$ and both a positive and a negative incident
contribution $Y_eT_e(w)$; call their number $R_\Gamma$.
All members of $S_\Gamma\setminus B$ are selected except possibly
vertices whose incident noise signs are all the same. A vertex in
this exception with $|D_v|<\overline\Gamma$ has unsigned degree
less than $\overline\Gamma$. For each fixed $k$, its degree
probability is $O(n^{-c}(\log n)^k)$ by
\eqref{eq:exact-degree-poisson}. Summing over finitely many $k$ and
$3n$ vertices, then using Markov's inequality, shows that the total
exception is $o_{\Pr}(r_n)$. Hence, uniformly in the budget,
\[
 R_\Gamma\ge |S_\Gamma|-o_{\Pr}(r_n).
\]

Choose one incident edge for each selected vertex; the edges are
pairwise disjoint and each meets $U$ only at that vertex. For a
positive selected degree $j=d_v^w$, choose a negative incident edge
and a dyadic $a_j$ with $j<a_j<\Gamma$. The two changes
$\xi_{e_v}=0$ and $\xi_{e_v}=a_jY_{e_v}$ give fields $j$ and
$j-a_j$, respectively. Both preserve the coefficient's original
sign. For a zero degree use
$\xi_{e_v}=\varepsilon\tau_vT_{e_v}(w)$, with
$\tau_v\in\{-1,+1\}$ and a dyadic
$0<\varepsilon<\min\{1,\Gamma\}$. The resulting field is
$\varepsilon\tau_v$, and coefficient signs again remain unchanged.

The matching controls every local load by $\Gamma$ and makes these
choices independent. Every global optimizer has a representative in
the original common domain, where each selected coordinate has its
unique prescribed sign. Two different task vectors therefore force
different products on a selected observed edge. Their global
optimizer sets of complete tensors are disjoint. This gives
$N_{\mathrm{hit}}\ge2^{R_\Gamma}$, while the trimmed restriction
gives $d_{\mathrm{fix}}\le|V_\Gamma|$. Together with
$\log_2N_{\mathrm{hit}}\le d_{\mathrm{fix}}$, we obtain
\[
 |S_\Gamma|-o_{\Pr}(r_n)
 \le\log_2N_{\mathrm{hit}}(\Gamma)
 \le d_{\mathrm{fix}}(\Gamma)
 \le|S_\Gamma|+o_{\Pr}(r_n).
\]
All exceptions are bounded by the same random quantity independent
of $\Gamma$ in the fixed envelope. This proves the uniform statement
and \eqref{eq:exact-common-dimension}. For a fixed budget the finitely
many dyadic changes have constant precision. Budgets approaching zero
or an integer from above may require increasing precision, which is
charged in their descriptions.
\end{proof}

For each integer $k\ge0$, $C(c,q,\Gamma)$ is constant on
$(k,k+1]$. At an integer $k\ge1$, the fields with $|D_v|=k$ can
be made to tie, but their original optimal signs still suffice.
They contribute only after the budget exceeds $k$. The corresponding
right jump has size
\[
 \frac{3(e^{\lambda k}+e^{-\lambda k})}
      {\sqrt{4\pi c\sqrt{q(1-q)}}}.
\]
At zero, $d_{\mathrm{fix}}=0$ and $N_{\mathrm{hit}}=1$ on every
input. Every positive budget exposes the zero-field contribution
$3/\sqrt{4\pi c\sqrt{q(1-q)}}$. These discontinuities depend on
the integer-valued original coefficients.

\begin{corollary}[Exact count in the singleton regime]
\label{cor:exact-singleton-dimension}
Fix $1/2<\alpha<1$ and $\overline\Gamma>0$. With probability
$1-o(1)$, simultaneously for $0\le\Gamma\le\overline\Gamma$,
\[
 d_{\mathrm{fix}}(H,Y;\Gamma)
 =\log_2N_{\mathrm{hit}}(H,Y;\Gamma)
 =\#\{v:\deg_H(v)>0,\ |D_v|<\Gamma\}.
\]
\end{corollary}
\begin{proof}
Here $\E B_L\le n^{1-2\alpha+o(1)}=o(1)$, so the set $B$ in
the preceding proof is empty with high probability. The trimming
therefore leaves exactly $S_\Gamma$ free. For the matching lower
bound, use every vertex of $S_\Gamma$, including the possible
one-sided incident-sign cases. At a positive degree $j<\Gamma$, any
incident edge can be given change $-a_jT_e(w)$, with
$j<a_j<\Gamma$, to reverse the field; the unmodified field selects
the opposite sign. The zero-degree construction is unchanged. These
changes belong to the stated budget class, although they can reverse
an original coefficient sign in the one-sided cases. The selected
edges form a matching, so the $2^{|S_\Gamma|}$ optimizer sets are
pairwise disjoint. The matching upper and lower bounds give the
identity. At $\Gamma=0$ all quantities are zero.
\end{proof}

The corollary's exact counting family need not preserve coefficient
signs. The theorem's sign-preserving family discards only
$o_{\Pr}(r_n)$ exceptional vertices and attains the same leading
constant throughout $0<\alpha<1$. Both results concern a fixed
query-independent answer domain, as in the definitions above.

\section{Gain, multiplicity, and the critical window}
\label{app:residual-laws}

The signed-degree distribution in Appendix~\ref{app:dimension-exact}
controls more than the size of a domain for future objectives. It also
controls the gain from fitting the realized errors and the number of
exact answers before the objective changes. The following results depend
only on the target law, not on the correctness of an acquired estimate.
Write $a=e^{-\lambda}$, $A=(4\pi c\sqrt{q(1-q)})^{-1/2}$, and let
$\mathcal O_n$ denote the set of distinct complete tensors maximizing
$F_Y$. Let $\mathcal O_n^H$ be their distinct product vectors on $H$.

\begin{theorem}[Empirical gain and optimal-answer multiplicity]
\label{thm:gain-entropy}
For fixed $c>0$, $q\in(0,1/2)$ and $0<\alpha<1$ under the target law,
\begin{align}
 \frac{G_n}{r_n}&\xrightarrow{\Pr}6A\frac{a}{(1-a)^2},
 &\frac{\log_2|\mathcal O_n|}{r_n},\quad
 \frac{\log_2|\mathcal O_n^H|}{r_n}&\xrightarrow{\Pr}3A.
 \label{eq:gain-entropy}
\end{align}
If $1/2<\alpha<1$, with probability $1-o(1)$ the exact counts are
$|\mathcal O_n|=2^{\#\{v:D_v=0\}}$ and
$|\mathcal O_n^H|=2^{\#\{v:D_v=0,\ \deg_H(v)>0\}}$.
In particular, for every fixed $0<\Gamma\le1$, the minimum common
coordinate dimension and the logarithm of the number of original
optima agree to $o_{\Pr}(r_n)$, and
$G_n/\log_2|\mathcal O_n|\xrightarrow{\Pr}2a/(1-a)^2$.
\end{theorem}
\begin{proof}
Apply Theorem~\ref{thm:order-residual} at $k=3$. Its interaction estimate takes a particularly simple form for the finite checks below. In a source-aligned flip set $S\subset W$, a triple meeting $S$ twice or three times contributes an error $4\eta_e$ to the sum of individual gains. With maximum pair codegree two, at most $2P_W$ such rows occur, where $P_W$ counts adjacent pairs in $W$. Thus
\begin{equation}\label{eq:gain-interaction-remainder}
 \left|-2J(S)+2\sum_{v\in S}D_v\right|\le8P_W.
\end{equation}
The $3A$ zero-field coefficient is positive. Dividing the gain limit by the log multiplicity limit gives the stated ratio. For $0<\Gamma\le1$ the only strictly reversible field is zero, giving the same leading coordinate dimension.
\end{proof}

\subsection{A Poisson window for the last empirical corrections}
The boundary $\alpha=1$ for fixed $c$ conceals a logarithmic correction.
Fix $q\in(0,1/2)$, put $h=1-2\sqrt{q(1-q)}$ and $c_*=1/h$, and now
allow the deterministic integer $m=m_n$ to vary so that
\begin{equation}
 h\frac{m}{n}=\log n-\frac12\log\log n+t+o(1),\qquad t\in\mathbb R.
 \label{eq:poisson-window}
\end{equation}
Use the same uniform size-$m$ mask and independent target errors.
Define $A_*=(4\pi c_*\sqrt{q(1-q)})^{-1/2}$ and
$\theta=3A_*e^{-t}$. Let $P_j$, $j\in\mathbb Z$, be independent
Poisson random variables of means $\theta e^{\lambda j}$.

\begin{theorem}[Critical-window distribution]
\label{thm:critical-window}
For each fixed finite collection of integer fields the counts
$\#\{v:D_v=j\}$ converge jointly in distribution to the corresponding
$P_j$. In the window \eqref{eq:poisson-window}, jointly for any fixed
finite collection of positive bounded budgets,
\begin{align}
 G_n&\Rightarrow 2\sum_{j\ge1}jP_{-j},\quad
 \log_2|\mathcal O_n|,\ \log_2|\mathcal O_n^H|&\Rightarrow P_0,\
 d_{\mathrm{fix}}(\Gamma),\ \log_2 N_{\mathrm{hit}}(\Gamma)
 &\Rightarrow \sum_{|j|<\Gamma}P_j.
 \label{eq:poisson-window-limits}
\end{align}
The infinite sum is almost surely finite. The probabilities that the
source is optimal and that it is the unique optimal tensor tend to,
respectively,
\begin{equation}
 \exp[-\theta a/(1-a)],\qquad \exp[-\theta/(1-a)].
 \label{eq:poisson-window-thresholds}
\end{equation}
This is a statement about the random target and the minimum dimensions;
it does not assert uniform finite acquisition bounds for varying density.
\end{theorem}
\begin{proof}
Corollary~\ref{cor:order-critical} with $k=3$ gives the joint limits, including the independent Poisson counts at positive and negative fields. Its uniform negative-field tail justifies the infinite weighted sum. The source is optimal exactly when the negative count is zero on the localization event, and it is unique when the zero count vanishes as well. Evaluating these two Poisson probabilities gives~\eqref{eq:poisson-window-thresholds}.
\end{proof}

\subsection{Finite checks and field-count diagnostics}
\label{app:residual-law-checks}
The executable \texttt{verify\_residual\_laws.py} fixes its seeds,
sizes and replicate counts before running. It independently enumerates
all factor assignments on 180 targets ($n=2,3,4$), scoring 280,320
assignments and checking the exact interaction identity for each.
The weak-set interaction bound is checked for 4,326 subsets. Among the
540 target--cutoff pairs tested ($L=0,1,2$), 325 satisfy the stated
localization and gauge-injectivity conditions; all satisfy the gain
and the two count bounds, and all 299 qualifying singleton cases obey
the exact count and gain formulas. The other 215 pairs are not used
as localization witnesses.

A separate diagnostic samples 128 uniform size-$m$ masks at each of
four $n$ values, with $q=.17$ and window offset $t=0$. It counts fields
and does not solve the large empirical objectives. Table~\ref{tab:field-window}
reports all replications, alongside the analytical limit. The sample
means and probabilities are descriptive; finite source optimality
is not inferred from absence of a negative field in this diagnostic.
\begin{table}[h]
\centering
\caption{Field-only diagnostics at the critical window. The last two columns are the numbers of trials (out of 128) with no negative and no nonpositive fields, respectively. The limit row reports expected counts on the same 128-trial scale.}
\label{tab:field-window}
\begin{tabular}{rr|rrr|rr}
\toprule
$n$ & $m$ & Mean $\#D=-1$ & Mean $\#D=0$ & Mean $\#D=1$ & No negative & No nonpositive\\
\midrule
128&2091&.336&.766&1.656&74&36\\
512&10957&.344&.805&1.695&75&37\\
2048&54416&.313&.734&1.523&75&37\\
8192&260570&.375&.625&1.445&63&34\\
Limit& &.312&.689&1.522&72.44&36.38\\
\bottomrule
\end{tabular}
\end{table}

\section{Finite certificates for a common restriction}
\label{app:robust-certificates}

The common probability event in Theorem~\ref{thm:common-reduction} proves
that a restriction is valid under the model. This appendix gives separate
finite certificates. Acceptance is deterministic and uniform over the
declared coefficient neighborhood. A rejected certificate need not indicate
that the restriction is invalid.

\subsection{Gauge, projection, and the exact robustness cost}

Fix a finite tripartite mask $H$, coefficients $Y_e$, a sign reference $r$,
and a proposed free set $U$. Coefficients may be arbitrary encoded rationals
in this appendix; the earlier model uses $Y_e\in\{-1,1\}$.
Here $\Gamma$ is one finite encoded budget; setting $\Gamma=\Gamma_n$
and $Y_e\in\{-1,1\}$ recovers the main model at that $n$. Select two
anchor vertices in different blocks, outside $U$, and define the canonical
domain $\mathcal X$ by fixing their signs to the corresponding entries of
$r$. Reversing any two whole blocks leaves every $T_e$ unchanged. Thus
every tensor has a representative in $\mathcal X$, simultaneously for
every coefficient array, and optimizing over $\mathcal X$ preserves the
unrestricted optimum. If necessary the two anchors can be removed from
the proposed free set before certification; this changes the proposed
restriction and must be included in its reported size.

Let $P=P_{U,r}$ retain the coordinates in $U$ and replace every other
coordinate by its reference sign. It maps $\mathcal X$ into itself and
is the identity on its image. Define
\[
 \Delta_P(x)=F_0(Px)-F_0(x),\qquad
 A_P(x)=\{e\in H:T_e(Px)\ne T_e(x)\}.
\]
For a subhypergraph $A\subseteq H$, write
\begin{equation}
 \nu^*(A)=
 \max\left\{\sum_{e\in A}t_e:
             t_e\ge0,\ \sum_{\substack{e\in A\\e\ni v}}t_e\le1
             \text{ for every }v\right\}.
 \label{eq:certificate-fractional-matching}
\end{equation}
The value is zero for the empty subhypergraph.

\begin{theorem}[Exact robust improvement condition]
\label{thm:robust-projection}
For every $x\in\mathcal X$ and every $\Gamma\ge0$,
\begin{equation}
 \min_{\xi:\,\max_v\sum_{e\ni v}|\xi_e|\le\Gamma}
       \{F_\xi(Px)-F_\xi(x)\}
 =\Delta_P(x)-2\Gamma\nu^*(A_P(x)).
 \label{eq:robust-projection-identity}
\end{equation}
Consequently, $P$ improves every assignment in $\mathcal X$ for every
allowed objective if and only if
\begin{equation}
 \Delta_P(x)\ge2\Gamma\nu^*(A_P(x))
 \quad\text{for every }x\in\mathcal X.
 \label{eq:robust-projection-criterion}
\end{equation}
When this condition holds, the restriction $x_{U^c}=r_{U^c}$ preserves
the global optimum of every allowed objective, including objectives
chosen after the certificate is verified.
\end{theorem}
\begin{proof}
The perturbation contribution is
$\sum_e\xi_e\{T_e(Px)-T_e(x)\}$. It vanishes outside $A_P(x)$;
on $A_P(x)$ its coefficient is either $+2$ or $-2$. An adversary may
therefore choose the sign of every nonzero $\xi_e$ independently to make
that term $-2|\xi_e|$. For $\Gamma>0$, setting
$t_e=|\xi_e|/\Gamma$ gives exactly
\eqref{eq:certificate-fractional-matching}. Conversely, a maximizing
fractional matching with these signs is an admissible perturbation, so
the bound is attained. The case $\Gamma=0$ follows directly.
This proves the identity and its equivalence with uniform improvement.
For any allowed objective choose a global optimizer, take its canonical
gauge representative $x$, and apply $P$. Its score cannot decrease,
so the restricted image contains a global optimizer.
\end{proof}

The last implication is sufficient for preserving optimum values, not
an equivalence. For example, $F(x,y)=x-y-2xy$ has unique maximum $4$ at
$(+1,-1)$. The restriction $x=+1$ preserves that maximum, but its projection
sends $(-1,+1)$, of score $0$, to $(+1,+1)$, of score $-2$.
This already separates the conditions at $\Gamma=0$.

Gauge selection cannot be omitted. For the one-triple objective $abc$,
reference $(+,+,+)$, and free set $\{a\}$, a raw projection sends the
optimal assignment $(-,-,+)$ to $(-,+,+)$ and decreases its score.
Fixing the two outside anchors $b=c=+1$ removes this irrelevant gauge
obstruction. The global optimum is unchanged by that canonicalization.

Equation~\eqref{eq:robust-projection-criterion} connects the uniform budget
to the established improving-mapping approach to persistency
\citep{shekhovtsov2016persistency}. Its fractional matching measures the
largest admissible adversarial mass on the observations that a particular
projection changes. Evaluating the matching for a given $x$ is a linear
program. Verifying the condition for every $x$ remains a separate
computational task.

\subsection{Polynomial-size certificates from conditioned relaxations}

For computation, the declared budget $\Gamma$ is also an encoded
nonnegative rational. The identity above itself requires no rationality
assumption on the budget.
Put
\[
 h_e(s)=Y_e\{T_e(Ps)-T_e(s)\},\qquad
 a_e(s)=\one\{T_e(Ps)\ne T_e(s)\},
\]
where $P$ acts on the coordinates of that triple in the same way as the
global projection. Consider an active vertex $w\notin U$ that is not an
anchor. Here active means incident to an observation. Impose the anchor
signs and $x_w=-r_w$.
Let $\mu_v$ be a distribution on the allowed signs of $v$, and let
$\mu_e$ be a distribution on the locally allowed sign patterns of $e$.
Require normalization, nonnegativity, and consistency:
\[
 \sum_{s:\,s_v=t}\mu_e(s)=\mu_v(t)
 \quad(e\ni v,\ t\text{ allowed at }v).
\]
These constraints define the usual local marginal relaxation; they do
not assert the existence of a joint distribution.
Add nonnegative variables $t_e$ and minimize
\begin{align}
 L_w(U)=\min_{\mu,t}\quad&
      \sum_{e,s}h_e(s)\mu_e(s)-2\Gamma\sum_e t_e,
       \label{eq:conditioned-certificate-lp}\\
 \text{subject to}\quad&
 t_e\le\sum_s a_e(s)\mu_e(s),\qquad
 \sum_{e\ni v}t_e\le1. \nonumber
\end{align}
If no active non-anchor vertex remains outside $U$, no such program is
needed.

\begin{proposition}[Conditioned finite certificate]
\label{prop:conditioned-certificate}
For each active non-anchor $w\notin U$, any verified lower bound
$\underline L_w\le L_w(U)$ is also a lower bound on
\[
 \min_{\substack{x\in\mathcal X\\x_w=-r_w}}
       \{\Delta_P(x)-2\Gamma\nu^*(A_P(x))\}.
\]
If every $\underline L_w$ is nonnegative, the restriction preserves
every allowed objective's global optimum.
There are at most $3n-2$ programs, each with $O(m+n)$ variables and
constraints for triple factors.
\end{proposition}
\begin{proof}
Each assignment in the displayed minimum supplies integral local
distributions. For these distributions, optimization over $t$ is exactly
the fractional matching on its affected triples: the negative coefficient of $\sum_e t_e$ converts that inner maximization into the joint minimization. Relaxing the
distributions can only lower the minimum.
Every assignment on which $P$ changes an active coordinate belongs to at
least one conditioned domain. If $P$ changes only isolated coordinates,
both objectives have exactly the same value before and after projection.
The claimed simultaneous improvement follows, and
Theorem~\ref{thm:robust-projection} gives the conclusion.
\end{proof}

An exact dual witness is easy to check without trusting a numerical
solver's reported optimality tolerance. For one conditioned domain, supply
rational numbers $\lambda_v,\rho_e\ge0$, messages $f_{ev}(t)$, and
constants $c_e,c_v$ satisfying
\begin{align}
 \rho_e+\sum_{v\in e}\lambda_v&\ge2\Gamma, \label{eq:robust-dual-cover}\\
 h_e(s)-\rho_ea_e(s)-\sum_{v\in e}f_{ev}(s_v)&\ge c_e
     &&\text{for every allowed }s, \nonumber\\
 \sum_{e\ni v}f_{ev}(t)&\ge c_v
     &&\text{for every allowed }t. \nonumber
\end{align}
Then
\begin{equation}
 \underline L_w=\sum_e c_e+\sum_v c_v-\sum_v\lambda_v
 \label{eq:robust-dual-value}
\end{equation}
is a valid lower bound. To see this, multiply the last two inequalities by
the local distributions and use their consistency to cancel the messages.
The first inequality and the capacity constraints bound the remaining
negative contribution by
$-\sum_e\rho_e\sum_s a_e(s)\mu_e(s)-\sum_v\lambda_v$.
This yields \eqref{eq:robust-dual-value}.
There are at most eight factor patterns and two unary states to check
per triple and vertex. Verification is polynomial in the supplied
rational encoding length. Supplying the equivalent ordinary
linear-programming dual is sufficient as well.

For clarity, if the assembled program is
$\min\{c^\top z:A_{\rm eq}z=b_{\rm eq},
A_{\rm ub}z\le b_{\rm ub},z\ge0\}$, the exact checks are
\[
 y\le0,\qquad
 c-A_{\rm eq}^{\top}u-A_{\rm ub}^{\top}y\ge0,\qquad
 b_{\rm eq}^{\top}u+b_{\rm ub}^{\top}y\ge0.
\]
All three are rational inequalities. A floating-point primal value near
zero is not a certificate. Rational reconstruction may conservatively
fail even when a numerical solver suggests acceptance.
The finite implementation verifies the reconstructed dual, including
any explicitly charged outward repairs, before accepting it.

\paragraph{Why conditioning is needed.}
The analogous unconditioned program has a structural limitation. Its
nonnegative test is equivalent to the test obtained by replacing the
local budget with the more conservative independent bound
$|\xi_e|\le\Gamma$ on every edge.
Indeed, write
$B(\mu)=\sum_{e,s}\{h_e(s)-2\Gamma a_e(s)\}\mu_e(s)$.
Every feasible unconditioned pair satisfies objective $\ge B(\mu)$.
Conversely, if a locally consistent $\mu$ has $B(\mu)<0$, mix it with
the integral distributions of the reference $r$, which is a projection
fixed point. For sufficiently small $\epsilon>0$ set
$t_e=\epsilon\sum_s a_e(s)\mu_e(s)$.
All vertex capacities are then satisfied, while the objective is
$\epsilon B(\mu)<0$.
Thus the unconditioned test cannot use the shared budgets to certify
anything beyond that independent-edge relaxation. Forcing a changed
coordinate removes the projection fixed points responsible for this
degeneracy.

\subsection{A limitation of the first local relaxation}

Conditioning repairs the fixed-point degeneracy but does not make the
first local relaxation exact.

\begin{proposition}[Locally satisfying marginals can defeat certification]
\label{prop:certificate-local-obstruction}
Normalize $r$ to all positive signs and absorb its products into $Y$.
Take $U=\varnothing$. Fix two gauge anchors to $+1$ and condition a
third, distinct coordinate $w$ to $-1$.
If every observed triple contains at most one of these three fixed
coordinates, then
\begin{equation}
 L_w(\varnothing)\le F_0(r)-\sum_{e\in H}|Y_e|.
 \label{eq:local-certificate-obstruction}
\end{equation}
In particular, for binary coefficients with at least one negative
observation, this bound is strictly negative for every $\Gamma\ge0$.
Under the random target law with fixed $q>0$, $\alpha>1$, and bounded
$\Gamma$, fixed anchors in the first two blocks and a fixed coordinate
in the third block satisfy this obstruction with probability $1-o(1)$,
even though the acquired empty restriction is globally exact with
probability $1-o(1)$.
\end{proposition}
\begin{proof}
Give each unfixed unary coordinate the uniform sign distribution.
Every factor has at least two unfixed coordinates. On that factor, take
the uniform distribution among its assignments satisfying
$T_e=\operatorname{sign}(Y_e)$ and respecting any fixed coordinate;
when $Y_e=0$, choose either parity.
Each unfixed coordinate remains marginally uniform, so all factor
distributions are locally consistent. Set every $t_e=0$.
Since $P$ maps to $r$, the resulting objective is
$\sum_e\{Y_eT_e(r)-|Y_e|\}$, proving
\eqref{eq:local-certificate-obstruction}.
For a uniform size-$m$ mask, any specified cross-block pair lies in an
observed triple with probability at most $m/n^2$.
A union bound over the three pairs gives exceptional probability
at most $3m/n^2=O(\log n/n)$.
With fixed $q>0$, the probability of no negative source-relative error is
$(1-q)^m=o(1)$.
For $\alpha>1$, the earlier acquisition, reuse, and source-optimality
theorems give $r$ equal to the source in its reference gauge and
$U=\varnothing$ with probability $1-o(1)$.
Intersecting these events proves the last statement.
\end{proof}

This is an obstruction to this particular factor/unary relaxation.
It is not a lower bound against arbitrary finite certificates or
against higher-order relaxations. Accurate additional measurements and
global certification impose different requirements in this example.

\paragraph{A finite robust restriction rejected by the relaxation.}
Take $n=4$ with indices $0,1,2,3$, the nine triples
$H=\{(i,j,0):i,j\in\{1,2,3\}\}$, and all positive coefficients except
$Y_{110}=-1$. Set $r$ to all positive signs, $U=\varnothing$, and fix
the isolated anchors $a_0=b_0=+1$. Every observed product pattern
different from the all-positive pattern disagrees with it in at least three
entries of this $3\times3$ rank-one sign matrix. Its original score
therefore lies at least two below $F_0(r)=7$. Since every edge contains
$c_0$, the total absolute perturbation mass is at most $\Gamma$.
Thus $r$ remains globally optimal for every allowed objective when
$\Gamma\le1$.

Conditioning $a_1=-1$ nevertheless gives
$L_{a_1}(\varnothing)=-2-2\Gamma$.
The locally satisfying marginals above contribute $-2$; assigning
$t_{110}=1$ and all other matching variables zero contributes
$-2\Gamma$. This attains the lower bound because the positive factors
have $h_e\ge0$, the negative factor has $h_{110}\ge-2$, and
$\sum_e t_e\le1$ follows from the capacity at $c_0$.
The explicit-cover check below accepts the empty restriction for every
$\Gamma\le1$: each changed product pattern has cover bound one,
and the unchanged pattern has cover bound zero. At $\Gamma=1/2$ the
relaxation value is exactly $-3$, although the certified restriction
has no free coordinate. The anchors in this example are declared as
part of the instance; no claim about a data-dependent anchor rule is
needed.

\subsection{An exact finite check using explicit vertex covers}

A less expensive sufficient bound than enumerating every assignment and
solving its fractional matching can be evaluated by exploiting the
tripartite form. Let $V_i^+$ denote the active vertices of block $i$ and
put $N_i=|V_i^+|$. For $m=0$ all objectives are zero and certification is
immediate; hence assume $m>0$. Every $V_i^+$ covers every observed
triple. Let $\mathcal G$ consist of the empty set and the three unions of
two active blocks.
For an assignment $x$, let
$S(x)=\{v\in V^+\setminus U:x_v\ne r_v\}$.
Every affected edge meets $S(x)$ oddly and therefore meets
$S(x)\mathbin{\triangle}g$ oddly for every $g\in\mathcal G$.
These sets are vertex covers of $A_P(x)$.
Weak duality for fractional matching gives
\begin{equation}
 \nu^*(A_P(x))\le
 C_U(x):=\min\left\{N_1,N_2,N_3,\
                 |S(x)\mathbin{\triangle}g|:g\in\mathcal G\right\}.
 \label{eq:certificate-explicit-cover}
\end{equation}
When all vertices are active this is
$\min\{n,|S(x)\mathbin{\triangle}g|:g\in\mathcal G\}$.

\begin{proposition}[An exact exponential certificate check]
\label{prop:cover-certificate-check}
The condition
\begin{equation}
 \min_{x\in\mathcal X}\{\Delta_P(x)-2\Gamma C_U(x)\}\ge0
 \label{eq:cover-certificate-test}
\end{equation}
is sufficient for a common exact restriction.
For rational input, its left side can be evaluated exactly in
$O(4^{n-1}(m+n\log(n+2)))$ arithmetic operations and $O(m+n)$ working
memory, apart from output.
The bound is explicitly exponential; it is a finite certificate check,
not an improvement to the near-linear acquisition theorem.
\end{proposition}
\begin{proof}
Sufficiency follows from
\eqref{eq:certificate-explicit-cover} and
Theorem~\ref{thm:robust-projection}.
Normalize $x_v$ by $r_v$, absorbing $T_e(r)$ into each base coefficient,
so the reference is all positive. Choose the first and second blocks
as the anchored blocks, relabeling blocks if necessary.
Enumerate their canonical assignments $a,b$, with their two anchor
signs positive. Let $a',b'$ be their projections, and define
\[
 h_k=\sum_{(i,j,k)\in H}Y_{ijk}a_i b_j,\qquad
 h'_k=\sum_{(i,j,k)\in H}Y_{ijk}a'_i b'_j .
\]
Let $a_-$ and $b_-$ count negative active outside coordinates in the
first and second blocks. If $c_-$ is the analogous count in the third,
enumerating the four gauges yields
\begin{align*}
 A&=\min\{a_-+b_-,\,N_1+N_2-a_--b_-\},\\
 B&=\min\{N_1+N_3-a_-+b_-,\,
          N_2+N_3+a_--b_-\},\\
 C_U(x)&=\min\{N_1,N_2,N_3,A+c_-,B-c_-\}.
\end{align*}
For $k\in U$ in the third block, the change in objective is
$(h'_k-h_k)c_k$, whose minimum is $-|h'_k-h_k|$.
For active $k\notin U$, it is $h'_k-h_kc_k$.
The all-positive outside-third assignment therefore gives the base value
\[
 B_0=\sum_{\substack{k\notin U\\k\in V_3^+}}(h'_k-h_k)
       -\sum_{k\in U\cap V_3^+}|h'_k-h_k|.
\]
Changing exactly $j$ active outside-third signs adds twice the sum of
the $j$ smallest $h_k$ among those coordinates. Let this prefix sum be
$H_j$, including $H_0=0$. The minimum for the current $a,b$ is exactly
\begin{equation}
 \min_{0\le j\le|V_3^+\setminus U|}
 \left[
 B_0+2H_j-2\Gamma
       \min\{N_1,N_2,N_3,A+j,B-j\}
 \right].
 \label{eq:certificate-third-block-sort}
\end{equation}
Isolated coordinates contribute zero and may be fixed to their reference
signs during the check.
Computing both field arrays costs $O(m+n)$ per first-two-block
assignment, sorting costs $O(n\log(n+2))$, and all prefix values cost
$O(n)$. There are $4^{n-1}$ canonical assignments.
Streaming them gives the stated memory bound.
All comparisons can be performed after clearing the encoded
denominators. Their bit costs depend on the supplied precision.
\end{proof}

The cover minimum and the projection fixed points make the left side
of \eqref{eq:cover-certificate-test} at most zero. Exact acceptance
therefore means it is zero, without a numerical tolerance.
A negative value supplies an assignment failing this sufficient test.
It does not by itself prove that the exact fractional-matching condition
fails, and even failure of that stronger mapping condition would not
prove that the restricted domain misses every global optimizer.

An adaptive procedure may release an active outside coordinate changed
in a failed sufficient-test assignment and repeat the test.
At least one such non-anchor coordinate exists whenever the displayed
value is negative. Releasing coordinates changes the projection, so
every proposed restriction must be checked anew; acceptance is not
assumed monotone.
After at most $3n-2$ releases the projection changes only anchors or
isolated coordinates on its canonical domain, and the test accepts.
This termination statement does not bound the final component sizes.
A declared component/work limit may stop the procedure earlier, reporting
that no efficiently usable certified reduction was obtained.
Every accepted certificate, however obtained, applies to the entire
declared perturbation family and can be reused without new source
readings or repeated certification.

\section{Exact commonness and the obstruction to a fixed reset}
\label{app:exact-common-certification}

The restriction in Theorem~\ref{thm:common-reduction} contains an optimum
of every allowed objective. A reset projection asks more: it must improve
every competing assignment while retaining all of its free coordinates.
The two properties differ on the generated instances themselves. We first
prove this obstruction, then characterize common exactness directly.

\subsection{A finite obstruction that survives every anchor choice}

Let $H$ be a finite tripartite mask with $Y_e\in\{-1,+1\}$, and write
$\Delta(H)=\max_v\deg_H(v)$ for its maximum unsigned degree. Fix a
domain $\mathcal D=\{y:y_{U^c}=r_{U^c}\}$ and its reset $P=P_{U,r}$.
The domain is assumed to preserve an optimum of every objective in
$\mathcal U_\Gamma(H)$ for the stated budget. Let $w\in\mathcal D$
maximize the original objective. A vertex $v\in U$ is a singleton if
no observed triple contains both $v$ and another vertex of $U$.

\begin{lemma}[Finite reset loss]
\label{lem:finite-reset-loss}
Suppose $Z\subseteq U$ consists of $R$ nonisolated singleton vertices
with $d_v^w=0$. Suppose one can select an incident triple $e_v$ for
each $v\in Z$ so that these triples form a matching. For
$0<\varepsilon<\min\{1,\Gamma\}$ there is one sign-preserving
perturbation $\xi\in\mathcal U_\Gamma(H)$ such that the following
holds for every choice of two anchors outside $U$ in different blocks.
There is an assignment $x$ with the prescribed anchor signs and
\begin{equation}
 F_\xi(Px)-F_\xi(x)
 \le -\frac{4\varepsilon R}{3}+4\Delta(H)+4\Gamma.
 \label{eq:finite-reset-loss}
\end{equation}
The perturbation and the pair of blocks used in the construction may
be chosen before the anchors.
\end{lemma}
\begin{proof}
Set $\xi_{e_v}=\varepsilon T_{e_v}(w)$ on the selected triples and
zero elsewhere. Their disjointness gives local perturbation load at
most $\varepsilon$. Since $\varepsilon<1$, no coefficient changes
sign. Each selected triple meets $U$ only at its selected vertex.
Consequently, the perturbed restricted objective changes only the
individual zero fields in $Z$, making $w_v$ strictly preferred there
with field $\varepsilon$. Its other component tables are unchanged.
Thus $w$ still maximizes the restricted objective, and common
exactness makes it a global optimizer of $F_\xi$.

Among the three pairs of blocks, choose one containing at least
$2R/3$ vertices of $Z$. Reverse those two blocks of $w$, obtaining
$g w$. Every observed product is unchanged, so
$F_\xi(gw)=F_\xi(w)$. For any two prescribed outside anchors, restore
each reversed anchor to its reference sign, producing $x$. This
changes at most two coordinates. Flipping a coordinate $a$ changes
the objective by at most
$2\sum_{e\ni a}|Y_e+\xi_e|\le2(\deg_H(a)+\Gamma)$.
Therefore
\[
 F_\xi(x)\ge F_\xi(w)-4\Delta(H)-4\Gamma.
\]

The reset $Px$ agrees with $r$ outside $U$ and with $gw$ on $U$,
because both restored anchors are outside $U$. At each selected
vertex in the reversed blocks, it has the sign opposite to the
strictly preferred $w_v$, losing $2\varepsilon$. The other restricted
components cannot gain relative to $w$, which maximizes each of
their tables. Hence
\[
 F_\xi(Px)\le F_\xi(w)-2\varepsilon(2R/3).
\]
Subtracting the preceding inequalities proves
\eqref{eq:finite-reset-loss}, simultaneously for all anchor choices.
\end{proof}

\begin{theorem}[Failure of every anchored reset]
\label{thm:anchored-reset-obstruction}
Under the retained-source assumptions, fix $0<\alpha<1$,
$\Gamma>0$, and an integer $J\ge2$ with $J>1/\alpha$. Let
$(U,r)$ be the common observable restriction of
Theorem~\ref{thm:common-reduction} for this budget. With probability
$1-o(1)$, the domain is common exact, but there is one sign-preserving
$\xi\in\mathcal U_\Gamma(H)$ for which every choice of two outside
anchors in distinct blocks admits a canonical assignment $x$ with
\begin{equation}
 F_\xi(Px)-F_\xi(x)
 \le -\Omega_{c,q,s,J,\Gamma}(r_n)+O_{c,\Gamma}(\log n)<0.
 \label{eq:asymptotic-reset-loss}
\end{equation}
In particular, the exact robust improvement condition
\eqref{eq:robust-projection-criterion} fails for this restriction.
\end{theorem}
\begin{proof}
Use the common event and the observable set $Z$ from
Appendix~\ref{app:dimension-optimality}. Its cardinality is
$\Theta(r_n)$, its vertices are nonisolated zero-field singletons
relative to a computed base optimizer $w$, and the selected incident
triples form a matching. Choose a fixed dyadic
$0<\varepsilon<\min\{1,\Gamma\}$.

The maximum unsigned degree is $O_c(\log n)$ with probability
$1-o(1)$. One direct fixed-size bound is as follows. A specified
vertex has $n^2$ possible incident triples. For integer $t\ge1$,
the probability that at least $t$ of them belong to the uniform
size-$m$ mask is at most
\[
 \binom{n^2}{t}\frac{(m)_t}{(n^3)_t}
 \le \left(\frac{e m}{nt}\right)^t.
\]
Here $(m)_t/(n^3)_t\le(m/n^3)^t$. Set
$t=\lceil A\log n\rceil$ with a sufficiently large constant $A$
depending only on $c$, and sum over $3n$ vertices. The resulting
probability tends to zero. Since $|U|=O(r_n)=o(n)$, every block
contains outside vertices for all sufficiently large $n$ on this
event.

Lemma~\ref{lem:finite-reset-loss} now gives
\eqref{eq:asymptotic-reset-loss}. Its negative term dominates because
$r_n/\log n=n^{1-\alpha}/(\log n)^{3/2}\to\infty$.
All events precede the choice of anchors, so data-dependent anchors
are covered as well.
\end{proof}

This theorem concerns the fixed reset $P$, even after fixing a
canonical gauge. A sound relaxation that converges to its exact
pointwise improvement test cannot accept the displayed domain on
this event. The obstruction does not apply to every improving map,
to every certificate of common exactness, or to every finite target.
The finite results in Appendix~\ref{app:certification-experiments}
must therefore distinguish a failure of the test from a failure of
the proposed restriction.

\subsection{An exact comparison using restricted assignments}

The next results are deterministic. Let $\mathcal X$ be a finite
assignment space, let $H$ be a finite collection of nonempty factor
scopes, and let $T_e:\mathcal X\to\{-1,+1\}$ be arbitrary signed
factor functions. Parity products are one instance. Put
$F_\xi(x)=\sum_e(Y_e+\xi_e)T_e(x)$, with the same local perturbation
set $\mathcal U_\Gamma(H)$. Let $\varnothing\ne\mathcal D\subseteq
\mathcal X$ be the proposed domain. For each competitor $x$, define
\begin{equation}
 g_{\mathcal D}(x)=
 \min_{\xi\in\mathcal U_\Gamma(H)}
       \left[\max_{y\in\mathcal D}F_\xi(y)-F_\xi(x)\right].
 \label{eq:exact-competitor-gap}
\end{equation}
The domain is common exact if and only if
$g_{\mathcal D}(x)\ge0$ for every $x\in\mathcal X$.
Indeed, this says exactly that its best value dominates every
competitor for each allowed objective. When common exactness holds,
$\min_x g_{\mathcal D}(x)=0$: choose an original global optimizer
and use $\xi=0$ for the reverse inequality.

For nonnegative edge weights $a=(a_e)$, define the weighted
fractional matching value
\begin{align}
 \nu^*(a)
 &=\max\left\{\sum_e a_et_e:t_e\ge0,
                        \ \sum_{e\ni v}t_e\le1\right\}\\
 &=\min\left\{\sum_v\lambda_v:\lambda_v\ge0,
                        \ \sum_{v\in e}\lambda_v\ge a_e\right\}.
 \label{eq:weighted-fractional-matching}
\end{align}
The equality is finite linear-programming duality.

\begin{theorem}[Exact mixed comparison]
\label{thm:exact-mixed-comparison}
For every competitor $x$ and every $\Gamma\ge0$,
\begin{equation}
 g_{\mathcal D}(x)
 =\max_{\pi\in\Delta(\mathcal D)}
 \left\{\E_\pi F_0(y)-F_0(x)-\Gamma\nu^*(a(\pi,x))\right\},
 \label{eq:mixed-common-identity}
\end{equation}
where $\Delta(\mathcal D)$ is the simplex of probability
distributions on $\mathcal D$ and
\begin{equation}
 a_e(\pi,x)=\left|\E_\pi T_e(y)-T_e(x)\right|
           =1-T_e(x)\E_\pi T_e(y).
 \label{eq:mixed-common-load}
\end{equation}
Equivalently, the right side of
\eqref{eq:mixed-common-identity} is the maximum of
\begin{equation}
 \E_\pi F_0(y)-F_0(x)-\Gamma\sum_v\lambda_v
 \label{eq:mixed-common-certificate-value}
\end{equation}
over $\pi\in\Delta(\mathcal D)$ and $\lambda_v\ge0$ satisfying
\begin{equation}
 \sum_{v\in e}\lambda_v
 \ge 1-T_e(x)\E_\pi T_e(y)\quad(e\in H).
 \label{eq:mixed-common-certificate-cover}
\end{equation}
Thus common exactness holds if and only if every competitor has
such a pair $(\pi,\lambda)$ with nonnegative value.
\end{theorem}
\begin{proof}
For fixed $\xi$, maximizing over the finite set $\mathcal D$ is
equivalent to maximizing its expected value over
$\Delta(\mathcal D)$. The two sets
$\mathcal U_\Gamma(H)$ and $\Delta(\mathcal D)$ are compact convex
polytopes, and the expected comparison is bilinear. Finite
linear-programming duality therefore gives
\[
 \min_{\xi}\max_{\pi}
  \E_\pi[F_\xi(y)-F_\xi(x)]
 =\max_{\pi}\min_{\xi}
  \E_\pi[F_\xi(y)-F_\xi(x)].
\]
For a fixed $\pi$, the adversary chooses each perturbation sign
opposite to $\E_\pi T_e(y)-T_e(x)$. Its allowed magnitudes are
$\Gamma$ times a fractional matching. The inner value is therefore
$\E_\pi F_0(y)-F_0(x)-\Gamma\nu^*(a(\pi,x))$; at $\Gamma=0$
this follows directly. The second equality in
\eqref{eq:mixed-common-load} uses $T_e(x)\in\{-1,+1\}$ and
$\E_\pi T_e(y)\in[-1,1]$. Applying the covering dual in
\eqref{eq:weighted-fractional-matching} proves the remaining
formulation and the equivalence.
\end{proof}

The distribution in this certificate can depend on the competitor
$x$. For a fixed $x$, it is chosen before the perturbation in the
right side of \eqref{eq:mixed-common-identity}. A nonnegative
expected comparison ensures that, for each perturbation, at least
one assignment in its support matches or exceeds $F_\xi(x)$; that
assignment can change with the perturbation. The minimax step does
not exchange the outer minimization over competitors with this
choice of distribution.

For parity products, both formulations depend only on observed
products. Replacing a competitor by any whole-block gauge
representative leaves its value unchanged. Replacing every member
of $\mathcal D$ by a gauge representative likewise preserves the
value. Anchors can therefore reduce enumeration without changing
the property being tested. This gauge invariance follows from the
objective itself and does not require one predetermined reset to
improve every representative.

Restricting $\pi$ to the point mass at $Px$ recovers
\eqref{eq:robust-projection-identity}, because
$a_e=2\one\{e\in A_P(x)\}$. Allowing mixtures over
$\{P(gx):g\text{ reverses zero or two whole blocks}\}$ gives an
intermediate sufficient test involving at most four assignments.
Enlarging this candidate set to all of $\mathcal D$ gives the exact
test. Neither the four-candidate test nor any other proper
subcollection is asserted to be complete.

\subsection{Exact component formulations and rational witnesses}

Now take $\mathcal X$ to be the binary assignment space, possibly
with prescribed anchor coordinates. Suppose $\mathcal D$ fixes
the coordinates outside $U$, with every prescribed anchor outside
$U$ and fixed to its prescribed sign. Remove an anchor from $U$
before forming the components if necessary. Let
$C_1,\ldots,C_h$ be the components of its induced factor interaction
graph. Each factor meets at most one component. Write
\[
 F_\xi(y)=B_\xi+\sum_{i=1}^h A_i^\xi(s_i),
 \qquad y\in\mathcal D,
\]
using the component tables of
\eqref{eq:common-component-tables} in the parity case. The same
decomposition holds for arbitrary finite signed factors.

For a fixed competitor $x$, \eqref{eq:exact-competitor-gap} is
exactly the linear program
\begin{align}
 \min_{\xi,u,t}\quad& B_\xi-F_\xi(x)+\sum_i t_i,
       \label{eq:exact-common-component-primal}\\
 \text{subject to}\quad& t_i\ge A_i^\xi(s)
                      &&(i,\ s\in\{-1,+1\}^{C_i}),\nonumber\\
 &-u_e\le\xi_e\le u_e,\quad u_e\ge0,\nonumber\\
 &\sum_{e\ni v}u_e\le\Gamma &&(v\in V).
       \nonumber
\end{align}
Minimization sets each epigraph variable $t_i$ to its component
maximum. It does not require a common maximizing configuration for
different coefficient arrays.

The mixed formulation is equally small. Replace the full joint
distribution $\pi$ by a probability vector $p_i(s)$ on each
component, with $p_i(s)\ge0$ and $\sum_s p_i(s)=1$. For a factor
meeting $C_i$, use
\[
 \overline T_e=\sum_s p_i(s)T_e(s,r_{U^c});
\]
for a factor missing $U$, use $\overline T_e=T_e(r)$. Substitute
these expectations in
\eqref{eq:mixed-common-certificate-value}--\eqref{eq:mixed-common-certificate-cover}.
Every joint distribution supplies these marginals. Conversely, the
product of any collection of the component marginals is a genuine
distribution on $\mathcal D$ with the stated expectations. Thus
this reduction is exact: it introduces no inconsistent local
marginals.

If the factor arity is bounded and $|C_i|\le K$, these programs have
$O(m+|V|2^K)$ variables and constraints and
$O((m+|V|)2^K)$ explicitly represented coefficients. For fixed $K$,
each competitor can therefore be checked in time polynomial in the
rational input length. A feasible rational collection $(p_i,\lambda)$
with nonnegative value is a certificate for that competitor; its
nonnegativity, normalization, cover inequalities, and objective
comparison can all be checked exactly.

A strictly negative value instead has an explicit witness: an
allowed rational perturbation $\xi$ and an assignment $x$ for which
the component-table maximum is strictly below $F_\xi(x)$. This
proves that the proposed domain fails for that objective. Merely
checking selected competitors, including witnesses from a failed
reset test, does not certify the domain against all competitors.
Canonical exhaustive checking still has up to $2^{3n-2}$
competitors in the tripartite model.

\begin{proposition}[Worst-case complexity of common-domain validity]
\label{prop:common-validity-conp}
For rational coefficients and budgets, explicitly encoded signed
factor truth tables on binary variables of fixed arity, and fixed-coordinate domains
with component sizes bounded by a fixed $K$, common-domain
validity belongs to coNP. For arbitrary signed factors it is
coNP-complete already at $\Gamma=0$ with a singleton domain and
factor arity at most four.
\end{proposition}
\begin{proof}
If the domain is invalid, some fixed-competitor program in
\eqref{eq:exact-common-component-primal} has a negative optimum.
For fixed arity and $K$, its description has polynomial length in
the input. The coefficient perturbations and their absolute-value
variables range over a bounded polytope; each component epigraph
has a finite lower bound. An optimal vertex therefore exists.
After clearing denominators in its active constraints, Cramer's
rule gives a rational vertex of polynomial encoding length:
a $d$ by $d$ determinant with integer entries of at most $B$ bits
has $O(d(B+\log d))$ bits, and both $d$ and $B$ are polynomial
in the input length.

Supply its competitor $x$ and perturbation $\xi$. A verifier checks
the budget using rational arithmetic and computes the restricted
maximum by enumerating at most $2^K$ states per component. It then
checks that this maximum is strictly smaller than $F_\xi(x)$.
The verifier and the witness have polynomial size, proving coNP
membership.

For hardness, let $f_1,\ldots,f_M$ be the satisfaction indicators
of a 3-CNF formula with $M\ge1$, put $B=M-1$, and introduce a
binary selector $t$. Define
\[
 G(t,z)=t\sum_{j=1}^M f_j(z)+(1-t)B.
\]
Each term $t f_j$ has arity at most four and the remaining term is
unary. A fixed assignment with $t=0$ has value $B$; it is a global
optimizer exactly when the formula is unsatisfiable. To express
this as signed factors, use $2t f_j-1$ with coefficient $1/2$ and
$1-2t$ with coefficient $B/2$. Their sum is
$G-(M+B)/2$, so it has the same optimizers. All truth tables have
constant size and the coefficients have polynomial encoding
length. The singleton domain has no free components. At zero
budget its common validity is therefore equivalent to
unsatisfiability, proving coNP-hardness.
\end{proof}

This proposition concerns arbitrary factor truth tables and the
zero-budget worst case. It gives no hardness result for
positive-budget planted XOR targets. The polynomial program for a
fixed competitor and the universal quantifier over competitors
are distinct computational requirements.

\section{Finite certificates that cover all competitors}
\label{app:common-search}

The program in \eqref{eq:exact-common-component-primal} checks one
competitor. This section gives a finite certificate for the universal
comparison. It optimizes every restricted component and uses a tree of
linear-programming bounds to cover sets of competitors at once. The
construction uses standard mixed-integer linearization and branch and
bound; rational verification separates a proposed bound from a trusted
one \citep{mccormick1976,cheung2017verification}. Its worst-case cost is
exponential. Neither the formulation nor a finite successful search
asserts a polynomial-time certificate for the planted model.

\subsection{An exact mixed-integer formulation}

Let $H$ be a finite tripartite mask on $N=3n$ vertices, with rational
coefficients $Y_e$ and rational budget $\Gamma\ge0$. Fix a reference
$r\in\{-1,+1\}^N$ and free vertex set $U\subseteq V$ (where $V$ is the set of all $N$ vertices). The domain is
$\mathcal D=\{y:y_{U^c}=r_{U^c}\}$. Prescribe two anchors outside
$U$ in different blocks, with signs given by $r$. Each whole-block
gauge orbit has exactly one representative with these anchor signs.
All observed products are unchanged by this normalization, so it
suffices to check the canonical competitors.

Use reference coordinates
\[
 b_v=\frac{1-x_vr_v}{2},\qquad
 g_e=Y_eT_e(r),\qquad
 z_e=\frac{1-T_e(x)T_e(r)}{2}.
\]
Thus $b_v$ is binary, the two anchor bits are zero, and $z_e$ is the
parity of the three incident bits. Normalize a coefficient
perturbation by $\xi_eT_e(r)=\Gamma\delta_e$. Introduce variables
$a_e$ for its magnitude and impose
\begin{equation}
 -a_e\le\delta_e\le a_e,\quad 0\le a_e\le1,
 \qquad \sum_{e\ni v}a_e\le1\quad(v\in V).
 \label{eq:common-search-budget}
\end{equation}
At $\Gamma=0$ these variables have no effect on the objective and
represent the sole allowed perturbation $\xi=0$.

For an edge $e=\{i,j,k\}$, set $q=(b_i,b_j,b_k,z_e)$ and impose
$0\le q_\ell\le1$ and the eight inequalities
\begin{equation}
 \sum_{\ell\in S}q_\ell-\sum_{\ell\notin S}q_\ell
 \le |S|-1
 \quad(S\subseteq\{1,2,3,4\},\ |S|\text{ odd}).
 \label{eq:common-search-parity}
\end{equation}
Every binary tuple of even parity satisfies these inequalities.
If the first three entries are binary, they force the fourth entry
to equal their parity. To check the latter assertion, take $S$ to
be the support of the first three entries when that support is odd;
its inequality forces $z_e\ge1$. When that support is even, add
the fourth index to $S$; its inequality forces $z_e\le0$.

Introduce $w_e$ for the product $\delta_ez_e$ and impose the
McCormick inequalities
\begin{equation}
 -z_e\le w_e\le z_e,\qquad
 \delta_e+z_e-1\le w_e\le\delta_e-z_e+1.
 \label{eq:common-search-product}
\end{equation}
They hold for every genuine product with
$\delta_e\in[-1,1]$ and $z_e\in[0,1]$, and force the product
exactly when $z_e\in\{0,1\}$. In particular, only the competitor
bits $b_v$ need integrality constraints.

Let $C$ range over the induced components of $U$, and let $H_C$
be the factors meeting $C$. No factor meets two such components.
Put $H_U=\bigcup_C H_C$ and, for a binary component state $s$, set
\[
 p_e(s)=(-1)^{\sum_{v\in e\cap C}s_v},\qquad
 A_C^\delta(s)=\sum_{e\in H_C}(g_e+\Gamma\delta_e)p_e(s).
\]
Introduce an epigraph variable $t_C$ with
\begin{equation}
 t_C\ge A_C^\delta(s)\quad(s\in\{0,1\}^C),\qquad
 -M_C\le t_C\le M_C,
 \quad M_C=\sum_{e\in H_C}(|g_e|+\Gamma).
 \label{eq:common-search-component}
\end{equation}
The bounds are valid for every component maximum. They also make
all continuous variables bounded, which will permit exact checks
of approximate dual proposals.

\begin{theorem}[Exact global adversary]
\label{thm:common-search-exact}
Under \eqref{eq:common-search-budget}--\eqref{eq:common-search-component},
the two anchor constraints, and $b_v\in\{0,1\}$, the minimum of
\begin{equation}
 \mathcal L=
 \sum_C t_C-\sum_{e\in H_U}g_e
 -\Gamma\sum_{e\in H_U}\delta_e
 +2\sum_{e\in H}g_ez_e+2\Gamma\sum_{e\in H}w_e
 \label{eq:common-search-objective}
\end{equation}
equals $\min_x g_{\mathcal D}(x)$, where $x$ ranges over all
assignments. Hence its value is zero exactly when $\mathcal D$ is
common exact, and is strictly negative otherwise.

If every free component has size at most $K$, the formulation has
$O(N+m)$ variables,
$O(N+m+\sum_C2^{|C|})$ inequalities, and
$O((N+m)2^K)$ explicitly represented coefficients.
\end{theorem}
\begin{proof}
Fix a canonical binary competitor $b$. The parity inequalities
force its true $z_e$, and \eqref{eq:common-search-product} then
forces $w_e=\delta_ez_e$. The feasible perturbations in
\eqref{eq:common-search-budget} are exactly
$\mathcal U_\Gamma(H)$ after reference normalization; at positive
budget take $a_e=|\delta_e|$ for the reverse inclusion. At zero
budget the statement follows because all perturbation terms vanish.

For this competitor, minimization sets each $t_C$ to the maximum
of its component table. Its stated upper bound never obstructs
that choice. The restricted optimum is therefore
\[
 \max_{y\in\mathcal D}F_\xi(y)
 =\sum_{e\notin H_U}(g_e+\Gamma\delta_e)
   +\sum_C\max_s A_C^\delta(s).
\]
Subtracting
$F_\xi(x)=\sum_e(g_e+\Gamma\delta_e)(1-2z_e)$
gives \eqref{eq:common-search-objective}. Optimizing over the
continuous variables gives $g_{\mathcal D}(x)$, and then
optimizing the binary variables gives its minimum over canonical
competitors. Gauge invariance makes this the unrestricted minimum.
It is always at most zero: at $\xi=0$, choose an original global
optimizer as the competitor. The common-exactness characterization
following \eqref{eq:exact-competitor-gap} proves the sign assertion.

There are $N$ competitor bits, four variables per factor, and one
epigraph variable per component. Each edge has a fixed number of
parity, product, and magnitude inequalities. The only component
enumeration uses $2^{|C|}$ rows for a single component at a time;
it never forms the Cartesian product of all component tables.
Counting the nonzero coefficients proves the size bounds.
\end{proof}

\subsection{Exact verification of a tree of lower bounds}

At a search node, fix any subset of the nonanchor competitor bits
and relax all the remaining bits to $[0,1]$. Every binary
completion and every allowed perturbation embeds into this node
program. A lower bound on its objective therefore applies to
\emph{every} competitor in the corresponding subcube.

The following elementary form of weak duality avoids treating a
floating solver's status or objective as a certificate. All numbers
in the statement are rational.

\begin{lemma}[A dual bound with an exact residual correction]
\label{lem:common-search-dual}
For
\[
 \min\{c_0+c^\top u:Au\le h,\ \ell\le u\le v\},
\]
and any $y\le0$, put $\rho=c-A^\top y$. Then
\begin{equation}
 c_0+h^\top y+
 \sum_j\min\{\rho_j\ell_j,\rho_jv_j\}
 \label{eq:common-search-dual-bound}
\end{equation}
is a valid lower bound. It can be checked by rational arithmetic,
without exact dual feasibility beyond $y\le0$.
\end{lemma}
\begin{proof}
For feasible $u$, the inequality $Au\le h$ and $y\le0$ imply
$y^\top Au\ge y^\top h$. Also
$\rho^\top u\ge\sum_j\min\{\rho_j\ell_j,\rho_jv_j\}$.
Adding these inequalities to $c_0$ proves the claim.
\end{proof}

The coordinate bounds used in this lemma are $[0,1]$ for
unfixed $b,z,a$, $[-1,1]$ for $\delta,w$, fixed singleton
intervals for assigned bits, and $[-M_C,M_C]$ for $t_C$.
Thus a rounded rational dual vector remains a sound proposal:
its residual is charged exactly. A negative corrected bound is
inconclusive, even if the floating program reports a nonnegative
optimum.

\begin{theorem}[Complete finite verification]
\label{thm:common-search-tree}
A rooted binary tree is a certificate of common exactness if it
satisfies the following conditions. The root fixes the two anchor
bits. Each internal node selects an as-yet-unfixed competitor bit
and has both children, fixing that bit to zero and one. Every leaf
either supplies a nonnegative rational bound
\eqref{eq:common-search-dual-bound} for its node program, or fixes
every observed coordinate outside $U$ to zero.

Such a certificate is sound. Its verification requires time
polynomial in the explicitly represented component tables and the
encoded tree and dual vectors. Conversely, under the same outside-anchor assumption, every rational
common exact instance has a finite certificate of this form, with at most
$2^{d+1}-1$ nodes, where $d$ is the number of nonanchor vertices
incident to an observed factor. If the instance is not common
exact, there is instead a rational allowed perturbation and a
canonical competitor for which the componentwise restricted
maximum is strictly smaller than the competitor's value.
\end{theorem}
\begin{proof}
The branching rules partition the canonical competitor cube into
the leaf subcubes. A verified dual leaf proves
$g_{\mathcal D}(x)\ge0$ for each of its completions. At the
second type of leaf, every completion has the same observed
products as some assignment in $\mathcal D$: keep its free
coordinates and replace any unobserved outside coordinates by
their reference signs. Its gap is therefore nonnegative for every
allowed perturbation. Every canonical competitor is covered, so
the domain is common exact.

For verification, reconstruct the components, their tables, and
the rational linear constraints from the instance. Check that the
tree contains both children at every branch and never reassigns
a bit along a path. Check every declared bound using
\eqref{eq:common-search-dual-bound}, including the node's fixed
coordinate bounds. These operations have polynomial cost in
their encoded input and proof length. They do not use a solver
status as evidence of validity.

For completeness, branch on every nonanchor observed bit. At a
leaf, all edge parities are integral, so
Theorem~\ref{thm:common-search-exact} identifies its relaxation
with the exact fixed-competitor program. This is a feasible
rational linear program on a compact box. If common exactness
holds, its optimum is nonnegative. Linear-programming duality
supplies rational multipliers attaining its optimum in
\eqref{eq:common-search-dual-bound}; the bound variables' dual
contributions are equivalently the displayed coordinate minima.
The complete binary tree has at most $2^{d+1}-1$ nodes.

If common exactness fails, some canonical competitor has a
negative fixed-competitor optimum. The corresponding bounded
rational program has a rational optimum. Its perturbation and
competitor form the stated witness, checked by local budget sums
and independent maximization of each component table.
\end{proof}

Completeness concerns the proof system. A bounded search may
return unresolved, and a floating solver may fail to propose an
exact nonnegative leaf bound. Both outcomes remain unresolved.
The certificate cost includes every explored node and every
explicit component state. A short successful tree can cover
exponentially many competitors, but there is no asserted polynomial
upper bound on its size for the random model. The use of optimized component
completions removes the logical restriction of a fixed reset;
it does not by itself resolve the finite acquisition or uniform
certification frontier.

\subsection{The SOS obstruction also forces branch depth}
\label{app:common-search-depth}

The global degree barrier has a consequence for the finite verifier
above. Conditioning a pseudoexpectation follows one path in a
branch tree. Until enough degree has been spent, neither kind of
accepted leaf can occur. The resulting lower bound is on maximum
depth; it is not an exponential lower bound on tree size.

\begin{proposition}[Degree forces branch depth]
\label{prop:common-search-depth}
For the domain $\mathcal D$ and free set $U$ in Appendix~\ref{app:common-search}, suppose $Y_e\in\{-1,+1\}$ and
$R_0=\max_{y\in\mathcal D}F_Y(y)<m$. Assume there is an anchored
degree-$D_0$ pseudoexpectation satisfying all clause identities
\eqref{eq:sos-all-clauses}, where $D_0$ is even. Let $K$ be the
largest free component size, taking $K=0$ when $U$ is empty. If
$D_0\ge2\max\{2,K\}$, every certificate tree in
Theorem~\ref{thm:common-search-tree}, at any perturbation budget,
has maximum depth $L$ satisfying
\begin{equation}
 L\ge \frac{D_0}{2}-\max\{2,K\}+1.
 \label{eq:common-search-depth-bound}
\end{equation}
In particular, it has at least $2L+1$ stored nodes.
\end{proposition}
\begin{proof}
Use reference coordinates $X_v=x_vr_v$, so the original clauses
read $\prod_{v\in e}X_v=g_e$, where $g_e=Y_eT_e(r)$. At a
degree-$D$ node, the two bit indicators
$\ell_\pm=(1\pm X_v)/2$ have nonnegative pseudoexpectations
whose sum is one. Select a sign with positive pseudoexpectation
and condition by
\[
 \widetilde{\mathbb E}_{\pm}f
 =\frac{\widetilde{\mathbb E}(\ell_\pm f)}
        {\widetilde{\mathbb E}\ell_\pm}.
\]
This gives a degree-$(D-2)$ pseudoexpectation with that bit fixed.
Indeed, $\ell_\pm^2=\ell_\pm$ makes positivity of a square
follow from positivity of $(\ell_\pm f)^2$ at degree $D$.
The same identity imposes all truncated equalities for the fixed
bit. Multiplying a new clause multiplier by $\ell_\pm$ stays
within the old degree bound, so all original clauses persist.
This is the one-bit version of the conditioning argument in
Lemma~\ref{lem:sos-gauge-anchors}.

Suppose, for a contradiction, that
$D_0-2L\ge D_*:=\max\{4,2K\}$. Follow a positive-probability
child at every split. At its eventual leaf, the conditional
pseudoexpectation has degree at least $D_*$ and satisfies both
the node's fixed bits and every original clause.

First consider a dual-bound leaf. A degree-four pseudoexpectation
with these clause identities induces a genuine distribution on
each triple, supported on its required parity. To see this, write
the triple variables as $X_1,X_2,X_3$, its sign as $g$, and
$m_i=\widetilde{\mathbb E}X_i$. The clause identity multiplied
by $X_3$ gives
$\widetilde{\mathbb E}X_1X_2=g m_3$. For each state $s$ with
$s_1s_2s_3=g$, put
\[
 \lambda_s
 =\widetilde{\mathbb E}
       \frac{(1+s_1X_1)(1+s_2X_2)}4
 =\frac{1+s_1m_1+s_2m_2+s_3m_3}{4}.
\]
The pair indicator has degree two and equals its own square on
the Boolean cube; degree-four positivity gives $\lambda_s\ge0$.
These four numbers sum to one and have coordinate means $m_i$.
They therefore make
\[
 b_v=(1-\widetilde{\mathbb E}X_v)/2,
 \qquad z_e=(1-g_e)/2
\]
feasible for every parity inequality
\eqref{eq:common-search-parity}. These means satisfy every
prescribed bit; an unfixed unobserved bit may instead be set to
zero. Set
$\delta=a=w=0$ and choose each $t_C$ to be its exact unperturbed
component maximum. All other node constraints hold, including
the prescribed bits. By \eqref{eq:common-search-objective}, this
feasible point has value $R_0-m<0$. A verified nonnegative lower
bound for the node is impossible.

At an identity leaf, every observed coordinate outside $U$ has
been fixed to its reference sign. The conditional functional has
degree at least $2K$. For each component and state, its indicator
\[
 p_{C,s}(X)=\prod_{v\in C}\frac{1+s_vX_v}{2}
\]
has nonnegative pseudoexpectation, since its square has degree at
most $2K$ and it is idempotent. These expectations form a genuine
probability distribution on the component. Substituting the
fixed outside signs into each factor is allowed by the truncated
coordinate equalities, as the factor degree is at most three.
Thus the pseudoexpectation of every component objective is at
most its table maximum, and
$\widetilde{\mathbb E}F_Y\le R_0<m$. The original clause
identities instead give $\widetilde{\mathbb E}F_Y=m$.
This also excludes an identity leaf.

Both alternatives contradict the certificate's leaf rule. Hence
$D_0-2L<D_*$, which is \eqref{eq:common-search-depth-bound}
because both degrees are even. A binary tree with a path of
length $L$ has at least the $L+1$ path vertices and $L$ distinct
sibling vertices, proving the node count.
\end{proof}

On the event of Theorem~\ref{thm:sos-certification}, its
constructed anchored pseudoexpectation has degree
$D_0=d_n^{\mathrm{SOS}}$, while $R_0\le\max_xF_Y(x)<m$.
For any fixed component cap $K$, every accepted tree of this
form therefore has depth
$\Omega_c(n/(\log^2 n\log\log n))$. This statement is uniform
over the reference, free domain, anchor choice, and perturbation
budget whenever the standing assumptions hold. It concerns the
specified parity relaxation and leaf rules. It does not bound
the depth of a different proof system, imply exponential tree
size, or supply an unrestricted running-time lower bound.

\subsection{An exponential size bound for the specified verifier}
\label{app:common-search-size}

The particular parity relaxation permits a stronger conclusion than
the degree-to-depth argument alone. At zero perturbation, a node
that closes by its LP bound also closes by ordinary propagation of
the original parity equations. We can therefore convert the
certificate tree into a tree-like resolution refutation, to which
the size--width theorem of \citet{bensasson2001short} applies. The
conclusion is specific to the inequalities and leaf rules in
Theorem~\ref{thm:common-search-tree}; it does not cover arbitrary
cuts, additional valid inequalities, or general mixed-integer proof
systems.

\begin{lemma}[Parity feasibility after propagation]
\label{lem:common-search-propagation}
Fix any competitor bits, and fix every edge bit to
$z_e=(1-g_e)/2$, where $g_e\in\{-1,+1\}$. The parity
constraints \eqref{eq:common-search-parity} are feasible if and
only if the following propagation does not find a contradiction:
whenever two bits of a triple are fixed, fix the third to satisfy
its required parity; declare a contradiction if a fully fixed
triple has the wrong parity.
\end{lemma}
\begin{proof}
Each forced bit is required by the parity inequalities when the
other two bits and the edge parity are integral. Hence a propagated
contradiction makes the linear constraints infeasible. Conversely,
suppose propagation terminates without contradiction. Every edge
then has zero, one, or three fixed vertex bits, and every fully
fixed edge has the correct parity. Set every remaining bit to
$1/2$. With no fixed bit, use the uniform distribution on the
four parity-consistent triple states; with one fixed bit, use the
uniform distribution on its two consistent completions. Their
coordinate means are the proposed bits and they satisfy every
parity inequality. Fully fixed edges already satisfy the required
parity. The common coordinate means are consequently feasible
for the whole linear system.
\end{proof}

\begin{lemma}[Resolution width after two gauge anchors]
\label{lem:common-search-resolution-width}
Assume the expansion hypothesis \eqref{eq:sos-vertex-expansion}
for an integer $S\ge1$ and $\gamma>0$. Suppose the parity
equations $\chi_e=Y_e$ are inconsistent. Encode each equation by
its four width-three clauses, and substitute arbitrary prescribed
signs at two vertices in distinct blocks. Every resolution
refutation of the resulting CNF formula has width strictly larger
than $\gamma S-2$.
\end{lemma}
\begin{proof}
For a clause $C$ on the remaining variables, let $\nu(C)$ be the
minimum number of original parity equations that, together with
the two anchor assignments, imply $C$. An input clause has
$\nu(C)\le1$. Resolution is sound, so $\nu$ is subadditive
under a resolution step; weakening cannot increase it.

Every set of at most $S$ original equations is satisfiable for
every choice of right sides. Indeed, the expansion hypothesis
gives a degree-one vertex in each of its nonempty edge subsets,
so repeated removal of a degree-one vertex and its edge, followed
by reverse assignment, solves the equations. A whole-block gauge
transformation then gives either prescribed pair of anchor signs
without changing any equation. Thus the empty clause has
$\nu(\bot)>S$.

Let $F$ be a minimum representing edge set for a clause $C$.
Every degree-one vertex of $F$ outside the anchors must occur
in $C$. Otherwise, remove its sole edge. By minimality, the
remaining equations and anchors have an assignment falsifying
$C$; changing that vertex alone satisfies the removed equation
without changing any other equation, anchor, or literal of $C$.
This contradicts the implication from $F$. When $|F|\le S$,
\eqref{eq:sos-boundary-expansion} therefore gives
\[
 |\operatorname{vars}(C)|\ge|\partial_1F|-2
                         \ge2\gamma|F|-2.
\]
In a refutation, consider the first line with $\nu>S$. It is a
resolution step whose predecessors have $\nu\le S$, and one
predecessor has $\nu>S/2$. Its width is strictly larger than
$\gamma S-2$, as claimed.
\end{proof}

\begin{theorem}[Size of a parity-LP certificate tree]
\label{thm:common-search-size}
Let $Y_e\in\{-1,+1\}$ and assume
$\max_xF_Y(x)<m$ and \eqref{eq:sos-vertex-expansion}. Let
$\mathcal D$ have two outside anchors in distinct blocks and
free components of size at most $K$, with $K=0$ when there are
no free coordinates. For any perturbation budget, every accepted
certificate tree of Theorem~\ref{thm:common-search-tree}, with
$T$ stored nodes, satisfies
\begin{equation}
 T>\frac{2^{\gamma S-5}}{2N+2^{K+1}},\qquad N=3n.
 \label{eq:common-search-size-bound}
\end{equation}
Consequently, on the event of
Theorem~\ref{thm:sos-certification}, every such certificate with
fixed $K$ has
\[
 T=\exp\!\left(\Omega_c\!\left(
           \frac{n}{\log^2 n\log\log n}\right)\right).
\]
The statement is uniform over the reference, proposed domain,
anchor choice, and budget under these assumptions.
\end{theorem}
\begin{proof}
Write $R_0=\max_{y\in\mathcal D}F_Y(y)<m$ and use reference
coordinates, with $g_e=Y_eT_e(r)$. At a dual-bound leaf, suppose
the partial assignment admits a feasible parity point with
$z_e=(1-g_e)/2$. Extend it by
$\delta=a=w=0$ and by the exact base component maxima $t_C$.
This is feasible for the full node relaxation and has objective
$R_0-m<0$, contradicting the leaf's verified nonnegative bound.
Thus Lemma~\ref{lem:common-search-propagation} gives a
propagation contradiction from the node's fixed bits.

Replace this leaf by an ordinary binary query tree. For every
propagated bit, the branch opposing its forced value immediately
falsifies a clause of the original width-three parity encoding;
the other branch continues propagation. At most $N$ new bits
are fixed. The resulting subtree has at most $2N+1$ vertices
and ends with an original clause falsified at every leaf.

At an identity leaf, all observed outside bits equal their
reference values. The restricted parity system is inconsistent,
because $R_0<m$. Either a factor missing $U$ already has the
wrong parity, or the equations of at least one free component
are inconsistent. In the latter case, query the at most $K$
unfixed bits of such a component, since all outside endpoints of its factors were fixed on the branch to the identity leaf. Each complete assignment
falsifies an original clause, giving a subtree of at most
$2^{K+1}-1$ vertices. Additional bits fixed along the path do
not change this conclusion.

Replacing every leaf in this way produces a binary query
refutation of the anchored CNF, of size at most
$(2N+2^{K+1})T$. Such a query refutation yields a tree-like
resolution refutation with no more vertices. To verify this
directly, label each leaf by its falsified initial clause and
work upward. If both child clauses contain the queried variable,
their literals have opposite signs, and resolve to a clause
falsified by the parent's partial assignment. If one child clause
omits the variable, retain that child's derivation. At the root
the resulting clause is empty. No derived clause is reused.

The anchored input formula has initial width at most three.
Lemma~\ref{lem:common-search-resolution-width} gives refutation
width greater than $\gamma S-2$. The tree-like size--width bound
\citep[Theorem~3.3 and Corollary~3.4]{bensasson2001short} states
that minimum refutation size is at least
$2^{w_{\rm ref}-w_{\rm input}}$. It is therefore strictly
larger than $2^{\gamma S-5}$. Comparing this with the converted
tree size proves \eqref{eq:common-search-size-bound}.

For the random mask, the same expansion event gives
$\gamma S=\Theta_c(n/(\log^2 n\log\log n))$, and the true
optimum is below $m$ with probability $1-o(1)$. For fixed $K$,
the denominator in \eqref{eq:common-search-size-bound} is
$O_K(n)$, whose logarithm is negligible relative to $\gamma S$.
The mask event, the true optimum bound, and gauge invariance
hold before the domain or reference is chosen. This proves the
uniform asymptotic statement.
\end{proof}

Thus the finite formulation is complete and can produce useful
small-instance proofs, but this particular tree verifier has
exponentially large proofs on the asymptotic instances where
acquired information makes optimization efficient. A method using
different relaxations, additional justified cuts, or a different
proof system is outside this lower bound. The conclusion is about
verification by the specified representation, not the difficulty
of optimizing the target or the impossibility of every efficient
common-domain certificate.

\section{A degree barrier for global SOS certificates}
\label{app:sos-certification}

This section proves Theorem~\ref{thm:sos-certification} directly for
the uniform size-$m$ tripartite mask. The underlying construction is
due to \citet{schoenebeck2008linear}, building on algebraic degree
lower bounds for parity \citep{grigoriev2001parity}: the absence of a narrow
XOR-resolution refutation gives a positive semidefinite character
moment matrix. The random-CSP bounds of \citet{kothari2017sum} give
the same density dependence in a more general setting. Recent work
of \citet{chan2026strongly} treats refutation tradeoffs without
uniformly random literals. We include
the argument to establish its precise scope here: sampling without
replacement, arbitrary target signs, two possibly data-dependent
gauge anchors, and the entire local perturbation family.

\subsection{The certificate class and its degree convention}

Write $V=A\sqcup B\sqcup C$, with $|A|=|B|=|C|=n$, and identify an
observed triple $e$ with its three vertices. For $S\subseteq V$, put
$\chi_S(x)=\prod_{v\in S}x_v$ and reduce polynomials modulo
$x_v^2=1$. A degree-$D$ pseudoexpectation, for even $D$, is a linear
functional on polynomials of degree at most $D$ satisfying
\[
 \widetilde{\mathbb E}1=1,
 \qquad
 \widetilde{\mathbb E}p^2\ge0
       \quad (\deg p\le D/2),
\]
with the Boolean identities imposed at all degrees where they are
defined. Its moment matrix is indexed by monomials of degree at most
$D/2$. Thus $D$ is the total polynomial degree, not the moment-matrix
level. The standard Boolean SOS upper bound is
\[
 \operatorname{SOS}_D(F)=\sup_{\widetilde{\mathbb E}}
                             \widetilde{\mathbb E}F.
\]
When anchors are prescribed, the supremum also imposes
$\widetilde{\mathbb E}[h(x)(x_a-t_A)]=0$ and
$\widetilde{\mathbb E}[h(x)(x_b-t_B)]=0$ for every multiplier $h$ with $\deg h\le D-1$, for $a,b$ in distinct
blocks and
$t_A,t_B\in\{-1,+1\}$.

For $D\ge4$, every feasible pseudoexpectation has
$|\widetilde{\mathbb E}\chi_e|\le1$. For example, if
$e=\{a_i,b_j,c_k\}$, the two polynomials $a_i$ and $b_jc_k$ have
squared norm one, so their moment-matrix Cauchy--Schwarz inequality
gives the claim. Equivalently, the trivial upper bound $m$ has the
degree-four SOS certificate
\begin{equation}
 m-F_Y(x)=\frac12\sum_{e=(i,j,k)\in H}
                  (a_i-Y_e b_jc_k)^2
                  \pmod{\{x_v^2-1:v\in V\}}.
 \label{eq:sos-trivial-upper}
\end{equation}
The quotient identity has a representation with all Boolean-ideal
terms of degree at most four. Consequently a feasible
pseudoexpectation with $\widetilde{\mathbb E}F_Y=m$ proves both that
the relaxation value equals $m$ and that no degree-$D$ SOS identity
can certify a smaller global upper bound.

\subsection{Expansion gives a pseudoexpectation for every sign array}

For $F\subseteq H$, let $V(F)$ be its incident vertices and let
$\partial_1F$ be the vertices incident to exactly one edge of $F$.
Let $\partial_{\rm odd}F$ be the vertices of odd degree in $F$.

\begin{lemma}[Finite expansion criterion]
\label{lem:sos-finite-expansion}
Let $S\ge1$ be an integer and $\gamma>0$. Suppose every nonempty
$F\subseteq H$ with $|F|\le S$ satisfies
\begin{equation}
 |V(F)|\ge(3/2+\gamma)|F|.
 \label{eq:sos-vertex-expansion}
\end{equation}
For every choice of signs $(Y_e)_{e\in H}$ and every even integer
$D$ with $4\le D\le\gamma S$, there is a degree-$D$
pseudoexpectation satisfying
\begin{equation}
 \widetilde{\mathbb E}
       [h(x)(\chi_e(x)-Y_e)]=0
 \quad(e\in H,\ \deg h\le D-3).
 \label{eq:sos-all-clauses}
\end{equation}
In particular, $\operatorname{SOS}_D(F_Y)=m$.
\end{lemma}
\begin{proof}
The degree sum gives
\begin{equation}
 |\partial_{\rm odd}F|\ge|\partial_1F|
      \ge2|V(F)|-3|F|\ge2\gamma|F|.
 \label{eq:sos-boundary-expansion}
\end{equation}
Choose an integer $w$ with $D\le w\le\gamma S$. A width-$w$
XOR derivation starts with the equations $\chi_e=Y_e$ and
$\chi_\varnothing=1$. It may multiply two previously derived
equations and cancel repeated variables, provided that the resulting
support has size at most $w$. Every line has support of size at most
$w$; here the initial supports have size three.

For a derived equation $C$, let $\nu(C)$ be the smallest number of
original equations whose product is $C$, with repeated equations
cancelled. This quantity is finite and subadditive under a derivation
step. If a contradiction $\chi_\varnothing=-1$ were derived, its
minimum representing set would have more than $S$ edges: otherwise
\eqref{eq:sos-boundary-expansion} would give a nonempty support.
Consider the first line in the derivation with $\nu>S$. Its two
predecessors have $\nu\le S$, and one of them has $\nu>S/2$ by
subadditivity. A minimum representing edge set for this predecessor
has odd boundary equal to its support. Equation
\eqref{eq:sos-boundary-expansion} makes its width strictly larger
than $\gamma S\ge w$, a contradiction. Thus the width-$w$ closure
contains no contradiction. In particular, it never derives two
opposite signs on the same support.

For $|R|\le D$, define
\[
 \psi(R)=
 \begin{cases}
  \sigma,&\text{if the closure derives $\chi_R=\sigma$,}\\
  0,&\text{if it derives neither sign.}
 \end{cases}
\]
Set $\widetilde{\mathbb E}\chi_R=\psi(R)$ and extend linearly.
Normalization follows from $\psi(\varnothing)=1$. To prove
positivity, index the moment matrix by $|I|,|J|\le D/2$ and write
\[
 M_{I,J}=\psi(I\mathbin\triangle J).
\]
Declare $I\sim J$ when the closure derives an equation on
$I\mathbin\triangle J$. This is an equivalence relation. For
transitivity, multiply the equations for $I\mathbin\triangle J$
and $J\mathbin\triangle K$; their product has support
$I\mathbin\triangle K$, of size at most $D\le w$, so it belongs
to the closure. The same operation multiplies the corresponding
signs. In each equivalence class, choose a representative index set $R$ of size at most $D/2$ from that class and
put $s_I=\psi(I\mathbin\triangle R)$. The matrix block for this
class is $(s_I s_J)_{I,J}$, while different classes have zero cross
entries. Each block is a rank-one positive semidefinite matrix.
This proves $M\succeq0$ and hence positivity on squares.

Finally, if $|R|\le D-3$, the closure derives an equation on $R$
if and only if it derives one on $R\mathbin\triangle e$: multiply
by the original equation for $e$ in either direction. Both supports
have size at most $D\le w$. Their signs differ by the factor $Y_e$,
and therefore
\[
 \psi(R\mathbin\triangle e)=Y_e\psi(R).
\]
Linearity proves \eqref{eq:sos-all-clauses}. Taking $h=1$ gives
$\widetilde{\mathbb E}F_Y=m$; combine this with
\eqref{eq:sos-trivial-upper}.
\end{proof}

The positive semidefiniteness step is essential. Merely excluding
small even covers, or checking consistency of small collections of
parities, would not establish a feasible SOS moment matrix.

\subsection{The fixed-size tripartite mask}

\begin{lemma}[Expansion at logarithmic density]
\label{lem:sos-tripartite-expansion}
Let $H$ be uniform among size-$m$ subsets of $[n]^3$, with
$n\le m\le n^3$, and define
\begin{equation}
 \begin{split}
  \Delta&=m/n,\qquad
  \gamma=\frac{1}{8\log(\Delta+e)},\qquad
  \beta=\frac12-\gamma,\\
  S&=\left\lfloor n(256\Delta)^{-1/\beta}\right\rfloor.
 \end{split}
 \label{eq:sos-finite-parameters}
\end{equation}
The probability that \eqref{eq:sos-vertex-expansion} fails for some
$1\le|F|\le S$ is at most
\begin{equation}
 \sum_{s=1}^{S}
    \left[64\Delta(s/n)^\beta\right]^s.
 \label{eq:sos-expansion-failure}
\end{equation}
If $m=\lceil cn\log n\rceil$ for fixed $c>0$, this probability is
$o(1)$, $S=\Theta_c(n/\log^2 n)$, and
\[
 \gamma S=\Theta_c\!\left(
                  \frac{n}{\log^2n\,\log\log n}\right).
\]
\end{lemma}
\begin{proof}
Put $a=3/2+\gamma<2$. If a set of $s$ edges violates
\eqref{eq:sos-vertex-expansion}, all its vertices lie in a set of
$L=\lceil as\rceil-1\le as$ vertices. A vertex set of size $L$
supports at most $(L/3)^3$ tripartite triples, by the arithmetic--
geometric mean inequality. For any fixed set of $s$ distinct triples,
\[
 \Pr(F\subseteq H)=\frac{(m)_s}{(n^3)_s}
                       \le(m/n^3)^s.
\]
The count is zero if the vertex set supports fewer than $s$ triples.
Otherwise the union bound and the usual binomial estimate give
\begin{align*}
 &\Pr\{\exists F:\ |F|=s,\ |V(F)|<as\}\\
 &\quad\le
   \binom{3n}{L}\left(\frac{eL^3}{27s}\right)^s
          \left(\frac{m}{n^3}\right)^s\\
 &\quad\le
   \left[\left(\frac{3e}{a}\right)^a
          \frac{ea^3}{27}\,
          \frac{m}{n}(s/n)^{2-a}\right]^s\\
 &\quad\le[64\Delta(s/n)^\beta]^s.
\end{align*}
For the second inequality, use the monotonicity of
$(3en/L)^L$ for $L\le3n$ and $L\le as\le2n$. The constant is
uniform: for $3/2\le a<2$,
$(3e/a)^a(ea^3/27)\le32e^3/27<64$. Summing over $s$ proves
\eqref{eq:sos-expansion-failure}.

At every $s\le S$, the bracket in
\eqref{eq:sos-expansion-failure} is at most $1/4$. For every fixed
$s$, it tends to zero when $\Delta=\Theta_c(\log n)$, since
$\beta\to1/2$. Dominated convergence, with summable bound $4^{-s}$,
makes the whole sum $o(1)$. Finally,
\[
 \frac{1}{\beta}=2+O(1/\log\Delta),
 \qquad
 (256\Delta)^{1/\beta}=\Theta(\Delta^2),
\]
with absolute asymptotic constants. This proves the claimed scales
without a constant that degenerates as $\gamma\to0$.
\end{proof}

\subsection{Two gauge anchors cost four degrees}

The tripartite gauge group has four elements. It multiplies every
coordinate in each block by $g_A,g_B,g_C\in\{-1,+1\}$ with
$g_Ag_Bg_C=1$. Every observed product, and therefore every objective
$F_\xi$, is invariant. Every assignment has a gauge representative
with any prescribed signs at two vertices in distinct blocks.
Consequently these anchors do not change the true optimum.

\begin{lemma}[Anchored pseudoexpectations]
\label{lem:sos-gauge-anchors}
Suppose a degree-$D$ pseudoexpectation satisfies
\eqref{eq:sos-all-clauses}, where $D\ge8$ is even. For any two
vertices in distinct blocks and any prescribed anchor signs, there
is a degree-$(D-4)$ pseudoexpectation satisfying the same clauses
and the anchor equalities. It still has objective value $m$.
\end{lemma}
\begin{proof}
Average the functional over the four gauge transformations. The
result, denoted $\overline{\mathbb E}$, remains normalized and
positive on squares, and still satisfies all the original clause
identities through degree $D$. Relabeling the blocks if needed,
let the anchors be $a\in A$, $b\in B$, with signs $t_A,t_B$.
Gauge averaging gives
\[
 \overline{\mathbb E}x_a=
 \overline{\mathbb E}x_b=
 \overline{\mathbb E}(x_a x_b)=0.
\]
The polynomial
\[
 p(x)=\frac{(1+t_A x_a)(1+t_B x_b)}4
\]
has degree two, satisfies $p^2=p$ on the Boolean cube, and has
$\overline{\mathbb E}p=1/4$. For $\deg f\le D-4$, set
\[
 \widehat{\mathbb E}f=4\overline{\mathbb E}(pf).
\]
It is normalized. If $\deg f\le(D-4)/2$, then
$\deg(pf)\le D/2$ and
\[
 \widehat{\mathbb E}f^2
   =4\overline{\mathbb E}(pf^2)
   =4\overline{\mathbb E}[(pf)^2]\ge0.
\]
The middle equality is within degree $D$: its two polynomial
representatives have degrees at most $D-2$ and $D$, respectively,
and agree modulo the Boolean identities.

For an original clause and a multiplier $h$ of degree at most
$(D-4)-3=D-7$, the multiplier $ph$ has degree at most $D-5$.
The original clause identity therefore gives
\[
 \widehat{\mathbb E}[h(\chi_e-Y_e)]
       =4\overline{\mathbb E}[ph(\chi_e-Y_e)]=0.
\]
For an anchor multiplier of degree at most $(D-4)-1=D-5$, the
polynomial $ph(x_a-t_A)$ has degree at most $D-2$ and is identically
zero modulo the Boolean equations; the same holds for $b$. Hence
all anchor equality tests at degree $D-4$ vanish. This checks every
truncated constraint, not only the first anchor moments.
Taking the clause multiplier $h=1$ retains objective value $m$.
\end{proof}

The argument is deterministic for each pair of anchors and signs.
It therefore covers data-dependent choices without a probability
union bound. It also explains why revealing an acquired reference
does not remove this obstruction: the coefficients of an SOS
certificate may depend arbitrarily on that reference, but every
valid global certificate is still tested by the same feasible
pseudoexpectation. Treating all reference coordinates as hard
equalities would instead change the optimization domain.

\subsection{The true value and uniform coefficient perturbations}

\begin{lemma}[The empirical value at logarithmic density]
\label{lem:sos-true-value}
For the planted signs with fixed $q\in(0,1/2)$ and
$\mu=1-2q$, conditional on any mask $H$ and source tensor,
\begin{equation}
 \begin{split}
  \mu m-\sqrt{2mu}\ &\le\max_x F_Y(x)\\
                   &\le\mu m+
                      \sqrt{2m(3n\log2+u)}
 \end{split}
 \label{eq:sos-true-value-bound}
\end{equation}
with probability at least $1-2e^{-u}$, for every $u>0$.
In particular, for $m=\lceil cn\log n\rceil$,
$\max_xF_Y(x)=\mu m+O_c(\sqrt{mn})=(\mu+o(1))m$
with probability at least $1-2e^{-n}$.
\end{lemma}
\begin{proof}
For every fixed assignment, the $m$ signed summands are independent,
belong to $\{-1,+1\}$, and have sum of expectations at most
$\mu m$. Hoeffding's inequality and a union bound over at most
$2^{3n}$ factor assignments give
\[
 \Pr\{\max_xF_Y(x)>\mu m+t\mid H,T\}
       \le2^{3n}\exp(-t^2/(2m)).
\]
The source assignment has expected score $\mu m$, and its lower
tail has the same bound without the factor $2^{3n}$. Choosing the
two values of $t$ in \eqref{eq:sos-true-value-bound} proves the
claim. Set $u=n$ for the last statement.
\end{proof}

\begin{theorem}[A global SOS certificate barrier]
\label{thm:sos-certification}
Fix $c>0$, $q\in(0,1/2)$ and any nonnegative sequence
$\Gamma_n=o(\log n)$. There is an even sequence
\[
 d_n^{\mathrm{SOS}}=\Theta_c\!\left(\frac{n}{\log^2 n\,\log\log n}\right)
\]
such that, with probability $1-o(1)$, every even $4\le D\le d_n^{\mathrm{SOS}}$
satisfies
\begin{equation}
 \operatorname{SOS}_D(F_Y)=m,
 \qquad \max_x F_Y(x)=(1-2q+o(1))m.
 \label{eq:sos-certification-gap}
\end{equation}
Moreover, uniformly over $\xi\in\mathcal U_{\Gamma_n}(H)$,
\[
 \operatorname{SOS}_D(F_\xi)-\max_x F_\xi(x)
   =(2q+o(1))m.
\]
\end{theorem}
\begin{proof}[Proof of Theorem~\ref{thm:sos-certification}]
Use the parameters in \eqref{eq:sos-finite-parameters} and, for
sufficiently large $n$, set
\begin{equation}
 d_n^{\mathrm{SOS}}=2\left\lfloor\frac{\gamma S}{2}\right\rfloor-4.
 \label{eq:sos-certificate-degree}
\end{equation}
On the expansion event, Lemma~\ref{lem:sos-finite-expansion} gives
a degree-$(d_n^{\mathrm{SOS}}+4)$ pseudoexpectation for every sign array.
Lemma~\ref{lem:sos-gauge-anchors} permits any optional choice of two
anchors at degree $d_n^{\mathrm{SOS}}$. Restriction to every even
$4\le D\le d_n^{\mathrm{SOS}}$ is still feasible. The trivial upper bound then
gives $\operatorname{SOS}_D(F_Y)=m$, simultaneously for all these
degrees and anchor choices. Lemma~\ref{lem:sos-true-value} supplies
the true optimum on an event of probability $1-o(1)$.

For the uniform perturbation statement, the local budget gives
\[
 3\|\xi\|_1
   =\sum_{v\in V}\sum_{e\ni v}|\xi_e|
   \le3n\Gamma_n.
\]
Both the true optimum and the SOS optimum are
$\|\xi\|_1$-Lipschitz in the coefficient array. For the former,
every $|\chi_e(x)|=1$; for the latter, every feasible
pseudoexpectation has $|\widetilde{\mathbb E}\chi_e|\le1$ when
$D\ge4$. These statements also hold with either choice of gauge
anchors. Therefore
\begin{align*}
 \sup_{\xi\in\mathcal U_{\Gamma_n}(H)}
  \big|\operatorname{SOS}_D(F_\xi)-\max_xF_\xi(x)-2qm\big|
  \le O_c(\sqrt{mn})+2n\Gamma_n=o(m).
\end{align*}
All choices may depend on the realized data. No independence from
the acquired reference is required.
\end{proof}

\paragraph{Interpretation and limits.}
The lower bound applies to global SOS upper bounds on the original
Boolean factor domain, optionally normalized by two gauge anchors.
It also excludes certificates of common-domain exactness obtained
by this relaxation together with the exact restricted optimum: at
the base objective, the two values already differ by
$(2q+o(1))m$. It does not exclude certificates in other proof systems,
certificates that use additional instance-specific valid inequalities, or direct verification of the probabilistic model's
structural conditions. Nor is it a target-only running-time lower
bound for unrestricted algorithms. The acquired optimization
theorem and this certificate barrier hold simultaneously, making
their separation a statement about the same realized objectives.

The full moment matrix at the displayed degree has
$\sum_{j=0}^{d_n^{\mathrm{SOS}}/2}\binom{3n}{j}$ rows. Its row count is
$\exp(\Theta_c(n/\log^2n))$, since
$d_n^{\mathrm{SOS}}=\Theta_c(n/(\log^2n\log\log n))$. This is the size of the
standard dense representation at that degree, not a lower bound on
the size of an arbitrary certificate or on arbitrary verification
algorithms.

\paragraph{Finite verification.}
The supplied checker constructs an unsatisfiable tripartite system
with $n=7$ and $m=14$. Every variable has degree two, and the product
of the observed signs is negative, so at least one parity must be
violated. The all-positive assignment violates exactly one and has
the exact score $12$. Its width-four XOR closure has no contradiction.
The resulting degree-four moment matrix has $232$ rows and an exact
integer Gram factorization, satisfies all $308$ admitted clause
multiplier tests, and gives pseudoexpectation score $14$. Separate
exact rational checks verify the two-anchor conditioning identities
for all $48$ choices of distinct-block anchors and signs on a
six-variable gauge orbit. These are finite checks of the algebraic
construction; the asymptotic mask event is established by
Lemma~\ref{lem:sos-tripartite-expansion}.

\section{Global parity consistency and syndrome relaxations}
\label{app:syndrome-barrier}
The branch lower bound applies to a specified parity-LP tree, not to
all verification representations. Gaussian elimination over the two-element field $\mathbb F_2$
recognizes inconsistency of the unrelaxed parity system in polynomial
time, without branching. It does not solve the minimum-error problem.
This distinction motivates a test of a stronger, natural relaxation
that imposes every redundant syndrome check separately.

Let $B\in\mathbb F_2^{m\times 3n}$ be the incidence matrix ($B_{ev}=1$ when $v\in e$, zero otherwise), $b_e=(1-Y_e)/2$, and $x$ now denote binary factor bits (a sign $(-1)^{x_v}$).
The error vector is $u=b+Bx$ over $\mathbb F_2$.
For every $a\in\ker B^\top$ it satisfies
$a^\top u=a^\top b$ in $\mathbb F_2$. Conversely those checks imply
$u-b\in(\ker B^\top)^\perp=\operatorname{im} B$.
A basis of the kernel therefore suffices for the exact binary
formulation, and gauge fixing at two anchors in distinct blocks
leaves its feasible error vectors unchanged. The weighted objective
is $\sum_e(1+Y_e\xi_e)(1-2u_e)$ in ordinary real arithmetic.

Following the separate parity-polytope construction of LP decoding
\citep{feldman2005lp}, define the all-check relaxation by $0\le u\le1$ and, for every nonzero
$a\in\ker B^\top$, require the restriction $u_{\operatorname{supp}a}$
to belong to the convex hull of the binary vectors on that support
having parity $a^\top b$. Let $L_{\rm syn}(F_\xi)$ be its maximum
weighted score. This includes the full parity polytope of every
redundant check, not only checks from a chosen basis.

\begin{theorem}[A gap for all separate syndrome checks]
\label{thm:syndrome-barrier}
If every nonzero vector in $\ker B^\top$ has Hamming weight at least
$g\ge2$, the all-check relaxation contains $u_e=1/g$ for every $e$.
In the uniform size-$m$ tripartite model with $m=\lceil cn\log n\rceil$,
fixed $c>0$ and $q\in(0,1/2)$, with probability $1-o(1)$,
\begin{equation}
 L_{\rm syn}(F_0)\ge m-O_c(\log^3 n),\qquad
 L_{\rm syn}(F_\xi)-\max_x F_\xi(x)=(2q+o(1))m
 \label{eq:syndrome-gap}
\end{equation}
uniformly for $\xi\in\mathcal U_{\Gamma_n}(H)$ with
$\Gamma_n=o(\log n)$. The bounds also hold with arbitrary two-block
anchors. They concern the intersection of the separate check
polytopes, not the convex hull of all simultaneous binary solutions.
\end{theorem}
\begin{proof}
For a check of length $\ell\ge g$, the all-$1/g$ vector has total
weight $\ell/g$ in $[1,\ell/2]$. The minimum allowed integer
weight of a given parity is either zero or one, and its maximum
is either $\ell$ or $\ell-1$. Since $\ell\ge2$, the interval between
these extrema contains $\ell/g$. The uniform distribution on binary
vectors of either extremal allowed weight has that weight divided
by $\ell$ as every coordinate mean. Mixing the two distributions
attains $1/g$ in each coordinate. (If the extrema coincide, use the
single distribution.) This proves membership in every check polytope
simultaneously; different checks need not use the same distribution.

A nonzero vector of $\ker B^\top$ is a set of observed edges with
even degree at every vertex. On the event of
Lemma~\ref{lem:sos-tripartite-expansion}, no nonempty set of at most
$S=\Theta_c(n/\log^2 n)$ edges can have this property: its number of
vertices would be at most $3|F|/2$, contrary to the expansion bound.
Take $g=S+1$. The uniform vector is feasible and has score
$m-2m/g$. The trivial box upper bound is $m$ at $\xi=0$, and
Lemma~\ref{lem:sos-true-value} gives the actual optimum
$(1-2q+o(1))m$. Both the relaxed and the exact score maxima change
by at most $\|\xi\|_1$ under coefficient edits. The sum of the
local budgets gives $\|\xi\|_1\le n\Gamma_n=o(m)$, proving the
uniform conclusion. Gauge invariance follows from the exact syndrome
formulation above.
\end{proof}

Even an exponential list of redundant checks can thus leave the
leading certification gap unchanged. Gaussian elimination identifies
that zero errors are impossible, while the relaxation still permits
an error mass of order at most $\log^3 n$. A certificate that uses
relations among several checks jointly, or uses other inequalities,
is outside this result. Related lower bounds for branch-and-bound
with variable disjunctions and with general disjunctions are developed
by \citet{glaser2024branch,dey2023branch}; the tree theorem here uses
only its explicitly specified parity LP and leaf rules.

\paragraph{A finite check.}
The complete $4\times4\times4$ grid has shortest even cover four:
parity of the total incidence count rules out odd cardinalities,
distinctness rules out two, and a rectangle in two blocks with a fixed
third coordinate gives four. A frozen seeded set of 64 sign arrays
is checked by enumeration of all 4,096 factor assignments per array.
One array has exact minimum error 22 (score 20), whereas the all-check
relaxation admits error mass $64/4=16$ (score 32). The supplied script
retains all 64 exact values and the witness labels. It verifies the
rational convex mixtures for each check length from 4 through 64 and
each parity (122 tests); the proof of Theorem~\ref{thm:syndrome-barrier}
then applies to every redundant check, without enumerating an
exponential list of checks. This example illustrates the relaxation
rather than its asymptotic onset under the planted model.

\section{Acquisition by reducing the order of the observations}
\label{app:one-fiber}
The support-growth schedule reads three fibers to obtain usable triples
with two known endpoints. A different use of the same measurement access
reduces the order of selected observations before estimating any of the
other coordinates. Only one fiber is read. Multiplying a triple's label
by its reading produces a noisy product of two latent signs. This gives
an ordinary signed bipartite graph on the two unread blocks.

\begin{theorem}[One-fiber acquisition]
\label{thm:one-fiber}
Fix the retained-source model of Section~\ref{sec:model}, with $m=\lceil cn\log n\rceil$,
$\mu=1-2q$, $\rho=1-2s$ and declared rational lower bounds $\mu_*>0$,
$\rho_*>0$ on these biases. Put $P=\lfloor m/2\rfloor$,
$K=\lceil32/(\mu_*\rho_*)^2\rceil$, and
\[
 k=\min\{n,\lceil Kn^2/P\rceil\}\quad(P>0),
\]
with $k=n$ if $P=0$. There is a randomized polynomial-time algorithm
using exactly $k$ readings, all in the first reference fiber, that
returns a feasible assignment on every input and an exact empirical
optimum with probability $1-o(1)$. Thus $k\sim 2K n/(c\log n)$.
For every fixed valid component bound $J>1/\alpha$, $J\ge2$, and deterministic local budget $\Gamma_n=o(\log n)$, the observable restriction in Theorem~\ref{thm:common-reduction} then preserves every optimum up to gauge, with probability $1-o(1)$. The matrix approximation step has polynomial cost;
the near-linear bound in Theorem~\ref{thm:geometric} is for the cascade.
\end{theorem}

\begin{proof}
Randomly order the target independently of its labels. Reserve the first
$P$ locations for projection and split the remainder into three deterministic
batches of sizes differing by at most one. Read the first $k$ entries of
$A$. For each projected location $(i,j,\ell)$ with $i\le k$, let
$R_i=A_i\epsilon_i$ and retain the pair $(j,\ell)$ with sign $R_iY_{ij\ell}$.
The matrix $W$ sums these signs at their pair addresses. In the analysis
write $L$ for the number retained. To avoid using any false independence
of labels sharing a reading, condition on all $\epsilon_i$ and the source.

First draw the first $P$ grid locations independently with replacement,
with a fresh target error at each draw. The chance that any address repeats
is at most $P(P-1)/(2n^3)=o(1)$. Conditional on no repetitions the ordered
sample and its errors have exactly the required law. In the iid experiment,
$L\sim\operatorname{Bin}(P,k/n)$. Conditional on $L$ and the readings, the
projected locations are iid uniform pairs, and their signs in source
coordinates are iid with mean $\delta=\mu k^{-1}\sum_{i\le k}\epsilon_i$.
Indeed the selected first index is uniform and independent of the pair;
its error is fixed by the conditioning. This remains true when a pair
appears more than once. The original model inherits any $1-o(1)$ event
in this experiment by the collision bound.

As $n\to\infty$, $k<n$, $k\to\infty$, and with probability $1-o(1)$,
\[
 L\ge .9Pk/n\ge .9Kn,\qquad \delta\ge .9\mu\rho.
\]
The two inequalities follow from binomial concentration and Hoeffding's
inequality for the independent reading errors. For signs $b,d$, define
$u=n^{-1}\sum_j b_j B_j$, $v=n^{-1}\sum_\ell d_\ell C_\ell$. The
projected score $S(b,d)=b^\top Wd$ is a sum of $L$ independent signs
of mean $\delta uv$ conditional on the above history and $L$ (before
revealing the projected pairs and labels). In particular,
$S(B,C)\ge .9\delta L$ except with probability at most
$\exp(-L\delta^2/200)$. For any fixed $b,d$ with $uv\le1/16$,
\[
 \Pr\{S(b,d)\ge .45\delta L\}
 \le\exp\{-L\delta^2(31/80)^2/2\}.
\]
Union over the $2^{2n}$ sign pairs. By the choice of $K$, the exponent
is at least $32(.9)^3(31/80)^2n/2>2n\log2$. Thus with probability
$1-o(1)$, every half-approximation to the maximum projected score has
$uv>1/16$ and hence $|u|,|v|>1/16$ with equal signs.

The bilinear sign maximum admits a randomized polynomial-time
approximation with expected ratio greater than $.56$ by the rounding
of Alon and Naor~\citep{alon2006cutnorm}, Theorem 5.3. Use a fixed
inverse-polynomial semidefinite programming (SDP) accuracy, absorbing an additive error smaller
than $.01$ when the integer optimum is positive, to obtain expected
ratio at least $.55$. In one trial the score is at most the optimum,
so the probability of reaching half the optimum is at least $.10$.
Taking the best of $\lceil100\log(n+2)\rceil$ independent trials fails
with probability $o(1)$. If the optimum is zero, every assignment is
optimal for the projected matrix. This uses polynomial work and only
observed projected scores to select the result.

After the projection, a simultaneous reversal of both estimated blocks
is a gauge transformation. For analysis choose that gauge so both
correlations exceed $1/16$. This choice depends only on the source and
the history before the fresh batches. In the first fresh batch, estimate
$A$ by the majority of $Y_{ij\ell}b_jd_\ell$ at each $i$, with
independent fair ties. For any fixed opposite estimates with positive
product correlation, the mean of a vote on the remaining opposite-pair
grid is at least $\mu(1/256-O(m/n^2))$. Each vertex has a hypergeometric
degree of mean $\Theta(\log n)$. The one-block proof of
Lemma~\ref{lem:cascade-majority} applies: it uses only the opposite-block
means and the fresh history. It gives first-block mean $1-o(1)$ in
probability. Thus after this batch all three means exceed a positive
constant with probability $1-o(1)$. In the second fresh batch apply
one simultaneous majority to all three blocks. The same lemma and
Proposition~\ref{prop:finite-majority-bound} give means at least
$1-\varepsilon_n$, for a deterministic $\varepsilon_n\to0$, with
probability $1-o(1)$. The analysis gauge is kept fixed; independent fair
ties have the same law after changing it.

Use the last fresh batch for the final refinement. On the pre-final
calibration event just obtained, Lemma~\ref{lem:direct-mixed-moment}
applies in the aligned gauge, with final fraction tending to $1/6$.
That lemma requires no independent calibration errors; it retains the
prescribed-location factors. Exclude failure of the calibration event
once, and apply the same root-star and incidence-tree witnesses as in
the proof of Theorem~\ref{thm:geometric}. In particular all vertices have
boundedly many erroneous neighbors before full-target reuse, and the
reused error set lies in a fixed-cutoff intrinsic weak set. The capped
repair around the reused estimate is exact with probability $1-o(1)$.
The argument for the uniform common restriction in
Appendix~\ref{app:common-reduction} uses precisely this localization and
therefore applies on the same event. Gauge changes preserve every
objective and every restricted-domain score. All batch scans and the
capped repair have polynomial all-input cost. For small inputs with
$P=0$ or empty projection a fixed feasible candidate suffices; the
asymptotic argument never relies on these cases.
\end{proof}

The preceding guarantee does not use growth of the set of known
coordinates. Its sufficient reading coefficient depends quadratically
on the inverse product of channel biases, rather than the fourth
power in the three-fiber cascade. The explicit concentration bound
still requires a large projected sample; it is not a finite onset
claim. The following distinction is specific to the projected record.

\begin{proposition}[Information lost under replacement]
\label{prop:projection-replacement}
Use the same one-fiber addresses, projection batch, and $k$ as in
Theorem~\ref{thm:one-fiber}. Keep only the number of projected pairs,
their addresses in the two unread blocks, and their projected signs;
remove their first-block indices. Under independent replacement, no
estimator based on this record and the replacement readings has a
fixed positive absolute correlation with an unread source block with
probability bounded away from zero. The claim holds even if the
estimator is also given the true first reference fiber.
\end{proposition}
\begin{proof}
Use the same iid-location coupling, whose failure is $o(1)$.
Let $R$ be the independent replacement readings. Conditional on the
first source fiber $A$ and on $R$, put $\epsilon_i=R_i A_i$. The remaining
reference-fiber entries of $B,C$ are independent fair signs except for
the shared center, determined by $A$. The signs of the compressed
record, conditional also on the unread source blocks and the pair
addresses, are independent with bias
$\delta=\mu k^{-1}\sum_{i\le k}\epsilon_i$ relative to $B_jC_\ell$.
In the reference experiment keep the source, readings, $L$, and pair
addresses under exactly the same law, but use independent fair
projected signs. The relative entropy of the joint laws is at most
$\E[L\delta^2]$, because a sign with bias $d$ has relative entropy to
a fair sign at most its chi-square divergence $d^2$.
The vector $A$ has independent fair entries, and is independent of
$R$, so $\E[(\sum_i R_i A_i)^2]=k$. Also $L$ is independent of these
variables and $\E L=Pk/n$. Hence the relative entropy is at most
$\mu^2P/n=O(\log n)$. Under the reference experiment, even given $A$,
any estimated unread block has correlation at least a fixed
$\gamma>0$ in absolute value with probability at most
$2\exp(-\Omega_\gamma(n))$ by Hoeffding's inequality on its $n-1$
independent signs. The event version of relative-entropy monotonicity
bounds its probability under the actual iid experiment by
$(O(\log n)+\log2)/\Omega_\gamma(n)=o(1)$.
Returning to the fixed-size mask adds at most the collision probability.
The result concerns the compressed record. It does not apply to an
algorithm that subsequently inspects the full target.
\end{proof}

\subsection{A reduced-reading finite comparison}
\label{app:one-fiber-experiments}
We examine a simple spectral implementation using the complete target for
projection and refinement, rather than the fresh batches and SDP of the
proof. The retrospective population consists of all 370 earlier targets
at $n=8,10,12$, with $q=.17$, $s=.10$, and the five recorded densities.
The seed, three reading budgets $k=\lceil n/4\rceil,\lceil n/2\rceil,n$,
four initializations, tie rules, twelve-sweep block-ascent limit, and
component-repair cap were fixed before execution. For each arm and
budget, multiply every target label in the first $k$ slices by the
corresponding reading, sum at its pair address, and form four pairs of
sign vectors from the first four singular-vector pairs of the resulting
matrix. At zero entries use common pre-generated fair signs. Complete
the first block by majority on the full target, run at most twelve
block-coordinate sweeps for each candidate, keep the largest empirical
score, and apply the existing capped component repair once. The cold
control uses four independent sign arrays, the same ascent cap and one
repair, but no projection or first-block completion. The retained and
replacement arms use the same reading count, addresses, SVD, candidate
count, and repair rules. Actual work can vary with the stopping tests.
We recompute all 370 reference optima by independent integer enumeration
of two blocks with exact maximization over the third. All target hashes,
source scores, and prior exact optimum values agree. Every feasible
answer is rescored directly on the original observations.

\begin{table}[h]
\centering\small
\caption{Exact empirical optima in the frozen one-fiber comparison.
The 124 source-suboptimal targets at $1/2<\alpha<1$ are a declared
subset of the complete 370-target population, not additional trials.
The cold control is the same run for every budget.}
\label{tab:one-fiber}
\begin{tabular}{lrrrrrr}
\toprule
&\multicolumn{3}{c}{All targets (370)}&\multicolumn{3}{c}{Suboptimal middle regime (124)}\\
Readings & Retained & Replacement & Cold & Retained & Replacement & Cold\\
\midrule
$\lceil n/4\rceil$&135&83&78&28&16&22\\
$\lceil n/2\rceil$&157&84&78&37&20&22\\
$n$&213&94&78&50&18&22\\
\bottomrule
\end{tabular}
\end{table}

Across all targets the retained-only versus replacement-only success
counts are $76:24$, $109:36$, and $148:29$, in budget order. At the largest
budget, the mean score deficits are $2.238$, $7.059$, and $8.422$ for
retention, replacement, and cold starts. In the source-suboptimal middle
regime the retained-only versus replacement-only counts are $20:8$,
$27:10$, and $41:9$. The older 32-start target-only
baseline finds 201 of 370 exact optima, including 61 of 124 in the
source-suboptimal middle regime. Thus the one-fiber procedure shows a
clear paired finite advantage over matched independent readings, while
the smaller budgets remain below the 32-start target-only baseline. That baseline uses a different work allocation and does not establish an equal-cost comparison.
These small-instance results do not establish a finite onset of the
asymptotic $O(n/\log n)$ guarantee, and the full-fiber and cascade
comparisons in earlier appendices remain distinct. The complete paired
records, witnesses, exact replay, and work counts are in
\texttt{results/one\_fiber/} and can be regenerated by
\texttt{code/one\_fiber\_projection.py}.

\section{Beyond parity: noisy planted MAX-2-SAT}
\label{app:max2sat}
The residual geometry need not originate in a parity predicate. This section gives a second random objective, with an asymmetric local observation law and no gauge symmetry. Its source is accessible through literal frequencies, so the acquisition cost differs from that of tripartite parity. We distinguish recovery of the generating assignment from exact empirical optimization, including densities at which the two disagree with high probability. The planted-clause framework of \citet{feldman2018planted} concerns recovery of a planted assignment; here the planted assignment may violate clauses and the task is to maximize the realized (possibly reweighted) objective.

\paragraph{Model.} Fix $0<q<1/4$ and $c>0$, and let $a=(1-q)/3$. On $n$ Boolean variables with a fixed unknown source $T$, include each unordered pair independently with probability $b_n=c\log n/n$ (for sufficiently large $n$). Each included pair has one clause, described by its unique forbidden assignment. In source-relative coordinates, its forbidden pattern is $00$ with probability $q$ and each of $01,10,11$ with probability $a$, independently. Put $F(x)=\sum_e f_e(x_e)$ with $f_e$ the satisfaction indicator, and
\[
 F_\xi(x)=\sum_e(1+\xi_e)f_e(x_e),\qquad
 \mathcal U_\Gamma=\{\xi:\sum_{e\ni v}|\xi_e|\le\Gamma\text{ for every }v\}.
\]
Weights are real, including negative weights; $\Gamma\ge0$ is fixed. In the algorithmic statements, arithmetic and comparisons on supplied weights are exact (rational bit complexity is polynomial in the input length). Even the unweighted simple-pair objective class is NP-hard: encode MAX-CUT \citep{garey1976} by the two clauses $(x_u\vee x_v)$ and $(\neg x_u\vee\neg x_v)$ for each graph edge. Their total satisfaction count is one plus the cut indicator. Replace each clause $(l_1\vee l_2)$ by $(l_1\vee z),(\neg z\vee l_2)$ with a separate new variable $z$. Maximization over $z$ gives one plus the original clause satisfaction, and all resulting unordered pairs are distinct. This is a worst-case reduction, not a hardness claim for the distribution above.

Write $D_v=F(T)-F(T^{\{v\}})$, where $T^S$ flips $S$. Define
\[
 h_2=(\sqrt a-\sqrt q)^2,\quad \alpha_2=ch_2,\quad
 \lambda_2=\tfrac12\log(a/q),\quad
 A_2=(4\pi c\sqrt{aq})^{-1/2},\quad r_2=\frac{n^{1-\alpha_2}}{\sqrt{\log n}}.
\]
For this model, $d_{\rm fix}(\Gamma)$ and $N_{\rm hit}(\Gamma)$ are defined as in the main text, over Boolean assignments with no gauge quotient. Let $G=\max_x F(x)-F(T)$ and $\mathcal O=\arg\max F$.

\begin{theorem}[Residual geometry for noisy clauses]\label{thm:max2sat}
For every fixed $c>0$, $0<q<1/4$ and fixed budget $\Gamma\ge0$, a target-only algorithm, using no additional source readings, either reports a component-cap failure or returns a reference and a fixed-coordinate domain with bounded-size induced components. With probability $1-o(1)$ it returns such a domain containing every optimizer of every $F_\xi$, $\xi\in\mathcal U_\Gamma$. On this event, component enumeration gives exact optimization in $O(n+|E|)$ arithmetic operations for each query, with constants depending on the fixed parameters. A component-size cap ensures the work bound on every input. For $0<\alpha_2<1$,
\begin{align}
 \frac{G}{r_2}&\xrightarrow{\Pr} A_2\frac{e^{-\lambda_2}}{(1-e^{-\lambda_2})^2},\qquad
 \frac{\log_2|\mathcal O|}{r_2}\xrightarrow{\Pr} A_2,\label{eq:sat-gain}\\
 \frac{d_{\rm fix}(\Gamma)}{r_2},\quad \frac{\log_2 N_{\rm hit}(\Gamma)}{r_2}&\xrightarrow{\Pr} A_2\sum_{j\in\mathbb Z:\,|j|<\Gamma} e^{\lambda_2j}.\label{eq:sat-dim}
\end{align}
If $\alpha_2\ge1$, the source is the unique optimizer simultaneously for every admissible perturbation with probability $1-o(1)$. The zero-budget minimum dimension is zero at every density.
\end{theorem}

\subsection{Signed margins and rare vertices}
For a fixed vertex, each of the $n-1$ possible pairs contributes to $D_v$ a $+1$ with probability $b_na$, a $-1$ with probability $b_nq$, and zero otherwise. Under the tilt with density proportional to $e^{-\lambda_2D_v}$, the positive and negative probabilities both become $u_n=b_n\sqrt{aq}/(1-b_nh_2)$. The normalizer is $(1-b_n h_2)^{n-1}=(1+o(1))n^{-\alpha_2}$. Under the tilt the characteristic function is $[1-2u_n(1-\cos t)]^{n-1}$. For large $n$ its base is positive, and on $[-\pi,\pi]$ it is bounded in magnitude by $\exp(-C(n-1)u_nt^2)$ for an absolute $C>0$. Fourier inversion, changing variables by $\sqrt{(n-1)u_n}$ and dominated convergence, gives for every fixed integer $j$
\begin{equation}\label{eq:sat-local-limit}
 \Pr(D_v=j)\sim A_2 e^{\lambda_2j}n^{-\alpha_2}/\sqrt{\log n}.
\end{equation}
The same domination gives the uniform bound $\Pr(D_v=j)\le C_{c,q}e^{\lambda_2j}n^{-\alpha_2}/\sqrt{\log n}$ for all integers $j$. Removing any fixed number of possible pairs preserves the asymptotic and bound. Summing the geometric tail gives $\Pr(D_v\le L)=O(n^{-\alpha_2}/\sqrt{\log n})$ for fixed $L$.

For distinct $v,w$, condition on their shared pair (absent or present with a specified clause); their other incident pairs are independent. For any fixed $j$, the joint probability of their two fields equalling $j$ is $(1+o(1))$ times the product of the marginal probabilities: the absent case has probability $1-b_n$ and the present cases contribute at most $O(b_n n^{-2\alpha_2}/\log n)$. Thus the variance of the field count is $O(r_2)+o(r_2^2)$. For $\alpha_2<1$ this proves joint convergence of the normalized counts to $A_2e^{\lambda_2j}$ on every finite set of $j$. The uniform tail bound gives $r_2^{-1}\E\sum_v(-D_v)\one_{D_v<-M}\le C\sum_{j>M}j e^{-\lambda_2j}$, justifying truncation.

For $W_L=\{v:D_v\le L\}$, a fixed number $t$ of specified vertices, after conditioning on a fixed number of prescribed pair states, have probability at most $n^{-t\alpha_2+o(1)}$ of all lying in $W_L$. Indeed, condition also on all pairs among the specified vertices; each field loses only a bounded number of contributions, and the remaining incident edge variables are independent across vertices. The bound is uniform over the conditioned states, and also holds if a bounded number of auxiliary vertices are involved in the prescribed pairs. Counting rooted trees on $t$ vertices gives an expected number of connected $t$-vertex weak sets at most $n(C\log n)^{t-1}n^{-t\alpha_2+o(1)}$ for fixed $t$. Consequently all weak components have fewer than any fixed integer $J>1/\alpha_2$ with high probability. Counting $k$-stars in the same way gives at most $n(C\log n)^k n^{-k\alpha_2+o(1)}$ stars with $k$ weak leaves, so every vertex has fewer than $k$ weak neighbors for fixed $k>1/\alpha_2$. Counting paths of length one and two with weak endpoints gives $n^{1-2\alpha_2+o(1)}$ such paths in expectation. This is $o(r_2)$ for $0<\alpha_2<1$, and is $o(1)$ for $\alpha_2>1/2$. In particular, all but $o_{\Pr}(r_2)$ weak vertices are separated from every other weak vertex by distance greater than two. For $\alpha_2\ge1$, the one-vertex tail already makes every fixed-threshold weak set empty with high probability.

\subsection{Uniform localization}
Let $J(S)=F(T)-F(T^S)$. For a specified $S$ of size $t$, each of the $M_t=t(n-t)+\binom{t}{2}$ potential pairs meeting $S$ contributes $+1,-1,0$ with the same probabilities as a field increment: the two assignments at that pair are distinct. If $S$ induces a connected component of the flip set of an optimizer of $F_\xi$, undoing $S$ cannot improve its objective, because no pair connects this component to another flipped vertex. Thus $J(S)\le\Gamma t$.

For $t\ge2$, require the $t-1$ edges of a prescribed spanning tree of $S$ to be present. Conditioning on their states costs at most a factor $e^{\lambda_2(t-1)}$ in the tilted moment, and the remaining potential pairs are independent. Summing over labeled trees and vertex sets bounds the probability of a connected improving set of size $t$ by
\[
 \binom nt t^{t-2} b_n^{t-1}e^{\lambda_2(\Gamma t+t-1)}(1-b_nh_2)^{M_t-t+1}
 \le n(C\log n)^t n^{-(\alpha_2/2-o(1))t}.
\]
Here $M_t\ge t(n-1)/2$, and the $o(1)$ in the exponent is uniform for $2\le t\le n$. For a fixed integer $J_0\ge2$ with $J_0>4/\alpha_2$ the sum for $t\ge J_0$ tends to zero. This event is uniform in $\xi$ because the necessary condition involves only $J(S)$ and the budget. In a remaining component of size $t<J_0$, undoing an individual flip at $v$ cannot improve the objective. The cost of flipping $v$ at the source differs from its cost in the presence of the other flips by at most $2(t-1)$, and reweighting changes this cost by at most $\Gamma$. Therefore every optimizer of every admissible objective differs from the source only in $W_{L_0}$ for $L_0=2(J_0-2)+\Gamma$. This argument covers all optimizers, including ties. No recovery assumption has entered the localization proof.

\subsection{An observable domain without readings}
Initialize $z_v$ as the value occurring less often as the forbidden literal at vertex $v$, breaking ties arbitrarily. In source coordinates the forbidden literal equals the source with probability $q+a$, and differs with probability $2a>q+a$. The wrong-sign event is a nonpositive sum of independent increments. After removing any bounded number of prescribed incident pairs its probability is at most $n^{-\beta+o(1)}$, with $\beta=c(1-2\sqrt{(q+a)2a})>0$, by the exponential-moment bound. The same conditional-star argument above shows that every vertex has fewer than a fixed $k_0>1/\beta$ erroneous neighbors with probability $1-o(1)$. This does not require independent initialization errors.

Compute one simultaneous best response $r$ from $z$ for the original objective. Because a wrong neighbor changes the difference of the two local scores by at most two, $r_v=T_v$ whenever $D_v>2k_0$. Hence $r$ and every perturbed optimizer agree with the source outside $W_L$ for $L=\max\{2k_0,L_0\}$. Choose fixed integers $J\ge2$ with $J>1/\alpha_2$, and $k>1/\alpha_2$. On the weak-component and neighbor events for $W_L$, every component of the difference between $r$ and a perturbed optimizer has size less than $J$. Let $d_v(r)=F(r)-F(r^{\{v\}})$. The individual best-response comparison at that optimizer now yields $d_v(r)\le B=2(J-2)+\Gamma$ for every differing vertex. The observable restriction
\[
 U=\{v:d_v(r)\le B\},\qquad x_{U^c}=r_{U^c}
\]
contains all optimizers for all perturbations. For $v\notin W_L$, the reference is correct and $|d_v(r)-D_v|\le2k$ because fewer than $k$ neighbors are in $W_L$. Thus $U\subseteq W_{\max\{L,B+2k\}}$, whose components have fewer than $J$ vertices with high probability. The algorithm checks this component cap before enumerating. A pair touches at most one free component; collecting its table costs a constant number of operations per pair. All construction and enumeration costs are $O(n+|E|)$ for the fixed parameters. On cap failure the algorithm may return unresolved. Even if the cap succeeds, the probabilistic domain guarantee is not an unconditional certificate. If $\alpha_2\ge1$, the containing weak set is empty on the same event, so every perturbed optimum equals $T$.

\subsection{Sharp improvement, multiplicity, and common dimension}
For any $S\subseteq W_L$ (increase $L$ to a fixed nonnegative cutoff containing the localized optimizers), $J(S)$ differs from $\sum_{v\in S}D_v$ only on pairs with both endpoints in $S$. Each discrepancy has absolute value at most one by the four forbidden patterns. The count of internal weak pairs is $o_{\Pr}(r_2)$. Comparing the all-negative flip set with an optimal flip set proves $G=\sum_v(-D_v)_++o_{\Pr}(r_2)$. Equation~\eqref{eq:sat-local-limit} and tail truncation give the first limit in~\eqref{eq:sat-gain}. Isolated vertices of the induced weak graph contribute independent binary choices precisely at field zero; the total number of vertices in nonsingleton weak components is $o_{\Pr}(r_2)$. The objective separates across components and there is no gauge quotient. This bounds $\log_2|\mathcal O|$ between the isolated zero-field count and that count plus the nonsingleton vertex count, proving the second limit.

For common dimension, fix all strong coordinates at the source and retain the interacting weak components. At each singleton of the weak graph, the unperturbed preference has signed field $D_v$. Its preferred value remains an optimizer for all budgets $\Gamma\le|D_v|$, including equality, because perturbations change this field by at most $\Gamma$. Retain the other singletons. This gives a common domain with dimension at most $\#\{v:|D_v|<\Gamma\}+o_{\Pr}(r_2)$, and the domain's entire cube is a hitting family. At $\Gamma=0$, an arbitrary original optimum alone suffices.

For fixed $\Gamma>0$, discard from the reversible-field count those vertices within distance two of another weak vertex and those with no incident pair contributing a nonzero increment to their field. The first loss is $o_{\Pr}(r_2)$; the expected second loss is $n(1-b_n(a+q))^{n-1}=n^{1-c(a+q)+o(1)}=o(r_2)$ since $a+q>h_2$. Every remaining vertex is a weak singleton with at least one incident active pair. Choose one such pair per vertex; these pairs form a matching. For each vertex, any desired preferred sign can be enforced strictly by perturbing just its selected weight (or leaving a strictly preferred sign unchanged): a field of size $|D_v|<\Gamma$ can be reversed with a weight change of magnitude in $(|D_v|,\Gamma]$, and zero can be biased either way. The matching ensures simultaneous budget feasibility. Uniform localization keeps all strong coordinates fixed, so every optimizer of each constructed objective must take the prescribed singleton values. For $R$ selected vertices, the $2^R$ choices therefore give pairwise disjoint optimizer sets. Both $d_{\rm fix}(\Gamma)$ and $\log_2N_{\rm hit}(\Gamma)$ are at least $R$. Applying~\eqref{eq:sat-local-limit} to the finite set of fields proves~\eqref{eq:sat-dim}.

The role of the source information is different in this model: the rare-margin geometry persists, but the literal-frequency initialization already identifies a reference close enough for exact local repair. The result establishes transfer of the residual laws, not a necessity of added measurements for every NP-hard generated problem.

\begin{corollary}[Isolated margins and the critical window]\label{cor:sat-critical}
For fixed $1/2<\alpha_2<1$, with probability $1-o(1)$,
\[
 G=\sum_v(-D_v)_+,\qquad \log_2|\mathcal O|=\#\{v:D_v=0\},
\]
and for every fixed $\Gamma>0$,
\[
 d_{\rm fix}(\Gamma)=\log_2N_{\rm hit}(\Gamma)
 =\#\{v:|D_v|<\Gamma,\ \text{some incident clause contributes nonzero to }D_v\}.
\]
Now replace the fixed density by
\[
 b_n=\frac{\log n-\tfrac12\log\log n+t+o(1)}{nh_2},\qquad t\in\mathbb R\text{ fixed},
\]
and set $A_{2,*}=(4\pi\sqrt{aq}/h_2)^{-1/2}$, $\Lambda=A_{2,*}e^{-t}$. Let $(Z_j)_{j\in\mathbb Z}$ be independent Poisson variables with means $\Lambda e^{\lambda_2 j}$. For any finite list of fixed budgets, the following convergence holds jointly:
\[
 \left(G,\log_2|\mathcal O|,d_{\rm fix}(\Gamma),\log_2N_{\rm hit}(\Gamma)\right)
 \xrightarrow{\mathrm d}
 \left(\sum_{j\ge1}j Z_{-j},Z_0,\sum_{|j|<\Gamma}Z_j,\sum_{|j|<\Gamma}Z_j\right).
\]
In particular, the source is optimal with limiting probability $\exp[-\Lambda e^{-\lambda_2}/(1-e^{-\lambda_2})]$ and uniquely optimal with limiting probability $\exp[-\Lambda/(1-e^{-\lambda_2})]$.
\end{corollary}
\begin{proof}
For fixed $\alpha_2>1/2$, take a weak threshold large enough for the given budget and localization. The path bound makes all weak vertices separated by distance greater than two with high probability. Thus all choices are independent singletons, and the gain and multiplicity identities are exact. If no incident clause contributes a nonzero field at a singleton, its two values remain tied under every admissible perturbation since all its neighbors are strong. Such a vertex may be frozen without loss of the optimum value. All remaining reversible fields can be biased strictly in either direction by matching weight perturbations, exactly as in the proof of Theorem~\ref{thm:max2sat}. The upper and lower dimension bounds now coincide exactly.

In the varying-density window, the constants in the localization and neighbor bounds can be chosen uniformly using any fixed positive lower bound on $c_n=nb_n/\log n$ for large $n$. The tilted local limit has the same proof, with $(1-b_nh_2)^{n-1}\sim e^{-t}\sqrt{\log n}/n$ and $(n-1)u_n\sim(\sqrt{aq}/h_2)\log n$. Thus $n\Pr(D_v=j)\to\Lambda e^{\lambda_2j}$, with the uniform tail bound $n\Pr(D_v=j)\le C e^{\lambda_2j}$. For any fixed number $s$ of distinct vertices and prescribed fields $j_1,\ldots,j_s$, condition on all $\binom{s}{2}$ internal pair states. When all are absent, the remaining fields are independent and their marginals have the same limits. The total probability of any internal pair being present is $O(b_n)$; conditional on such states the joint event is $O(n^{-s})$ by the uniform local bound with bounded shifts. Therefore the joint probability is asymptotic to $n^{-s}\prod_{i=1}^s \Lambda e^{\lambda_2 j_i}$. This gives all mixed factorial moments of any finite list of field counts. The factorial-moment characterization of independent Poisson variables proves their joint convergence; the limiting distribution has finite exponential moments and is moment determinate.

The distance-two and localization estimates again give the isolated-singleton event with probability $1-o(1)$. The expected number of vertices without a nonzero incident contribution is $n^{1-(a+q)/h_2+o(1)}=o(1)$, so the additional exclusion in the dimension formula disappears. For a finite list of budgets use their maximum in the localization event. The tail of the weighted negative-field sum has expectation bounded by $C\sum_{j>M}j e^{-\lambda_2j}$, tending to zero as $M\to\infty$. This extends the finite-list count convergence to the claimed joint law. Optimality of the source is the absence of negative fields, and uniqueness the absence of nonpositive fields, on the same event. Summing their independent Poisson intensities gives the two limits. The target-only solver also has uniform constants in this window: the proof uses only fixed lower bounds on $\alpha_2$ and $\beta$ and an upper bound on $c_n$.
\end{proof}

\subsection{Finite target-only diagnostic}
The finite experiment uses 180 independently generated targets, 30 per pair $(n,c)$ in $\{12,16\}\times\{1.5,3,4\}$, at $q=0.05$ (seed 2026112402). The source is uniformly random and never supplied to the algorithm. Of these targets, 94 have a strictly suboptimal source. The implementation takes the literal-frequency reference and one simultaneous best response, as in the proof. The theoretical constants in the component and field caps are conservative; this diagnostic instead uses the smaller cutoffs $B_0\in\{0,1,2\}$, keeping $U=\{v:d_v(r)\le B_0+\Gamma\}$ and refusing components larger than eight. For each target, in addition to the original objective, it evaluates one randomly generated reweighting at each of $\Gamma=1/2,3/2$. The perturbed edges form a greedy random matching and each selected weight changes by $+\Gamma$ or $-\Gamma$ with equal probability. This is a test of sampled queries, not a certificate of common exactness.

\begin{center}
\begin{tabular}{rrrrrrr}
\toprule
$\Gamma$ & $B_0$ & Refused & Exact hits & Domain hits & All optima & Free fraction\\
\midrule
0 & 0 & 1 & 160 & 160 & 72 & .237\\
0 & 1 & 49 & 126 & 172 & 133 & .506\\
0 & 2 & 116 & 64 & 179 & 172 & .721\\
$1/2$ & 0 & 1 & 153 & 153 & 73 & .237\\
$1/2$ & 1 & 49 & 125 & 171 & 139 & .506\\
$1/2$ & 2 & 116 & 64 & 178 & 172 & .721\\
$3/2$ & 0 & 49 & 102 & 140 & 107 & .506\\
$3/2$ & 1 & 116 & 61 & 168 & 158 & .721\\
$3/2$ & 2 & 171 & 9 & 179 & 174 & .868\\
\bottomrule
\end{tabular}
\end{center}
Each row has denominator 180. Exact hits count successful nonrefused component enumeration; domain hits count domains containing at least one global optimizer, irrespective of the cap. All optima counts domains containing every global optimizer of the sampled query. Free fraction is averaged over all 180, including refusals. At the smallest cutoff and zero perturbation, 160/180 targets are solved exactly, including 78/94 source-suboptimal targets, with a mean free fraction of .237. Raising the cutoff improves domain coverage but increases cap refusals. All 540 query optima are computed independently by exhaustive integer enumeration (18,800,640 assignment scores); the algorithm does not use these optima to choose its reference or cutoff. All 952 nonrefused restricted scores match the independently enumerated restricted maximum. The code also checks 5,040 direct flip differences, 507 deterministic gain inequalities conditional on at least one optimizer lying in the intrinsic weak set, and 345 multiplicity inequalities conditional on all optimizers lying in it. All checks pass. The generated clause lists, source assignments, half-unit weights, per-query outcomes, field checks, and reproduction script are supplied with the code and data. These small tests illustrate the finite cutoff tradeoff, not the onset of the asymptotic sharp constants.

A separate finite fixture tests common dimension rather than sampled-query success. Its fourteen-vertex core is divided into two blocks of seven. Within each block all clauses forbid $11$; across blocks an oriented edge forbids tail $1$, head $0$. Orient four of the seven cyclic cross edges out of each vertex of the first block, and the remaining three out of each vertex of the second. Attach four leaves with source fields $-1,0,1,0$; the last leaf has no incident nonzero field contribution, while the other three do. All five attachment edges have distinct core endpoints in the first block. For a core flip set with $s_1,s_2$ flipped vertices in the two blocks and $s=s_1+s_2>0$, the number of violated core clauses is at least
\[
 4s_1+3s_2-s_1s_2+\binom{s_1}{2}+\binom{s_2}{2}
 =\tfrac52s+s_1+\tfrac12(s_1-s_2)^2.
\]
Resetting the core to the source can lose at most $s_1$ base units on the attachment edges, and at most $\Gamma s$ from perturbations. For every $\Gamma\le3/2$ the reset therefore improves strictly, for any leaf assignment. The four leaves decouple uniformly over the entire budget family, not merely the tested queries. The matching construction in the proof gives exact common dimensions $1,1,3$ at budgets $1/2,1,3/2$, respectively: the inactive zero field need not be free, and ties at equality do not require freedom. A separate exact enumerator checks all $16,383$ nonempty core flip sets and checks the conservative reset inequality; it then evaluates all $2^{18}$ assignments for each of the 12 disjoint-answer queries, or 3,145,728 assignment scores. The minimum reset margin is three half units in the finite check, and every selected leaf sign is forced as prescribed. The fixture is constructed for the deterministic identity, not sampled from the random model.

\section{A clause objective with hidden parity information}
\label{app:max3sat-transfer}
The literal-frequency estimator for MAX-2-SAT uses information absent from noisy parity. A different planted clause law has no information about any individual literal or pair of literals. Nevertheless, a parity statistic of each clause can be used to acquire a small domain for the original, non-parity satisfaction objective. The distinction between learning a source and fitting its realized observations is particularly important here: even exact knowledge of the latent tensor does not determine which of its four factor orientations gives the best clause score.

\paragraph{Model.} Retain the tripartite mask, independent uniform source factors $T$, and noisy fiber-reading law of Section~\ref{sec:model}. At each observed triple $e$, draw a forbidden sign pattern $L_e\in\{-1,1\}^3$ independently given $T,H$. Each of the four patterns with product equal to the source product $T_e$ has probability $b=q/4$; each of the other four has probability $a=(1-q)/4$. Let $f_e(x)=\one\{x_e\ne L_e\}$ denote clause satisfaction, where $x_e$ here is the ordered triple of factor signs, not their product. The objective is $F_\xi(x)=\sum_e(1+\xi_e)f_e(x)$, with the local budget $\mathcal U_{\Gamma_n}(H)$ of Section~\ref{sec:model}. The original MAX-3-SAT objective is $F=F_0$; neither projection nor acquisition adds clauses to it. The planted-clause distribution framework is related to \citet{feldman2018planted}; our objective is exact empirical fitting rather than source identification.

Write $Y_e=-\prod_{v\in e} L_{ev}$. Conditional on $T,H$, the $Y_e$ are precisely independent noisy parity labels with error probability $q$. Conditional on $Y,H,T$, each $L_e$ is uniform on the four patterns of product $-Y_e$, independently of $T$. Thus the complete clause record and the parity record are equivalent for inference about $T$ (with the same optional readings): one is a deterministic projection, the other a randomized simulation independent of the source. This equivalence does not identify their optimization objectives. Every one- and two-literal marginal is exactly uniform even conditional on the source. In particular, the estimator of Appendix~\ref{app:max2sat} has no population signal in this law.

Let $\mathcal A(T)$ consist of the four assignments obtained from $T$ by reversing any even number of entire blocks. Unlike parity, clause satisfaction is not invariant under these reversals. For each $g\in\mathcal A(T)$ put $D_v(g)=F(g)-F(g^{\{v\}})$, with superscripts denoting sign flips. Set
\[
\begin{gathered}
 h_3=(\sqrt a-\sqrt b)^2=\tfrac14(1-2\sqrt{q(1-q)}),\quad
 \alpha_3=ch_3,\\
 \lambda=\tfrac12\log(a/b),\quad
 A_3=(4\pi c\sqrt{ab})^{-1/2},\quad r_3=n^{1-\alpha_3}/\sqrt{\log n}.
\end{gathered}
\]

\begin{theorem}[Acquisition transfers from parity to clauses]\label{thm:clause-transfer}
Fix $c>0$, $q,s\in(0,1/2)$ and a deterministic $\Gamma_n=o(\log n)$. Given the declared positive channel-bias bounds of Theorem~\ref{thm:geometric}, a randomized algorithm reads $O(n/\log n)$ noisy fiber entries and constructs four fixed-coordinate domains. With probability $1-o(1)$, their union contains every optimizer of every $F_\xi$, $\xi\in\mathcal U_{\Gamma_n}(H)$; each domain has bounded-size induced components. Exact optimization on the union, including acquisition, takes $O(n\log^2 n)$ word operations for the unweighted problem, on every input, with a fixed component cap. For real reweightings the subsequent bound is $O(m+n)$ arithmetic operations per query (for rational weights, polynomial bit complexity). The algorithm always supplies a feasible assignment; an exceeded cap need not yield an exact answer. For $0<\alpha_3<1$,
\begin{align}
 \frac{\max_x F(x)-\max_{g\in\mathcal A(T)}F(g)}{r_3}
 &\xrightarrow{\Pr}3A_3\frac{e^{-\lambda}}{(1-e^{-\lambda})^2},\label{eq:clause-transfer-gain}\\
 \frac{\log_2|\arg\max_x F(x)|}{r_3}&\xrightarrow{\Pr}3A_3.\label{eq:clause-transfer-entropy}
\end{align}
Moreover, an exact target-only clause optimizer with probability bounded away from zero yields, by randomized polynomial-time postprocessing of the noisy parity record alone, an estimate whose tensor disagreement fraction is $o(1)$ with at least that probability minus $o(1)$.
\end{theorem}

\subsection{Four neighborhoods of the population optima}
For a fixed $g\in\mathcal A(T)$ and flip set $S$, write $J_g(S)=F(g)-F(g^S)$. If a grid triple has odd intersection with $S$, its cost is $+1,-1,0$ with probabilities $a,b,1-a-b$. If the intersection has size two, the cost is $+1,-1,0$ with probabilities $b,b,1-2b$. Empty intersections contribute zero. In particular the expected cost on odd intersections is $a-b>0$. The perturbation changes $J_g(S)$ by at most $\Gamma_n|S|$, as each changed clause meets $S$ and its change in satisfaction has absolute value at most one.

Every assignment farther than $\delta n$ Hamming distance from all four anchors, for fixed $0<\delta<1/2$, has tensor disagreement fraction at least $\delta/3$ by the block-correlation argument of Appendix~\ref{app:local}. Its expected loss relative to any fixed anchor is at least $(a-b)\delta m/3$. For completeness, for a random grid edge the mean of $e^{-\theta J_e}$ is at most $1-\theta(a-b)\delta/3+C\theta^2$ for $0<\theta\le1$, by $|J_e|\le1$. Choose a sufficiently small fixed $\theta$ so this is less than $1-\epsilon$ for fixed $\epsilon>0$. Conditional on the mask, labels are independent. Maclaurin's inequality applied to the grid moments bounds the probability of loss at most $3\Gamma_n n$ by $e^{3\theta\Gamma_n n}(1-\epsilon)^m$. Union over $2^{3n}$ assignments gives $o(1)$. Thus uniformly over all admissible perturbations, every optimizer is within $\delta n$ of some anchor. The same argument for each fixed $\delta>0$ yields $o_{\Pr}(n)$ distance for every unperturbed optimizer.

We sharpen this to bounded components. At $-\lambda$ the cost moment is $1-h_3$ on odd intersections, $1+2b(\cosh\lambda-1)$ on double intersections, and one on empty intersections. With $t=|S|\le\delta n$, the fraction of triples meeting exactly one vertex is at least $(t/n)(1-t/n)$ and the fraction meeting at least two is at most $(t/n)^2/2$. Hence, by choosing $\delta$ smaller if necessary, the grid mean moment is at most $1-h_3t/(2n)$. Require a prescribed set $E$ of $s_0\le t-1$ distinct observed triples. The inclusion probability is at most $p^{s_0}$, where $p=m/n^3$. Conditional on inclusion, deleting the forced edges changes the grid mean by $O_q(s_0/n^3)$, and their factors are bounded by $e^{\lambda s_0}$. Maclaurin's inequality and $m-s_0$ remaining draws give
\[
 \Pr\{E\subset H,\ J_g(S)\le\Gamma_nt\}
 \le p^{s_0}\exp[-(h_3/2)mt/n+O_q(s_0)+\lambda\Gamma_nt].
\]
Here the constants are uniform for $t\le\delta n$. The rooted incidence-tree enumeration in Appendix~\ref{app:local} gives at most $3n4^{t+s_0}(6n^2)^{s_0}$ witnesses with $t$ variables and $s_0$ triple nodes. Summation bounds the probability of a connected improving set of size $t$ by $3n(C\log n)^t n^{-(\alpha_3/2-o(1))t}$. A fixed $J_0>4/\alpha_3$ makes the sum over $J_0\le t\le\delta n$ vanish. The event holds for all four anchors. For an optimizer near $g$, its difference components are additive and undoing any one cannot improve the perturbed score, so each has fewer than $J_0$ vertices. With probability $1-o(1)$ pair codegrees are at most two (Appendix~\ref{app:local}). Undoing a single changed sign then gives
\begin{equation}\label{eq:clause-transfer-localization}
 S\subseteq\{v:D_v(g)\le4(J_0-2)+\Gamma_n\}.
\end{equation}
Indeed each edge shared with another changed vertex alters the single-flip loss by at most two, and the local budget alters it by at most $\Gamma_n$. The argument is simultaneous over all admissible weights, with no assumptions on initialization.

\subsection{Mixed rare fields and an observable union}
Let $P_v=\sum_{e\ni v}Y_eT_e$ be the parity field. For a deterministic $L_n=o(\log n)$, consider the union $W$ of the vertices where $P_v\le L_n$ or $D_v(g)\le L_n$ for any of the four anchors. For any fixed $t$ specified distinct vertices and fixed $s_0$ forced distinct triples, choose one of the five field types at each vertex. An edge meeting exactly one chosen vertex has tilted moment $1-h_3$ for a clause field or $1-4h_3$ for a parity field, at the same $-\lambda$. Edges meeting multiple chosen vertices have bounded moments; they occupy $O_t(n)$ grid locations. Removing the forced edges changes the mean by $O_{t,q,s_0}(n^{-3})$. The grid mean is at most $1-th_3/n+O_{t,q,s_0}(n^{-2})$. Markov and Maclaurin, followed by union over the $5^t$ choices, yield
\begin{equation}\label{eq:mixed-clause-weak}
 \Pr\{E\subseteq H,\ v_1,\ldots,v_t\in W\}\le p^{s_0} n^{-t\alpha_3+o(1)}.
\end{equation}
The error is uniform over labeled witnesses and the anchors, conditional on the source. The union of fields is essential: clause and parity estimation errors on the same edges need not be independent. Summing the incidence-tree witnesses for a fixed integer $J>1/\alpha_3$, $J\ge2$, gives no connected $J$-vertex subset of $W$ with high probability. Prescribing stars with $k>1/\alpha_3$ distinct weak neighbors of an arbitrary root gives at most $n(C\log n)^k n^{-k\alpha_3+o(1)}$ expected witnesses (each distinct incident triple supplies at most two neighbors). Thus all vertices have fewer than $k$ neighbors in $W$. These conclusions apply at any deterministic sublogarithmic threshold, including a larger threshold chosen in advance.

Run the parity cascade on $Y$ and keep its first all-target iterate $z^1$ rather than an arbitrary parity optimizer. By Appendix~\ref{app:recycling}, for fixed parameters, in its true reference-fiber gauge $E(z^1)\subseteq\{v:P_v\le C\}$ for a fixed $C$ with high probability. The reference-fiber gauge is one of the four anchors. Take all four even-block reversals $r$ of $z^1$. On the preceding event these correspond one-to-one to the anchors $g$ with $\{v:r_v\ne g_v\}\subset W$. Define the observable domains by freeing
\[
 U(r)=\{v:F(r)-F(r^{\{v\}})\le 4(J-2)+\Gamma_n\}.
\]
Choose $L_n\ge\max(C,4(J_0-2)+\Gamma_n)$ in the first weak-set event. Any optimizer near $g$ differs from $r$ only inside $W$ by~\eqref{eq:clause-transfer-localization}. Every connected component of that difference has fewer than $J$ vertices. Comparing its single-sign undo with the reference as before implies that every difference vertex is in $U(r)$. For $v\notin W$, $r_v=g_v$ and its margin differs from $D_v(g)$ by at most $4k$ (at most two edges per erroneous neighbor, each changing a margin by at most two). Consequently $U(r)$ is contained in the weak union at threshold $\max(L_n,4(J-2)+\Gamma_n+4k)$. Apply~\eqref{eq:mixed-clause-weak} again at this deterministic enlarged threshold. All four induced domains have components of fewer than $J$ vertices on the same event. This proves common exactness of their union, not of any one of them. A fixed cap $J-1$ and the four domain searches cost $O(m+n)$ arithmetic operations, by local truth-table enumeration; if a cap fails, keep the reference as a feasible candidate. Integer margins and scores fit in words for the original objective. The stored acquisition and reuse cost and reading bound give the stated all-input complexity. Simulating the clauses from a parity record requires only two independent fair bits per edge, and the localization already proved makes any exact optimizer a near-source tensor. This establishes the last claim of the theorem.

\subsection{Sharp residual laws}
Fix an anchor $g$. Its field at any vertex is a sum of independent three-valued increments conditional on its degree, with probabilities $a,b,1-a-b$. Conditional on degree $d$, tilting at $-\lambda$ changes the two nonzero masses to $\sqrt{ab}/(1-h_3)$, with the remaining probability at zero. The unconditional degree is hypergeometric. Multiplicative Poisson approximation uniformly for $d=O(\log n)$, with a negligible large-degree tail, as in Appendix~\ref{app:dimension-exact}, replaces the positive and negative counts by independent Poisson variables of means $a m/n,b m/n$. Tilting balances both means at $\sqrt{ab}m/n$. Fourier inversion of the resulting difference gives
\[
 \Pr\{D_v(g)=j\}\sim A_3e^{\lambda j}n^{-\alpha_3}/\sqrt{\log n}
\]
for each fixed integer $j$. In the Bernoulli mask with probability $p$, the same formula follows by tilting the independent grid increments, whose positive and negative probabilities before tilt are $pa,pb$. Within a factor block the fields are independent in that mask. For each fixed $j$, the count in each block has diverging mean of order $r_3$ if $\alpha_3<1$; multiplicative Chernoff bounds give failure probability $\exp[-\Omega(r_3)]$ for any fixed relative error. Conditioning the Bernoulli mask on size $m$ costs only $O(\sqrt m)$ by Stirling's formula. Thus the normalized count over the three blocks converges in probability to $3A_3e^{\lambda j}$, jointly over finitely many $j$ and all four anchors.

The tilted three-valued sum at degree $d$ has maximal point mass $O_q((d+1)^{-1/2})$ by Fourier domination (the increment has a positive zero mass and span one). Multiplying by $(1-h_3)^d e^{\lambda j}$ and averaging over the hypergeometric degree gives the uniform bound $C_{c,q}e^{\lambda j}n^{-\alpha_3}/\sqrt{\log n}$ for all integer $j$: split degrees at a sufficiently small constant times $\log n$ and use the tilted degree moment as in Appendix K. This justifies truncation of the negative weighted tail by $\sum_{j>M}j e^{-\lambda j}$.

For the unweighted objective take $W_g=\{v:D_v(g)\le L\}$ with fixed $L\ge\max(0,4(J_0-2))$. The weak-field witness bound implies that the expected number of observed edges with two or more vertices in $W_g$ is at most $n^{1-2\alpha_3+o(1)}$: prescribe one edge and two weak endpoints, sum over the $n^3$ grid triples. Hence this count is $o_{\Pr}(r_3)$. The number of interacting weak vertices is at most three times the count. Also $|W_g|=O_{\Pr}(r_3)=o_{\Pr}(n)$ by the tail bound, so the four domains freezing the anchors outside $W_g$ are disjoint with high probability. All optima belong to their union by~\eqref{eq:clause-transfer-localization}. For each restricted domain the gain over its anchor is $\sum_v(-D_v(g))_++o_{\Pr}(r_3)$: the discrepancy between the gain of any subset of $W_g$ and its additive single-flip gain is bounded by four times the internal-edge count, and flipping precisely the negative fields achieves the additive maximum. Within each domain the base-two logarithm of the number of restricted maximizers differs from the number of zero fields by at most the number of interacting weak vertices: singleton fields with nonzero margin fix a sign, singleton zero fields give two choices, and each remaining component contributes at most one bit per variable. Zero fields are all in $W_g$. Both estimates hold jointly across the four anchors. The maximum of the four gains added to the corresponding anchor scores exceeds the maximum anchor score by the same leading constant, because the four normalized gains converge to that constant. The global optimizer count is the sum of the restricted optimizer counts of the winning domains (at least one and at most four). All four have the same leading log count, so choosing a winning orientation introduces no additional asymptotic assumption. The local count limit and the geometric sum $\sum_{j\ge1}je^{-\lambda j}=e^{-\lambda}/(1-e^{-\lambda})^2$ prove~\eqref{eq:clause-transfer-gain}--\eqref{eq:clause-transfer-entropy}.

\subsection{When a common coordinate domain becomes extensive}
The clause objective separates two notions that agree to leading order for the parity model. Each of the four neighborhoods is small, but a single coordinate cube must allow changes between macroscopically distinct orientations. We use $d_{\rm fix}$ and $N_{\rm hit}$ as in Section~\ref{sec:exact-common-dimension}, here over actual factor assignments without a gauge quotient.

\begin{theorem}[Four small domains versus one large domain]\label{thm:clause-four-domains}
Under the clause law above, for every fixed $\Gamma>0$,
\[
 \frac{d_{\rm fix}(\Gamma)}{3n}\xrightarrow{\Pr}1.
\]
If $0<\alpha_3<1$, then
\[
 \frac{\log_2N_{\rm hit}(\Gamma)}{r_3}
 \xrightarrow{\Pr}3A_3\sum_{j\in\mathbb Z:\,|j|<\Gamma}e^{\lambda j}.
\]
For $\alpha_3\ge1$, with probability $1-o(1)$, $d_{\rm fix}(\Gamma)=3n$ and $N_{\rm hit}(\Gamma)=4$. At zero budget $N_{\rm hit}=1$ and $d_{\rm fix}=0$. In contrast to the minimum single-cube dimension, the union of four observable domains in Theorem~\ref{thm:clause-transfer} can be searched in $O(m+n)$ arithmetic work per query.
\end{theorem}
\begin{proof}
Fix $\Gamma>0$ and a deterministic cutoff $L\ge\max(\Gamma,4(J_0-2)+\Gamma)$. Put $W_g=\{v:D_v(g)\le L\}$ and $W^\cup=\bigcup_g W_g$. With high probability all optimizer differences from some anchor lie within its $W_g$. The maximum unsigned degree is $O_c(\log n)$, and $|W^\cup|=O_{\Pr}(r_3)$ for $\alpha_3<1$; for $\alpha_3\ge1$ the same uniform tail bound makes $W^\cup$ empty with high probability. Thus uniformly over all allowed weights and optimizers in the neighborhood of $g$, the objective advantage over $g$ is at most $(1+\Gamma)\Delta(H)|W_g|=o_{\Pr}(n)$, and in the subcritical case it is $O_{\Pr}(r_3\log n)$.

Each $g$ violates a distinct class of clauses, of size $V_g=\sum_e\one\{L_e=g_e\}$. For fixed source and mask, the four counts and the count of remaining clauses have a multinomial law with probabilities $(b,b,b,b,1-4b)$. In particular $V_g=(b+o_{\Pr}(1))m$ jointly and all unperturbed anchor-score differences are $O_{\Pr}(\sqrt m)$. Choose $0<\epsilon<\min(\alpha_3,1/2)$ and set
\[
 \zeta^{g}_e=-\frac{n^{-\epsilon}}{\log n}\one\{L_e=g_e\}.
\]
This uses local budget $O(n^{-\epsilon})$ with high probability. It lowers the weights of clauses violated at $g$, which are satisfied at every other anchor $h$. It therefore increases the score difference $F(g)-F(h)$ by $n^{-\epsilon}V_g/\log n=\Theta_{\Pr}(n^{1-\epsilon})$. This dominates both $\sqrt m$ and $r_3\log n$ (and dominates zero when the weak sets are empty). Consequently, with probability $1-o(1)$ every optimizer of $F_{\zeta^g}$ lies in the neighborhood of $g$, simultaneously for the four choices. The construction is a witness in the perturbation family, not an acquisition algorithm; it may use the source. Any coordinate outside $W^\cup$ has opposite values at some pair of anchors. A fixed-coordinate domain containing an optimizer for all four of these objectives must therefore free every coordinate outside $W^\cup$. The full cube is an upper bound. This proves the dimension claim and its exact version when $W^\cup$ is empty.

For the hitting-set upper bound, freeze each singleton of $W_g$ with $|D_v(g)|\ge\Gamma$ to its unperturbed preferred sign. It remains a maximizing choice even when the perturbation creates a tie. Retain the interacting vertices and the singletons with $|D_v(g)|<\Gamma$. The frozen singleton coordinates take their preferred signs; a negative field can therefore be fixed opposite to $g$. The resulting cube contains an optimum of every objective whose global optimum occurs in the neighborhood of $g$, because its component scores are independent conditional on the strong coordinates. Its dimension is at most $\#\{v:|D_v(g)|<\Gamma\}$ plus the number of interacting weak vertices. Taking the union of the four cubes gives a hitting set. In the subcritical regime, the local limit and the $o_{\Pr}(r_3)$ interaction bound give the claimed upper constant; the factor four contributes at most two additional bits.

For the matching lower bound, work at one anchor $g$. Keep vertices with $|D_v(g)|<\Gamma$ that have an active incident clause (a nonzero single-flip loss) and no other vertex of $W_g$ at distance at most two in the interaction graph. There are $R=(3A_3\sum_{|j|<\Gamma} e^{\lambda j}+o_{\Pr}(1))r_3$ such vertices. Indeed, the no-active-clause probability is at most $[1-(a+b)/n]^m$ by the hypergeometric moment bound; since $c(a+b)>\alpha_3$, the expected inactive count is $o(r_3)$. To count pairs of weak vertices at distance one or two, prescribe the one or two connecting triples and apply~\eqref{eq:mixed-clause-weak} with two weak endpoints. The number of labeled paths with these prescribed triples, weighted by their inclusion probabilities, is $O(n\log^2 n)$, by the same rooted incidence-tree count (a path with both steps in one triple is counted as distance one). Thus their expectation is at most $n^{1-2\alpha_3+o(1)}=o(r_3)$. Dropping their endpoints is negligible.

Choose one active incident clause at each retained vertex; their scopes are disjoint. Under the steering perturbation $\zeta^g$, the single-flip margin changes by $o(1)$ uniformly, because its local budget is $O(n^{-\epsilon})$. The finitely many integers $j$ with $|j|<\Gamma$ have a positive gap to the boundary. Choose $\delta_0>0$ so $|j|+2\delta_0<\Gamma$ for each. On the selected clause with base increment $s_e\in\{-1,1\}$, add a coefficient $\sigma s_e(|D_v(g)|+\delta_0)$, where $\sigma=\pm1$ is the desired sign of the perturbed margin. These added coefficients respect the remaining local budget for all sufficiently large $n$, and force the desired sign strictly. Their total absolute mass is at most $\Gamma R=O_{\Pr}(r_3)$, which cannot undo the steering gap of order $n^{1-\epsilon}$. The previous uniform localization still holds, so all global optima are in $g$'s neighborhood. The retained vertices have no weak neighbor; their incident clauses therefore have all other coordinates fixed at $g$ in every optimizer in this neighborhood. Their forced choices are independent. The $2^R$ sign patterns give pairwise disjoint global optimizer sets, and every hitting set requires at least $2^R$ assignments. This proves the leading lower constant. If $\alpha_3\ge1$, all four anchors suffice and the four steering witnesses force each separately, giving $N_{\rm hit}=4$. At $\Gamma=0$, any one optimizer suffices by definition.
\end{proof}

\subsection{Finite diagnostic}\label{app:clause-transfer-finite}
We generate 90 targets at $n=4,5,6$, $c=3,6$, with 15 independent targets per cell, $q=.17$, and sensor error $.10$. Exhaustive integer enumeration of the $2^{3n}$ assignments supplies the exact optimum. At each of $\lceil n/4\rceil,\lceil n/2\rceil,n$ readings, compare matched retained and replacement prefixes of the first fiber. The finite heuristic uses four sign pairs from a singular-value decomposition of the full-target projected matrix, two sequential parity sweeps per start, and eight clause-coordinate sweeps for each of the four orientations. A target-only comparator uses four independent initial sign arrays and the same refinement budget, without the matrix decomposition. This is a diagnostic for the original clause objective, not an implementation of the split-batch proof. The exact success counts (retained, replacement, target-only) are $(87,87,88)$, $(87,88,88)$, and $(87,87,88)$ out of 90, respectively. Among the 29 targets for which even the best true anchor is suboptimal, the counts are $(26,27,27)$ at every budget. Thus this small-instance comparison shows no retained-source advantage; the target-only solver is already highly successful. Theoretical acquisition order does not predict the finite advantage of this heuristic.

As a separate geometry diagnostic, the union of four true-anchor weak cubes at cutoffs $0,1,2$ attains the unrestricted optimum on 82, 88, and 90 targets. This uses the unknown source and exhaustive cube comparisons and is not an observable solver. All 5,400 direct margin comparisons and 1,080 restricted gain and entropy bounds hold exactly; the reference enumeration evaluates 8,970,240 assignment scores. A separately seeded full-mask fixture tests the extensive-dimension conclusion by explicitly steering four distinct objectives; at $n=5$ with all 125 triples, the four unique optima require all 15 coordinates free at budget 5. The enumerator checks 131,072 assignment scores on the first accepted draw and the local budgets. The fixture is selected by a declared finite search, not a random-frequency estimate. All trial records, feasible witnesses, exact reference scores, and the fixture are supplied in the code/data supplement.

\section{Two scales of residual freedom in the clause model}
\label{app:sector-window}
The minimum hitting-set size for a family of nearby objectives depends both on which signs can change locally and on which of the four source orientations can become optimal. At the density where only finitely many weak fields remain, we can distinguish these two effects exactly. The scores of the four orientations fluctuate on the scale $\sqrt m$, whereas each local margin is an integer. A vanishing local perturbation budget can change a zero margin, but moving between orientations requires a total coefficient change comparable with $\sqrt m$.

Use the law of Appendix~\ref{app:max3sat-transfer}. Label the four even-block orientations $g=1,2,3,4$ in a fixed order relative to the source. Write $V_g$ for the number of clauses violated at anchor $g$, $D_v(g)$ for its single-flip loss, and $N_{gj}=\#\{v:D_v(g)=j\}$. An incident clause is active at $(g,v)$ when changing only $v$ changes its satisfaction. Define
\[
 R_g(\Gamma)=\#\{v:|D_v(g)|<\Gamma,\ \text{some incident clause is active at }(g,v)\}.
\]
A positive budget is understood throughout; zero budget is considered separately.

\begin{theorem}[Exact common freedom in the singleton regime]\label{thm:clause-singleton-exact}
At fixed density $\alpha_3>1/2$ and any fixed $\Gamma>0$, with probability $1-o(1)$,
\begin{equation}\label{eq:clause-singleton-exact}
 d_{\rm fix}(\Gamma)=3n,\qquad N_{\rm hit}(\Gamma)=\sum_{g=1}^4 2^{R_g(\Gamma)}.
\end{equation}
These equalities hold jointly for any fixed finite collection of positive budgets. In particular, the exact dimension conclusion of Theorem~\ref{thm:clause-four-domains} extends throughout the singleton regime, including densities with a diverging number of local decisions.
\end{theorem}

\begin{proof}
Fix a field cutoff $L$ large enough for the uniform localization in~\eqref{eq:clause-transfer-localization} for the largest budget. Let $W_g=\{v:D_v(g)\le L\}$ and $W=\bigcup_g W_g$. For distinct physical vertices, the mixed-field witness bound~\eqref{eq:mixed-clause-weak} and the one- and two-edge path count used in Theorem~\ref{thm:clause-four-domains} give at most $n^{1-2\alpha_3+o(1)}$ expected pairs of weak vertices at distance at most two. The same vanishing bound holds for a vertex weak in two different orientations. Indeed, on an edge incident to a specified vertex $v$, the two single-flip losses for $g,h$ are nonzero on disjoint forbidden patterns: the pairs of patterns are $(g,g^{\{v\}})$ and $(h,h^{\{v\}})$. The first pattern in each pair has product equal to the source product, the second has opposite product, and all four patterns are distinct. At the common tilt $-\lambda$, the joint moment is $1-2h_3$. Maclaurin's inequality over the grid and Markov's inequality therefore bound $\Pr(D_v(g),D_v(h)\le L)$ by $e^{2\lambda L}\exp(-2h_3m/n)$. Sum over the $3n$ vertices and the six pairs of orientations. Thus, on one high-probability event, no physical vertex belongs to two of the $W_g$, and no two vertices in $W$ lie at distance at most two.

Every optimizer is in the union of the four domains that fix $g$ outside $W_g$. Within each domain the objective is separable. A coordinate with $|D_v(g)|\ge\Gamma$ can be frozen to its preferred unperturbed sign (at equality this sign may tie). An inactive coordinate cannot change its preference because its neighbors in the domain are fixed at $g$. Only the $R_g(\Gamma)$ active, strictly reversible coordinates need both signs. The resulting four trimmed cubes contain at least one optimizer for every admissible objective. They are disjoint because their anchors differ in $2n$ coordinates and $|W|=o_{\Pr}(n)$. This gives the upper bound in~\eqref{eq:clause-singleton-exact}.

For the lower bound, the steering perturbation and the local sign-forcing perturbations in Theorem~\ref{thm:clause-four-domains} now apply to \emph{all} active, reversible fields in each orientation, without dropping any weak neighbors. The total mass of the local sign-forcing perturbations is at most $\Gamma |W|$, and is negligible relative to the steering gap by the argument of that theorem. Each sign pattern in each sector forces a disjoint set of optima, and sets forced in different sectors are disjoint. Every hitting set therefore has at least $\sum_g2^{R_g(\Gamma)}$ elements. For the dimension, at every physical coordinate there are at least three orientations in which it is outside the weak set. Any three of the four anchors contain both signs at that coordinate. The four steering queries therefore force both signs at every coordinate even if one of the sectors can modify that coordinate. A single common cube must be the full cube. All estimates can use the largest cutoff and the minimum positive boundary slack over the given finite collection of budgets.
\end{proof}

\subsection{Joint Gaussian and Poisson fluctuations}
Let $m=m_n$ be deterministic and satisfy
\begin{equation}\label{eq:clause-critical-window}
 h_3 m/n=\log n-\tfrac12\log\log n+t+o(1),\qquad t\in\mathbb R.
\end{equation}
Put $c_*=1/h_3$, $A_*=(4\pi c_*\sqrt{ab})^{-1/2}$ and $\theta=3A_*e^{-t}$. Let $P_{gj}$ be independent Poisson variables of means $\theta e^{\lambda j}$ for all four $g$ and integers $j$. Independently let $Z\sim\mathcal N(0,b I_4-b^2\mathbf 1\mathbf 1^\top)$, and write $g_*=\arg\min_g Z_g$ (almost surely unique).

\begin{theorem}[Joint orientation and residual limit]\label{thm:clause-window}
In the window~\eqref{eq:clause-critical-window}, the vectors $((V_g-bm)/\sqrt m)_{g=1}^4$ and $(N_{gj})_{g,j\in I}$ converge jointly to $Z$ and $(P_{gj})_{g,j\in I}$ for any fixed finite set $I$ of integer fields. With $G_n=\OPT-\max_g F(g)$ and $\mathcal O=\arg\max F$,
\begin{equation}\label{eq:clause-window-unweighted}
 \left(\frac{\OPT-(1-b)m}{\sqrt m},G_n,\log_2|\mathcal O|\right)
 \Rightarrow \left(-\min_g Z_g,\sum_{j\ge1}jP_{g_*,-j},P_{g_*,0}\right).
\end{equation}
For each fixed $\Gamma>0$, jointly with~\eqref{eq:clause-window-unweighted},
\begin{equation}\label{eq:clause-window-fixed-budget}
 d_{\rm fix}(\Gamma)=3n\ \text{with probability }1-o(1),\qquad
 N_{\rm hit}(\Gamma)\Rightarrow \sum_{g=1}^4 2^{\sum_{|j|<\Gamma}P_{gj}}.
\end{equation}
The best source orientation is optimal with limiting probability $\exp[-\theta/(e^\lambda-1)]$. The probability of a unique global optimum tends to $e^{-\theta}$; the probability that the best source orientation is itself uniquely optimal tends to $\exp[-\theta/(1-e^{-\lambda})]$.
\end{theorem}

\begin{proof}
Condition on the source. All forbidden patterns, written relative to it, are independent with four probabilities $b$ and four probabilities $a$. The orientation violation counts have the stated multinomial covariance, and the ordinary multivariate central limit theorem gives the Gaussian marginal. The field-count marginal requires a rare-event argument because a fixed-size mask cannot be replaced by a Bernoulli mask merely by conditioning a high-probability event.

The one-vertex tilting calculation in Appendix~\ref{app:max3sat-transfer} is uniform for $m/n\sim c_*\log n$. In particular, for fixed $j$,
\[
 \Pr(D_v(g)=j)\sim A_*e^{-t}e^{\lambda j}/n,\qquad
 \Pr(D_v(g)=j)\le C_q e^{\lambda j}/n
\]
for all integers $j$ in the second bound and sufficiently large $n$ (the constant can depend on $t$). The latter follows by splitting the unsigned degree at $\kappa\log n$ for a sufficiently small fixed $\kappa$, just as in Appendix~\ref{app:max3sat-transfer}; all degree constants can be chosen on a compact neighborhood of $c_*$. It yields a uniformly summable negative tail.

For a fixed number $k$ of distinct physical vertices with specified orientations and integer fields, partition the grid into private, shared and remaining locations as in the proof of Theorem~\ref{thm:critical-window}. Each private group has $n^2+O_k(n)$ locations; $O_k(n)$ locations are shared. Conditional on no shared location being observed, the mask is uniform on the remaining grid. The multiplicative hypergeometric-to-Poisson approximation of that proof applies unchanged to the private counts, with a degree cutoff $B_k\log n$ and discarded probability $o(n^{-k})$. Conditional on these private counts the labels are independent. Tilting each private group as above gives joint mass
\[
 (1+o(1))n^{-k}\prod_{i=1}^k A_*e^{-t}e^{\lambda j_i}.
\]
The contribution from a shared location is $o(n^{-k})$: the joint exponential moment over unforced grid locations is at most $1-kh_3/n+O_k(n^{-2})$, since a private location contributes $1-h_3$, and the shared factors are bounded. Requiring one particular shared location costs $p=m/n^3$ and a bounded factor. Summing over all shared locations gives $O_k(n^{-k-1}(\log n)^{k/2+1})$, by~\eqref{eq:clause-critical-window}. The estimates are uniform in the specified labels of the vertices and orientations.

The factorial moments of $(N_{gj})$ also include terms with repeated physical vertices from distinct orientations. Repetition within one orientation and different fields is impossible; within the same count, falling factorials exclude repetitions. If there are $k$ field requirements at $r<k$ distinct physical vertices, a location private to a vertex with $\ell$ orientation requirements has joint tilt moment $1-\ell h_3$: the $2\ell$ nonzero patterns are disjoint. Shared locations again have bounded factors, so the mean moment is at most $1-kh_3/n+O_k(n^{-2})$. Markov and Maclaurin bound the probability of these requirements by $O_k(n^{-k}(\log n)^{k/2})$ (constants also depend on the prescribed fields). Since there are only $O_k(n^r)$ such tuples, their total contribution vanishes. The distinct-vertex masses give precisely the products of the Poisson factorial moments, with the factor $3$ for the three blocks.

To prove independence from $Z$, multiply a factorial-moment term by any bounded Lipschitz test function of the normalized violation counts. For distinct vertices condition on the mask and all labels on incident edges. Their number is at most $kB_k\log n$ except on an event of probability $o(n^{-k})$, by the degree tail bound. The remaining labels are independent multinomial trials of the same five categories (four anchor violations and the remainder). The conditional normalized violation vector differs from the vector based on these remaining trials, centered at their own mean, by $O_k(\log n/\sqrt m)=o(1)$. Its conditional test expectation consequently converges to the Gaussian test expectation uniformly on the truncated event, by the central limit theorem for all sample counts $m-O_k(\log n)$. The repeated-vertex terms and shared terms remain negligible by boundedness. Thus every mixed factorial moment factors into the Gaussian test expectation and the Poisson factorial moment. Tightness follows from the marginal limits, and higher factorial moments give uniform integrability of every fixed polynomial in the counts. In a subsequential joint limit, weighting by a nonnegative bounded Lipschitz function of the Gaussian coordinates with positive expectation leaves the factorial moments of the counts unchanged after normalization. The product Poisson distribution is moment determinate (its moment generating function is finite in a neighborhood of zero). These weighted distributions are therefore the same product Poisson law, which proves the asserted independence.

The localization proof~\eqref{eq:clause-transfer-localization} holds uniformly for bounded budgets in this window. Indeed $m/(n\log n)$ stays in a compact positive interval, $h_3m/(n\log n)\to1$, and a fixed $J_0$ suffices in the connected-set bound. The pair-codegree and maximum-degree bounds are unchanged. At a fixed sufficiently large cutoff $L$ the total weak count across the four orientations is $O_{\Pr}(1)$ by the uniform field tail. The path-witness bound is $O(n\log^2 n\exp(-2h_3m/n))=O(\log^3 n/n)$ for distinct weak vertices within distance two. The same-vertex double-orientation bound in Theorem~\ref{thm:clause-singleton-exact} is $O(\log n/n)$. Also the expected number of inactive vertex--orientation pairs is at most $12n\exp[-(a+b)m/n]=o(1)$, since $a+b>h_3$. Thus with probability $1-o(1)$ all weak vertices have active incident clauses, belong to only one orientation, and have disjoint neighbor sets. Every localized domain is separable. Its unperturbed gain is exactly $\sum_v(-D_v(g))_+$ and its maximizing set has size $2^{N_{g0}}$.

For any fixed bounded budget the local improvements over the anchors are uniformly $O_{\Pr}(1)$, by the negative-tail bound and the bounded number of weak vertices. The winning orientation of the unweighted objective therefore agrees with that of the best anchor with probability $1-o(1)$: the Gaussian limit has no ties, and $\sqrt m\to\infty$. Truncating the negative fields at $-M$ costs probability $O(e^{-\lambda M})$, uniformly in $n$, and the limiting Poisson array has the same summable tail. Applying the joint limit and then sending $M\to\infty$ proves~\eqref{eq:clause-window-unweighted}. For fixed positive budgets the steering argument of Theorem~\ref{thm:clause-four-domains} is uniform in this window, since its gap of order $n^{1-\epsilon}$ dominates $\sqrt m$ and all local gains. The proof of Theorem~\ref{thm:clause-singleton-exact} gives the exact identities with $R_g(\Gamma)=\sum_{|j|<\Gamma}N_{gj}$ on the same high-probability event. This gives~\eqref{eq:clause-window-fixed-budget}. Finally $g_*$ is independent of all Poisson counts. The best anchor is optimal when its negative-field count is zero, and its local maximizing set is a singleton when its zero-field count is zero. The geometric sum of the negative-field means gives the three probabilities.
\end{proof}

\subsection{The perturbation scale for changing orientation}
Consider a strictly positive deterministic budget sequence $\Gamma_n\to0$ with $n\Gamma_n/\sqrt m\to u\in[0,\infty]$. The capacity bound $\|\xi\|_1\le n\Gamma_n$ follows by summing the local budgets over all three blocks; each clause is counted three times. It bounds the change in every pairwise anchor-score difference. The following argument shows that the full scale is achievable, rather than just an upper bound.

\begin{lemma}[Asymptotically full reweighting capacity]\label{lem:sector-capacity}
For each $g$, there are nonnegative weights $w^g_e$ supported on clauses violated at $g$ and avoiding all vertices in $W$, with $\sum_{e\ni v}w^g_e\le1$ for every vertex and $M_g=\sum_e w^g_e$, with $M_g/n\to1$ in probability. Here $W$ is the weak union at any fixed cutoff in the window~\eqref{eq:clause-critical-window}. Consequently, for any $\tau\in[0,\Gamma_n]$, the perturbation $\xi_e=-\tau w^g_e$ increases the score of $g$ relative to every other anchor by $\tau M_g$, without changing any weak margin.
\end{lemma}
\begin{proof}
Write $d=bm/n=\Theta(\log n)$ and $d_g(v)$ for the number of incident clauses violated at $g$. Conditional on unsigned degree, it is binomial with probability $b$; the unsigned degree is hypergeometric. Thus $\mathbb E d_g(v)=d$ and $\operatorname{Var}(d_g(v))\le d$. Choose $\delta_n=d^{-1/4}$. Chebyshev and Cauchy--Schwarz give
\[
 \mathbb E[d_g(v)\mathbf 1_{d_g(v)>(1+\delta_n)d}]
 \le d\Pr(d_g(v)>(1+\delta_n)d)+\sqrt{d\Pr(d_g(v)>(1+\delta_n)d)}=o(d).
\]
Remove clauses incident to any such high-degree vertex. Their expected number is $o(nd)=o(m)$ by summing this bound. Markov gives $o_{\Pr}(m)$ removed clauses. Delete also clauses meeting $W$, at most $|W|\Delta(H)=O_{\Pr}(\log n)$, hence negligible. There were $(b+o_{\Pr}(1))m$ clauses violated at $g$. Assign weight $1/((1+\delta_n)d)$ to each remaining clause in this class. The local sum is at most one by the degree deletion, and the total is $(1-o_{\Pr}(1))n$ by the retained count. The bound $M_g\le n$ follows from summing the local sums. All four constructions hold jointly. Each affected clause is satisfied at all other anchors and violated at $g$, giving the claimed score difference. No affected clause meets a weak vertex, so the single-flip losses there are unchanged.
\end{proof}

\begin{theorem}[Macroscopic freedom at a vanishing budget]\label{thm:sector-budget-scale}
In the density window~\eqref{eq:clause-critical-window}, let
\[
 I_u=\{g:Z_g-\min_h Z_h\le u\},\qquad K_u=|I_u|,
\]
where $I_\infty=\{1,2,3,4\}$. For every strictly positive $\Gamma_n\to0$ with $n\Gamma_n/\sqrt m\to u$,
\begin{equation}\label{eq:sector-budget-limit}
 \left(\frac{d_{\rm fix}(\Gamma_n)}{3n},N_{\rm hit}(\Gamma_n)\right)
 \Rightarrow \left(\phi(K_u),\sum_{g\in I_u}2^{P_{g0}}\right),
 \qquad \phi(1)=0,\ \phi(2)=2/3,\ \phi(3)=\phi(4)=1.
\end{equation}
The convergence is joint with~\eqref{eq:clause-window-unweighted}. The Gaussian variables determine the accessible orientations, independently of the Poisson local freedoms. For $u=0$, only the best orientation is accessible asymptotically; for $u=\infty$, all four are accessible and $d_{\rm fix}=3n$ with probability $1-o(1)$. At $\Gamma_n=0$ exactly, $d_{\rm fix}=0$ and $N_{\rm hit}=1$ regardless of the density.
\end{theorem}
\begin{proof}
Work on the localization, singleton and activity events of Theorem~\ref{thm:clause-window}, with a cutoff covering every $\Gamma_n\le1$. Write $B_g=V_g-\min_h V_h$, and let $I_n$ be the set of orientations whose localized domain contains an optimizer for at least one admissible query. The local gain relative to each anchor is uniformly $O_{\Pr}(1)$ over this budget family, by the same bound as in the proof of Theorem~\ref{thm:clause-window}. Let $C_n=O_{\Pr}(1)$ be an upper bound on all these local improvements. If $B_g>n\Gamma_n+C_n$, every candidate in sector $g$ is strictly inferior to a best unperturbed anchor under every query.

Conversely, suppose $B_g$ is smaller than $n\Gamma_n$ by a fixed positive multiple of $\sqrt m$ and $0<u<\infty$. Lemma~\ref{lem:sector-capacity} gives $M_g/n\to1$ in probability, so for some fixed fraction $r<1$ the steering perturbation with $\tau=r\Gamma_n$ produces a positive score gap of order $\sqrt m$ over all other anchors. More precisely, on $B_g/\sqrt m<u-\epsilon$ one may choose $r$ fixed with $ru>u-\epsilon/2$, for any $0<\epsilon<u$. The gap dominates all localized corrections. Since steering avoids $W$, it leaves the local fields there unchanged. Reserve the remaining local budget $(1-r)\Gamma_n>0$. Every zero-field vertex in $W_g$ has an active incident clause, and the chosen incident scopes are disjoint. On each selected clause add a coefficient of magnitude $(1-r)\Gamma_n/2$ and the sign that forces the desired local sign. This changes no other weak margin in $g$, and uses the reserved budget. Its total absolute mass is $O_{\Pr}(\Gamma_n)$, negligible compared with the sector gap. All nonzero margins are integer and cannot change sign at a budget below one. Thus every one of the $2^{N_{g0}}$ assignments obtained by flipping all negative fields in $g$ and choosing the zero fields freely is the \emph{unique} global optimum of some admissible query. The argument holds for all these sign choices on the same event, without a union bound over queries.

When $u=0$, the best unperturbed anchor has a gap of order $\sqrt m$ in probability in the following sense: the probability of a gap below $\epsilon\sqrt m$ tends to zero as $\epsilon\downarrow0$. It already wins after any localized correction, and the tiny sign-forcing perturbations alone expose all its zero-field choices. No other orientation can be optimal at this budget by the capacity bound. When $u=\infty$, steering with $\tau=\Gamma_n/2$ has advantage $(1-o_{\Pr}(1))n\Gamma_n/2\gg\sqrt m$ and leaves half the budget for all zero-field choices. The same construction exposes every such candidate in all four orientations.

For finite positive $u$, the Gaussian difference distribution is continuous, so the probability that a nonzero normalized deficit lies within $\epsilon$ of $u$ vanishes as $\epsilon\downarrow0$ after taking $n\to\infty$. The upper and lower arguments imply that the accessible set agrees with $\{g:B_g\le n\Gamma_n\}$ with probability $1-o(1)$ (the best anchor is always included on the high-probability gap event). This also holds in the two endpoint regimes above. All nonzero fields have fixed preferred signs within an accessible sector and only its zero fields can vary. Consequently the union of the exposed candidate sets contains every optimizer of every admissible query. Each candidate is uniquely forced by some query, so the minimum hitting set is exactly this union. The sets are disjoint across orientations. The minimum common coordinate cube is the smallest cube containing the union, obtained by fixing precisely its constant coordinates. If one orientation is accessible, its dimension is $N_{g0}$. If two are accessible their anchors differ on $2n$ coordinates; if three or four are accessible the anchors differ on all $3n$ coordinates in the sense that each coordinate takes both signs. The local changes affect at most $|W|=O_{\Pr}(1)$ coordinates, giving the normalized dimension law. With all four accessible, the no-double-orientation property ensures at least three unchanged anchor signs at every coordinate, hence the dimension is exactly $3n$. The joint Gaussian--Poisson limit of Theorem~\ref{thm:clause-window}, and the absence of boundary ties, now give~\eqref{eq:sector-budget-limit} jointly with the unweighted statistics.
\end{proof}

\paragraph{An explicit distribution.}
The differences of the Gaussian scores have the same law as differences of four independent $\mathcal N(0,b)$ variables: subtract their mean, then add a common independent centered normal of variance $b(1-4b)/4$ to obtain the stated covariance. Let $f_b$ and $F_b$ denote the density and distribution of $\mathcal N(0,b)$. For finite $u\ge0$ and $k=1,\ldots,4$,
\begin{equation}\label{eq:sector-k-law}
 \Pr(K_u=k)=4\binom{3}{k-1}\int_{\mathbb R} f_b(x)[F_b(x+u)-F_b(x)]^{k-1}[1-F_b(x+u)]^{4-k}\,dx.
\end{equation}
Condition on which score is the minimum and on its value $x$; exactly $k-1$ of the remaining three must lie between $x$ and $x+u$. The formula includes $u=0$ with the convention $z^0=1$, and tends to a point mass at four as $u\to\infty$. Given $K_u=k$, the limiting hitting count is the sum of $k$ independent powers $2^{P_0}$, $P_0\sim\operatorname{Poisson}(\theta)$. Thus the ability to vary individual signs and the ability to change a macroscopic orientation are controlled by distinct probability laws. In particular a budget tending to zero can make a common cube extensive, although the exact number of mandatory candidate answers remains tight.

\subsection{Field and steering diagnostics at the critical density}
We draw 128 independent fixed-size masks at each of $n=128,512,2048$, with $q=.17$ and $t=0$, rounding $m$ to the nearest integer in~\eqref{eq:clause-critical-window}. The source is fixed to all positive signs by changing coordinates, and all forbidden patterns are sampled from their specified law. This diagnostic counts fields and evaluates anchor scores; it does not solve the unrestricted MAX-3-SAT objectives at these sizes.

Table~\ref{tab:sector-window-diagnostic} compares the mean counts per orientation with the Poisson means. We also record, for the best anchor (ties broken in a fixed order), the number of trials having no negative field and no nonpositive field. At finite $n$ these events do not by themselves establish global optimality. Of the 128 trials per size, ties in the raw anchor scores occur in 1, 3 and 2 trials; a change of winning orientation after adding the additive negative-field gains occurs in 5, 3 and 0 trials. A physical vertex weak at cutoff one in more than one orientation appears in 16, 4 and 1 trials. These finite deviations are retained.
\begin{table}[h]
\centering
\caption{Critical-window field diagnostics, 128 trials per size. The first three means average over all four orientations. The last two counts use the best anchor. The limit row gives limiting means and expected counts on the same 128-trial scale.}
\label{tab:sector-window-diagnostic}
\begin{tabular}{rr|rrr|rr}
\toprule
$n$ & $m$ & $\#D=-1$ & $\#D=0$ & $\#D=1$ & No negative & No nonpositive\\
\midrule
128&8362&.361&.732&1.627&84&44\\
512&43828&.322&.738&1.592&90&44\\
2048&217663&.367&.799&1.566&70&30\\
Limit&&.312&.689&1.522&72.44&36.38\\
\bottomrule
\end{tabular}
\end{table}

For a finite check of the reweighting capacity, construct the weights of Lemma~\ref{lem:sector-capacity} with integer denominator $C=\lceil(1+d^{-1/4})d\rceil$, $d=bm/n$. Remove edges meeting a vertex of violation degree exceeding $C$, or meeting the diagnostic weak union at cutoff one. Give each remaining edge in the chosen violation class weight $1/C$. The independent replay verifies the resulting loads using integer arithmetic. The construction is feasible but deliberately conservative: the mean fraction $M_g/n$ grows from $.314$ to $.426$ over the tested sizes. It need not approximate one at these sizes, and cutoff one need not cover all globally optimal differences. For each $u$ in Table~\ref{tab:sector-steering-diagnostic}, the number of anchors with deficit strictly less than $\Gamma_n M_g$ (together with the anchor-score maximizers) is a lower bound on the number of attainable anchor-score maximizers, and the number with deficit at most $n\Gamma_n$ is an upper bound. Here $\Gamma_n=u\sqrt m/n$. These are finite anchor-score bounds only, not certificates for the full optimization objectives. The table reports the average bounds across every trial and the Gaussian prediction for the asymptotic number of accessible orientations. The slow finite increase of the lower bound leaves room for stronger finite fractional matching constructions.
\begin{table}[h]
\centering
\caption{Mean attainable-anchor count brackets at three budget scales. The interval endpoints are the means of the feasible steering lower bound and the total-mass upper bound. The last column uses~\eqref{eq:sector-k-law}.}
\label{tab:sector-steering-diagnostic}
\begin{tabular}{r|rrr|r}
\toprule
$u/\sqrt b$ & $n=128$ & $n=512$ & $n=2048$ & Limit mean $K_u$\\
\midrule
.5&[1.12,1.46]&[1.23,1.59]&[1.31,1.69]&1.573\\
1&[1.25,2.08]&[1.45,2.20]&[1.59,2.38]&2.208\\
2&[1.61,3.24]&[1.86,3.31]&[2.16,3.39]&3.291\\
\bottomrule
\end{tabular}
\end{table}

All 384 trial records are retained. The diagnostic includes 576 direct field checks. A separate incidence-pattern replay verifies every trial, all 1,536 rational capacity witnesses, and all 1,152 budget brackets using squared integer comparisons. The integral in~\eqref{eq:sector-k-law} is evaluated by numerical quadrature; the probabilities sum to one at each tested scale. Neither this diagnostic nor the limit theorems extend the fixed-density acquisition work guarantee to varying density without an additional algorithmic argument.

\section{The reading threshold for seeded projection}
\label{app:seeded-projection}
The one-fiber construction in Appendix~\ref{app:one-fiber} reduces the order of the target observations, but its sufficient constant comes from a uniform score bound over all pairs of sign vectors. A vanishing number of readings in the other two fibers allows a sharper analysis using propagation on the projected graph. The relevant parameter is the mean projected degree times the squared sign bias, as in the censored block model \citep{saade2015spectral}. The contribution here is the threshold in returned readings and its use in exact fitting of the original third-order objective.

\begin{theorem}[Acquisition just above the projection threshold]
\label{thm:seeded-projection}
In the retained-source model, fix rational channel biases $\mu=1-2q>0$, $\rho=1-2s>0$, and a rational projection fraction $\beta\in(0,1)$. Let $P=\lfloor\beta m\rfloor$. For any fixed rational $d_0>(\mu\rho)^{-2}$, read $k=\min\{n,\lceil d_0 n^2/P\rceil\}$ signs in the first fiber (take $k=n$ if $P=0$). There is an algorithm using at most $O(n/\log^2 n)$ further readings in the other two fibers and $O((m+n)\log(n+2))$ word operations on every input. With probability $1-o(1)$ it returns an exact empirical optimizer and the observable common restriction of Theorem~\ref{thm:common-reduction}, under the same conditions on $J,\Gamma_n$. In particular, for any fixed $\epsilon>0$, the coefficient of $n/\log n$ in the reading bound can be made at most $(1+\epsilon)/(c\mu^2\rho^2)$ by choosing $\beta$ and $d_0$ appropriately.
\end{theorem}

\begin{proof}
Choose a fixed rational $d\in((\mu\rho)^{-2},d_0)$. Randomly order the observations, reserve the first $P$ for projection, and split the rest into three batches of asymptotically equal size. The first-fiber prefix produces the projected pairs and signs as in Theorem~\ref{thm:one-fiber}. In the iid-location experiment, coupled to the original model with error $O(P^2/n^3)=o(1)$, condition on the first-fiber reading errors and the number of projected observations. The ordered pairs are iid uniform, and their signs relative to the two unread reference fibers are iid with bias
\[
 \delta_n=\mu k^{-1}\sum_{i\le k}\epsilon_i.
\]
With probability $1-o(1)$ there are at least $\ell=\lfloor dn\rfloor$ projected observations and $|\delta_n-\delta|\le(\log n)^{-2}$, where $\delta=\mu\rho$. These events use binomial and independent sensor concentration; here $k\asymp n/\log n$. Retain just the first $\ell$ projected observations, discarding their first indices. Their graph is a bipartite multigraph on two sets of $n$ vertices. All calculations below condition on a good first-fiber error vector; their bounds are uniform on this event. The iid coupling is used only for this first-stage result, not for independence of later target batches.

Independently mark vertices in the two unread blocks with a dyadic probability $p_n\asymp(\log n)^{-2}$, and read their signs through the return channel. If their number exceeds $4np_n+4$, stop and return an arbitrary feasible assignment. This event has probability $o(1)$; it is a deterministic cap on reading count. Independent marks at the two centers can also be read with independent sensor errors. Work in the reference-fiber gauge. A marked vertex $v$ provides $R_v=X_v\zeta_v$, where $X_v$ denotes its source sign, and $\E\zeta_v=\rho$. Put $R_v=0$ at unmarked vertices. Let $r=d\delta^2>1$, and take an integer $t_n=\lceil3\log\log(n+e)/\log r\rceil$. The census at a root $v$ is the sum, over all length-$t_n$ nonbacktracking edge walks from $v$, of the terminal reading times the product of projected edge signs. Denote this integer by $S_v$. It can be evaluated simultaneously for all roots by directed-edge messages: start with $R_v$ at each tail, and at every step sum incoming signed messages except the message on the reverse of the recipient edge. Treat parallel edges as different edges and exclude only the exact reverse. The final root sums use all incoming messages.

We first analyze a Poisson($d$) Galton--Watson tree with edge error bias $\delta_n$, independent Bernoulli($p_n$) leaf marks, and leaf-reading bias $\rho$. Products along paths cancel every internal source sign, including the terminal sign when multiplied by its reading. Thus the root-signed census $X_vS_v$ has mean
\[
 M_n=p_n\rho(d\delta_n)^{t_n}.
\]
The second moment counts ordered pairs of leaves. The diagonal contributes $p_n d^{t_n}$. For two distinct leaves whose last common ancestor has depth $j$, the expected number of ordered pairs is $d^{2t_n-j}$, because the Poisson offspring factorial second moment is $d^2$. Their common path cancels and their distinct paths have combined expected sign $\delta_n^{2(t_n-j)}$. Consequently
\begin{equation}
\label{eq:seeded-moment}
 \frac{\E (X_vS_v)^2}{M_n^2}
 =\frac{1}{p_n\rho^2(d\delta_n^2)^{t_n}}
 +\sum_{j=0}^{t_n-1}(d\delta_n^2)^{-j}.
\end{equation}
The first term tends to zero and the sum stays bounded, uniformly on the good error event. Normalize by the known $\overline M_n=p_n\rho(d\delta)^{t_n}$; $M_n/\overline M_n=1+o(1)$. There is a fixed integer $C_0$ such that the normalized census has second moment at most $C_0$ and mean at least $1/2$ for large $n$. Round to a sign with conditional mean $\operatorname{clip}(S_v/(4C_0\overline M_n),-1,1)$ using independent randomness at each root. Since clipping $z$ changes it by at most $z^2$, the expected root correlation on the tree is at least $1/(16C_0)$. This rounding, unlike an unproved assertion about the sign of a census, follows directly from the two moments.

Here are the local approximation details for the growing depth. A graph made of $N\sim\operatorname{Poisson}(dn)$ iid uniform pairs has independent Poisson($d/n$) counts at pair addresses. Explore from one or two specified distinct roots up to depth $t_n$. Stop if more than $M=(\log n)^B$ vertices or edges have been exposed, for a fixed $B$ sufficiently larger than $3\log(d+1)/\log r+3$. The expected number of tree vertices through depth $t_n$ is at most $(t_n+1)\max(1,d)^{t_n}$, so the probability of this stop on the comparison tree is $o(1)$. Until the stop, the unexplored pair counts incident to a new vertex are independent Poisson variables. The number directed to fresh vertices differs from Poisson($d$) by a coupling error at most $dM/n$ per vertex, and the total probability of a return to an exposed vertex or a repeated new endpoint is $O_d(M^2/n)$. With two roots, include both in the initially exposed set; the same bounds couple the two neighborhoods to independent trees. Marks, sensor errors, projected edge errors, and rounding bits agree under this coupling. No assumption on independence of source signs is needed for the census in source coordinates, since they cancel along each path.

For the deterministic $\ell$-edge experiment, couple using a common iid sequence of edges and the independent Poisson count $N$. With probability $1-o(1)$, $|N-\ell|\le n^{2/3}$. Conditional on $N$, the smaller graph has at most $\ell+n^{2/3}$ edges. The expected size of its root neighborhood through depth $t_n$ is bounded by the simple-path count $\sum_{h\le t_n}(d+o(1))^h$ (also for two roots). Hence it is at most $M$ with probability $1-o(1)$. The additional iid edges hit this neighborhood with probability at most $2M n^{2/3}/n=o(1)$. This transfers the one- and two-root couplings. In the graph, the rounded signs are bounded; the one-root coupling gives the same positive mean up to $o(1)$ and the two-root coupling gives covariance $o(1)$ uniformly for distinct vertices, conditional on the first-fiber errors. The empirical mean of the error signs in each unread block therefore exceeds a fixed positive constant with probability $1-o(1)$ by Chebyshev's inequality. The cap on the number of readings affects this event only by $o(1)$.

For the all-input operation bound, abort the projection stage if the maximum projected degree exceeds $\lceil\log^2(n+2)\rceil$. This is an $o(1)$ event by binomial tails and a union bound, and the absolute integer messages on any other input are bounded by $(\lceil\log^2(n+2)\rceil)^{t_n}$, requiring $O((\log\log n)^2)$ bits. The rational normalization and rounding threshold have $O(\log\log n)$ bits for fixed rational parameters and dyadic $p_n$; rounding to accuracy $n^{-2}$ with bounded fair bits preserves the result. Computing the depth and caps with integer upper and lower approximations or using equivalent fixed rational slack changes neither estimates nor cost. The graph construction and $t_n$ message rounds cost $O(m+n\log\log n)$ word operations, plus at most $O(n\log n)$ bit draws and the random target ordering. On aborts return a feasible assignment; no exactness is asserted on such inputs.

Return to the actual fixed-size target, on the projection success event. Use the first fresh batch to estimate $A$ by the majority given the two unread-block estimates, the second to update all three blocks, and the third for final refinement, as in Theorem~\ref{thm:one-fiber}. The batches are conditionally uniform over the remaining grid, with independent target errors; the seed readings are independent of their errors. The majority estimates have block means tending to one on events of probability $1-o(1)$. The mixed-moment bound of Lemma~\ref{lem:direct-mixed-moment} applies uniformly on this pre-final calibration event, which is excluded only once. The root-star and incidence-tree argument and full-target reuse then give exact capped repair and the common restriction of Appendix~\ref{app:common-reduction}, for the same valid component cap. These stored stages have the word bound in Theorem~\ref{thm:geometric}. The reading count is $k+O(n/\log^2 n)\sim d_0 n/(\beta c\log n)$; the stated coefficient follows by taking $\beta$ close enough to one and $d_0$ close enough to $(\mu\rho)^{-2}$, with fixed slack. No assertion at equality is needed.
\end{proof}

\begin{proposition}[Below the projection threshold]
\label{prop:seeded-projection-converse}
Keep only the compressed projected pair record of Appendix~\ref{app:one-fiber}, with $P=\lfloor\beta m\rfloor$ and a deterministic first-fiber prefix of length $k$ such that $Pk/n^2\to d_0< (\mu\rho)^{-2}$. Also allow the true first fiber, its sensor errors, and true unread-fiber signs at any $o(n)$ addresses chosen independently of the projected graph and unread source signs. No estimator from this record has a nonvanishing probability of fixed positive absolute overlap with either unread block. The same holds with noisy rather than true additional seed signs. This is a converse for the compressed record, not for arbitrary access to the original target.
\end{proposition}
\begin{proof}
Use the iid coupling and condition on the first-fiber history and projected edge count $L$. When $d_0>0$, with probability $1-o(1)$, $L/n\to d_0$ and $\delta_n\to\mu\rho$. Thus $a=(L/n)\delta_n^2\le1-\kappa$ for a fixed $\kappa>0$ on a good event. When $d_0=0$ the same conclusion follows from $L/n=o_{\Pr}(1)$ and $|\delta_n|\le1$, including an empty prefix. Conditional on these data the graph is independent of the unread source spins and its edge channels are independent binary symmetric channels of bias $\delta_n$. The two shared centers can be handled by initially taking all $2n$ spins independent fair, including the centers in the revealed seed set $S$, and then conditioning both centers to equal the known first-fiber center. The conditioning costs a factor at most four in any nonnegative expectation bound. The other spins have exactly the original conditional law.

Apply the information-percolation inequality of \citet{abbe2020percolation} for symmetric channels in its chi-squared information form to the independent-spin experiment. For multiple projected observations at the same pair, introduce a fresh independent uniform auxiliary spin for each observation, put a noiseless product observation on the first half of its subdivided edge and a channel of bias $\delta_n$ on the second half. Their product is the original observation, and the extra noiseless signs are independent fair bits given the original record and source. The percolation graph after contracting the noiseless half has independent open edges of probability $\delta_n^2$, one per observation. To incorporate the known seed spins, introduce one further independent spin $W$ and noiseless products $WX_v$ for $v\in S$. Conditional on $W=+1$ these are the true seed signs. Posterior pair products are invariant under simultaneous sign reversal of all spins, so their posterior expectation does not change when $W$ is revealed. The squared posterior pair correlation is the conditional chi-squared information in the independent-spin experiment, by global sign symmetry before revealing $W$. Its expectation is bounded by the probability of open connection in this augmented graph. This connection is possible only through the ordinary projected graph or through the extra seed vertex.

For distinct fixed vertices $u,v$, the expected number of open simple paths of length $h\ge1$ between them in the ordinary projected graph is at most $n^{h-1}(L)_h n^{-2h}\delta_n^{2h}\le a^h/n$: the alternating intermediate vertices have at most $n$ choices each and distinct observation indices have $(L)_h$ choices. Summing gives $O_\kappa(1/n)$. For $u\notin S$, the probability of connection to the seed set is at most $|S|O_\kappa(1/n)$ by the same path count. For $u\in S$ use the bound one. The average augmented connection probability over ordered pairs in either unread block is thus $O_\kappa((1+|S|)/n)=o(1)$. For any sign estimator $\widehat X$, expand its squared overlap and bound every term by the expected absolute posterior pair correlation. Cauchy--Schwarz followed by Jensen bounds the mean squared overlap by the square root of the average connection probability. Revealing $W$ and conditioning the centers as above preserve the $o(1)$ bound. Randomized estimates can be conditioned on their independent randomness. Markov's inequality, the good first-fiber event, and the fixed-mask coupling complete the proof.
\end{proof}

The two results identify the strict threshold $d_0\mu^2\rho^2=1$ for the compressed graph with a vanishing fraction of independently located seeds. The upper result also solves the unchanged empirical objective after fresh refinement. The converse does not constrain a solver that uses the original triples before reaching positive overlap, and does not settle the critical point itself.

\subsection{Finite sparse-message comparison}
\label{sec:seeded-projection-experiment}
For a finite diagnostic, generate 16 independent targets at each $n=4096,16384$, with $q=.17$, $s=.10$, $m=\lceil2.5n\log n\rceil$. The fixed protocol projects the first $80\%$ of the randomly ordered target with a first-fiber prefix chosen for three nominal ratios $d_0(\mu\rho)^2$. In each of the other fibers mark each noncenter vertex with probability $1/\log^2 n$. Use all projected observations, six nonbacktracking rounds with sign rounding, one full-target first-block update, then eight full-target alternating majority sweeps. This finite variant uses cyclic walks, not the tree rounding and fresh-batch proof. Retained and independent replacement readings share addresses and sensor error draws, with identical operations. A cold start uses the same tie vector and eight sweeps without the projection or first-block update. A post-hoc stronger cold comparator runs eight independent starts (including the original start), eight sweeps each. It has a larger refinement budget, not matched wall time.

\begin{table}[ht]
\centering\small
\caption{Finite seeded projection: mean signed initial overlap of the two unread blocks and mean final empirical score divided by $m$ (16 targets per row). The reading fraction is total readings divided by $n$, not by $3n$. The nominal ratio is $d_0(\mu\rho)^2$.}
\label{tab:seeded-projection}
\begin{tabular}{rrrrrrrr}
\toprule
$n$ & Ratio & Reads/$n$ & \multicolumn{2}{c}{Initial overlap} & \multicolumn{3}{c}{Final score/$m$}\\
 & & & Retained & Replaced & Retained & Replaced & Cold\\
\midrule
4096 & .7 & .180 & .019 & .001 & .326 & .326 & .326\\
4096 & 1.3 & .309 & .147 & -.001 & .326 & .326 & .326\\
4096 & 2.2 & .503 & .550 & .001 & .659 & .325 & .326\\
16384 & .7 & .151 & .020 & .002 & .304 & .304 & .304\\
16384 & 1.3 & .261 & .140 & -.002 & .304 & .304 & .304\\
16384 & 2.2 & .428 & .514 & .001 & .660 & .304 & .304\\
\bottomrule
\end{tabular}
\end{table}
At ratio 2.2 the retained arm strictly improves the generating source's score in all 32 targets (mean gains 14.875 and 27.625 score units at the two sizes), beating both the replacement arm and eight-start cold comparator on every target. The eight-start comparator's mean scores divided by $m$ are .329 and .306, respectively. At ratios .7 and 1.3 no arm reaches the source score, despite the positive initial retained overlap at ratio 1.3. These results show a finite benefit of retention at reduced budgets and a gap between weak recovery and useful finite refinement. They do not verify the global optimum: the source can itself be suboptimal, and no large-instance exact reference optimum is used. All 96 target--budget records and 288 feasible scores are checked by direct parity evaluation; the 32 cold-comparator records preserve the eight individual scores. The scripts and seed are supplied with the code/data.

\section{Finite certificates for improving-flip radius}
\label{app:local-radius}
A large feasible score does not show that the empirical optimum has been
attained. A verifier can instead ask how many coordinates would have to
change simultaneously to improve that score. This question is independent
of the generating source. The following criterion applies to any signed
tripartite target, not just the random model.

For a candidate $x$ put $\eta_e=Y_eT_e(x)$, $d_v=\sum_{e\ni v}\eta_e$,
and let $x^S$ reverse the coordinates in $S$. Define the signed flip cost
\begin{equation}
 J_x(S)=\sum_{e\in H}\eta_e\mathbf1\{|e\cap S|\text{ odd}\},
 \qquad F_Y(x^S)-F_Y(x)=-2J_x(S).
 \label{eq:radius-cost}
\end{equation}
For an integer radius $R\ge1$, the claim to verify is $J_x(S)\ge0$ for
all $|S|\le R$. Define the positive pair codegrees and the integer
requirements
\begin{equation}
 w_{uv}=\#\{e:u,v\in e,\ \eta_e=1\},\qquad
 k_v=\max\{0,\lfloor d_v/2\rfloor+1\}.
 \label{eq:radius-capacities}
\end{equation}
For a finite list of nonnegative numbers $a$, write $B_j(a)$ for the sum
of the largest $j$ entries, padding with zeros when necessary. The
positive interaction graph contains precisely the pairs with $w_{uv}>0$.

\begin{proposition}[Vertex and pair elimination]
\label{prop:radius-peeling}
Start with all vertices and the positive interaction graph. In the
following rules, removed vertices and edges have zero capacity; all
other capacities retain their original values.
\begin{enumerate}
\item Delete $v$ if $B_{R-1}((w_{vu})_{u\ne v})<k_v$.
\item For $R\ge2$, delete a remaining edge $uv$ if
\begin{equation}
 2w_{uv}+B_{R-2}((w_{uz}+w_{vz})_{z\notin\{u,v\}})<k_u+k_v.
 \label{eq:radius-pair-test}
\end{equation}
\end{enumerate}
Apply the rules until neither changes the graph. It suffices to evaluate
$J_x(S)$ for all nonempty subsets of size at most $R$ contained in a
single remaining component, using the original signed observations.
If all evaluated costs are nonnegative and every component is checked,
the candidate is optimal within Hamming radius $R$. A negative cost is
an exact improving witness. A component or search cap yields no positive
claim unless every component is checked. In particular an empty remaining
vertex set certifies the radius without enumeration.
\end{proposition}
\begin{proof}
If some improving set has size at most $R$, choose an inclusion-minimal
one $S$ within it. For each $v\in S$, $J_x(S)<0\le J_x(S\setminus\{v\})$.
The difference of these costs is
\[
 d_v-2\sum_{e\ni v:\ |(e\setminus\{v\})\cap S|\text{ odd}}\eta_e<0.
\]
Dropping negative summands and counting each remaining incident edge
through at least one other selected vertex shows that
\begin{equation}
 \sum_{u\in S\setminus\{v\}}w_{vu}\ge k_v.\label{eq:radius-requirement}
\end{equation}
The positive part in the definition of $k_v$ is legitimate because the
left side is nonnegative. Initially every vertex and positive pair of
$S$ is present. Inductively the first rule cannot remove a vertex of
$S$, since at most $R-1$ other vertices contribute to its requirement.
If a surviving positive pair $uv$ lies in $S$, adding the two requirements
gives the upper capacity in \eqref{eq:radius-pair-test}, since at most
$R-2$ other vertices contribute. Thus the second rule cannot remove a
positive pair of $S$ either. Both arguments remain valid for simultaneous
deletions based on the current graph. At termination all vertices and
original positive pairs of $S$ survive.

Partition $S$ by its connected components in the remaining graph. Every
positive observation intersects at most one of these parts, because
its selected vertices would otherwise share a surviving positive pair.
For a negative observation the indicator of odd total intersection is
at most the sum of the odd-intersection indicators of the parts.
Multiplying by $-1$ reverses the inequality. Hence $J_x(S)$ is at least
the sum of the part costs. If $S$ improves, one part improves, and
minimality implies that $S$ lies in one component. The stated enumeration
therefore cannot miss every improving witness. Every witness is checked
in the original objective, so any negative cost is conclusive even when
other components are not searched. Finiteness follows from monotone
deletions. The case $R=1$ needs only the vertex rule and singleton costs.
\end{proof}

This is a deterministic certificate of a restricted optimization claim.
Taking $R=3n$ would verify global optimality, but the remaining components
may then be too large. The rule uses signs at the candidate, not its
distance from a latent source. A record of deletions and original
component costs can be checked using integer arithmetic. In the
implementation the vertex bound is computed as
$B_j(a)=\sum_{\ell\ge1}\min(j,\#\{i:a_i\ge\ell\})$ for integer capacities.
The pair rule uses the merged sorted neighbor lists. Each pass is finite;
no near-linear worst-case bound is asserted for repeated pair elimination.

\begin{corollary}[Local verification in the localized regime]
\label{cor:radius-localized}
Under the random target law of Theorem~\ref{thm:common-reduction}, fix
$J>1/\alpha$ with $J\ge2$, a fixed nonnegative $L$, and deterministic
integer radii $J-1\le R_n=o(\log n)$. There is an event of probability
$1-o(1)$ on which the following holds simultaneously for every candidate
$x$ having a source-gauge representative that differs from the source
only in $W_L^*$. Testing radius $R_n$ by the first vertex filter and
component enumeration decides whether $x$ is globally optimal on this
event. The test either certifies that radius or supplies a strict
improving witness. Its components contain fewer than $J$ vertices and
its work is $O((m+n)\log(m+n+2))$ with deterministic sorting and a
component-wise enumeration allowance of $2^J$ states.
\end{corollary}
\begin{proof}
Choose the fixed threshold $L_0$ from the uniform optimizer localization
in Appendix~\ref{app:common-reduction} at zero budget, and a fixed
$k_*>1/\alpha$. Impose the pair-codegree bound two and the geometry of
Lemma~\ref{lem:sublog-weak-geometry} for the deterministic threshold
\[
 L_n'=\max\{L,4L_0,4(R_n-1)+4k_*\}.
\]
Every vertex outside the candidate's error set has candidate field
within $4k_*$ of its source field by the same incident-pair count used
in Appendix~\ref{app:common-reduction}. The initial filter retains only
vertices with $d_v<4(R_n-1)$, since at most $R_n-1$ pair capacities
of at most two can contribute; any vertex with $d_v<0$ is also included
by this inequality. Retained vertices in the error set already lie in
$W_L^*$. Thus the initial retained set lies in $W_{L_n'}^*$ and every
component has fewer than $J$ vertices. Proposition~\ref{prop:radius-peeling}
needs only the first filter, not repeated elimination: its proof already
places every minimal improving set in one surviving positive component.
Sorting the observations and pair incidences and enumerating the constant-size
components give the work bound (each observation touches at most three
components). This argument is invariant under the whole-block gauge.

If $x$ is suboptimal, choose a global optimizer with a source-gauge
representative supported relative to the source in $W_{4L_0}^*$.
The difference between it and the localized candidate lies in
$W_{L_n'}^*$. Partition those changed coordinates into components of
the full observation interaction graph; the score increments add because
no observation meets two parts. One part strictly improves $x$ and has
size at most $J-1$. Conversely global optimality implies radius optimality.
The probability event does not depend on the particular candidate.
\end{proof}

The localized regime includes the estimates of Theorem~\ref{thm:recycling}
and their bounded-component repairs on the common probability event. The corollary distinguishes a finite,
source-independent radius certificate from a global conclusion conditional
on a high-probability structural event. The latter does not provide an
instance-wise certificate of that event, nor circumvent the global
proof-system barriers of Appendices~\ref{app:sos-certification} and
\ref{app:syndrome-barrier}.

\subsection{Retrospective verification and repair at large sizes}
\label{sec:radius-diagnostic}
Replay all 32 frozen targets of Appendix~\ref{app:seeded-projection} at
nominal ratio 2.2, keeping each of the retained, replaced, and cold
candidate scores unchanged before verification. Each score agrees with
the earlier record. Use vertex peeling at radii $1,2,4,8,16,32$ with
component cap 18 and a cap of 200,000 enumerated subsets per test.
All 32 retained answers pass radius one, but nine have exact
improving two-coordinate witnesses (six at $n=4096$, three at
$n=16384$). Every replacement and cold answer has a strict single-coordinate
improvement. The radius-two result is decisive: nine earlier high-score
answers are not global optima.

For a separate post-hoc repair of the retained candidates, apply at most
16 strict improving witnesses, testing radii $1,2,4$ in order on each
current candidate. Stop when none is found or a cap intervenes. The 15
accepted flips improve nine candidates by 40 score units in total (34 and
6 at the two sizes). No extra readings or latent signs enter this repair.
Table~\ref{tab:radius-results} reports the number of radius certificates
on the 32 repaired candidates. The pair rule is tested at radii $10,12,14$;
all unresolved counts in the table arise from caps, not counterexamples.

\begin{table}[ht]
\centering\small
\caption{Number passing the stated finite-radius test out of 16 repaired candidates at each size.
A dash denotes a radius not tested by that method. All noncertified
entries are unresolved; no improving witness is found in these tests.}
\label{tab:radius-results}
\begin{tabular}{rr|rrrrrr}
\toprule
$n$ & Method & 4 & 7 & 8 & 10 & 12 & 14\\
\midrule
4096 & Vertex & 16 & 16 & 0 & 0 & 0 & --\\
4096 & Vertex+pair & -- & -- & -- & 16 & 0 & 0\\
16384 & Vertex & 16 & 16 & 16 & 0 & 0 & --\\
16384 & Vertex+pair & -- & -- & -- & 16 & 16 & 14\\
\bottomrule
\end{tabular}
\end{table}

The joint test certifies a larger neighborhood than the vertex test.
At radius ten, all 32 answers are verified; in the 4096 cases the
remaining graph is empty. At radius fourteen two of the larger cases
are unresolved. Neither the repaired scores nor these radius tests
establish global optimality on the finite targets. An independent small
check uses 180 random signed tripartite multigraphs with arbitrary
candidate signs and $n=2,3,4$. Direct enumeration of all 280,320 flip
assignments checks both tests at every radius (3,240 comparisons): 196
positive certificate comparisons and 3,044 improving-witness comparisons
agree exactly. Independent parity evaluation also checks all 15 stored
repair witnesses and their strict gains. The scripts retain every cap
outcome and the original frozen denominator.

\section{Balanced acquisition on the original factor graph}
\label{app:balanced-acquisition}
Projection obtains information by first reading one fiber and discarding one
index of selected triples. Keeping all three indices gives a different
threshold. In this section the reading budget is distributed across the
three fibers. The proof uses posterior second moments on a tree rather
than a sign census. The subsequent fitting and reuse stages are unchanged.

\begin{lemma}[Combining independent binary channels]
\label{lem:binary-energy-study}
Let $X$ be uniform on $\{-1,1\}$. Observations $Z_i$ are independent
conditional on $X$; the marginal law of $M_i=\E[X\mid Z_i]$ is symmetric
about zero. Write $v_i=\E M_i^2$ and
$v=\E(\E[X\mid Z_1,\ldots,Z_l])^2$. Then
\[
 \sum_i v_i-2\sum_{i<j}v_iv_j\ \le v\le\ \sum_i v_i.
\]
In particular, for a binary symmetric return channel of bias $\rho$, one
reading at a coordinate with current posterior mean $m$ increases its
conditional expected squared posterior mean by
$\rho^2(1-m^2)^2/(1-\rho^2m^2)\le\rho^2$.
\end{lemma}
\begin{proof}
Let $m$ be the mean after the previous channels. With respect to the
unconditional law of a new observation $Z$, its conditional law given the
previous observations has likelihood ratio $1+mM$, where
$M=\E[X\mid Z]$. Bayes' rule gives the new mean
$m'=(m+M)/(1+mM)$. The posterior mean is a martingale. In the following calculation the expectation on the left is conditional on the past; the expectations on the right are under the symmetric unconditional marginal of $M$, after canceling one factor $1+mM$ against the likelihood ratio. By symmetry,
\begin{align*}
 \E[(m'-m)^2\mid m]
 &=(1-m^2)^2\E\frac{M^2}{1+mM}\\
 &=(1-m^2)^2\E\frac{M^2}{1-m^2M^2}.
\end{align*}
This lies between $(1-2m^2)\E M^2$ and $\E M^2$, since
$1-m^2\le 1-m^2M^2\le1$. The formulas extend to degenerate channels by
limits. Adding the upper bounds gives $v\le\sum_i v_i$. At the $i$th
addition, use the preceding upper bound on $\E m^2$ in the lower bound
and sum. For a binary symmetric channel, $M=\pm\rho$ with equal marginal
probabilities, giving the stated formula. \end{proof}

\begin{theorem}[Sharp total reading coefficient for local tree inference]
\label{thm:balanced-acquisition}
Fix the retained-source law with known rational biases $\mu=1-2q$,
$\rho=1-2s$ and $m=\lceil cn\log n\rceil$. For every fixed
$\epsilon>0$, there is an algorithm returning a feasible assignment on
all inputs, an empirical optimum and the common restriction of
Theorem~\ref{thm:common-reduction} with probability $1-o(1)$, and using at
most
\begin{equation}
 Q_n\le \frac{3/4+\epsilon}{c\mu^2\rho^2}\frac{n}{\log n}
 \label{eq:balanced-budget-study}
\end{equation}
for sufficiently large $n$, within $O((m+n)\log(n+2))$ word operations.
For the common restriction, fix $J>1/\alpha$, $J\ge2$, and any
deterministic budget $\Gamma_n=o(\log n)$ as in Theorem~\ref{thm:common-reduction}.
The reading addresses are independently marked, with equal marking
probability in the three reference fibers, subject to a deterministic cap.

Conversely, consider the first-stage record of $P\asymp n\log n$ triples,
with $d=P/n$ and independent Bernoulli reading marks of probabilities
$p_A,p_B,p_C$ in the three fibers. If
\begin{equation}
 \mu^2d\rho^2(p_A+p_B+p_C)\le 3/4,
 \label{eq:balanced-converse-study}
\end{equation}
no estimator using only each root's radius-$t$ factor neighborhood,
$t=O(\log\log n)$ (one radius unit is a variable-to-variable step), and its readings, with randomness independent of the data and no data-dependent global advice, has fixed positive absolute block overlap with nonvanishing
probability. This is a converse for local inference with independent
marks on the original factor graph, not a lower bound for unrestricted
computation or target-adaptive addresses.
\end{theorem}
\begin{proof}
Randomly order the target. Fix a constant $\beta\in(0,1)$ and take
$P=\lfloor\beta m\rfloor$, $d=P/n$, $\lambda=\mu^2d$.
If $P=0$ return an arbitrary feasible assignment. For the upper
bound choose a fixed rational $a>1/4$ and mark each vertex independently
with probability $p=a/(\lambda\rho^2)$ (a dyadic approximation of error
$O(n^{-3})$ is sufficient). Use $p=\min(1,p)$ for small $n$. Read each
marked reference-fiber sign once. Abort before reading if the number of
marks exceeds $3np+\delta n/\log n$ for any fixed positive slack
$\delta$; binomial concentration makes the abort probability $o(1)$.
The upper bound is proved for the uncapped experiment and the cap then
imposed. The readings at the three centers, if marked, are independent
sensor outputs even though their latent products coincide.

\paragraph{A recursion for information on a tree.}
First consider the three-type tree with independent $\operatorname{Pois}(d)$
child factors at each variable, one child variable of each of the two
other types at each factor, and independent source signs, observations,
marks and return errors. The binary observation at a factor has
likelihood $(1+\mu Y X_1X_2X_3)/2$. Conditional on the two child
subtree records, the observation and records together form a binary
channel for the parent. Its posterior mean under a uniform parent prior
is $M_f=\mu Y M_jM_k$, where $M_j,M_k$ are the posterior means of the
children. Squaring removes $Y$; independence of the child records gives
$\E M_f^2=\mu^2 x_jx_k$ if $x_j=\E M_j^2$ and $x_k=\E M_k^2$.
The marginal law is symmetric because under the uniform parent prior
$Y$ is fair conditional on the two child records. The marked unary
channel has energy $p_i\rho^2$; the channels at a variable are independent
conditional on its source sign. The degree is observed and independent
of the sign. Let $x_{i,t}$ be the posterior energy at type $i$ for a tree
of depth $t$, with $x_{i,0}=p_i\rho^2$. Lemma~\ref{lem:binary-energy-study}
and the Poisson first and second factorial moments give
\begin{align*}
 s_{i,t}&=p_i\rho^2+\lambda x_{j,t}x_{k,t},\\
 s_{i,t}-s_{i,t}^2&\le x_{i,t+1}\le s_{i,t}.
\end{align*}
Indeed the expected pair correction is
$2p_i\rho^2d\mu^2x_{j,t}x_{k,t}+(d\mu^2x_{j,t}x_{k,t})^2\le s_{i,t}^2$.
If $s_{i,t}>1/4$ while $p_i\rho^2\le1/4$, independently thin the
child factors to make the total expected energy $1/4$. The thinned
record is less informative (conditional Jensen); its lower bound is
$3/16$. Hence in both cases
\begin{equation}
 x_{i,t+1}\ge\tfrac34\min(s_{i,t},1/4).
 \label{eq:balanced-energy-lower-study}
\end{equation}
The tree calculation is exact; no Gaussian approximation is used.

With equal marks let $x_t=x_{i,t}$ and $z_t=\lambda x_t$. For every fixed
$t$, as $n\to\infty$, the preceding bounds give convergence to the
iteration
\begin{equation}
 u_0=a,\qquad u_{t+1}=a+u_t^2.
 \label{eq:balanced-recursion-study}
\end{equation}
This follows inductively since for bounded $z_t$ the scaled lower bound
is $a+z_t^2-(a+z_t^2)^2/\lambda+o(1)$. If $a>1/4$, the iteration is
increasing and unbounded: a bounded limit would solve $u=a+u^2$, which
has no real root. Thus at a fixed depth $T$, $z_T\ge4$ for large $n$.
Equation~\eqref{eq:balanced-energy-lower-study} subsequently implies
$z_{t+1}\ge(3/4)\min(z_t^2,\lambda/4)$. Before reaching
$z_t^2\ge\lambda/4$, this iteration at least doubles, and in fact squares
$(3/4)z_t$ at each step. It reaches that threshold in $O(\log\log\lambda)$
steps. The next step gives $x_t\ge3/16$. Posterior energy is nondecreasing
with depth by conditional Jensen, so depth
$t_n=\lceil\log\log(n+e^e)\rceil$ suffices eventually.

For the converse put $a_i=\lambda p_i\rho^2$,
$z_{i,t}=\lambda x_{i,t}$, $A=\sum_i a_i$, and $S_t=\sum_i z_{i,t}$.
The upper energy bound gives, without requiring equal marks,
\[
 S_{t+1}\le A+z_{A,t}z_{B,t}+z_{B,t}z_{C,t}+z_{C,t}z_{A,t}
 \le A+S_t^2/3.
\]
If $A\le3/4$, the scalar map has a bounded least fixed point
$S_-=(3/2)(1-\sqrt{1-4A/3})$. Since $S_0=A\le S_-$, induction gives
$S_t\le S_-$ at all depths. Each posterior energy is therefore
$O(1/\lambda)$, and the expected absolute posterior mean is at most
$O(\lambda^{-1/2})$. This upper bound holds for any rootwise estimator
based on the same tree observations, since the conditional mean maximizes
expected signed correlation among sign estimates. The tree threshold
$3/4$ is sharp across all allocations of independent marks: the symmetric
allocation realizes it from above, while the sum bound excludes every
allocation at or below it.

\paragraph{Transfer from trees to the uniform mask.}
In an iid-location experiment, the first $P$ triples agree with the
uniform mask by conditioning on no repeated grid address; the collision
probability is $O(P^2/n^3)=o(1)$. Poissonize the number of triples with
mean $P$. Counts at grid addresses are independent Poisson variables of
mean $d/n^2$. Explore the neighborhoods of one or two specified distinct
vertices to depth $t=O(\log\log n)$. The expected size on the comparison
tree is at most $2\sum_{j=0}^t(2d)^j=n^{o(1)}$. Stop if more than
$M=n^{1/20}$ vertices or factors have been encountered; this has
probability $o(1)$ on the trees. At a new variable, the counts whose other
two addresses are fresh are independent Poisson counts with total mean
$d(1-O(M/n))$. Counts meeting an already exposed vertex, or repeated
children among new endpoints, produce total coupling error
$O(d M^2/n+M^2/n)=o(1)$ before the stop: for any prescribed exposed
opposite vertex the mean number of incident edges is $d/n$, and the
new endpoints are uniform within the unused blocks. Previously explored
counts are omitted, except for the parent factor. This couples both
neighborhoods to independent three-type trees; their marks and target and
return errors agree. Equivalently, sample fresh uniform child indices on
the comparison trees and exclude collisions and the three centers. The
probability of encountering a center is $O(M/n)$ for noncenter roots.
The source signs away from the centers are independent uniform under the
reference-fiber law. Alternatively, for the upper bound the entire
message distribution in source coordinates is the same for each fixed
source, by sign equivariance.

De-Poissonize using a common iid sequence and an independent count
$N\sim\operatorname{Pois}(P)$. With probability $1-o(1)$,
$|N-P|\le n^{2/3}$. Conditional on the count, in the smaller graph the
expected number of vertices within $t$ steps of one or two roots is
bounded by the simple-path count $2\sum_{j\le t}(2(d+o(1)))^j=n^{o(1)}$.
Markov's inequality gives the same size cap with probability $1-o(1)$.
The extra iid edges hit this neighborhood with probability at most
$3Mn^{2/3}/n=o(1)$. Thus the one- and two-root couplings hold also at
fixed $P$ and for the original mask. All errors are uniform for distinct
noncenter roots. Source-independent indices in the local observations
may be retained: the comparison trees can carry iid uniform indices,
with collision and center encounters excluded at the same cost.

On the graph compute the depth-$t_n$ sum-product messages. On a tree,
these are the exact posterior means \citep{kschischang2001factor}.
Round each final mean $M_v$ to a sign with conditional mean $M_v$ using
independent bits. Its mean correlation on the tree is
$\E[X_vM_v]=\E M_v^2\ge3/16$. The one- and two-root couplings give mean
at least $3/16-o(1)$ and covariance $o(1)$ for the bounded source-signed
outputs. Chebyshev's inequality gives all three block means bounded
below by a positive constant with probability $1-o(1)$. The center
coordinates affect each block mean by at most $O(1/n)$.
For the converse, conditional on any independent estimator randomness,
the two-root coupling bounds the expected product of the source-signed
outputs at distinct noncenter vertices in the same block by
$O(1/\lambda)+o(1)$, since on independent trees their means have absolute
values at most $\sqrt{x_{i,t}}$. The diagonal and centers add $O(1/n)$.
Thus the mean squared block overlap is $o(1)$, proving the asserted
converse by Markov's inequality. This argument concerns outputs computed
from local records, not estimators that exchange information outside the
specified neighborhoods.

\paragraph{Work and exact fitting.}
For a variable-to-factor message, combine the unary mean $r_v=\rho R_v$
($0$ at an unmarked vertex) with incoming factor means
$u_{e\to v}=\mu Y_e\prod_{w\in e\setminus v}M_{w\to e}$ excluding the
recipient. The combined mean is $(A-B)/(A+B)$ with
$A=(1+r_v)\prod(1+u_{e\to v})$ and
$B=(1-r_v)\prod(1-u_{e\to v})$. The final vertex mean includes all
factors. This is the same update as adding half log-likelihood ratios.
All factors in the products are positive and bounded away from zero by
constants depending on $\mu,\rho$. Cap the maximum variable degree at
$\lceil\log^2(n+2)\rceil$, an $o(1)$ abort event by binomial tail bounds
and the mask coupling. Compute products using a binary mantissa with
$O(\log n)$ bits and an integer exponent; each multiplication and
exclusion division has relative error at most $O(n^{-4})$, so all
products have relative error $O(\log^2 n\,n^{-4})$. This uses a constant
number of $O(\log n)$-bit words per number; exponents are $O(\log^2 n)$.
Normalization changes the mean by at most a constant times that relative
error. In its equivalent hyperbolic-tangent form, each combined mean is
Lipschitz in the incoming means with total constant $O(\log^2 n)$, since
$|u_{e\to v}|\le\mu<1$. Factor means have bounded Lipschitz constants.
Clipping numerical means to $[-1,1]$ cannot enlarge the error. Over
$t_n$ rounds the amplification is at most $(O(\log^2 n))^{O(\log\log n)}=n^{o(1)}$, so the accumulated error is $n^{-4+o(1)}=o(1)$.
Products including all neighbors can be computed once and divided by
one positive factor for each cavity message. Thus each round takes
$O(P+n)$ word operations; the degrees, marks, rounding and random target
ordering fit the $O((m+n)\log(n+2))$ bound. On any abort return a feasible
assignment. The cap and rounding change the success probability by $o(1)$.

Now split the remaining target into three batches as in the proof of
Theorem~\ref{thm:one-fiber}. Conditional on the first-stage record and
readings, the remaining locations are uniform on the remaining grid,
with independent errors. The first two fresh majority batches bring
all three means to $1-o(1)$ in probability. The last reserved batch,
Lemma~\ref{lem:direct-mixed-moment}, full-target reuse and capped repair
then give exact fitting and the observable common restriction, exactly
as in Theorem~\ref{thm:one-fiber}. No independence of the first-stage
coordinate errors is required. The reading cap is
$3a n/(\beta c\mu^2\rho^2\log n)+\delta n/\log n+o(n/\log n)$.
Choose $a$ above $1/4$, $\beta$ below one, and $\delta$ positive with
sufficiently small fixed slacks to obtain \eqref{eq:balanced-budget-study}.
\end{proof}

The threshold above concerns the original third-order record, unlike the
single-fiber projection threshold in Appendix~\ref{app:seeded-projection}.
The reduction in its leading coefficient comes from reading all three
fibers: in the weak-information limit each unread variable gets a product
of the information in the two other blocks. The converse does not
establish that no global algorithm can improve on this coefficient.

\paragraph{A bound on repeated return readings.}
In the auxiliary experiment of independent fair source bits, allow
addresses to depend on earlier return readings and independent randomness,
but not on target labels. Let $V_Q$ be the sum of squared posterior means
after at most $Q$ readings. The posterior still factorizes given the
history: the conditional likelihood multiplies by the likelihood at the
chosen coordinate, and the address rule itself reveals no new information.
Lemma~\ref{lem:binary-energy-study} bounds the conditional drift of $V$ by
$\rho^2$. Each increment lies in an interval of length one conditional on
the history. Padding a stopped procedure with zero increments and using
the conditional Hoeffding bound gives
\[
 \E V_Q\le Q\rho^2,\qquad
 \Pr\{V_Q>Q\rho^2+u\}\le \exp(-2u^2/Q)\quad(Q>0).
\]
For $Q\le n$, reading distinct bits achieves $V_Q=Q\rho^2$ exactly in a
single fiber. This is a bound on the information delivered by a
return-only policy, not on readings chosen using the target.

\subsection{Finite soft amplification with an independent batch}
\label{app:soft-amplification-study}
The effect of the first inference stage depends on the confidences it
passes to the next stage, not only on its rounded overlap. Condition on the source, a
history independent of the remaining target errors, and bounded weights $u_j,v_k\in[-1,1]$ in the two opposite blocks.
For a given vertex, let $\mathcal P$ be the remaining available opposite
pairs and let a fresh batch contain $l$ pairs sampled uniformly without
replacement from $\mathcal P$. The target errors on these pairs are
independent and fresh. Put
\[
 a=\frac{\mu}{|\mathcal P|}\sum_{(j,k)\in\mathcal P} B_jC_k u_jv_k,
 \qquad b=\frac1{|\mathcal P|}\sum_{(j,k)\in\mathcal P}u_j^2v_k^2.
\]
For $a>0$, the probability that the weighted field is nonpositive in
source coordinates is at most
\begin{equation}
 \exp\{-l[(a+b)\log(1+a/b)-a]\}.
 \label{eq:soft-fresh-bound-study}
\end{equation}
To see this, for a signed vote $Z=\eta B_jC_k u_jv_k$, $|Z|\le1$,
$e^{-h Z}\le1-h Z+(e^h-1-h)Z^2$ for $h\ge0$ by the power series.
Independence of the target errors and Maclaurin's inequality for the
positive moment factors from the available pairs bound the moment of
the sum by $\exp(l[-ha+b(e^h-1-h)])$. Set $h=\log(1+a/b)$ and use
Markov's inequality. This includes ties as errors and conditions on the
available grid, not on a fictitious independent full grid.

On the full opposite grid, $a^2/b$ factors as
$\mu^2(\overline{B u})^2/\overline{u^2}\cdot
(\overline{C v})^2/\overline{v^2}$. This is the leading weak-signal
quantity in the bound. If a feature $Z$ is available for a randomly
chosen coordinate with source sign $X$, then among weights $w(Z)$
with nonzero second moment,
\[
 \frac{(\E[Xw(Z)])^2}{\E[w(Z)^2]}
 \le \E[\E(X\mid Z)^2].
\]
The proof is conditional expectation followed by Cauchy--Schwarz;
posterior-mean weights attain equality. The best sign-only weights give
$(\E|\E(X\mid Z)|)^2$, no larger and strictly smaller for nonconstant
posterior magnitudes (with nonzero posterior energy). Thus calibrated confidences maximize this
second-moment transfer quantity on a fresh full grid. This statement is
not a claim that loopy messages on the reused target are calibrated.

\subsection{Finite full-record and projected soft messages}
\label{app:soft-experiment-study}
The finite algorithm differs from the split-sample proof. It reuses the entire
original target in 24 damped sum-product rounds (new-message weight $0.8$),
rounds the means to signs, and performs eight sequential block-majority sweeps,
retaining the old sign on ties in each sweep. The projected initialization uses
24 additional pairwise message rounds on triples touching the read prefix of
one fiber. It polishes the projected means directly and, separately, runs full-record messages initialized at those same projected means before polishing. It selects the better of these two scores. The direct-only ablation instead initializes full-record messages at the unary reading priors, not at the projected means. This third candidate is not included in the two-candidate maximum and can score higher despite using fewer message rounds. Neither method uses the source to select an answer.
These fixed-round algorithms are diagnostics, not the capped exact algorithms
of the theorems.

The retrospective comparison uses the 32 frozen targets of
Appendix~\ref{app:seeded-projection}, at projected ratio $1.3$. The new method
strictly improves on the source in all 32 cases, whereas the previous method
at that budget did not reach its score. The new method uses more message work;
this is not an equal-work comparison with the earlier solver. The replacement
and cold arms of the new projected method use the same message and sweep caps.

The fresh fixture uses $m=\lceil 2.5n\log n\rceil$, $q=.17$, $s=.10$,
with eight independent targets at each of $n=4096,16384$. The distinct triple
mask is uniformly sampled; source and replacement reference fibers, target
errors, and reading noise are generated independently (the same return-error
array is used in the two paired arms). Prefix reading uses
$k=\lceil r n^2/(\mu^2\rho^2 P)\rceil\wedge n$ with
$P=\lfloor .8m\rfloor$, together with independent marks of probability
$1/\log^2 n$ in each of the other two fibers, excluding their centers.
Here $r$ labels the nominal projected ratio even for the direct method, which
uses no projected edges. The balanced method instead marks every fiber with
probability $\min(1,A/(3\mu^2\rho^2(m/n)))$, nesting budgets using the same
uniform draws. Actual read counts, including marks, are recorded. The same 16
fresh targets are reused across the three diagnostics and budgets; they are
not 144 independent instances. In the direct and balanced methods, replacement
and cold arms use the identical addresses, 24 rounds and eight sweeps, with
replacement-source readings or independent fair signs, respectively. In the
projected diagnostic the cold arm uses fair signs in the projection and the
same sparse unary seed mask; its full-record unary prior is zero. That
comparison controls message caps, not all prior choices or actual runtimes.
No exact-reference optimum is available at these sizes.

\begin{center}
\begin{minipage}{\linewidth}
\centering\small
Frozen projected comparison. Each arm entry gives mean $F/m$ and strict improvements over the source.\par\medskip
\begin{tabular}{rrrlllrr}\toprule
$n$&$r$&$N$&Retained&Replacement&Cold&Win R&Win C\\\midrule
4096&1.3&16&0.6590 (16)&0.3365 (0)&0.3389 (0)&16&16\\
16384&1.3&16&0.6602 (16)&0.3151 (0)&0.3147 (0)&16&16\\
\bottomrule\end{tabular}\end{minipage}\end{center}
\begin{center}
\begin{minipage}{\linewidth}
\centering\small
Fresh projected initialization (the separate unary-initialized retained ablation is in the last column). Each arm entry gives mean $F/m$ and strict improvements over the source.\par\medskip
\begin{tabular}{rrrrlllrrr}\toprule
$n$&$r$&$N$&$Q/n$&Retained&Replacement&Cold&Win R&Win C&Direct\\\midrule
4096&0.7&8&0.178&0.4150 (2)&0.3344 (0)&0.3460 (0)&4&2&7\\
4096&1.0&8&0.243&0.6205 (7)&0.3350 (0)&0.3375 (0)&8&8&8\\
4096&1.3&8&0.308&0.6605 (8)&0.3385 (0)&0.3374 (0)&8&8&8\\
16384&0.7&8&0.150&0.3558 (1)&0.3120 (0)&0.3230 (0)&6&1&8\\
16384&1.0&8&0.206&0.6609 (8)&0.3139 (0)&0.3135 (0)&8&8&8\\
16384&1.3&8&0.261&0.6609 (8)&0.3148 (0)&0.3142 (0)&8&8&8\\
\bottomrule\end{tabular}\end{minipage}\end{center}
\begin{center}
\begin{minipage}{\linewidth}
\centering\small
Fresh direct full-record messages. Each arm entry gives mean $F/m$ and strict improvements over the source.\par\medskip
\begin{tabular}{rrrrlllrr}\toprule
$n$&$r$&$N$&$Q/n$&Retained&Replacement&Cold&Win R&Win C\\\midrule
4096&0.3&8&0.092&0.3372 (0)&0.3387 (0)&0.3390 (0)&3&3\\
4096&0.5&8&0.135&0.3462 (0)&0.3456 (0)&0.3440 (0)&3&5\\
4096&0.7&8&0.178&0.6209 (7)&0.3552 (0)&0.3551 (0)&7&7\\
16384&0.3&8&0.076&0.3148 (0)&0.3152 (0)&0.3152 (0)&3&3\\
16384&0.5&8&0.113&0.3211 (0)&0.3206 (0)&0.3203 (0)&5&5\\
16384&0.7&8&0.150&0.6609 (8)&0.3313 (0)&0.3318 (0)&8&8\\
\bottomrule\end{tabular}\end{minipage}\end{center}
\begin{center}
\begin{minipage}{\linewidth}
\centering\small
Fresh balanced marks on the full record. Each arm entry gives mean $F/m$ and strict improvements over the source.\par\medskip
\begin{tabular}{rrrrlllrr}\toprule
$n$&$A$&$N$&$Q/n$&Retained&Replacement&Cold&Win R&Win C\\\midrule
4096&0.6&8&0.106&0.3419 (0)&0.3424 (0)&0.3412 (0)&5&4\\
4096&0.9&8&0.159&0.4719 (3)&0.3553 (0)&0.3611 (0)&7&4\\
4096&1.2&8&0.211&0.6605 (8)&0.3289 (0)&0.3259 (0)&8&8\\
16384&0.6&8&0.088&0.3176 (0)&0.3177 (0)&0.3177 (0)&4&5\\
16384&0.9&8&0.132&0.6609 (8)&0.3345 (0)&0.3340 (0)&8&8\\
16384&1.2&8&0.176&0.6609 (8)&0.3055 (0)&0.3055 (0)&8&8\\
\bottomrule\end{tabular}\end{minipage}\end{center}

Win R and Win C count strict wins over the replacement and cold arms. The
projected ablation's last column counts strict improvements by direct retained
messages. All trials are included, including failure to improve the source.
At the direct ratio $0.7$, radius-two verification on the 16 retained outputs
finds two improving witnesses at $n=4096$; the remaining 14 pass radius-two screening. Thus even improvement over the source does not certify global
optimality. The balanced comparisons improve over the source in 11/16 at
$A=.9$ and 16/16 at $A=1.2$, with 0/16 in either control at each budget.
The finite transition is not an estimate of the asymptotic sharp coefficient.

The independent audit reconstructs 480 feasible assignment scores (115,836,720
observation parities) from the packed fresh witnesses, checks 273 exact
return-channel drift identities and 256 tree marginals, and checks 485
fresh-batch bounds by enumeration of 230,040 subset/noise terms. The frozen
projected solver also checks selected scores by direct parity evaluation.
The supplementary scripts retain all objective traces and fresh witnesses. In the 48 fresh target--budget cases the unary-initialized retained ablation ties the projected method in 34 and improves on it in 14. The projected method never scores higher in this comparison. The selection audit checks the maximum of the two projected-branch scores in each of the 144 arm--budget cases and reconstructs 192 selected witness scores, including the direct ablation, from the original labels. The 16 overlapping direct-only runs at ratio $0.7$ match the ablation traces and packed witnesses exactly.

\section{Global optimality of high-scoring local optima}\label{app:score-basin}
The radius certificate of Appendix~\ref{app:local-radius} can be tested on the observed objective, but the hypothesis of Corollary~\ref{cor:radius-localized} refers to the unknown source. A score screen removes that hypothesis. We prove that every one-flip-stable assignment of sufficiently high observed score lies in the same intrinsic weak set as the global optima. The statement is uniform over assignments, including those produced using the entire target and adaptive readings.

\begin{theorem}[Observable score and local optimality]\label{thm:score-basin}
Under the tripartite parity target law, let $\mu=1-2q$, $h=1-2\sqrt{q(1-q)}$, $\alpha=ch$, and fix an integer $J\ge2$ with $J\alpha>1$. There is a constant $\delta\in(0,1/2)$ depending only on $c,q,J$ such that, with probability $1-o(1)$, the following statements hold simultaneously for all assignments $x$ with
\begin{equation}\label{eq:score-screen28}
 F_Y(x)\ge \mu(1-\delta/3)m.
\end{equation}
If $x$ is one-flip stable, it has a source-gauge representative supported on $W_{4(J-2)}^*=\{v:D_v\le4(J-2)\}$ relative to the source. Every component of this weak set has fewer than $J$ vertices. Consequently $x$ is globally optimal if and only if it admits no strict improvement by at most $J-1$ flips. For $\alpha>1/2$ one may take $J=2$: on the same event, every one-flip-stable assignment passing \eqref{eq:score-screen28} is globally optimal.
\end{theorem}
The score screen and radius test require no latent signs. The score threshold assumes known $q$. This is a model-based probability guarantee for a false acceptance anywhere among the assignments, not an instance-wise proof that the random structural event holds. It adds no readings to the acquisition procedure.

\begin{proof}
Put $\lambda=\frac12\log((1-q)/q)$ and define
\[
 M=\max\{1,(1-q)e^{2\lambda}+qe^{-2\lambda},
                (1-q)e^{-3\lambda}+qe^{3\lambda}\},\qquad
 k(\delta)=h(1-\delta)-(M-1)\delta/3.
\]
Choose $\delta\in(0,1/2)$ so that $cJk(\delta)>1$. Such a choice exists by continuity at zero. All probability estimates below are for the fixed-size mask, conditional on any source tensor.

\paragraph{A score screen implies source proximity.}
Let $d$ be the fraction of the complete grid on which a fixed tensor differs from the source. If its smallest source-gauge flip set has size at least $\delta n$, then $d\ge\delta/3$. Indeed, $1-2d$ is the product of the three source correlations; if $d<\delta/3$, their absolute values exceed $1-2\delta/3$ and an even number of whole-block sign changes makes all three positive. The sum of the three flipped fractions is then less than $\delta$.
For any positive $u$, the exponential moment of the score of the fixed tensor is at most
\[
 \E e^{uF_Y(x)}\le[\cosh u+\mu(1-2d)\sinh u]^m.
\]
Integrate the independent signs and apply Maclaurin's inequality to the uniform distinct grid locations. The bracket is the moment of a sign of mean $\mu(1-2d)$. The usual bounded-variable exponential inequality gives
\[
 \Pr\{F_Y(x)\ge\mu(1-\delta/3)m\}
 \le\exp(-m\mu^2\delta^2/18)
 \quad(d\ge\delta/3).
\]
Union over at most $2^{3n}$ factor assignments. Since $m\asymp n\log n$, the probability that any distant tensor passes the screen tends to zero. This bound does not condition on the candidate or its acquisition history.

\paragraph{A stable flip set cannot contain a large component.}
Use a source-gauge representative with flip set $S$ of size less than $\delta n$. For a connected component $A$ of the interaction graph induced by $S$, the stability of every flipped vertex implies
\begin{equation}\label{eq:stable-sum28}
 V(A):=\sum_{e\in H}|e\cap A|(-1)^{|e\cap A|}\eta_e\ge0.
\end{equation}
Each summand is the contribution to the sum of final fields over $A$. No edge meets two components of $S$, so replacing $S$ by $A$ leaves every field in that sum unchanged. The expression for $V(A)$ differs from the flip cost of Appendix~\ref{app:local}; stability rather than comparison with the source is the hypothesis here.

For a fixed set $A$ of $t\le\delta n$ vertices, write $a_i=|A\cap V_i|/n$. The fraction of grid triples having exactly one endpoint in $A$ is at least $(1-\delta)t/n$, since $(1-a_j)(1-a_k)\ge1-a_j-a_k$. The fraction having at least two endpoints is at most $\sum_{i<j}a_i a_j\le(t/n)^2/3\le\delta t/(3n)$. The moment at $\lambda$ for one endpoint is $1-h$; the moments for two and three endpoints are bounded by $M$. Thus the average exponential moment of a single grid location is at most $1-k(\delta)t/n$.

Write $N=n^3$, $p=m/N$ and fix $s$ distinct required triples $E$. Their inclusion has probability at most $p^s$. The product of their moments is at most $M^s$. Conditional on their inclusion, the remaining mean moment is at most $N(1-k(\delta)t/n)/(N-s)$. For $s\le m$, Maclaurin's inequality gives
\[
 \E[\one\{E\subseteq H\}e^{\lambda V(A)}]
 \le p^s M^s\exp[-k(\delta)mt/n+(1+k(\delta)\delta)s].
\]
Here $(m-s)\log(N/(N-s))\le s$ and $t/n\le\delta$; the cases $s>m$ have probability zero. For a connected $A$ the incidence-tree enumeration in Appendix~\ref{app:local} has at most $4^{t+s}$ shapes, $3n$ root labels, and $(6n^2)^s$ label completions, with $1\le s\le t-1$. With $B=24M e^{1+k(\delta)\delta}m/n$, summing the resulting probability bounds and using $t\le2^t$ yields
\begin{equation}\label{eq:stable-bound28}
 \Pr\{\exists A\text{ connected}, |A|=t, V(A)\ge0\}
 \le 3n[8\max(1,B)]^t e^{-k(\delta)mt/n}.
\end{equation}
Because $cJk(\delta)>1$, summing this bound from $J$ to $\lfloor\delta n\rfloor$ is $o(1)$. The bound applies to all connected sets, not to the distribution of the output of a particular solver.

\paragraph{From stability to exactness.}
Impose the event of pair codegree at most two and the weak-component bound of Lemma~\ref{lem:sublog-weak-geometry} for the fixed cutoff $4(J-2)$. Both hold with probability $1-o(1)$ since $J>1/\alpha$. On the event of the preceding two paragraphs, each component of the one-flip-stable candidate's source-gauge support has fewer than $J$ vertices. At a flipped vertex $v$, its final field is $-D_v+2\sum_e\eta_e$, where the correction sums incident observations with an odd number of other flipped endpoints. At most $2(J-2)$ observations contribute. Nonnegativity of the final field therefore implies $D_v\le4(J-2)$. The whole support lies in $W_{4(J-2)}^*$.

If a candidate passing the screen is suboptimal, a global maximizer also passes it and is one-flip stable. The two source-gauge representatives differ only in the same weak set. Partition their difference into components in the observation interaction graph. The objective gains add over components, each of size at most $J-1$; one is strictly positive. Gauge invariance transfers that improving flip to the candidate's original representative. Conversely no strict improvement is possible at an optimum. This proves the equivalence.
\end{proof}

\paragraph{Computational and finite probability scope.}
For a fixed $R\ge J-1$ with $R=o(\log n)$, test one-flip stability, the score screen and the radius-$R$ condition. On the event above and the additional weak-neighbor event of Corollary~\ref{cor:radius-localized}, the first vertex filter of Proposition~\ref{prop:radius-peeling} reduces the radius test to components of fewer than $J$ vertices. Sorting and bounded enumeration require $O((m+n)\log(m+n+2))$ word operations. One may cap enumeration at $J-1$ vertices per component and return unresolved on cap failure; this makes the operation bound valid on every input. A separate direct test at radius two uses signed pair sums: once all fields are nonnegative, the cost of a pair is $d_u+d_v-2\sum_{e\supseteq\{u,v\}}\eta_e(x)$, and pairs with no observed edge cannot strictly improve.

For clarity, an explicit upper bound on the probability of the structural failure in Theorem~\ref{thm:score-basin} is the minimum of one and the sum of four terms. The first two are $2^{3n}e^{-m\mu^2\delta^2/18}$ and the sum of \eqref{eq:stable-bound28} over $J\le t\le\lfloor\delta n\rfloor$. The codegree term is $3n^2\binom n3p^3$. For the fourth, put $L=4(J-2)$, $M_w=\max(1,\E e^{-2\lambda\eta},\E e^{-3\lambda\eta})$, $k_w=h(1-J/n)-(M_w-1)J/(3n)$, and $B_w=24M_w e^{1+k_w J/n}m/n$. For sufficiently large $n$ that $J/n<1/2$ and $k_w>0$, the fourth term is
\[
 3n[8\max(1,B_w)]^J\exp[-k_w mJ/n+\lambda LJ].
\]
To obtain it, a weak component of size at least $J$ contains a connected $J$-set with $\sum_{v\in A}D_v\le LJ$. Its exponential moment for the negative sum has single-endpoint moment $1-h$ and other moments bounded by $M_w$. Apply the same conditional mask and witness count as above. All four terms tend to zero at the stated fixed density. At the finite sizes of the following experiment this bound is vacuous; it does not certify the realized instances. It controls the unconditional probability of false acceptance, not the probability of error conditioned on acceptance.

\subsection{An observable screening diagnostic on frozen large candidates}\label{app:score-basin-checks}
Use the 48 balanced and 48 direct target--budget records of Appendix~\ref{app:soft-experiment-study}; both protocols use the same 16 targets. For $q=.17,c=2.5,J=2$, choose $\delta=.02$, which gives $k(\delta)>.223$ and $cJk(\delta)>1$. The score threshold is $.6556m$. Decode all 288 stored factor witnesses and independently recompute their scores. Apply the screen before any repair or source-distance evaluation. For accepted candidates, repeatedly test radius two, applying a strict improving witness if found, with a cap of 32 moves. The exact radius-two test above uses only signed fields and signed pair incidences.
\begin{table}[ht]
\centering\small
\caption{Score screen and radius verification on frozen records. Each arm has 48 records per protocol; no new targets are introduced. The distance is to the nearest source-gauge representative, evaluated only for the diagnostic.}
\begin{tabular}{llrrrr}
\toprule
Protocol & Arm & Screen passes & Radius two passes & Repaired & Max. distance/$n$\\
\midrule
Balanced & Retained &27&27&0&.003662\\
Balanced & Replacement &0&0&0&--\\
Balanced & Cold &0&0&0&--\\
Direct & Retained &15&15&1&.003906\\
Direct & Replacement &0&0&0&--\\
Direct & Cold &0&0&0&--\\
\bottomrule
\end{tabular}
\label{tab:score-basin28}
\end{table}
All 42 screen passes lie within the declared source-gauge distance before and after repair. The direct protocol has one strict repair of gain two. All final accepted candidates pass the exact radius-two test. These outcomes show that the screen distinguishes the high-score region on these records without latent signs. The finite observations do not imply global exactness, since the probability bound above is vacuous here. The tests are retrospective and the 288 records are not independent samples.

\subsection{The sharp radius: isolated weak paths}\label{app:radius-sharp28}
The preceding theorem gives a sufficient radius. A matching lower bound follows from an explicit local configuration that is invisible to every smaller-radius improvement test. It concerns the discrete empirical objective, where the source itself may be suboptimal. For comparison, \citet{mcrae2025benign} study second-order critical points of a continuous low-rank relaxation for pairwise synchronization near exact recovery. Here the stationary points and the required radius are defined directly on the original Boolean assignments.

\begin{theorem}[Sharp radius for high-scoring local optima]\label{thm:sharp-radius28}
Fix an integer $R\ge1$ and any constant $\varepsilon>0$. If $\alpha(R+1)<1$, with probability $1-o(1)$ there is an assignment $x$ such that
\[
 F_Y(x)\ge (\mu-\varepsilon)m,\qquad
 \OPT-F_Y(x)=2,
\]
and every flip of at most $R$ coordinates is nonimproving. The same conclusion holds if $\alpha(R+1)=1$ and $R\ge2$. At $R=1,\alpha=1/2$ the probability of existence is bounded away from zero. Together with Theorem~\ref{thm:score-basin}, the least fixed positive radius sufficient with high probability for the uniform local-to-global implication at some constant score cutoff below $\mu m$ is
\begin{equation}\label{eq:sharp-radius28}
 R_*(\alpha)=\max\{1,\lfloor1/\alpha\rfloor\}.
\end{equation}
\end{theorem}

\begin{proof}
Theorem~\ref{thm:order-radius} covers $k=3$, including all reciprocal-integer boundaries. For the tripartite diagnostics the path has $t=R+1$ alternating first- and second-block vertices, with fields $0,2,\ldots,2,1$ and distinct third-block completions on its $t-1$ positive connecting observations. A subset $A$ of its vertices has half-score cost
\begin{equation}\label{eq:path-flip28}
 J(A)=\sum_{v\in A}D_v-2\#\{i:v_i,v_{i+1}\in A\}.
\end{equation}
The entire path has cost $-1$ and every proper subset has nonnegative cost. In~\eqref{eq:order-path-mean}, independence of the third indices on the forced rows makes their distinctness condition cost a factor $1-O_t(1/n)$. Conditioning all addresses to be distinct preserves the field asymptotic by the bounded-density tilt in~\eqref{eq:order-mass} (repetitions involving a forced row add $O_t(m/n^2)$). At $m=\lceil cn\log n\rceil$ the expected number of eligible paths is therefore
\begin{equation}\label{eq:path-mean28}
 (1+o(1))[c(1-q)]^{t-1} A^t e^{-\lambda}\,n^{1-t\alpha}(\log n)^{t/2-1},
 \qquad A=(4\pi c\sqrt{q(1-q)})^{-1/2}.
\end{equation}
Isolation from other weak vertices is essential to turn this local cost into a two-unit global gap. The proof of Theorem~\ref{thm:order-path-window} constructs the competing assignment and excludes off-path improvements.
\end{proof}

The radius law is sharp for the uniform implication based on a high score and absence of a strict improving flip. It is not a lower bound for a verifier that uses additional global information. In particular it does not conflict with the exact component search after acquisition, or show that all local search algorithms encounter these configurations. The counterexamples differ from an optimum by only two score units, so an arbitrarily narrow fixed normalized score band cannot eliminate them asymptotically.

\paragraph{Finite path diagnostics.}
At $q=.17$, independently generate 16 fixed-size masks at each of $n=1024,4096,16384$ and $\alpha\in\{.25,1/3,.5,.65\}$ (192 targets). Count the ordered eligible paths of two and three vertices, using exact integer fields, positive pair incidences and distinct third-block completions. Table~\ref{tab:path-count28} shows the largest-size results. The mean formula is asymptotic and the counts can be rare. In particular, none of the 48 masks at $\alpha=.5$ contains the specified two-vertex path. This does not contradict its positive limiting mean; the finite sample is small. These are counts before isolation from the rest of the weak set; they are not global local-optimum counts. The constructed finite fixture has two weak vertices, each of field two, four positive shared triples, and no other vertex of field at most four. Exhaustive evaluation of all 4096 assignments gives source score 36 and maximum 44: the source is one-flip stable, while the pair flip attains the optimum. The path cost identity and its strict full-path improvement are separately checked for every one of 8188 subsets at lengths two through twelve. Independent direct enumeration of candidate paths checks the counting code on 60 small masks.

\begin{table}[ht]
\centering\small
\caption{Mean eligible-path counts at $n=16384$ ($16$ targets per density). Predictions use \eqref{eq:path-mean28}; the sample means are not fits.}
\begin{tabular}{r|rr|rr}
\toprule
 &\multicolumn{2}{c|}{Two vertices}&\multicolumn{2}{c}{Three vertices}\\
$\alpha$&Observed&Prediction&Observed&Prediction\\\midrule
$.25$&9.438&10.186&1.250&1.074\\
$1/3$&2.438&2.021&.063&.110\\
$.5$&0&.080&0&.001\\
$.65$&0&.004&0&$1.52\times10^{-5}$\\
\bottomrule
\end{tabular}\label{tab:path-count28}
\end{table}

\section{The critical window for one-coordinate verification}
\label{app:pair-window}
The radius law in Appendix~\ref{app:score-basin} identifies a transition at $\alpha=1/2$. At that density the last interacting weak pairs govern the validity of one-coordinate verification. We can classify all of their local minima and determine the limiting probability of false acceptance, rather than count only a sufficient obstruction.

Fix $q\in(0,1/2)$, put $h=1-2\sqrt{q(1-q)}$, $\mu=1-2q$, and $r=\sqrt{(1-q)/q}$. Use the same fixed-size uniform mask and suppose
\begin{equation}\label{eq:pair-window-density}
  2hm/n=\log n+u+o(1),\qquad u\in\mathbb R.
\end{equation}
For an assignment $x$ with no strictly improving one-coordinate flip, call its regret $\OPT-F_Y(x)$. For a fixed score cutoff $\tau m$, let $M_n$ be the maximum regret among such assignments passing the cutoff, with maximum zero if the set is empty.

\begin{theorem}[One-coordinate verification window]\label{thm:pair-window}
There exists $\tau_0<\mu$ depending only on $q$ such that for each fixed $\tau\in[\tau_0,\mu)$,
\[
 M_n\ \Rightarrow\ 2P_1+4P_2,
\]
where $P_1,P_2$ are independent Poisson variables with means $K_1e^{-u},K_2e^{-u}$, and
\begin{align}
 K_1&=\frac{3}{4\pi\sqrt{q(1-q)}}\,2(r+r^{-1}),\\
 K_2&=\frac{3}{4\pi\sqrt{q(1-q)}}\left[(1-q)(r^2+r^{-2})+2q\right].
\end{align}
In particular, the probability that every one-coordinate-stable assignment passing the score cutoff is globally optimal converges to
\begin{equation}\label{eq:pair-window-prob}
 \exp\{-(K_1+K_2)e^{-u}\}.
\end{equation}
The probability is over the target, and the implication is uniform over assignments. The result does not require any acquired readings. It is a limiting probability of validity of this test, not a finite deterministic certificate or a claim about the distribution of a particular algorithm's output.
\end{theorem}

\begin{proof}
\par\medskip\noindent\textbf{Reduction to independent singleton and pair objectives.}\quad
The proof of Theorem~\ref{thm:score-basin} with $J=3$ works uniformly under \eqref{eq:pair-window-density}: $m/(n\log n)$ tends to $c_*=(2h)^{-1}$, so choose $\delta>0$ fixed with $3c_*k(\delta)>1$ strictly. Every bound in that proof retains a strict exponent margin for $c$ in a sufficiently small compact neighborhood of $c_*$. Take $\tau_0=\mu(1-\delta/3)$. With probability $1-o(1)$, every passing one-flip-stable assignment and every maximizer (which passes with probability $1-o(1)$ since $F_Y(T)/m\to\mu$) have a source-gauge representative with support in $W_4^*$. The finite moment in Lemma~\ref{lem:sublog-weak-geometry} applies without change for this sequence of masks; its exponent is $-s(\log n+u)/2+O(1)$ at fixed cutoff. With $L=13$, $J=k_*=3$, the same union bounds give components of $W_L^*$ of size at most two and fewer than three neighbors in $W_L^*$ at every vertex. Pair codegree is at most two with probability $1-o(1)$ by the stored bound. Moreover, no pair of weak vertices has codegree two or more: there are at most $3n^4$ choices of a pair and two distinct incident shared triples; the fixed weak moment bounds their expected eligible count by $O(n^4(m/n^3)^2 n^{-1+o(1)})=o(1)$.

On this common event, restricted flip costs on $W_L^*$ add over its components. Every choice of locally stable configurations of these component objectives extends to a one-flip-stable assignment: outside $W_L^*$ the source sign is unchanged and its signed field is at least $D_v-4k_*>0$, by the codegree and weak-neighbor bounds. Conversely every passing one-flip-stable assignment is supported there after a gauge change, and its component configurations must be locally stable. Since a global optimum is supported there, maximizing regret reduces to choosing the worst local minimum in each component's flip cost. A singleton has only optimal local minima. A pair with fields $a,b$ and a unique shared triple of source sign $s\in\{-1,1\}$ has the four costs
\[
  0,\ a,\ b,\ a+b-2s.
\]
These follow directly by summing the odd-incidence cost. Two adjacent local minima have equal costs. For $s=1$ the only possible unequal opposite local minima are $00,11$, which coexist exactly if $a,b\in\{0,1,2\}$. For $s=-1$ they are $10,01$, coexisting exactly if $a,b\in\{-2,-1,0\}$. Thus the cost regret of the worst local minimum is
\begin{equation}\label{eq:pair-regret}
 g(a,b,s)=\begin{cases}
 |a+b-2|&s=1,\ 0\le a,b\le2,\\
 |a-b|&s=-1,\ -2\le a,b\le0,\\
 0&\text{otherwise}.
 \end{cases}
\end{equation}
The fields are integers, so the only nonzero costs are one and two. Let $N_i$ count pairs with a unique shared triple and $g=i$, including those not isolated. All their endpoints belong to $W_L^*$, so on the common structural event they are exactly the pair components with that regret. Choosing worst local minima loses $2N_1+4N_2$ score units relative to a maximizer. We show next that these counts are tight, which ensures that this constructed assignment passes every fixed cutoff below $\mu m$ with probability $1-o(1)$. Consequently $M_n=2N_1+4N_2$ with probability $1-o(1)$.

\par\medskip\noindent\textbf{Intensity of the two kinds of pair.}\quad
There are $3n^3$ choices of an unordered cross-block pair and a possible shared triple. Write $A_*=(4\pi c_*\sqrt{q(1-q)})^{-1/2}$ and $p=m/n^3$. Conditional on inclusion of the specified triple with sign $s$, the joint field mass at $a,b$, with no other shared observed triple, is
\[
 (1+o(1))\frac{A_*^2}{\log n}\exp(-2hm/n)r^{a+b-2s}.
\]
The forced private-group local limit in the proof of Theorem~\ref{thm:sharp-radius28} is uniform for $m/(n\log n)\to c_*$: the private Poisson means are $m/n+O(\log n/n)$ and the exponential tilt normalizer is $\exp(-hm/n)$, as in Appendix~\ref{app:residual-laws}. The extra shared-edge correction is $o(1)$ relative to this mass by the fixed weak-field moment there. Multiplying by the inclusion probability $p$, the sign probability, and $3n^3$ gives a prefactor tending to $3c_*A_*^2 e^{-u}$. For $s=1$, regret one has field sums one and three, each with multiplicity two; regret two has sums zero and four, each with multiplicity one. For $s=-1$, regret one has sums $-3,-1$, each with multiplicity two, and regret two has sum $-2$ with multiplicity two. Summing $r^{a+b-2s}$ over these lists yields $\E N_i\to K_i e^{-u}$.

\par\medskip\noindent\textbf{Joint factorial moments.}\quad
For any fixed orders $j_1,j_2\ge0$, expand $\E[(N_1)_{j_1}(N_2)_{j_2}]$ over ordered marked pair--triple choices. Identical choices of different types are incompatible. For the choices with disjoint marked endpoints and no forced triple meeting another marked pair, the forced-field local limit factors into the product of the single-pair limits. These choices comprise a $1-o(1)$ fraction of all label choices. Extra sampled shared triples between marked endpoints have negligible relative contribution by the same fixed-cutoff moment argument as in the path proof. Shared unmarked third endpoints require no independence assumption: the private groups depend only on the marked endpoints and have the same uniform asymptotics. These terms tend to $(K_1 e^{-u})^{j_1}(K_2 e^{-u})^{j_2}$.

For the remaining choices let $v$ be the number of distinct marked endpoints, $e$ the number of distinct forced triples, and $b$ the number of those triples containing three marked endpoints. Each of the other $e-b$ triples requires at most one further label. For each fixed pattern there are $O(n^{v+e-b})$ choices, and the moment bound for $v$ weak vertices and $e$ forced triples is $p^e n^{-v/2+o(1)}$. Its total is at most $n^{v/2-e-b+o(1)}$. Partition the marked vertices into connected components using the forced triple incidences. In a component of $v'$ vertices the sum of $|H_i\cap V'|-1$ over its forced triples is at least $v'-1$, i.e., $e'+b'\ge v'-1$. The exponent for this component is at most $1-v'/2$. A component of two vertices can represent only one compatible marked choice, because every marked pair requires a unique shared triple. A configuration not of the preceding disjoint form therefore has a component with at least three vertices. All components have at least two; the total exponent is strictly negative. The finite number of patterns makes their aggregate $o(1)$. This establishes all joint factorial moments of independent Poisson variables. The factorial-moment characterization of the Poisson law gives joint convergence. In particular $N_1+N_2=O_{\Pr}(1)$, completing the preceding regret argument. Because all regrets are nonnegative even integers, convergence in distribution gives the claimed zero-regret probability by evaluating the distribution function at one.
\end{proof}

\par\medskip\noindent\textbf{Finite counts.}\quad
For $q=.17$, $K_1=3.383875\ldots$ and $K_2=2.899591\ldots$. Thus at $u=0$ the limiting uniform validity probability is approximately $.001867$; at $u=\log 4$ it is $.207865$. We count the unique-shared-triple pairs in \eqref{eq:pair-regret} on 384 independently generated fixed-size masks, 64 at each combination of $n\in\{512,2048,8192\}$ and $u\in\{0,\log4\}$. At $n=8192$ the two mean counts are $3.094,2.766$ at zero, versus $3.384,2.900$ predicted, and $.859,.750$ at $\log4$, versus $.846,.725$. The zero-count totals are $0/64$ and $16/64$, respectively. None of the counted pairs on any of the 384 targets is isolated at the conservative cutoff $L=13$. This negative result matters: the limiting weak-component event is not demonstrated at these sizes, and the observed counts cannot be interpreted as the finite maximum regret or false-acceptance probability. The implementation checks the entire four-state classification on 578 signed field pairs and compares vectorized counts to direct enumeration on 120 additional small masks. The source-gauge fields are used only for this diagnostic.

\section{Two-hop capacity certificates for finite neighborhoods}
\label{app:twohop-radius}
A vertex with a large positive field can participate in a minimal improving
flip only if sufficiently many of its positively coupled neighbors also
flip. Those neighbors incur their own requirements. Accounting for these
requirements jointly can rule out a flip set far larger than the radius
excluded by a single-vertex or pair test. The one-hop threshold test resembles heterogeneous core pruning \citep{baxter2011core}. Here the constraints arise from minimal improving flips and share a total cardinality budget. The following result is entirely
deterministic and uses neither the generating source nor the noise rate.

Use the positive pair capacities $w_{uv}$ and integer requirements $k_v$
from \eqref{eq:radius-capacities}. Throughout the elimination procedure,
removed vertices and edges have zero capacity. Fix a remaining vertex $v$
and write $U_v=\{u:w_{vu}>0\}$ for its current positive neighborhood. For
$u\in U_v$ and $z\notin U_v\cup\{v\}$, define
\begin{align}
 b_{vu}&=\left(k_u-w_{vu}-\sum_{z'\in U_v}w_{uz'}\right)_+,
 & c_{vz}&=\sum_{u\in U_v}w_{uz}.
 \label{eq:twohop-burden30}
\end{align}
Here $b_{vu}$ is the minimum external capacity needed by $u$ even if it
receives all the available capacity inside the closed neighborhood of $v$.
For $0\le t\le \min(R-1,|U_v|)$ let
\begin{equation}
 L_v(t)=\min\left\{\sum_{u\in T}b_{vu}:T\subseteq U_v,\ |T|=t,
 \ \sum_{u\in T}w_{vu}\ge k_v\right\},
 \label{eq:twohop-L30}
\end{equation}
with $L_v(t)=+\infty$ when the family is empty. As before, $B_j$ is the sum
of the $j$ largest capacities, padded with zeros.

\begin{proposition}[Two-hop capacity elimination]
\label{prop:twohop30}
Delete $v$ and its incident positive pairs whenever, for every integer
$0\le t\le\min(R-1,|U_v|)$,
\begin{equation}
 L_v(t)> B_{R-1-t}\big((c_{vz})_{z\notin U_v\cup\{v\}}\big).
 \label{eq:twohop-delete30}
\end{equation}
The deletion preserves all inclusion-minimal strictly improving sets of
size at most $R$. The rule may be iterated, combined with either rule in
Proposition~\ref{prop:radius-peeling}, and followed by exact component
search. An empty remaining vertex set certifies radius $R$ without any
flip enumeration. In particular, the test includes the one-hop vertex
rule and can make strictly more deletions.
\end{proposition}
\begin{proof}
Let $S$ be inclusion-minimal with $J_x(S)<0$ and $|S|\le R$. At every
stage before a vertex or pair of $S$ has been deleted, the necessary
inequality \eqref{eq:radius-requirement} holds with the current capacities
for every vertex of $S$. Suppose $v\in S$, and partition its remaining
selected vertices into $T=S\cap U_v$ and $Z=S\setminus(U_v\cup\{v\})$.
The inequality at $v$ gives $\sum_{u\in T}w_{vu}\ge k_v$; thus $T$ is a
feasible choice in \eqref{eq:twohop-L30}. For each $u\in T$, the same
inequality at $u$ gives
\[
 \sum_{z\in Z}w_{uz}\ge
 \left(k_u-w_{vu}-\sum_{z'\in T}w_{uz'}\right)_+\ge b_{vu}.
\]
Consequently, since $|Z|\le R-1-|T|$,
\[
 L_v(|T|)\le \sum_{u\in T}b_{vu}
 \le \sum_{z\in Z}\sum_{u\in T}w_{uz}
 \le B_{R-1-|T|}\big((c_{vz})_z\big).
\]
This contradicts \eqref{eq:twohop-delete30}; therefore a vertex of $S$
cannot be deleted. Induction establishes the invariant, also when all
valid deletions in a pass are applied simultaneously. The argument in
Proposition~\ref{prop:radius-peeling} then places a minimal improving set
in a single surviving positive component. If no such component contains
an improving flip of size at most $R$, no such flip exists anywhere.
The old vertex test is the case in which no $t$ admits enough root
capacity in \eqref{eq:twohop-L30}. All proof inequalities are necessary
conditions, not sufficient tests for an improving set.
\end{proof}

\paragraph{Computation and a local quadratic radius.}
The minima in \eqref{eq:twohop-L30} are integer knapsack values with a
cardinality constraint. Index the dynamic program by the number $t$ of
selected neighbors and their total root capacity clipped at $k_v$.
Initialize the value at $(0,0)$ to zero and all other values to infinity.
For each $u$ retain the current value or add $b_{vu}$ and advance the
state by $(1,w_{vu})$, clipping the second coordinate at $k_v$.
The final value at $(t,k_v)$ is $L_v(t)$. With $d_v^+=|U_v|$ and
$T_v=\min(R-1,d_v^+)$, the number of state updates is
$O(d_v^+(T_v+1)(k_v+1))$. All capacities are bounded by the number of
observations, so this is polynomial for the unweighted target; for
binary-encoded integer weights this bound is pseudopolynomial. Computing
$b$ and $c$ scans the incident positive pairs of $U_v$; sorting the
external capacities supplies every required $B_j$. There are at most
$3n+1$ vertex-only passes, including the final pass without deletions.
No near-linear worst-case bound is claimed. If all root pair capacities
are one, the smallest feasible $t$ is $k_v$, and the minimum at that
$t$ is the sum of the $k_v$ smallest burdens. Since the burdens are
nonnegative and $B_{R-1-t}$ decreases with $t$, testing only this $t$
is exact for the deletion rule.

The local improvement over one-hop tests can be quadratic in the
requirements. Suppose all capacities incident to the closed neighborhood
of $v$ are at most one, each neighbor of $v$ has at most one neighbor
inside $U_v$, and every vertex outside the closed neighborhood receives
at most one unit of capacity from $U_v$. Assume $k_v\ge k$ and
$k_u\ge k$ for all $u\in U_v$, where $k\ge2$. Every candidate $T$ then
has at least $k$ vertices and total external burden at least $t(k-2)$.
The outside capacity is at most $R-1-t$. Thus an inclusion-minimal
improving set containing $v$ must satisfy
\begin{equation}
 R\ge 1+k(k-1).
 \label{eq:twohop-quadratic30}
\end{equation}
For comparison, the first vertex test by itself requires only
$R\ge k+1$ when $k_v=k$ and $v$ has at least $k$ unit-capacity neighbors.
The quadratic exclusion is a deterministic consequence of the two-hop
geometry; it is not asserted to hold uniformly over random targets.

\subsection{Finite verification of acquired candidates}
Replay the 32 repaired candidates of Appendix~\ref{app:local-radius},
with no further source readings, optimization, or changes to their scores.
The two-hop algorithm uses the same component cap 18 and subset cap
200,000 as the pair verifier. Each cap refusal is unresolved. Table~\ref{tab:twohop30}
reports all 16 targets at each size. Both methods are replayed at radii
14, 18 and 24 for the smaller cohort; the larger cohort uses the new
method at radii 14, 24 and 32. The previous larger-cohort radius-14 pair
result is retained in Appendix~\ref{app:local-radius} (14 of 16 pass).
At size 4096 the pair rule leaves a giant component at all tested radii,
whereas two-hop elimination deletes every vertex at radius 24. At size
16384, only two targets have any surviving vertices at radius 32, each
with one two-vertex component. Six nonempty flip subsets in total are
checked at that radius. All 32 candidates therefore have exact radius-24
certificates. The larger cohort also has exact radius-32 certificates.
These certificates exclude all strictly improving flips within the
stated radius, not just the ones encountered during search. They do not
exclude improvements outside the radius, and do not certify the model's
local-to-global event on a realized instance.

\begin{table}[ht]
\centering\small
\caption{Deterministic radius verification on frozen repaired candidates.
Each row contains 16 targets; ``pass'' means every flip within the
stated radius is nonimproving. The pair comparator at the larger size
is not rerun at radii 24 and 32.}
\begin{tabular}{rr|rr|rr}
\toprule
$n$ & $R$ & \multicolumn{2}{c|}{Pair rule} & \multicolumn{2}{c}{Two-hop rule}\\
&& Pass & Largest component & Pass & Largest component\\\midrule
4096 &14&0&8398&16&0\\
4096 &18&0&12089&16&0\\
4096 &24&0&12288&16&0\\
16384&14&14&13007&16&2\\
16384&24&--&--&16&2\\
16384&32&--&--&16&2\\\bottomrule
\end{tabular}
\label{tab:twohop30}
\end{table}

The finite audit generates 90 independently seeded small signed targets
at sizes $n=2,3,4$ and enumerates 140,160 flip scores. Direct enumeration
of every neighbor subset independently checks the dynamic program's
fixed point at 600 target--radius pairs. It also checks preservation of
all 3,447 qualifying minimal-improvement set--radius pairs, and the final
radius result against complete enumeration. The integer checks apply to
arbitrary candidate signs, not only candidates supplied by acquisition.

\section{When every short observation hides the noise}\label{app:lower}
\paragraph{Computational comparison under replacement.}\label{sec:comparison}
For the tripartite decision problem, suppose the threshold is supplied to both arms. The signed answer is $+1$ if the optimum meets the threshold, $-1$ otherwise.
\begin{theorem}[Independent returns at a balanced decision]\label{thm:main}
Take the score lattice point nearest $\E\OPT$, with ties rounded down. Each answer has probability $1/2+o(1)$ at every fixed density. The retained solver succeeds with probability $1-o(1)$. Under independent replacement with addresses chosen without target labels, set $D_n=\lfloor n^3/(16e^2m^2)\rfloor$. Every polynomial of target-label degree at most $D_n$, allowing arbitrary dependence on the mask and returns in its coefficients, has squared-error risk at least $1-o(1)$. Every adaptive tree reading at most $D_n$ target labels succeeds with probability at most $1/2+o(1)$. For $\alpha\ge1/2$ these conclusions also hold at
\begin{equation}
 K_m=m-2\lfloor qm\rfloor.\label{eq:balanced}
\end{equation}
\end{theorem}

Define the even-cover size $g(H)$ as the minimum number of distinct rows of the binary incidence matrix whose sum over $\mathbb F_2$ is zero, or infinity if there is no such collection. For a uniform mask with $m=\lceil Cn\log n\rceil$, put
\[
 D=\left\lfloor\frac{n^3}{16 e^2 m^2}\right\rfloor.
\]
Then $\Pr\{g(H)\le D\}=o(1)$. To prove it, first draw all rows independently with replacement. A fixed set of $t$ draws can be an even cover only for even $t$. In each mode its $t$ coordinate occurrences must admit a pairing of equal indices. For a fixed pairing this has probability $n^{-t/2}$; there are $(t-1)!!\le t^{t/2}$ pairings. The modes are independent. A union bound over the $\binom mt$ subsets gives
\[
 \Pr\{\text{even cover of size }t\}
 \le \left(\frac{em\sqrt{t}}{n^{3/2}}\right)^t.
\]
For $2\le t\le\sqrt D$ the base tends uniformly to zero, so this range sums to $o(1)$ by geometric summation. For $\sqrt D<t\le D$ the base $em\sqrt t/n^{3/2}$ is at most $1/4$. Distinctness of all draws has probability at least $1-\binom m2/n^3=1-o(1)$. Conditioning gives the claimed bound for the actual mask law. 

For a fixed good mask $H$, every nonempty product
$\prod_{e\in S}Y_e$ with $|S|\le D$ has conditional mean zero given the
entire target-noise vector $\eta$: some latent vertex sign occurs an odd
number of times. Equivalently, every such label subvector is jointly
uniform conditional on $\eta$. The same statement holds after conditioning
on an independent replacement record $Z'$ and its addresses.

We state the transfer to the optimum decision at every fixed positive
density. Let $U_n\in\{-1,1\}$ be the optimum decision at the common threshold,
and let $V_n$ be a function only of the independent target-error signs such
that
\[
 \mathbb E V_n=o(1),\qquad
 \varepsilon_n=\Pr(U_n\ne V_n)=o(1).
\]
For the nearest score-lattice point to $\mathbb E\mathrm{OPT}$, take
\[
 V_n=2\mathbf1\!\left\{\sum_{e\in H}\eta_e\ge m(1-2q)\right\}-1.
\]
Appendix~\ref{app:fluct} proves the required decision agreement for every
$c>0$. The binomial central limit theorem gives its vanishing mean.
If $\alpha\ge1/2$, the explicit threshold $K_m$ also permits
$V_n=2\mathbf1\{B\le\lfloor qm\rfloor\}-1$, where
$B=\sum_e\mathbf1\{\eta_e=-1\}$; Appendix~\ref{app:local} proves the same
agreement in that regime. In both cases $V_n$ is independent of $H,Z'$.

Let $P$ be a polynomial of target-label degree at most $D$, with arbitrary
coefficients depending on $H,Z'$. Multilinearization is exact because
target labels are signs. On a good mask, averaging the latent signs first
shows that $P$ is orthogonal to $V_n-\mathbb E V_n$, conditionally on
$(H,Z')$. For $\|P\|_2\le1$, Cauchy--Schwarz on the bad masks and the
agreement failure gives
\begin{align}
 |\mathbb E[P(U_n-\mathbb E U_n)]|
 &\le 2\sqrt{\Pr(g(H)\le D)}
 +2\sqrt{\varepsilon_n}+2\varepsilon_n.
 \label{eq:proj}
\end{align}
The three terms respectively bound the exceptional masks, $U_n-V_n$,
and their difference in means. The polynomial space contains constants.
Its best squared-error risk is therefore $\operatorname{Var}(U_n)$ minus
the squared norm of the projection of $U_n-\mathbb E U_n$ onto its closure.
Equation~\eqref{eq:proj} makes that norm $o(1)$, while
$|\mathbb E U_n|\le|\mathbb E V_n|+2\varepsilon_n=o(1)$.
The risk is consequently at least $1-o(1)$.

A depth-$D$ decision tree reading target labels is a sum of leaf indicators
of degree at most $D$, conditionally on $H,Z'$. Fixing private coins proves
the same statement for randomized trees. Its sign output $P$ has
$\|P\|_2=1$, and its classification success is
$(1+\mathbb E[PU_n])/2\le1/2+o(1)$ by the same bound. This is a restriction
on target-label degree or query depth; it is not a lower bound for all
polynomial-time algorithms.

\subsection{Replacement records selected from a short target pilot}
\label{app:adaptive-replacement-study}
An acquisition transcript chosen after looking at target labels is not
independent of the target, even when its physical source is independent.
A bounded pilot can be charged explicitly in polynomial degree.

\begin{lemma}[Charging a target-dependent acquisition pilot]
Fix a mask $H$. Let $P=P(H,\omega)\subseteq H$, $|P|\le h$, where the
pilot-selection coins $\omega$ are independent of target signs and noise.
Let $Z$ be independent of target latent signs and target errors, conditional
on $(H,\omega)$. Suppose the complete replacement acquisition record,
including its addresses, length and stopping indicators, is
\[
 A=\mathcal F(H,\omega,Y_P,Z).
\]
Consider any statistic having a representation
\[
 Q(Y,A)=\sum_{|S|\le d}g_S(H,\omega,A)\prod_{e\in S}Y_e.
\]
Conditional on $(H,\omega,Z)$, this statistic is a polynomial of degree at
most $d+h$ in the target labels. Consequently, on every mask with
$g(H)>d+h$, it is orthogonal to every centered function of the complete
target error vector, whenever the required expectations exist.
\end{lemma}

\begin{proof}
For fixed $(H,\omega,Z)$, each coefficient is a function on at most $h$
binary pilot labels. Its exact multilinear expansion has degree at most
$h$. Substitution into the displayed representation gives degree at most
$d+h$; repeated target signs reduce by $Y_e^2=1$. For a good mask, every
nonempty target monomial of degree at most $d+h$ has conditional mean
zero given all target errors, because some latent coordinate occurs an
odd number of times. The constant monomial is orthogonal to a centered
function of the errors. Integration over the independent coins proves
the assertion.
\end{proof}

Thus the replacement-record argument of Appendix~\ref{app:lower} remains
valid for target degree $d\le D_n-h$, with exactly the same exceptional-mask
and decision-agreement bounds in~\eqref{eq:proj}. Here the coefficient norm is the joint
$L^2$ norm under the actual adaptive transcript distribution. It need not
be uniformly bounded separately for each $Z$; the original global
Cauchy--Schwarz bounds suffice.
A depth-$d$ target-label decision tree after acquisition likewise becomes
a degree-at-most-$d+h$ polynomial after fixing $(H,\omega,Z)$, by expanding
its leaves and its pilot-dependent coefficients. This yields the same
classification bound with the shifted budget.

The hypothesis concerns which original target labels the acquisition
policy inspects. It permits arbitrary computation on the mask, pilot and
independent replacement data, but does not cover additional acquisition
choices depending on uncharged target labels. A full replacement tensor,
all prospective independent sensor errors and private coins can be drawn
in advance as $Z$, so repeated addresses and variable transcript lengths
cause no problem. For the calibration policy above $h=O(\log^2 n)$,
whereas $D_n=\Theta(n/\log^2n)$; the lower-bound scale is unchanged.

\section{Why calibration alone needs more readings}
\label{app:source-only}
The original observations can replace repeated calibration. The following
bound applies when those observations are unavailable. It allows arbitrary
adaptive choices of the queried grid addresses.

\begin{proposition}[Recovery from the source alone]
\label{prop:source-only}
Let the three factor vectors be independent uniform signs. Suppose each
query returns one selected tensor entry through an independent binary
symmetric channel with fixed error probability $s\in(0,1/2)$.
An algorithm may use private randomness and choose each address from its
previous replies, but it has no target labels or other observations of the
source. If it uses at most $Q$ queries on every execution, its probability
of recovering the entire factor tensor is at most
\[
 \left(1+\exp\left[-\frac{Q}{n}
                 \log\frac{1-s}{s}\right]\right)^{-n}.
\]
In particular, $Q=o(n\log n)$ gives success probability tending to zero.
For any fixed $\delta\in(0,1)$, success at least $1-\delta$ requires
\[
 Q\ge\frac{n}{\log((1-s)/s)}
 \log\frac{1}{(1-\delta)^{-1/n}-1}
 =\frac{n\log n}{\log((1-s)/s)}-O_{s,\delta}(n).
\]
The order $n\log n$ is attained by repeated fiber measurements.
\end{proposition}

\begin{proof}
Give the algorithm the true second and third factor vectors. This extra
information can only help. After multiplying a query by the corresponding
known factor signs, the reply is one noisy copy of the selected first-block
bit $a_i$. Full tensor recovery also recovers every $a_i$, since
$a_i=T_{i11}b_1c_1$.

Condition on the private random coins and a complete query transcript.
Its likelihood, as a function of the independent uniform bits $a_i$,
factors over $i$. The address-selection and stopping probabilities do not
depend further on the unknown bits once the preceding replies are fixed.
Consequently the posterior also factors over $i$, even for an adaptive
policy. If coordinate $i$ was queried $t_i$ times, its posterior log odds
have absolute value at most $Lt_i$, where $L=\log((1-s)/s)$.
The largest posterior probability of one value of $a_i$ is therefore at
most $(1+e^{-Lt_i})^{-1}$.

The best conditional probability of recovering all the bits is the product
of their largest posterior probabilities. The function
$x\mapsto-\log(1+e^{-Lx})$ is increasing and concave on $[0,\infty)$.
Since $\sum_i t_i\le Q$, this product is at most
$(1+e^{-LQ/n})^{-n}$ on every transcript. Averaging over transcripts and
private coins proves the bound. Its consequences follow by taking
logarithms. The bound is for a fixed worst-case query budget; it does not
assert a sharp constant for an expected-length stopping rule.
\end{proof}

This is an elementary specialization of the noisy-query setting studied
by \citet{feige1994}; see also \citet{gu2026}. The target-assisted algorithm
uses the original noisy constraints as additional evidence, so the premise
excluding target labels is essential. The proposition does not lower-bound
the optimal number of auxiliary reads when those constraints are available.

\section{Experimental protocols and resource bounds}
\label{app:erasure-experiments}

\subsection{Targets, acquisition laws, and exact references}

The 440-target exact study uses \(q=.17\), five densities
\(\alpha=.30,.55,.75,.95,1.25\), and \(30,24,20,12,2\) independent
targets per density at \(n=8,10,12,14,16\), respectively.
Thus each density contains 88 targets with identical size proportions.
The target has \(m=\lceil\alpha n\log n/h(q)\rceil\) uniformly sampled
distinct locations. Its generator draws independent source signs and label
flips and records a SHA-256 hash of the ordered signed clauses.
Every compared method receives those original clauses.

The stored target seed is the NumPy SeedSequence
\([20261122,5,n,\operatorname{round}(1000d),t,j]\), where \(d\) is the
stated density, \(t\) the trial index, and \(j=0\) for the exact study
or \(1\) for source-recovery diagnostics. The target-only baseline appends
99. The original one-pilot method appends 33; the new exact replays append
501, 502, 503, or 504 for the prior fixed, prior growing, fixed erasure, or
growing erasure schedule, respectively. The linear method needs no private
initialization coins. Retained and replacement arms use the same algorithm
seed. The independently generated replacement source has the same marginal
factor law, sensor error rate, measured addresses, and reading count.

The reference optimizer fixes the first signs of two factor blocks, enumerates
all assignments of those two blocks, and maximizes the remaining block exactly.
Gray-code updates use integer sums. The existing \(n\le16\) references are
unchanged. The reference solver also returns an optimal witness and permits \(n=18\).
Before the new grid, it was compared against a separate NumPy exhaustive
implementation on 84 random and boundary cases at \(n=1,\ldots,7\), including
empty and complete observation sets. All comparisons passed.

\subsection{Eight exact-reference targets}

The size-18 grid was fixed before execution: \(n=18\),
\(\alpha\in\{.30,.55,.75,1.25\}\), two targets per density,
\(q=.17\), and \(s=.10,.20\). Its target seed is
\([20261122,10,18,\operatorname{round}(1000\alpha),t]\).
Sensor coins use this key followed by \([200,1000s]\).
The retained and replacement source readings share these noise coins,
while their latent sources are independent. Private method seeds append
\([1000s,b]\), where \(b=301,302,303,304\) identifies the four schedules
in the same order as above. The baseline again appends 99.
Changing the sensor condition leaves the target unchanged.

\begin{table}[ht]
\centering
\caption{Exact references. All eight optimal witnesses are stored and
their integer scores independently checked.}
\label{tab:erasure-n18-oracles}
\begin{tabular}{rrrr}
\toprule
\(\alpha\)&Trial&Source score&\(\OPT(H,Y)\)\\
\midrule
.30&0&25&55\\
.30&1&39&55\\
.55&0&70&82\\
.55&1&76&82\\
.75&0&109&109\\
.75&1&111&111\\
1.25&0&178&178\\
1.25&1&192&192\\
\bottomrule
\end{tabular}
\end{table}

Each reference enumerates \(2^{34}\) assignments. The eight jobs ran
concurrently and required approximately 339--345 seconds each in the recorded
environment. These exponential validation costs are outside the tested
polynomial procedures. The four suboptimal sources have gaps \(30,16,12,6\).

\begin{table}[ht]
\centering
\caption{Complete retained-source procedures on the same eight \(n=18\)
targets. Entries count exact empirical optima. The baseline solves two in
both sensor conditions. All replacement counts are zero.}
\label{tab:erasure-n18-methods}
\begin{tabular}{lrrr}
\toprule
Method&Reads&\(s=.10\)&\(s=.20\)\\
\midrule
Linear&54&5&0\\
Four batches, \(a=2/3\)&18&0&0\\
Erasure, \(a=3/4\)&15&0&1\\
Prior growing&54&0&0\\
Growing erasure&54&3&0\\
\bottomrule
\end{tabular}
\end{table}

At \(\alpha=.55,t=1,s=.10\), the linear method reaches the exact score 82,
above the source score 76 and baseline score 72. No new erasure success in
this eight-target grid exceeds its source score. The two sensor conditions
are 16 evaluations of eight targets, not 16 independent targets.

The fixed $a=3/4$ experiment uses two propagation batches and a final update;
 the round condition permits equality. The linear comparator uses the same $3n$
readings as the growing schedules at these sizes. Its replay matches every
retained and replacement one-read score in the 440-target panel.

\subsection{Exact outcomes and portfolios}

\begin{table}[ht]
\centering
\caption{Complete \(s=.10\), one-read comparisons on 440 targets.
The first pair gives uncombined exact counts; the second pair adds the
common target-only baseline. R and I denote retained and replacement readings.}
\label{tab:raw-acquisition}
\setlength{\tabcolsep}{3pt}
\begin{tabular}{lrrrr}
\toprule
&\multicolumn{2}{c}{Alone}&\multicolumn{2}{c}{With baseline}\\
Method&R&I&R&I\\
\midrule
Linear&235&4&306&215\\
One pilot&111&3&241&214\\
Four batches&8&7&216&214\\
Fixed erasure&30&6&222&217\\
Prior growing&15&3&220&216\\
Growing erasure&119&3&252&215\\
\bottomrule
\end{tabular}
\end{table}

The target-only method uses 32 independent starts, with at most 25 alternating
block sweeps per start and ties retaining their prior signs. A portfolio
selects the larger feasible score and retains the baseline on a tie.
It uses no reference optimum in selection. At fixed iteration limits the
baseline adds \(O(m+n)\) word operations and no readings. The reported
portfolios are ordinary feasible-score selection, with the work of both
procedures included.

The fixed erasure method reaches six of the 241 optima above the source score;
growing erasure reaches 32; the linear method reaches 68. Their retained
portfolios add \(8,38,92\) exact targets, respectively, to the baseline's
214. At small sizes the growing schedules use all \(3n\) source readings.
The new result concerns their asymptotic acquisition law; the linear method
performs better on this grid at the same reading count.

The initial one-pilot portfolio analysis was retrospective. Its original
sensor conditions \(s=.10,.20\), with \(r=1,3,7\) readings per coordinate,
are retained in the cumulative trial files. The newer four-schedule replay
was fixed at \(s=.10,r=1\). It regenerates all 440 original targets and
checks every method's returned witness against the original exact optimum.
Including the linear comparison gives 4,400 method witnesses. These are
additional evaluations of existing targets and introduce no new instances.

\subsection{The larger source-recovery diagnostic}

The fixed erasure diagnostic reuses the prior four-batch grid:
\(n=1024\), \(c=20,100\), \(q=.17\), \(s=.10,.20\), and
40 targets per cell. Its target hashes and one-read sensor prefixes match
the stored records. New private split and reservoir coins append 1093
to the target seed. With \(a=3/4\), the exact schedule has
\(\ell_n=11\), \(d_n=7\), \(k_n=147\), and three target batches.
It therefore reads 441 bits. The prior four-batch and one-pilot methods
read 615 and 1,167 bits, respectively.

All 160 output arrays are stored. Their target scores and source-tensor
equality indicators are checked independently by regenerating each target.
There are 80 distinct targets, each observed at two sensor error rates.
The procedure stops before component repair. No independently computed
empirical optimum is available at this size. Table~\ref{tab:bootstrap-recovery}
reports every cell, including all zero-success erasure outcomes.

\begin{figure*}[t]
\centering
\includegraphics[width=\textwidth]{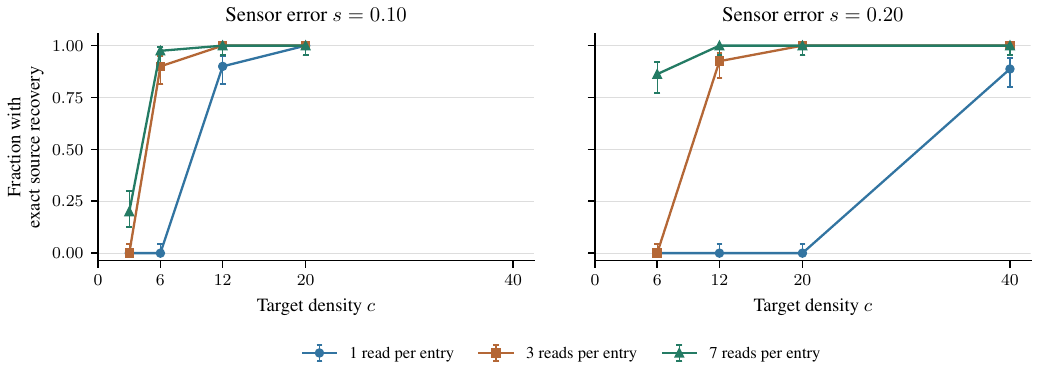}
\caption{All-coordinate source recovery after one target update
at \(n=1024\). Each point has 80 independent targets; bars are pointwise
95\% Wilson intervals. Read budgets are \(3nr\). The horizontal scale is
shared: the left panel samples \(c=3,6,12,20\), the right
\(c=6,12,20,40\). These experiments do not verify empirical optimality.}
\label{fig:v5-source}
\end{figure*}

The older all-coordinate source experiments remain in the package.
At \(n=1024,s=.10\), one reading at each of \(3n\) fiber entries recovers
none of the complete sources before refinement. A synchronous full-target
update recovers \(0,0,72,80\) of 80 sources at \(c=3,6,12,20\),
respectively. Figure~\ref{fig:v5-source} displays these measurements, whose
reference is also the latent tensor.

\subsection{Work and bit accounting}

Each update in these historical comparators is simultaneous and retains its
previous sign at a zero field. The later calibration cascade instead uses
independent fair ties in its calibration stages.
During erasure propagation, eligibility and predictions use the arrays at
the start of the batch. A later record cannot use a sign created earlier in
the same batch. Each unknown coordinate uses a uniform reservoir of eligible
records; unresolved coordinates remain zero until the final fair fill.

Component repair uses integer signed fields, adjacency lists, graph traversal,
and a Walsh transform to optimize a component with \(d\) variables over its
\(2^d\) assignments. The cutoff ranges from zero through
\(\max\{12,\lceil\log_2(n+1)\rceil\}\), which is also the component-size
cap. The declared repair budget is
\[
\left\lceil64(m+3n)\log(n+2)\right\rceil.
\]
A unit covers a bounded constant number of array operations, rather than a
CPU instruction. The budget is checked before scans and table allocation;
an incomplete stage leaves the previous feasible incumbent unchanged.
All original clauses enter repair, including clauses in unused propagation
batches when the growing erasure schedule stops early.

Fixed schedules initialize in \(O_a(m+n)\) operations. The growing erasure
implementation uses \(O(m+Bn)\), where its prescribed number of batches is
\(B=O(\log\log\log n)\). Every experimental work cap passed. Address,
score, and counter words need \(w=O(\log(m+n+2))\) bits. With random-access
storage, charging a conservative \(O(w^2)\) elementary bit operations for
each exact arithmetic operation gives
\[
 O\bigl((m+n)\log(n+2)\log^2(m+n+2)\bigr),
\]
including the bounded random-bit sampling below. This is
\(O(n\log^4 n)\) at \(m=\Theta(n\log n)\).
A Turing-machine implementation additionally charges its memory-access
simulation; ordinary polynomial simulation suffices for tractability.
Fiber addresses are generated in linear time in
the number of readings. Explicitly recording them costs one address word
per reading. Sensor latency, energy, and physical source retention are
separate costs of implementing the stated acquisition experiment.

The fixed and growing erasure schedules use integer comparisons and bit
lengths. Uniform draws may be implemented with fair random bits by rejection
sampling, allowing at most \(8\lceil\log_2(n+m+2)\rceil\) attempts
per draw. Each attempt accepts with probability at least one half, and there
are \(O(m+n)\) draws. The total cap-failure probability is therefore
\(o(1)\). Returning an initial feasible assignment on this event preserves
the all-input work bound. The polynomial estimator is set to zero on this
mask-and-private-coin event before score composition. The experiments use NumPy's seeded generator;
objective scores and search comparisons are exact integers in their represented
range.

\subsection{Verification and reproduction}

The larger exact-reference study comprises eight exact-reference targets,
16 corresponding sensor-condition rows, 160 complete method witnesses on
those rows, 4,400 method witnesses on the 440 stored exact targets, and
160 new source-only witnesses on 80 stored larger targets. The eight
exact-reference witnesses are checked separately. No two sensor conditions or
methods sharing a target are counted as independent targets.

The exhaustive reference program enumerates its full declared space. Its 84 small
cross-checks supplement the direct enumeration argument; feasible-witness
checking alone is not an optimality proof. The experimental audit regenerates
targets, checks their hashes, recomputes all integer scores, verifies work
caps, and reconstructs the reported aggregates. It records the exact
algorithm-source hash frozen before the experiments.

The reproducibility supplement includes the original masks, labels,
readings, returned witnesses and reference solver. From its
\texttt{auxiliary\_tripartite} directory, run \texttt{python3 -B replay.py}
to check the saved references and witness scores. The directory's
README gives the separate command that regenerates the eight exponential
reference solves.

The finite evidence leaves a clear unresolved problem: improve reduced-reading
performance without sacrificing the exact empirical guarantee or using the
unknown source during optimization. None of these measurements determines
the practical onset of the asymptotic theorems.

\subsection{Geometric acquisition: frozen implementation and replay}
\label{app:geometric-experiments-study}

The geometric implementation uses the fixed schedule and full-fiber branch specified in the experiment protocol.

The declared-bound replay uses the exact rational biases
$\underline\mu=33/50$ and $\underline\rho=4/5$, corresponding to $q=.17$
and $s=.10$. The learned-bound replay uses only its pilot observations to
compute the bounds. Both reuse the 440 original exact-reference targets.
Final seed readings are paired with the stored linear procedure; the
pilot uses a separate fresh sensor stream. All target labels, including
pilot and unused propagation labels, remain in the scored objective.

There are four method/source combinations and 1,760 returned witnesses.
Each witness is rescored on the original target and checked against its
stored optimum. Every run selects the full-fiber branch, and every witness
is identical to the corresponding linear-method witness. Both retained
methods attain 235 exact optima, including 68 of the 241 source-suboptimal
targets; their portfolios attain 306. Both replacement methods attain four,
including one source-suboptimal target; their portfolios attain 215.
The declared-bound method spends 24--48 readings. Including all pilot visits,
the learned-bound method spends 28--100.

The learned bounds are both positive on 244 retained arms and 114 replacement
arms. These are branching outcomes, not evaluations of
Appendix~\ref{app:pilot-distinguishability-study}'s rank-checked diagnostic.
The program's \texttt{pilot\_passed} flag checks only positivity of the two
estimated bounds. It neither checks incidence rank nor certifies retention.
Even a positive pilot requests the entire fiber at these finite sizes.

\paragraph{Checks and reproduction scope.}
The implementation suite contains 15,063 finite checks: 2,000 integer-root
cases, 2,304 schedules, 10,415 rational pilot-statistic comparisons, 32 exact
pilot moments, 150 comparisons with a separate scalar initializer, 82
small exact-reference witness checks, and 80 full-fiber identity checks.
Partial-seed tests exercise propagation mechanics without asserting that their
artificial seed counts meet the theorem's sufficient constant.
An independent mathematical program contributes 94,818 checks of the support
recurrence, noise bounds, finite pilot laws, adaptive degree composition and
rank-conditioned uniformity, including counterexamples to omitted assumptions.
These totals count comparisons and enumeration cells, not independent trials
or a formal proof certificate. Wolfram separately returns \texttt{True} for
three displayed algebraic identities; its exact input and output are archived.

The study replay checks the 1,760 witnesses against the 440 target
records. For the separate eight-target study at $n=18$, it regenerates the
targets and scores the stored optimal and method witnesses. Recomputing
the exhaustive optima requires the separate reference-solver run
described in the \texttt{auxiliary\_tripartite} README.

\subsection{Prospective exact-optimization study}\label{sec:finite-study}\label{app:finite-grid-experiments}

The 96-target protocol and algorithm were hashed before generating the targets
or observing any evaluation outcome. No development-target outcome was used to
choose this heuristic, and neither an algorithm nor a target was changed after
the freeze. The grid uses $n\in\{12,14,16\}$,
$\alpha\in\{.30,.55,.75,1.25\}$ and eight trials per cell, with
$q=.17$ and $s=.10$. The NumPy SeedSequence is
\[
 [20261123,12,120125,n,\operatorname{round}(1000\alpha),t],
 \qquad t=0,\ldots,7.
\]
The generator draws independent uniform source signs, a uniform set of distinct
triples, independent target errors, and an independent replacement source.
Sensor coins append 200 to the key; private method coins append 300. All methods
receive the same original clauses, and all original clauses remain in the score.
The integer reference program fixes two gauge signs, exhaustively enumerates
the other signs in those factors, and maximizes the third factor exactly.
All 96 reference computations completed. Their exponential work is evaluation
cost and never enters an optimizer.

The reduced method reads the first
$k=\lceil n/\lceil\log_2(n+2)\rceil\rceil$ coordinates of each reference
fiber, once each. Its budgets are $9,12,12$ at the three sizes. The full method
reads all $3n$ coordinates, costing $36,42,48$ readings. Reduced readings are
prefixes of the full readings. Retained and replacement arms share sensor-error
coins. Distinct visits to a common physical address, including the three fiber
origins, have separate noise variables and are all charged. The optimizer
receives neither source signs nor $c,q,s,\alpha$ or the reference optimum.

\paragraph{The fixed finite heuristic.}
Unknown coordinates remain zero until a nonzero majority of original
observations with exactly one unknown endpoint supplies an estimate. A tie
remains unknown. Each round uses the same input array for all votes and retains
old nonzero estimates. There are at most $\lceil\log_2(n+2)\rceil$ rounds,
with early stopping when none activates a coordinate. This heuristic reuses
observations between rounds; it is separate from the fresh-batch theorem.

Every arm uses exactly 32 initializations and at most 25 block-coordinate
sweeps per initialization, followed by the same capped repair of its best
assignment. Sixteen initializations use the raw readings and sixteen their
propagated extension. The first in each group preserves available signs; in
other initializations, each available sign is independently flipped with
probability $1/8$. Unknown signs are fair. The target-only arm uses 32 fair
initializations. A sweep updates the three blocks successively, retaining the
old sign at a zero field. The best completed assignment is scored on all
original labels, and no target-only solution is included implicitly in a
source arm. All arms share fixed-size private random arrays. Source arms
add at most $\lceil\log_2(n+2)\rceil$ propagation scans; equal polishing caps
do not mean equal initialization work.

\begin{table}[ht]
\centering
\caption{Complete exact-success counts by size for the fixed grid.
Each row has 32 targets. The 96-target aggregate appears in
Table~\ref{tab:finite-study}.}
\label{tab:finite-study-by-size}
\begin{tabular}{rrrrrr}
\toprule
$n$ & Target only & Reduced retained & Reduced replacement & Full retained & Full replacement\\
\midrule
12 & 11 & 16 & 17 & 20 & 11\\
14 & 7 & 16 & 11 & 19 & 7\\
16 & 8 & 10 & 10 & 19 & 7\\
\bottomrule
\end{tabular}
\end{table}

The mean integer objective deficits for target-only, reduced retained,
reduced replacement, full retained and full replacement are respectively
$6.6042$, $4.1250$, $6.4167$, $1.6875$ and $7.8125$.
Reduced retention has 22 paired wins and six losses against target only:
$16.67$ percentage points, with descriptive 95\% bootstrap interval
$[6.25,27.08]$. Against reduced replacement it has 20 wins and 16 losses:
$4.17$ points, with interval $[-8.33,16.67]$. Thus this experiment does not
establish a retention advantage at the reduced budget. Independent replacement
records could also be simulated under their declared law; their success
shows why the comparison with fair target-only initialization cannot by itself
identify an advantage from retained information. Full retention has 34 paired
wins and two losses against target only: $33.33$ points, with interval
$[22.92,43.75]$.

The intervals use 100,000 paired target-row bootstrap resamples with fixed seed
202611231212. They and the Wilson bars are descriptive for the fixed grid,
not population-uniform guarantees. The explicitly doubled portfolio of reduced
retention and target only reaches 48 optima. It uses 64 initializations and a
1,600-sweep cap and is excluded from the primary five-arm comparison.
Wall times were recorded during concurrent reference jobs and are not controlled
speed measurements. Reading counts, scan counts and declared polishing caps
provide the cost comparisons.

\paragraph{Independent checks.}
We retain every target, source and replacement source for reproducibility,
all raw readings, all exact-reference and algorithm witnesses, hashes and seed keys.
A scalar implementation independent of the optimizer's evaluator checks all
576 witness scores, 96 target hashes and 8,064 fiber-reading identities.
Seventy small independent exhaustive comparisons validate the reference
program. An additional 240 scalar propagation comparisons and 60 small
exhaustive comparisons check the finite heuristic. A complete rerun reproduces
all 480 method witnesses, scores and metadata. The exact-reference computation,
feasible-witness checking, and implementation checks have separate roles;
a feasible score alone does not certify optimality.

\subsection{Cascade implementation and branch checks}
\label{app:cascade-implementation-study}

The implementation uses the integer schedule in
Appendix~\ref{app:geometric}, including $K=\lceil5/(\mu_*\rho_*)^4\rceil$,
the enlarged batch horizon, and the $32n$ size guard for calibration.
Each executed calibration update draws a full $3\times n$ array of fair signs
before evaluating any fields. Thus its tie coins are fixed coordinate by
coordinate when private randomness is fixed; they are not consumed only at
label-dependent ties. Calibration batches are executed or skipped according
to their preassigned sizes. The final update retains its previous sign at a
tie. The conservative pilot and full-fiber branch are unchanged, and all
original observations remain in the empirical objective.

The independent implementation suite checks 444 exact schedules, 180 scalar
fair-majority updates, 180 complete scalar initializations, and 128 small
solver outputs against exhaustive optima. Eleven of the known-bound small
executions take the actual nonfallback branch; 53 full-fiber executions return
the exact stored witness. The small solver tests include 64 adaptive
executions. Separate direct-initializer tests force selected guards in local
schedule copies to exercise branches; they do not change the public algorithm
or assert the theorem for artificial schedules.

Two actual nonfallback executions check larger branches. A noiseless $n=64$
complete-mask case uses three readings, two propagation updates and five
calibration updates, and returns score $m=262144$. The bound $F_Y\le m$
certifies that particular optimum. It is a dense noiseless mechanics check.
At $n=128,c=100,q=.17,s=.10$, the algorithm uses 207 readings, compared with
384 for full fibers, and executes one calibration update. It returns feasible
score 40826; no independent empirical optimum is available for that case.
These branch checks are separate from the frozen 96-target evaluation.

The cascade was also evaluated on all 440 targets used for the geometric schedule, using the
same seed readings and separate fresh pilot noises as the earlier geometric
replay. All 1,760 runs take the full-fiber branch and reproduce the corresponding
linear witnesses. The retained counts remain 235 alone and 306 in the charged
portfolio, versus four and 215 under replacement. Learned bounds incur the
same 28--100 total readings. These are repeated evaluations of existing
targets, not new independent instances.

The finite conditional-majority suite compares 376 exact hypergeometric/binary symmetric channel (BSC)
convolutions and 408 conditional joint laws with rational enumeration over
33,495 global subsets. It includes deleted histories, zero-noise and fair-noise
boundaries, and the dependence counterexample when majority predictions are
conditioned only on occupied reservoirs rather than their full counts.
These executed checks complement the finite-law proof.

\subsection{Source recovery before component repair}

\begin{table}[ht]
\caption{Exact source recoveries out of 40 targets per cell at $n=1024$.
The columns use 1,167, 615 and 441 readings, respectively. The same 80 targets
appear under two sensor conditions. No empirical optimum is computed.}
\label{tab:bootstrap-recovery}
\centering
\begin{tabular}{rrrrr}
\toprule
$c$ & $s$ & One pilot & Four batches & Fixed erasure\\
\midrule
20 & .10 & 0 & 1 & 0\\
20 & .20 & 0 & 0 & 0\\
100 & .10 & 32 & 40 & 0\\
100 & .20 & 0 & 40 & 0\\
\bottomrule
\end{tabular}
\end{table}

The 80 targets in this diagnostic were scored for source recovery.
The 96-target study above used exhaustive empirical optima. Its 96
reference witnesses and 480 method witnesses can be checked with
\texttt{python3 -B replay.py} in the supplement's
\texttt{auxiliary\_tripartite} directory. The same directory includes
the reference solver, separate scalar checks and commands for rerunning
the optimizations.

\subsection{Noisy executions with independently certified optima}
\label{app:certified-cascade-finite}

We also test the executable fresh-batch cascade on 48 fully observed noisy
tensors. The grid has $n\in\{32,48\}$, $q\in\{.10,.17\}$, sensor noise
$s=.10$, and 12 trials per cell. The protocol and program were hashed before
target generation. Each target contains all $n^3$ addresses, so this is a
denser finite experiment than the sparse regime of Theorem~\ref{thm:geometric}.
The supplied rational channel bounds equal the true biases.

Retained and replacement measurements use the same locations and sensor error
signs. A matched target-only arm supplies independent fair prefix signs and
uses the same cascade schedule. All three arms share the randomized target
partition and the private random stream. A separate target-only method uses
the three contractions described in the certification construction.
It supplies a useful comparison because these dense targets may already be
easy without further readings.

\begin{table}[h]
\centering
\small
\begin{tabular}{rrrcccc}
\toprule
$n$ & $q$ & Readings & Retained & Replaced & Fair prefix & Dense target \\
\midrule
32 & .10 & 12 & 8 & 6 & 9 & 12 \\
32 & .17 & 27 & 3 & 8 & 6 & 12 \\
48 & .10 & 9 & 10 & 9 & 10 & 12 \\
48 & .17 & 18 & 9 & 5 & 6 & 12 \\
\midrule
\multicolumn{3}{l}{Total out of 48} & 30 & 28 & 31 & 48 \\
\bottomrule
\end{tabular}
\caption{Exact empirical optima attained on the frozen dense grid. Each row
contains 12 targets. Full fibers would use $3n=96$ or $144$ readings;
both target-only arms use zero. Three calibration updates execute at $n=32$
and four at $n=48$. All 48 optima have independent exact certificates.}
\label{tab:certified-cascade-finite}
\end{table}

For each returned candidate, the observed target and its factor signs define
a signed bipartite Laplacian. An exact integer test of its Schur complement
certifies global optimality over the larger flattened problem, and hence over
tripartite assignments. The test receives no latent source. All 48 targets
have an accepted certificate, and every reported score is evaluated directly.
The source happens to be optimal on every target in this dense grid.

\paragraph{Why each accepted certificate proves optimality.}
For a candidate $(\widehat a,\widehat b,\widehat c)$, flatten the last
two modes and form the observed matrix
$S_{i,(j,k)}=Y_{ijk}\widehat a_i\widehat b_j\widehat c_k$, with zero at
unobserved locations. Define the signed Laplacian
\[
 L(S)=\begin{pmatrix}
 \operatorname{diag}(S\mathbf1)&-S\\
 -S^\top&\operatorname{diag}(S^\top\mathbf1)
 \end{pmatrix}.
\]
If $L(S)$ is positive semidefinite, the candidate is a global maximizer.
Indeed, for arbitrary left and right sign vectors $u,v$,
\[
 \sum_{i,j} S_{ij}-\sum_{i,j}S_{ij}u_iv_j
 =\frac12 (u,v)^\top L(S)(u,v)\ge0.
\]
Every competing tripartite assignment is among these flattened competitors.
This is the standard signed-Laplacian certificate
\citep{abbe2014,hajek2016}; it uses only the observed target and the candidate.
The implementation removes isolated vertices, handles an empty target
directly, and tests positive definiteness after grounding one active vertex.
Because $L(S)\mathbf1=0$, passing this stricter test implies the required
positive semidefiniteness. When all active right signed degrees are positive,
the same test reduces by a Schur complement to
\[
 Q=\operatorname{diag}(S\mathbf1)
       -S\operatorname{diag}(1/d_j)S^\top,
 \qquad d_j=(S^\top\mathbf1)_j.
\]
Positive denominator clearing and integer elimination make acceptance exact.
A failed test is an abstention; it does not show that the candidate is
suboptimal. Appendix~\ref{app:triple-calibration-reuse} supplies the denser-law threshold analysis, which
is not needed for this deterministic soundness argument.

The reduced-reading retained cascade attains 30 certified optima; replacement
attains 28 and the matched fair-prefix arm 31. Retained versus replacement
has 10 wins and 8 losses on discordant targets. The separate dense target-only
method attains all 48 optima. Thus the experiment verifies noisy exact
executions of the actual reduced-reading cascade, while showing no advantage
from retention on this grid. It does not test the sparse regime in which the
source is suboptimal.

An independent audit replays all 144 cascade executions, verifies 240 integer
scores and 1,584 sensor signs, and checks 137 accepted certificates through a
different exact integer rounding argument. All six frozen hashes match.
The certificate implementation also agrees with exhaustive binary optimization
on 879 small matrices, with no false acceptances. Complete targets, readings,
witnesses, negative outcomes, and reproduction commands accompany the source.

\section{Verification of full-target reuse}
\label{app:reuse-checks}

The executable program is \texttt{code/recycling\_two\_pass.py}.
Its acquisition functions expose the stored cascade estimate before any
component enumeration. They preserve the original sensor calls, seed arrays,
batch capacities, and random choices. Starting from the acquired estimate $z^0$, two full-target synchronous updates
produce $z^1=\mathcal M(z^0)$ and $z^2=\mathcal M(z^1)$; the general-density program applies the existing capped
repair to $z^1$. Its two-pass mode omits that search entirely.
Both modes score candidates on the same original target and retain the best.

\paragraph{Exact deterministic checks.}
The independent checker represents every possible observation at each of the
eight locations in a $2\times2\times2$ target by $-1$, $0$ (absent), or $+1$.
It enumerates all $3^8=6,561$ such targets and all $2^6=64$ factor assignments.
For these 419,904 configurations it verifies the field perturbation identity,
the strict-sign sandwich, and the empirical-gain identity used in
Corollary~\ref{cor:correction-scale}. There are 2,519,424 individual vertex
fields; these are nested checks of the same configurations.

For each target and threshold in $\{0,1,2,3,4\}$, the checker also identifies
cases in which the weak vertices are independent, contain a global optimum,
and satisfy the deterministic outside-field condition. Its 1,692 qualifying
cases all attain the enumerated optimum, including 108 cases with a
suboptimal source. Zero fields are retained. A separate 11,456 calls compare
the actual two-update function with the independently computed fields.
These finite identities check the deterministic steps of the proof; they
do not estimate an asymptotic success probability.

\paragraph{Acquisition and work paths.}
Ninety comparisons check agreement with the stored acquisition in seed
arrays, measurement calls, and the random generator's final state; seventeen
take a reduced-reading path. Eight complete two-pass executions replace both
component-enumeration entry points with functions that raise an error if
called. None calls either entry point. Twenty-four further small-target
checks compare the returned feasible score and incumbent preservation with
exhaustive reference values.

A simultaneous update need not improve an arbitrary input. The checker
retains a seven-observation example whose scores are $-1,-7,7$ along its
three iterates. Selecting the highest score preserves the initial candidate.
No score-monotonicity premise is used in the theorem.

\paragraph{Exact finite execution gates.}
The separate schedule checker compares the two gate formulas in
Proposition~\ref{prop:finite-gate} with 165,237 integer schedules, including
several rational final fractions. It also checks the factorization defining
the relaxed minimum at 1,001 rational points. High-precision evaluation
supplies the displayed necessary scales. These counts describe program and
algebra checks, rather than generated optimization trials.

\paragraph{Signed-degree probabilities.}
A third checker verifies 3,255 exact rational tilting identities, 3,255
conditional geometric-tail bounds, 165 normalized finite signed-degree
distributions, and 2,136 identities for the disjoint fixed-shift witnesses.
It also evaluates 36 rare-event cases at 90 and 130 decimal digits, with
144 cross-precision comparisons. These computations test the finite
identities used in the tail proof; the asymptotic orders follow from the
uniform analytic estimates in Appendix~\ref{app:recycling}.

From \texttt{triples\_and\_clauses} in the reproducibility supplement,
\texttt{python3 run\_checks.py --jobs 2} runs the field, finite-gate,
and signed-degree checks and replays the stored score-decrease example.

\subsection{Retrospective ablation of the two full-target updates}
\label{app:reuse-ablation}

The two-update theorem motivates a direct test on the 440 historical exact
targets and the 96 targets from the later prospective experiment.
We freeze a new analysis protocol before computing the new method outcomes;
this is a retrospective ablation of those datasets, not an extension of
their prospective protocols. We use every target, retain all failures, and
reuse the stored exhaustive optimum values. No new target or exponential
reference computation is introduced.

Both source arms use their original noisy full-fiber arrays, with
$q=.17$ and $s=.10$. The 440-target dataset uses independent sensor streams
for retention and replacement; the 96-target dataset uses a common sensor
error array. These stored pairings are preserved. All 1,072 acquisitions
take the full-fiber branch at the declared bounds
$\underline\mu=33/50$, $\underline\rho=4/5$, and $\tau=1/2$.
Each arm uses $3n$ readings, ranging from 24 to 48 across targets and
totaling 17,832 readings over the 536 targets.

Let $r$ denote the raw full-fiber array. The acquisition estimate is
$z^0=\mathcal M(r)$, followed by
$z^1=\mathcal M(z^0)$ and $z^2=\mathcal M(z^1)$.
Thus two reused updates mean three target updates in total from $r$.
We record every iterate, the best of these three candidates, and capped
repair separately around $z^0$ and $z^1$. Every witness is scored on the
unchanged objective. Exact reference values are used only to evaluate the
results, never to choose a candidate or stopping rule.

\begin{table}[h]
\caption{Retrospective exact-success counts by sampling exponent.
Every entry is retained/replacement. $R_i$ denotes capped repair around
$z^i$. The separate last row selects the source-suboptimal targets with
$1/2<\alpha<1$.}
\label{tab:reuse-all-strata}
\centering
\begin{tabular}{lrrrrrr}
\toprule
Subset & $N$ & $z^0$ & $z^1$ & $z^2$ & $R_0$ & $R_1$\\
\midrule
$\alpha=.30$ &112&2/0&2/1&5/1&10/1&10/3\\
$\alpha=.55$ &112&7/0&16/0&27/0&40/0&41/2\\
$\alpha=.75$ &112&17/0&33/1&43/1&61/1&66/2\\
$\alpha=.95$ &88&18/0&39/0&46/0&69/0&67/1\\
$\alpha=1.25$ &112&29/0&67/0&79/0&99/2&103/6\\
All &536&73/0&157/2&200/2&279/4&287/14\\
Source suboptimal &305&13/0&32/1&48/1&84/1&89/5\\
Source suboptimal, middle &180&10/0&26/0&38/0&63/0&67/2\\
\bottomrule
\end{tabular}
\end{table}

The 440-target subset gives retained exact counts $62,129,164$ for
$z^0,z^1,z^2$, and replacement counts $0,2,2$.
The corresponding counts on the 96-target subset are $11,28,36$ and
$0,0,0$. These differ from the older multistart experiments because the
present ablation executes the displayed simultaneous updates rather than
their heuristic initialization and polishing routines.

Across all 536 targets, the retained second iterate is exact while its
replacement counterpart is not on 199 targets; the reverse occurs on one.
The remaining 336 outcomes tie on the exact-success indicator.
Within $1/2<\alpha<1$, the counts are 115, zero, and 197.
Among the 180 source-suboptimal targets in that regime, they are 38, zero,
and 142. These are matched, retrospective comparisons. The data support a
benefit from source retention at full-fiber cost, while the small sample
sizes and full-fiber execution leave the sparse sublinear-reading theorem
untested. The stored target-only methods use different work budgets:
their aggregate exact counts are $214/440$ and $26/96$.

Synchronous majority is not ascent. In the retained arm, the score falls
from $z^0$ to $z^1$ on 59 targets and from $z^1$ to $z^2$ on 48 targets.
The corresponding replacement counts are 136 and 163.
Keeping the best candidate prevents these decreases from reducing the
reported solver score. On this dataset, that selection has the same exact
success count as $z^2$, but this equality is an observed outcome rather
than an algorithmic identity.

\paragraph{Full-fiber fallback and two-update comparison.}
A full-fiber acquisition followed by two extra updates and repair of $z^1$
need not return the direct repair of $z^0$.
For example, take $n=5$, the following zero-based target locations,
labels, and raw sensor array:
\[
H=\begin{pmatrix}
4&2&1\\1&3&4\\3&2&0\\3&0&0\\1&2&4\\
2&1&2\\4&1&0\\0&3&3\\3&4&1\\1&3&3
\end{pmatrix},\qquad
y=\begin{pmatrix}1\\1\\-1\\-1\\1\\1\\-1\\-1\\1\\1\end{pmatrix},\qquad
r=\begin{pmatrix}1&1&1&-1&-1\\1&-1&-1&-1&1\\1&-1&-1&1&1\end{pmatrix}.
\]
The direct full-fiber procedure obtains score 8, whereas the extra-update
variant obtains score 2 from the same readings. When acquisition requests
full fibers, the default algorithm uses direct repair of $z^0$, bypassing
the extra updates and preserving its tie breaking and work budget. The explicitly
requested two-update variant remains separate and never enumerates
components, including after full-fiber acquisition.

The branch-check program verifies this counterexample, 116 complete
full-fiber equivalences on arbitrary signed targets, ten reduced-branch
executions, 126 address/RNG comparisons, and twelve tests that trap any
attempt to enumerate in the two-update variant. The retrospective replay
checks complete fallback equivalence on all 1,072 source-arm pairs and
evaluates 8,576 recorded method witnesses. Every reference file is
hash-matched before and after the computation; an independent replay
regenerates all method records identically.

\subsection{Output names and the two target populations}
Each target records sixteen assignments: the retained and replacement source
arms each produce \texttt{z0}, \texttt{z1}, \texttt{z2},
\texttt{best\_two\_pass}, \texttt{repair\_z0}, \texttt{repair\_z1},
\texttt{two\_update\_recycling}, and \texttt{fallback\_default}.
The historical population contains 440 targets; the later prospectively
generated population contains 96. Applying new solver variants to either
population is retrospective. The historical comparisons use independent
sensor-error draws across source arms; the later population uses common
sensor-error draws. These original pairing rules are retained.

\begin{table}[h]
\centering\small
\caption{Exact-success counts split by target population. A slash
separates retained and replacement readings. All these replays use full fibers.}
\begin{tabular}{lrrrrr}
\toprule
Population & $N$ & $z^0$ & $z^1$ & $z^2$ & Repair at $z^0$ / $z^1$\\
\midrule
Historical &440&62/0&129/2&164/2&235/4; 241/14\\
Later &96&11/0&28/0&36/0&44/0; 46/0\\
Historical, middle &264&39/0&76/1&98/1&148/1; 152/5\\
Later, middle &48&3/0&12/0&18/0&22/0; 22/0\\
\bottomrule
\end{tabular}
\end{table}
The middle regime is $1/2<\alpha<1$. Its historical target-only successes
are $122/264$, and its later target-only successes are $8/48$, giving the
reported $130/312$. Among source-suboptimal middle targets, these counts
are $63/145$ and $6/35$. The corresponding retained $z^2$ successes are
$28/145$ and $10/35$, and retained repair at $z^1$ gives $55/145$ and
$12/35$. Different historical baseline work budgets remain part of the
comparison and are not treated as matched solver costs.

\section{Experiments with a common reduction and changing objectives}
\label{app:common-experiments}

\subsection{Frozen populations, queries, and comparisons}

The experiment uses every one of the 150 $n=8$ targets in the historical
440-target dataset: thirty targets at each
$\alpha\in\{.30,.55,.75,.95,1.25\}$, with target noise $q=.17$ and
sensor noise $s=.10$. The targets and source readings are fixed; the weighted objectives
and exact weighted reference solves are evaluated on this panel. No target is excluded after observing an outcome. The protocol and
its $\Gamma=2$ extension were frozen before computing any common-reduction
outcome. Their SHA-256 digests, timestamps, input hashes, and complete
text are supplied in \texttt{provenance/common\_reduction\_protocol/common\_reduction/}.

Each source arm uses its original first full-fiber array, comprising
$3n=24$ readings. As in the preceding retrospective ablation,
$z^0$ already includes one unweighted target update from those readings.
The immutable reduction reference is the next unweighted iterate $z^1$.
Its equality to the stored unweighted witness is checked for every
target and both arms. The common reduction receives no further source
readings and never uses the generating source or a reference optimum.

We prescribe $J=\max\{2,\lfloor1/\alpha\rfloor+1\}$ using exact rational
values of $\alpha$. Thus $J=4$ at $.30$ and $J=2$ at the other four
densities. The cutoff is $4(J-2)+\lfloor\Gamma\rfloor$. The component
guard admits only components of size strictly less than $J$. A rejected
reduction produces no common-optimizer answer in these comparisons;
its queries are retained in every population denominator. The general
implementation also provides the explicitly labeled feasible reference
fallback. Neither that fallback nor structural admission is counted as a
global-optimality certificate.

For each target, twelve distinct edge indices are sampled using its
original seed extended by \texttt{[1616,1]}. Each selected edge receives
one query with a negative coefficient change and one with a positive
change. The magnitude is $1/2$ for $\Gamma=1/2$ and $3/2$ for
$\Gamma=2$; all other coefficients retain their original values. The
same edges and weighted queries are supplied to both source arms.
Each budget therefore has $150\times24=3,600$ target--query pairs.
The $3/2$ general probes can reverse a coefficient's sign. The separate
adaptive probes below preserve signs.

The repeated weighted comparator starts from the unchanged acquisition
$z^0$, applies two weighted synchronous updates, and retains the best
of $z^0,z^1_\xi,z^2_\xi$. We also record its capped component repair.
The target-only warm start begins at all-positive signs for each target
and budget, then uses the preceding returned assignment. It performs
strict improving coordinate flips in vertex order for at most twelve
sweeps. Its adjacency construction and incident-term visits are charged.
This is a specified feasible baseline, not a claim about the best
target-only solver. Exact-reference work is excluded from all algorithmic
work comparisons. Different work counters are not treated as equivalent
operations, and the descriptive timings are not a matched-runtime
comparison across programming languages.

\subsection{Complete fixed-query results}

\begin{table}[h]
\caption{All fixed weighted queries, with 720 queries in each density/budget
row. $A_R,A_I$: admitted targets from retained/replacement readings;
$C_R,C_I$: exact common-reduction queries; $T_R,T_I$: exact best-of-two-pass
queries; $P_R,P_I$: exact capped-repair queries; $W$: exact target-only
warm-start queries. Common-reduction counts do not include rejected
queries. Their admitted denominators are $24A_R$ and $24A_I$.}
\label{tab:common-all-strata}
\centering\small
\begin{tabular}{rrrrrrrrrr}
\toprule
$\Gamma$&$\alpha$&$A_R/A_I$&$C_R$&$C_I$&$T_R$&$T_I$&$P_R$&$P_I$&$W$\\
\midrule
$1/2$&.30&0/0&0&0&24&0&54&5&50\\
$1/2$&.55&16/1&186&0&188&0&301&42&52\\
$1/2$&.75&9/2&167&24&202&24&372&40&166\\
$1/2$&.95&16/0&312&0&328&0&537&0&104\\
$1/2$&1.25&19/1&425&0&421&0&592&111&121\\
$1/2$&All&60/4&1,090&24&1,163&24&1,856&198&493\\
\midrule
$2$&.30&0/0&0&0&29&0&83&12&71\\
$2$&.55&2/0&46&0&174&3&294&36&89\\
$2$&.75&4/1&95&24&194&23&370&39&196\\
$2$&.95&10/0&213&0&321&0&539&3&95\\
$2$&1.25&16/0&382&0&416&4&581&95&171\\
$2$&All&32/1&736&24&1,134&30&1,867&185&622\\
\bottomrule
\end{tabular}
\end{table}

For $\Gamma=1/2$, 42 of the 60 admitted retained reductions are exact on
all their fixed queries. There are 350 inexact answers among the 1,440
admitted queries, as well as 2,160 queries whose targets are rejected.
For $\Gamma=2$, 26 of the 32 admitted retained reductions are exact on
all their fixed queries. There are 32 inexact answers among 768 admitted
queries and 2,832 rejected queries. The replacement arm has 72 admitted
inexact answers at $\Gamma=1/2$ and none among its 24 admitted queries at
$\Gamma=2$. All non-admissions in this population are due to the prescribed
component-size condition. The larger cutoff at $\Gamma=2$ admits fewer
targets. Its conditional success fraction uses a different selected
population and is not evidence that larger perturbations improve accuracy.

A separate explicit example shows that even an empty common set is not
a finite certificate. Take all 27 triples at $n=3$, the all-positive
reference, and negative coefficients precisely at
\[
 (0,1,0),(0,1,1),(0,2,0),(0,2,1),
 (1,0,0),(1,0,2),(2,0,0),(2,0,2).
\]
All other coefficients are positive. Every signed degree is positive,
so $J=2$, $\Gamma=1/2$ gives $U=\varnothing$ and passes the guard.
The restricted value is eleven, whereas enumeration of all $2^9=512$
factor assignments gives the global optimum nineteen.
The fixture and independent scalar enumeration are in
\texttt{code/audit\_common\_reduction\_independent.py}.

The source is suboptimal on 83 of these 150 original targets. On their
1,992 queries per budget, retained common reductions attain 120/456
admitted optima for $\Gamma=1/2$ and 74/96 for $\Gamma=2$; the remaining
queries are rejected. The corresponding retained two-pass counts are
182 and 200, capped-repair counts are 453 and 502, and warm-start counts
are 109 and 209, each out of 1,992. On the 67 source-optimal targets,
retained common reductions attain 970/984 and 662/672 admitted query
optima. The retained initialization remains a material finite limitation,
especially on source-suboptimal instances.

For a single-edge change of magnitude $1/2$, every new global optimum
is also an optimum of the original objective: an original score gap is
an even integer, while the change can improve a comparison by at most one.
For magnitude $3/2$, a new optimum can lose two original-score units.
This occurs on 612 of the 3,600 general queries, including 509 queries on
source-suboptimal targets. Of those 612, the retained reduction rejects
588, solves 15 exactly, and returns an inexact restricted answer on nine.
The claim concerns every optimizer: equality between a zero-gap and a
two-gap assignment after a half-integer single-edge perturbation of this
magnitude is impossible by parity of their scaled score difference.

The exhaustive solver returns multiple distinct optimal observed-edge
patterns across the fixed queries on 46 targets at $\Gamma=1/2$ and on
117 at $\Gamma=2$. The sums of distinct returned patterns across targets
are 253 and 741, respectively; the largest per-target counts are seven
and thirteen. These counts describe returned witnesses. The stronger
claim of incompatible optimizer sets is checked separately below.

Across all admitted retained fixed queries, sparse updates read 2,100
coefficient replacements at $\Gamma=1/2$ and 1,120 at $\Gamma=2$.
The totals of table entries updated are 214 and 266; entries scanned
are 212 and 260. A changed edge outside the common set changes only
the maintained constant, so many queries touch no component table.
Component maximizers change on 53 and 36 queries, respectively.
These counts charge undoing the preceding perturbation when the selected
edge changes; they are not counts of distinct observations acquired.

\subsection{Adaptive families with incompatible optimizer sets}

All choices in these probes use the observed target and computed
reduction. No generating signs or exhaustive outcomes select a component.
The probes are adaptive and are therefore not included in the fixed
retained-versus-replacement comparison.

For $\Gamma=1/2$, select at most six zero-field singleton components of
positive unsigned degree, in increasing vertex order, and take the first
incident edge at each selected vertex. If $\ell$ is the maximum selected
edge load at any vertex, set
$\varepsilon=2^{-\lceil\log_2(2\ell)\rceil}$.
Change each selected edge by $\pm\varepsilon$ times its reference product.
All $2^r$ sign patterns are visited in Gray-code order. This produces
20 eligible families and 52 tasks; the family dimensions are sixteen
with $r=1$, three with $r=2$, and one with $r=3$. There are 23 exact
common answers and eleven families exact on every task. Twelve families
have pairwise disjoint global optimizer-pattern unions; eleven satisfy
both exactness and disjointness. All other outcomes remain reported.
These probes select zero-margin coordinates of the restricted original
objective; finite admission does not ensure they are ties of the
unrestricted optimum.

For $\Gamma=2$, first optimize the original objective over the admitted
restriction, obtaining $w$. Select singleton components with signed
degree one relative to $w$ and a negative incident signed clause. Greedily
choose the first such edge disjoint from all previously selected edges,
with at most six selections. A task bit $\tau_v\in\{0,1\}$ replaces
the chosen coefficient by $Y_e(1+3\tau_v/2)$.
The coefficient sign is preserved, the local load is at most $3/2$, and
the represented local margin changes from one to $-1/2$ when the bit
turns on. These choices yield eight families: seven with two tasks and
one with four. All 18 tasks are exactly solved, and all eight families
have pairwise disjoint global optimizer sets as witnessed by disjoint
unions of selected observed-edge patterns. The ten nonzero task patterns
produce optimizers with original-score loss $2\sum_v\tau_v$.
No extra tie-breaking perturbation is silently added to this finite test.

The reference solver computes the union of selected observed-edge patterns
over \emph{all} global optimizers, including every relevant zero-field
completion in the third factor. Disjoint unions prove that no assignment
optimizes two tasks in the family; different arbitrary solver outputs
would not suffice. An independent full tensor enumeration reproduces
all seventy adaptive optimum values and pattern unions.

\subsection{A deterministic workload with certified exact answers}

The following construction tests update cost after a correct reduction
has been supplied. It does not model noisy acquisition or assert a
random-model finite onset. Let $n\in\{8,16,32,64\}$ and $t=r=n/2$.
Include every positive core triple $(i,j,k)$ with $i<t$ and $j,k<n$.
For each weak first-factor vertex $i=t+\ell$, add the two triples
$(i,\ell,0)$ with coefficient $+1$ and $(i,\ell,1)$ with coefficient $-1$.
There are $m=tn^2+2r$ triples. Supply the all-positive reference,
$J=2$, and $\Gamma=1/2$. Its common set consists of the $r$ singleton
weak vertices. Change each positive weak-edge coefficient by
$\pm\varepsilon$, where $\varepsilon=1/(2r)$.

The optimum is exactly
\[
 tn^2+r\varepsilon.
\]
Indeed, if the complete core tensor is not identically positive, its
score loses at least $2tn$. The weak triples can gain at most $2r$
relative to an aligned-core optimum, and $tn>r$. Thus every optimum
keeps the core tensor positive. Each weak vertex then contributes
$\varepsilon$ after choosing its forced sign. This also certifies the
incompatibility of different weak-edge sign patterns without an
exponential reference solve at the larger sizes.

For each size, we execute 64 queries in Gray-code order on the first
$\min\{r,6\}$ weak vertices and hold the other perturbation signs positive.
There are sixteen distinct objectives at $n=8$, repeated four times, and
64 distinct objectives at each larger size: 256 executions of 208 distinct
objectives. All four procedures---incremental tables, dense table rebuild,
two weighted passes, and the specified warm start---attain the exact
value on every execution. All 64 $n=8$ queries are additionally solved
by both independent exhaustive programs.

The initial query supplies $r$ coefficient replacements. Every subsequent
query supplies one replacement, changes one singleton table, updates two
entries and scans two entries. Dense rebuilding supplies and visits all
$m$ coefficients at every query. The common algorithm times only its
scalar-value and changed-component output. A full assignment requires
another $3n$ output signs and is timed separately. The table below records
the median of all 63 subsequent queries, with no discarded warmup.

\begin{table}[h]
\caption{Constructed workload: preprocessing in milliseconds and median
subsequent-query times in microseconds, from the supplied Python
implementations on one execution environment. These descriptive times
do not replace the exact work counts. Full-output time is additional to
the compact incremental update. Dense rebuilding also returns a scalar;
the two-pass and warm methods materialize their witnesses.}
\label{tab:common-mechanism}
\centering\small
\begin{tabular}{rrrrrrrr}
\toprule
$n$&$m$&Preprocess (ms)&Update ($\mu$s)&Output ($\mu$s)&Rebuild ($\mu$s)&Two passes ($\mu$s)&Warm ($\mu$s)\\
\midrule
8&264&0.51&9.47&2.73&188&388&197\\
16&2,064&3.5&14.1&4.34&1,608&3,192&1,635\\
32&16,416&29.2&17.8&6.1&12,814&25,279&14,105\\
64&131,136&326&28.4&14.1&108,839&207,708&122,583\\
\bottomrule
\end{tabular}
\end{table}

The sparse update uses indexed access to an already stored target. The
cost of choosing a new objective is outside the timed update, and a newly
supplied dense coefficient array has input cost $\Omega(m)$. Preprocessing
builds the common partition and all tables once. The warm baseline rebuilds
its adjacency lists per query as specified; it is not an optimized dynamic
coordinate-ascent implementation. Its comparison therefore makes no claim
of optimal dynamic performance against all competing methods.

\subsection{Executable checks and reproducibility}

\texttt{common\_reduction.py} stores exact integer numerators with one
common denominator. Edge indices point to a component, a local parity mask,
and a reference edge product. Coefficient replacements are validated
transactionally: accumulated absolute perturbations at each affected
vertex must remain within the original local envelope. Invalid updates
leave coefficients, tables, maxima, and budgets unchanged. Component
discovery uses a linear graph traversal, and touched components are visited
in first-occurrence order without a sort. Component and growing work guards
are checked before exponential table allocation; an optional caller cap
may impose an additional explicit resource limit.

The C++ exhaustive solver fixes two gauge signs, enumerates the first
factor, traverses the second factor in Gray-code order, and optimizes
the third factor exactly. The independent Python/NumPy audit instead
forms the complete tensor of fields for all first- and second-factor
assignments by direct contraction. It shares neither Gray-code updates
nor common-table computations with the reference program.
The initial experiment makes 7,484 exact-reference calls: 150 original
targets, 7,200 fixed weighted queries, seventy adaptive queries, and
64 constructed queries. The independent audit rechecks every one of
these values, all adaptive optimizer-pattern unions, 75,050 stored
feasible witness scores, and 2,328 common sparse-update outputs.
The unweighted and weighted reference records are supplied with the experiment.

Separate small scalar enumeration tests cover 744 restricted-table and
sparse-update cases, 42 weighted-reference and pattern-union comparisons,
46 guarded non-admissions with feasible fallback checks,
38 rejected-transaction rollback checks,
24 cases with at least 85 bits of integer precision, and an explicit
empty target. Four allocation-guard fixtures check the growing default,
an optional fixed cap, rejection before an oversized shift, and budget
saturation. The growing-default fixture admits 524,289 singleton
components, requiring 1,048,578 table entries, without actually allocating
those tables; the optional former $2^{20}$ entry cap rejects the same
size description. These verify finite arithmetic and program identities;
the probability statement still rests on the mathematical proof.
The incremental path avoids sorting and records witness-output time separately.
The default table cap grows with the input guard. The independent audit
reproduces the scientific outcomes at the fixed cutoffs and queries.

Run the following in a working copy of \texttt{triples\_and\_clauses} in the reproducibility supplement.
\begin{verbatim}
python3 code/check_common_reduction.py
python3 code/audit_common_reduction_independent.py
python3 code/experiment_common_reduction.py
python3 code/audit_common_reduction.py
\end{verbatim}
Machine-readable outcomes are in
\texttt{results/common\_reduction/}. The commands above regenerate the unit and independent data audits.

\section{Finite experiments with certified restrictions}
\label{app:certification-experiments}

\subsection{Frozen population and information available to the methods}

This is a retrospective evaluation on every one of the 150 historical
$n=8$ targets in Appendix~\ref{app:common-experiments}, using both
reading arms and $\Gamma\in\{1/2,2\}$. It contains 600
(target, arm, budget) cases. Each arm keeps its original 24 source
readings and its already computed unweighted reference $r=z^1$ (the first full-target majority update of the acquired estimate).
No additional reading, target, or global reference solve is introduced.
The same 7,200 weighted objectives are evaluated with both arms, giving
14,400 arm--query rows. Each target has 24 queries per budget: twelve
prespecified edges receive the two changes $\pm1/2$ or $\pm3/2$.
All targets, failed certificates, and resource refusals remain in the
reported denominators. The generating source and exact optimum are
used only for final evaluation and stratification.

The protocol was frozen before any certificate outcome on this
population. Its text, timestamp, SHA-256 digest, and input hashes are in
\texttt{provenance/certification\_protocol/certification/}.
The fixed input snapshots separate the certificate experiment from regenerated timing measurements.

In relative coordinates, $g_e=Y_eT_e(r)$ and the reference is all $+1$.
In each of the first two blocks, choose the smallest-index vertex
with maximum original signed degree as a gauge anchor. These are
\emph{observed, data-dependent} anchors; they are distinct from the
fixed anchors in the asymptotic obstruction of
Proposition~\ref{prop:certificate-local-obstruction}.
Every signed objective is invariant under the two-block gauge, so
fixing these two relative signs to $+1$ preserves a global optimum.
The initial free set is
\[
 U_0=\{v:D_v(r)\le 4(J-2)+\lfloor\Gamma\rfloor\}
       \setminus\{\text{two anchors}\},
 \qquad
 J=\max\{2,\lfloor1/\alpha\rfloor+1\}.
\]
Removing the anchors defines a candidate restriction; it is not itself
an inference of global safety.

\subsection{Acceptance, enlargement, and finite resource limits}

The two certificate procedures start independently from $U_0$.
The local-marginal procedure orders eliminated non-anchor vertices
by increasing original signed degree, breaking ties by index, and
forces each in turn to relative sign $-1$ in
\eqref{eq:conditioned-certificate-lp}. It stops at the first uncertified
condition. No condition is required for the identity projection.
For the minimization problem with nonnegative variables, an accepted
rational dual must satisfy
\[
 y\le0,\qquad
 c-A_{\rm eq}^{\mathsf T}u-A_{\rm ub}^{\mathsf T}y\ge0,
 \qquad b_{\rm eq}^{\mathsf T}u+b_{\rm ub}^{\mathsf T}y\ge0.
\]
The rational reconstruction tries denominator bounds $10^3,10^6,10^9$.
Any negative reduced cost is repaired conservatively by lowering the
appropriate load or normalization multiplier. Every inequality and
its resulting lower bound are then checked with exact fractions.
The floating solver's status never supplies a certificate. Negative
relaxation values are accompanied, where reconstructible, by exact
feasible rational primal witnesses; this succeeds on every population
LP rejection.

The separate cover procedure evaluates
\eqref{eq:cover-certificate-test} by integer arithmetic, using all four
gauge covers, all three whole-block covers, and removing isolated
vertices from the cover counts. It enumerates the canonical first two
factors and sorts the third-factor fields as in
\eqref{eq:certificate-third-block-sort}. When the bound is negative,
it releases the changed eliminated coordinate of smallest original
signed degree, breaking ties by index. It then rebuilds the proposed
restriction and checks it again. Validity is not assumed to persist
under enlargement. The witness proves failure of this sufficient
bound; it need not exhibit an admissible perturbation that invalidates
the restriction.

Both methods require every free interaction component to have at most
12 vertices for an executable reduced solve. The adaptive procedure
stops immediately when this cap is exceeded. Initial cap refusals consume
no cover call; a release that causes a later cap refusal is not followed
by another call. Without this cap, at most
$3n-2$ releases reach the canonical identity projection; that endpoint
has no elimination benefit. We count a certified executable reduction
only if it also fixes at least one observed, non-anchor coordinate.
A refused case provides only the original feasible reference, without
claiming global exactness. This cap differs from the earlier
$|C|<J$ guard, so the two admission counts are not a matched comparison
of computational budgets.

\subsection{Complete population results}

There are no nontrivial LP certificates in the population. Exactly
473 cases have an eliminated non-anchor coordinate, and their first
conditional LP has an exactly verified negative primal witness.
The remaining 127 cases are identity restrictions and are excluded
from useful admission. None of these outcomes is a numerical solver
failure. Table~\ref{tab:certification-strata} gives the cover results;
all four groups retain their full 150-target denominator.

\begin{table}[t]
\caption{Cover-certified targets at $\Gamma=1/2$. Each density has
30 targets and 720 queries per arm. The source column counts optimal
and suboptimal generating sources. Parentheses count accepted targets
whose source is suboptimal. Every $\Gamma=2$ entry is zero.}
\label{tab:certification-strata}
\centering\small
\setlength{\tabcolsep}{5pt}
\begin{tabular}{rcrr}
\toprule
$\alpha$&Source O/S&Retained&Replacement\\
\midrule
$.30$&$1/29$&$0\;(0)$&$0\;(0)$\\
$.55$&$7/23$&$0\;(0)$&$0\;(0)$\\
$.75$&$14/16$&$1\;(0)$&$1\;(0)$\\
$.95$&$20/10$&$8\;(0)$&$0\;(0)$\\
$1.25$&$25/5$&$14\;(1)$&$0\;(0)$\\
\midrule
All&$67/83$&$23\;(1)$&$1\;(0)$\\
\bottomrule
\end{tabular}
\end{table}

At $\Gamma=1/2$, retention certifies 22 of 67 source-optimal targets
and one of 83 source-suboptimal targets. Replacement certifies one
of 67 and zero of 83, respectively. The sole accepted
source-suboptimal retained target has $\alpha=1.25$; there is no
accepted source-suboptimal case in $1/2<\alpha<1$. The two arms'
accepted target sets are disjoint. Nineteen of the 23 retained
restrictions have no free coordinate and four have one; the replacement
restriction has none. All 552 retained and 24 replacement weighted
answers equal their exact saved global optima. The other 13,824
arm--query rows receive no certified answer.

Every acceptance occurs before the first enlargement.
Table~\ref{tab:certification-work} accounts for every rejection and
adaptive release. The 2,097 releases add no accepted case under the
prespecified component cap. These are failures of this frozen
LP/adaptive protocol to extend admission. They do not establish
that another enlargement rule, a tighter certificate, or a larger
computational budget would fail.

\begin{table*}[t]
\caption{Certificate construction and resource refusals. R and I denote
retained and independent replacement readings. Each row has 150 targets.
Initial and later refusals occur when a free component exceeds 12
vertices. Every cover call enumerates $4^7=16,384$ first--second factor
assignments. The LP column counts attempted conditional programs;
all have exact negative primal witnesses.}
\label{tab:certification-work}
\centering\small
\setlength{\tabcolsep}{8pt}
\begin{tabular}{ccrrrrrr}
\toprule
$\Gamma$&Arm&LP calls&Cover calls&Released&Initial refusal&Later refusal&Certified\\
\midrule
$1/2$&R&120&906&883&35&92&23\\
$1/2$&I&120&390&389&66&83&1\\
$2$&R&118&754&754&56&94&0\\
$2$&I&115&71&71&130&20&0\\
\midrule
All&&473&2,121&2,097&287&289&24\\
\bottomrule
\end{tabular}
\end{table*}

\subsection{Work accounting and independent verification}

Cover construction enumerates 34,750,464 first--second assignments in
total. Its $O(4^{n-1}(m+n\log(n+2)))$ preprocessing is explicitly
exponential; it is not a polynomial finite certificate or evidence
for the asymptotic acquisition onset. The implementation is capped at
$n\le12$ and uses bounded integer numerators, with all arithmetic
ranges checked. The local programs have 261--705 variables,
147--336 equalities, and 9--108 inequalities. Exact dual/primal checking
costs and unsuccessful constructions are included in their recorded
construction times. Wall times are descriptive; they are not compared
as asymptotic work measures.

The query implementation rebuilds exact component tables for each
accepted objective. It does not claim the incremental-update cost of
Theorem~\ref{thm:common-reduction}. Across 552 retained queries it reads
41,736 coefficients, builds 192 table entries, and evaluates 1,008
edge--table terms; the 24 replacement queries read 1,224 coefficients
and require no nonconstant table. The additional interaction-graph
scan and direct score verification each visit the same number of
edges as the coefficient scan. Complete witness materialization
writes 13,824 signs in total. These scans and output costs are recorded
separately from table arithmetic; a zero-dimensional restriction does
not make this implementation's score evaluation free.

The small-instance checks cover 36 seeded instances at $n=2,3,4$ and
14,016 canonical assignments. Direct enumeration independently agrees
with the sorted cover minimum and returned minimizing witness.
Four accepted positive-instance LP conditions have exact rational
duals; deliberately corrupting each causes rejection. A nine-edge
$n=4$ example has a valid robust empty restriction at $\Gamma=1/2$
but a conditional LP value $-3$, illustrating a false negative of the
relaxation. Here the two fixed anchors are deliberately isolated
vertices, unlike the observed-degree population anchors. The analytic
argument in Appendix~\ref{app:robust-certificates} covers the full
budget family; eighteen individual weighted objectives are also
checked exhaustively. The stored $n=3$, $m=27$ invalid empty
restriction is rejected by both procedures. Adaptive enlargement
certifies it only after releasing every non-anchor coordinate, so
this fixture achieves safety without compression; six weighted
queries are checked at that endpoint.

The population audit is independent of the primary LP assembly and
sorted C++ checker. It reads each negative LP witness as rational
vertex and factor distributions and edge loads, checking normalization,
consistency, capacities, and objective value directly. It verifies
all 2,097 negative cover witnesses by explicit gauge-cover sets and
direct projection scores, including every release choice and cap.
For each of the 24 accepted restrictions it enumerates \emph{all}
$2^{3n-2}$ canonical assignments, including every third-factor sign
vector: 100,663,296 assignments overall. It separately verifies all
576 weighted witness scores against the frozen exact references.
A second full population run reproduces every scientific case,
release trace, and query output. The reports retain all 14,400 query
rows and every failed or refused target.

The offline entry points are
\texttt{check\_certification.py},
\texttt{experiment\_certification.py}, and
\texttt{audit\_certification.py} in \texttt{code/}.
They require Python, NumPy, SciPy, and a C++17 compiler.
The exact witnesses, traces, full stratified summary, and audit report
are in \texttt{results/certification/}.

\section{Exactness and projection diagnostics}
\label{app:projection-diagnostic-study}

\subsection{A complete retrospective comparison}
The distinction between common exactness and pointwise improvement motivates
a new diagnostic of the frozen finite population. We retain all 150 targets,
both reading arms, the two budgets, and the original 24 weighted objectives
per target and budget. This gives 600 initial restrictions, 7,200 distinct
objectives, and 14,400 arm--query rows. The new protocol was frozen before
computing these diagnostic outcomes. It changes no target, reading, query,
reference solve, or earlier protocol.

For each case, we reconstruct the original observed-degree anchors and
initial free set. In particular, we retain identity restrictions and those
that exceeded the former twelve-vertex component cap. We compute the
restricted optimum and the exact minimum of
$F_w(Px)-F_w(x)$ over every canonical assignment for each fixed query $w$.
The stored global optimum is used only after the restriction and diagnostics
have been fixed. A negative reset minimum with an exact restricted answer
proves a discrepancy for that queried objective. Successful answers on this
finite query list do not prove validity throughout $\mathcal U_\Gamma$.

\begin{table}[t]
\caption{All fixed-query rows. The last column requires both exactness and
a strictly negative reset loss. Every row contains 3,600 queries.}
\label{tab:projection-diagnostic-full-study}
\centering\small
\begin{tabular}{clrrr}
\toprule
$\Gamma$&Arm&Exact&Negative reset&Both\\
\midrule
$1/2$&Retained&2,437&1,680&517\\
$1/2$&Replacement&1,101&2,856&357\\
$2$&Retained&2,691&2,152&1,243\\
$2$&Replacement&1,627&2,736&763\\
\bottomrule
\end{tabular}
\end{table}

Across the entire population, 7,856 rows preserve the queried optimum and
9,424 have a negative reset loss. Their intersection contains 2,880 rows.
There are 271 initial restrictions that solve all 24 queries, including
77 that fail the reset comparison on at least one of those queries.
The 3,048 identity-on-active rows are all exact and contribute no
discrepancy. Among the 7,512 rows with a proper active restriction and
largest component at most twelve, 3,673 answers are exact and 1,745 are
exact despite a negative reset loss. These groups are reported separately
so that identity restrictions and unrestricted exponential solves do not
inflate the interpretation of useful reductions.

\begin{table}[t]
\caption{Targets with a suboptimal source and $1/2<\alpha<1$.
Each arm--budget group contains 49 targets and 1,176 queries. The final
two columns restrict to proper domains with largest component at most
twelve; ``both'' again means exact answer and negative reset loss.}
\label{tab:projection-middle-study}
\centering\small
\begin{tabular}{clrrrr}
\toprule
$\Gamma$&Arm&Exact&Both&Cap rows&Cap both\\
\midrule
$1/2$&Retained&291&174&1,104&129\\
$1/2$&Replacement&63&63&864&63\\
$2$&Retained&473&384&768&299\\
$2$&Replacement&296&248&120&13\\
\bottomrule
\end{tabular}
\end{table}

\paragraph{Exact computation and work.}
For this calculation, express signs relative to the reference $r$ and
multiply each coefficient $w_e$ by $T_e(r)$. The fixed reference signs
are then all $+1$.
Fix the first two factor sign vectors. Let $h_k$ be the original weighted
field at the third-factor coordinate $k$, and let $h_k^P$ be its field after
resetting eliminated coordinates in the first two factors. Minimizing over
the third factor gives the exact reset loss
\[
 -\sum_{k\in U_3}|h_k^P-h_k|
 +\sum_{k\notin U_3}(h_k^P-|h_k|).
\]
The restricted optimum is obtained by enumerating only first--second
assignments compatible with the restriction and replacing each free
third-factor contribution by $|h_k|$. All calculations use integer
numerators with denominator two. Both optimizations retain a full attaining
witness, checked by direct score evaluation.

The primary evaluator visits 9,830,400 canonical first--second assignments
across the 600 domains and 235,929,600 weighted query--assignment pairs.
Its shared base-field scans evaluate 1,010,565,120 edge terms; third-factor
query calculations evaluate 1,887,436,800 field terms. It reads 30,840 base
integer coefficients and the 14,400 query perturbations; direct output
checks evaluate another 2,220,480 edge terms. These are explicitly
exponential diagnostics. They are not an acquisition algorithm, an efficient
certificate, or an estimate of the asymptotic execution threshold.

Before the population computation, independent Python enumeration verifies
36 deterministic small instances, 144 fixed queries, and 56,064 full
canonical assignments. A second implementation reconstructs all 600 domains
and independently recomputes every restricted optimum and every reset
minimum using integer arrays. It checks all 28,800 witnesses and recounts
every reported stratum. The frozen input hashes, all outputs, and both
implementations accompany the supplement.

\subsection{Uniform common validity with a failing reset}
\label{app:mixed-finite-study}
A distinct-triple example separates the properties for an entire budget
class. Take $n=3$, the all-positive reference, anchors $A_0,B_0$, and
$U=\{A_1,B_1,C_1\}$. The fourteen observed triples and signs are
\[
\begin{array}{c|rrrrrrr}
(a,b,c)&101&120&221&220&201&002&202\\
Y&+&+&-&+&+&+&+\\
\hline
(a,b,c)&210&012&022&200&021&000&110\\
Y&-&+&+&+&-&+&+
\end{array}
\]
where $101$ denotes $(1,0,1)$, and similarly for the other triples.
The three free vertices form one interacting component. Direct enumeration
of the 128 canonical assignments gives original optimum eight and exactly
three original optima, all in the restriction. The canonical assignment
\[
x=(1,-1,1;\ 1,1,-1;\ 1,1,1)
\]
has $F_0(Px)-F_0(x)=-2$. Thus the reset fails already at the unperturbed
objective.

For $\Gamma=1/2$, all 128 competitors have nonnegative, exactly checked
rational component certificates. Their complete enumeration establishes
common validity for every perturbation in this budget class, and hence for
every smaller budget. A separate rational checker constructs the eight
restricted assignments directly, verifies all probabilities and cover
inequalities, and verifies that the competitor list is complete. All 128
deliberately corrupted probability normalizations are rejected. An unsafe
empty restriction is also rejected. The elementary original-gap argument
alone establishes validity up to $\Gamma=1/3$; the rational certificates
verify the larger budget without relying on a floating solver status.

\paragraph{Why mixtures can be necessary.}
On the same graph, consider the single competitor
\[
x=(1,-1,-1;\ 1,-1,-1;\ -1,-1,-1)
\]
at budget $\Gamma=5/4$. Put probability $1/4$ on the all-positive restricted
assignment and $3/4$ on that assignment with $C_1$ reversed. The vertex
cover weights, in block order, are
\[
\lambda=(1,0,0;\ 1,0,0;\ 2,1/2,0).
\]
The exact mixed comparison is at least $3/8$. Yet each of the eight pure
restricted assignments is defeated by a feasible perturbation. Exact
adversarial witnesses give negative comparisons
\[
-3/2,\ -1/4,\ -1,\ -7/2,\ -5,\ -1,\ -11/2,\ -2/3,
\]
in lexicographic sign order with $+$ before $-$. Their perturbations,
local-budget checks, and direct score comparisons are supplied. This
proves that mixing can be necessary for a fixed competitor. The entire
domain is not valid at $5/4$; its uniform-validity example uses $1/2$.

\section{A frozen search for common component certificates}
\label{app:component-certification-study}

\subsection{Population, information, and fixed resource limits}

We evaluate the finite proof system of
Appendix~\ref{app:common-search} on all 600 initial domains from
Appendix~\ref{app:certification-experiments}: the same 150 historical
$n=8$ targets, retained and independent-replacement reading arms,
and budgets $\Gamma\in\{1/2,2\}$. There are 30 targets at each of
$\alpha=0.30,0.55,0.75,0.95,1.25$. Each arm retains its original
24 source readings and reference $r=z^1$ (the first full-target majority update). We introduce no target,
source reading, adaptive release, or global reference solve.
This is a retrospective study with a fixed protocol.

The protocol, sanitized inputs, search implementation, and verifier were
fixed before evaluating the 600 domains. For every case we reproduce the original observed-degree
anchors and initial free set $U_0$. The search receives only the target
identifier, arm, budget, $n$, observed scopes, relative coefficients
$g_e=Y_eT_e(r)$, free coordinates, and anchors. Its input contains no
generating source, exact optimum, fixed-query answer, previous certificate
outcome, or density stratum. The latter information is attached only
after all 600 search outcomes have been archived.

An identity-on-active domain receives a structural proof without
enumerating its component tables. A proper domain is eligible for search
only if its largest free component has at most 12 vertices. Among all
600 domains, 127 are identity restrictions, 160 proper domains exceed
this component cap, and 313 proper domains are eligible. Every case
remains in the denominator.

For an eligible domain, the implementation explicitly constructs
\eqref{eq:common-search-budget}--\eqref{eq:common-search-objective}
and permits at most 512 LP proposal calls. It uses depth-first search,
branching first on greatest fractionality, then preferring an eliminated
coordinate, greater original unsigned degree, and lower vertex index.
It visits the rounded LP bit first, choosing one at a tie. Four spawned
workers process independent cases; their completion order cannot change
a case's rule or budget. Each numerical proposal uses the HiGHS LP
solver through SciPy, with a 30-second and 100,000-iteration limit.
No larger secondary search is performed.

\subsection{Proposals, proofs, and refusals}

Numerical optimization proposes a branch or a witness. It never supplies
acceptance. At a node whose numerical lower bound is at least $-10^{-7}$,
we reconstruct nonpositive inequality multipliers with denominator limits
$64,1024,10^6,10^9$, in that order. A leaf is accepted only when exact
rational evaluation of \eqref{eq:common-search-dual-bound}, including
the finite-bound residual correction, is nonnegative. The structural
identity leaf of Theorem~\ref{thm:common-search-tree} is also allowed.
A positive outcome requires a complete binary tree covering every
canonical competitor.

For a numerical lower bound below $-10^{-7}$, the implementation rounds
the competitor bits and reconstructs normalized perturbations with
denominator limit $10^6$. If needed, it scales the entire perturbation
down to satisfy every local budget. It then computes the restricted
component maximum and competitor score exactly. A strictly negative
rational difference is an invalidity witness. Failure to find either
kind of witness leaves the search running until its fixed cap; a resource
or reconstruction refusal makes no claim of invalidity.

Every conclusive record contains its instance, budget, and complete proof
tree or negative witness. A separate verifier reconstructs components
and rational inequalities without importing the optimizer, model builder,
or original factor-table code. It checks all branches, every rational
leaf bound, perturbation feasibility, and each claimed negative
comparison. Structural identity proofs have a direct verification path
that avoids unnecessary component enumeration.

\subsection{Complete population results}

Table~\ref{tab:component-certification-all-study} reports all outcomes.
There are 22 proper common certificates: 20 retained and two replacement,
all at budget $1/2$. Seven have interacting free components, with largest
component size 11. All 22 require branching: they use 21--495 LP proposals,
with median 155. No proper domain is accepted at the root relaxation.
There are 156 exact invalidity witnesses, 160 component-cap refusals,
and 135 search-cap refusals. No other solver or arithmetic refusal occurs.

\begin{table}[t]
\caption{All 600 domains. Each arm--budget row has 150 cases.
``Proof'' means a proper common certificate; identity domains are separate.
``Negative'' means an exact invalidity witness. The last two columns
are unresolved component-cap and LP-call-cap refusals.}
\label{tab:component-certification-all-study}
\centering\small
\setlength{\tabcolsep}{3pt}
\begin{tabular}{clrrrrr}
\toprule
$\Gamma$&Arm&Identity&Proof&Negative&Comp.&Search\\
\midrule
$1/2$&Retained&30&20&53&5&42\\
$1/2$&Replacement&30&2&59&36&23\\
$2$&Retained&32&0&32&24&62\\
$2$&Replacement&35&0&12&95&8\\
\midrule
All&Both&127&22&156&160&135\\
\bottomrule
\end{tabular}
\end{table}

The earlier cover test accepted 24 proper domains. Ten overlap the new
22, giving 12 additional certificates; all seven interacting acceptances
are new. Fourteen earlier cover acceptances are not recovered within
the 512-call cap. Thus the capped search does not dominate the previous
test. The two archived methods together certify 36 distinct proper
domains, whereas the present procedure alone certifies 22. This union
uses both methods and their respective work.

Table~\ref{tab:component-certification-middle-study} retains the full
source-suboptimal population in $1/2<\alpha<1$: 49 targets in every
arm--budget group. At budget $1/2$, three retained domains are certified
and no replacement domain is certified. The retained cases have
$(\alpha,m,\OPT(H,Y),|U_0|)$ equal to
$(.55,37,35,1)$, $(.55,37,35,2)$, and $(.75,51,41,0)$.
They require 21, 25, and 171 LP proposals, respectively; the second
has one interacting two-vertex component. These are finite examples
in the source-suboptimal regime, not evidence of a high admission rate
or the onset of a sparse-acquisition schedule.

\begin{table}[t]
\caption{Source-suboptimal targets with $1/2<\alpha<1$.
Each row contains all 49 targets. Column meanings match
Table~\ref{tab:component-certification-all-study}.}
\label{tab:component-certification-middle-study}
\centering\small
\setlength{\tabcolsep}{3pt}
\begin{tabular}{clrrrrr}
\toprule
$\Gamma$&Arm&Identity&Proof&Negative&Comp.&Search\\
\midrule
$1/2$&Retained&0&3&35&3&8\\
$1/2$&Replacement&0&0&32&13&4\\
$2$&Retained&2&0&21&15&11\\
$2$&Replacement&2&0&4&42&1\\
\bottomrule
\end{tabular}
\end{table}

\paragraph{An explicitly post-hoc structural description.}
After the complete search and exact replay, we use the saved base-target
optima to describe satisfiability and the size of accepted restrictions.
This analysis changes no input, selection, branch, cap, or outcome.
All 22 proper certificates concern unsatisfiable base targets:
$\OPT(H,Y)<m$. They correspond to 20 distinct targets, because two targets
have both arms accepted. In fact, all 313 proper cap-eligible domains
have unsatisfiable base targets. The successes therefore do not arise
from a satisfying assignment whose optimality follows merely from
preserved coefficient signs.

Each accepted domain has 24 active vertices and two gauge anchors.
It retains 0--11 free coordinates and fixes 11--22 of the 22 active
nongauge coordinates; the median number fixed is 22. Twelve accepted
domains are single assignments, three have one free coordinate, and
seven have interacting components. The latter retain 2--11 coordinates
and fix 11--20. None is a nearly identity restriction. Their common
exactness is certified for the entire budget family, regardless of
whether the restricted domain contains one or many assignments.

\paragraph{Common validity and fixed-query evidence.}
After the search, every positive domain agrees with all 24 archived exact
query answers. This is a coherence check; the proof trees establish
the universal guarantee. Five proper accepted domains have a previously
verified negative anchored-reset loss on at least one of those queries
(Appendix~\ref{app:projection-diagnostic-study}). Hence those five domains
are common exact although their fixed reset is not pointwise improving.
Conversely, 22 of the new exact invalidity witnesses concern domains
that answered all 24 frozen queries exactly: 13 at budget $1/2$ and nine
at budget $2$. Their rational perturbations strictly defeat the restricted
optimum. Success on this finite query list did not establish common
validity, and a reset failure did not establish its absence.

\subsection{Work, constructed checks, and replay}

The search uses 91,260 LP proposals, counting negative outcomes and every
capped run. The 135 search-cap cases alone consume 69,120 proposals.
The 22 proper proof trees contain 4,332 nodes and 2,177 rational dual
leaves; their subcubes cover $22\cdot2^{22}$ canonical competitors
without enumerating that many separate assignments. Component-table
construction enumerates 174,340 states and 7,968,022 factor terms.
Summed across LP calls, the submitted matrices contain 2,555,764,825
nonzero entries. This is an input-size count, not a count of internal
solver pivots. The 47,144 attempted exact adversary checks evaluate
1,154,726,174 component factor terms and 3,130,703 competitor score terms.

Elapsed search time is 696.94 seconds with four workers; the sum of
per-case worker walltimes is 2,582.85 seconds. Independent exact replay
takes 23.76 seconds in the recorded environment. Timings are descriptive
and include no claim of polynomial scaling. The original reading cost
remains 24 per arm and target, or 7,200 readings across both arms and
all 150 targets, shared between the two budget evaluations. There are
no additional readings or reference solves.

The compressed archive contains all 600 instances and outcomes, including
refusals: 335,963 bytes compressed and 3,909,167 bytes uncompressed.
Its positive proofs contain 269,296 nonzero rational multipliers, with
at most seven bits in an absolute numerator or denominator. Negative
perturbations contain 1,949 nonzero coordinates in total; their
perturbation and gap numerators and denominators have at most nine bits.

On the fourteen-observation fixture in
Appendix~\ref{app:mixed-finite-study}, the interacting three-coordinate
domain at $\Gamma=1/2$ has a 49-node proof with 25 rational leaves and
depth six. It covers all 128 canonical competitors. Model construction
takes about $0.002$ seconds and search about $0.13$ seconds. This is a
finite comparison with the earlier 128 fixed-competitor LP checks;
the differently sized LPs are not interchangeable units of work, and
the comparison asserts no asymptotic speedup. At $\Gamma=5/4$ the new
search instead produces an exactly verified gap $-1$ between a restricted
maximum and a competitor. The witness need not minimize that gap globally.

The independent proof verifier checks the fixture at budgets
$0,1/2,5/4$ and rejects 17 deliberately corrupted witnesses or trees.
A separate deterministic model audit checks 15 constructed $n=2,3$
instances at budgets $0,1/2,2$. It compares 912 fixed-competitor node
LPs with the independent epigraph formulation
\eqref{eq:exact-common-component-primal}, with maximum numerical
discrepancy $1.8\times10^{-15}$, and checks 304 unweighted values by
direct integer enumeration. Every branch conclusion in that audit
is independently verified in exact arithmetic. The numerical comparisons
check model translation; they do not replace rational proof verification.

The population replay verifies 149 positive proof objects, comprising
127 identities and 22 proper certificates, all 156 negative witnesses,
and the membership and work records of all 295 refusals. It checks
the immutable input and code hashes and the complete 600-case denominator.
In the supplement's \texttt{triples\_and\_clauses} directory,
\texttt{python3 run\_checks.py --jobs 2} performs this replay. The
numerical proposal search can be run separately with the same fixed caps.
The final completion receipt binds the archived search results, passed
replay claims, and separately labeled post-hoc reports. The study establishes
finite common certificates with component completion, while leaving
scalable general certification and practical sparse acquisition unresolved.

\section{Other acquisition designs and direct score estimates}
\label{app:other-acquisition}
The main algorithm uses the original sample both to expand and to refine the source readings. The results in this section concern simpler procedures that reconstruct the reference fibers from the additional readings alone, or estimate a threshold directly from those fibers. Their larger budgets are not lower bounds for the main algorithm.

For odd $r$, read each reference entry $r$ times and take its majority. Write $e_r=\Pr\{\mathrm{Bin}(r,s)>r/2\}$. The reconstructed tensor is correct exactly when the error signs are constant within each block and the three block signs multiply to $+1$. Independence of the $3n$ majority errors gives
\begin{equation}
\Pr(\widehat T=T)=(1-e_r)^{3n}+3(1-e_r)^n e_r^{2n}.
\label{eq:readbudget}
\end{equation}
For fixed $s\in(0,1/2)$, Stirling's formula gives $e_r=\Theta_s(r^{-1/2}e^{-I_s r})$, where $I_s=-\tfrac12\log(4s(1-s))$. Complete reconstruction from these fibers changes from vanishing to asymptotically certain as $I_s r-\log n+\tfrac12\log\log n$ goes from $-\infty$ to $+\infty$. A direct bias-corrected score can decide a balanced threshold with $r=O(\log\log n)$; Appendix~\ref{app:partial-reads} proves this for that statistic. The target-assisted procedure in the main theorem uses a smaller acquisition budget.

\begin{proposition}[Single-use, disjoint measurements]\label{prop:single}
If $n^2>3m+3n(r-1)$, a deterministic design based only on $H$ acquires $(3n-2)r$ distinct auxiliary triples, none in $H$. Under coupled reads it reconstructs the latent tensor with failure at most $3n\exp[-r(1-2s)^6/2]$ in $O(m+nr)$ word operations. The independent-record lower bound holds for this design.
\end{proposition}
In each round the design selects three intersecting fibers at a new center. Products of three readings cancel the unknown center signs. A majority across rounds estimates each reference sign. Appendix~\ref{app:distinct} constructs the centers and accounts for every address.

\section{Algorithms with coupled measurements}\label{app:upper}
The noiseless arrays on the three reference fibers are
\[
 A_i=a_i b_1 c_1,\qquad B_j=a_1 b_j c_1,\qquad C_k=a_1 b_1 c_k .
\]
Consequently $A_iB_jC_k=T_{ijk}$. Each individual fiber entry has $r$ independent reads at error probability $s<1/2$, including separate reads at the shared center. The union bound and Hoeffding's inequality give
\begin{equation}
 \Pr\{(\widehat A,\widehat B,\widehat C)\ne(A,B,C)\}
 \le 3n\exp[-r(1-2s)^2/2]=:J_{n,r}.\label{eq:readbound}
\end{equation}
Use odd $r$. Evaluate $\widehat F=\sum_{e=(i,j,k)\in H}Y_e\widehat A_i\widehat B_j\widehat C_k$ in one pass. It is a feasible score on every input. Equations~\eqref{eq:optbound} and~\eqref{eq:readbound} show that $\Pr(\widehat F\ne\mathrm{OPT})\le E_{n,m}+J_{n,r}$. This one event handles every threshold, including a threshold chosen after the reads. At $K_m=m-2\lfloor qm\rfloor$, let $U=2\one\{\OPT\ge K_m\}-1$, $B=\sum_e\one\{\eta_e=-1\}$ and $V=2\one\{B\le\lfloor qm\rfloor\}-1$. The binomial Berry--Esseen bound gives
\[
 \Pr(U=1)=\tfrac12+O_q(m^{-1/2})+O(E_{n,m}).
\]
Hence for each sign the conditional failure probability is at most $(2+o(1))(E_{n,m}+J_{n,r})$. In fact a false YES is impossible. Taking $A>1$ in the choice of $m,r$ gives summable failure probabilities. On any common coupling across growing sizes the procedure is eventually correct almost surely, by the first Borel--Cantelli lemma.

\paragraph{An upper bound in the same polynomial model.}
Fix $\varepsilon>0$. Let $G\sim N(0,1)$ and choose a fixed polynomial $Q$ of degree $k$ for which $\mathbb E(Q(G)-\operatorname{sign}G)^2<\varepsilon/4$; polynomials are dense in Gaussian $L^2$. With $\sigma=2\sqrt{q(1-q)}$, put $T_m=(m-2B-m(1-2q))/(\sigma\sqrt m)$. The central limit theorem and uniformly bounded moments of every fixed order (by Hoeffding tails) give $\mathbb E(Q(T_m)-V)^2\to\mathbb E(Q(G)-\operatorname{sign}G)^2$. Rounding $qm$ affects the normalized threshold by $O(m^{-1/2})$. The discontinuity at zero has zero Gaussian mass.

Replace $m-2B$ in $T_m$ by $\widehat F$, obtaining $\widehat T_m$. On correct fiber recovery this substitution is exact. Both normalized scores are $O_q(\sqrt m)$ on every realization. Thus $\|Q(\widehat T_m)-Q(T_m)\|_2^2\le O_{Q,q}(1+m^k)J_{n,r}$. Choose the constant in $r=O_{q,s,\varepsilon}(\log n)$ large enough to make the right side $o(1)$. Also $\|U-V\|_2\le2\sqrt{E_{n,m}}$. The triangle inequality proves a limiting risk below $\varepsilon$. Each majority is a multilinear polynomial of degree at most $r$ in its input bits. Consequently $Q(\widehat T_m)$ has total label degree at most $k(3r+1)$. This upper bound and the independent-source lower bound compare the same squared-error criterion.

\paragraph{Calibration drift.}
There is also a continuously varying source law. In each round independently perturb each of the $3n$ latent signs by a fresh multiplier of mean $\rho\in(0,1]$, then take the three fibers with fresh independent sensor noise of mean $1-2s$. For a fixed entry its three multipliers are independent, including at the center because the modes are distinct. Conditional on the original signs the $r$ reads are independent across rounds, with bias $(1-2s)\rho^3$ about the desired bit. Equation~\eqref{eq:readbound} holds with that bias. Within-round dependence is unrestricted by the union bound. This is a sufficient drift bound. At $\rho=0$ this round-resampled donor differs from the fixed independent donor in the main comparison, but both transcripts are independent of the target.

\paragraph{Read budget for the fiber subproblem.}
Even with all other latent blocks and $a_1$ revealed, reconstructing the remaining $a_i$ from fiber data alone reduces to $n-1$ independent binary channels with $r$ repetitions. Under the uniform prior, optimal all-bit success is $(1-e_r)^{n-1}$, where $e_r=\Pr\{\mathrm{Bin}(r,s)>r/2\}$ for odd $r$. Since $e_r\ge s^r$, $r=o(\log n)$ gives success tending to zero for fixed $s>0$. This necessity concerns complete reconstruction from the chosen probe design. It leaves open the optimal acquisition rate for the decision given the target and probes together.

\subsection{Finite acquisition before record loss}\label{app:retention}
The repeated-read theorem specifies a number of sensor queries. To convert that count to a time deadline, consider $M=3n$ stored reference signs, each with an independent rate-$a$ Poisson read clock and an independent exponential erasure clock of rate $\nu$. Separate cells at the common center can carry the same sign. Detection of erasure is immediate. Each completed read has the independent sign noise of the fiber experiment. The deadline applies to the procedure requiring $r$ readings of each cell; its aggregate read bandwidth is $Ma$. For a single cell, the $r$th read time has density $a^r v^{r-1}e^{-av}/(r-1)!$. Integrating this density times $e^{-\nu v}$ up to $t$ and taking the product across cells gives
\[
 P_{\rm complete}(t)=
 \left[\left(\frac{a}{a+\nu}\right)^r
       \Pr\{\mathrm{Pois}((a+\nu)t)\ge r\}\right]^M .
\]
The infinite-deadline ceiling is $(1+\nu/a)^{-Mr}$. A ceiling of at least $1-\gamma$ requires and suffices that $\nu/a\le(1-\gamma)^{-1/(Mr)}-1$. Equality attains the target only in the infinite-time limit. Conditional on completion, sign errors have the original independent law, so the joint probability of completion and exact tensor reconstruction is $P_{\rm complete}(t)$ times~\eqref{eq:readbudget}. This supplies measurable retention and bandwidth requirements for this specific parallel reader.

\subsection{A threshold decision from partial reconstruction}\label{app:partial-reads}
Let $e_r$ be the one-cell majority error, $b_r=1-2e_r$, and $S_*=F(T)$. In the planted coordinates the three arrays of majority error signs are independent over all $3n$ cells and independent of $H,\eta$. For each $e=(i,j,k)$ write $L_e$ for the product of its three error signs. Then $F(\widehat T)=\sum_{e\in H}\eta_e L_e$. Conditional on $H,\eta$, the centered errors $L_e-b_r^3$ have mean zero and variance $1-b_r^6\le12e_r$. Distinct disjoint triples give independent products. For intersecting triples Cauchy--Schwarz bounds their absolute covariance by $12e_r$. Two distinct uniform grid triples share a vertex with probability at most $3/n$, by a union bound over their coordinates. Hence
\[
 \E[(F(\widehat T)/b_r^3-S_*)^2]
 \le 12 e_r b_r^{-6}\{m+3m(m-1)/n\}.
\]
If $e_r\log n\to0$, the centered score error is $o_{\Pr}(\sqrt m)$ at every fixed logarithmic density. Appendix~\ref{app:fluct} proves that $\OPT-\E\OPT$ and $S_*-\E S_*$ differ by $o_{\Pr}(\sqrt m)$. The normal limit gives vanishing classification error using $F(\widehat T)/b_r^3$ thresholded at the score lattice point nearest $\E S_*$ (either rounding convention has $O(1)$ effect) to predict the decision centered at $\E\OPT$. The same test works for~\eqref{eq:balanced} when $\alpha\ge1/2$. This test uses the known fixed $s$ to compute $b_r$, and the threshold at the source mean uses $q$. Fixed rational rates permit exact binomial arithmetic with bit lengths polynomial in $r+\log m$. Each sign has conditional failure $o(1)$ by balance. For any $r$ odd with $r\sim A\log\log n$ and $A I_s>1$, the binomial tail has $e_r\log n\to0$ and $ne_r\to\infty$. Equation~\eqref{eq:readbudget} then gives $\Pr(\widehat T=T)\to0$. The degree upper bound in Appendix~\ref{app:sublinear-degree} uses its own polynomial approximation and the sublinear acquisition record.

\subsection{The number of task-relevant directions}\label{app:quotient}
Represent the source as $3n$ bits $\theta$ over $\mathbb F_2$. A noiseless observed row gives its inner product with $\theta$. After measuring rows $B_P$, compatible sources form a translate of $\ker B_P$. The target responses $B_H\theta$ vary over a translate of $B_H\ker B_P$, of dimension $\operatorname{rank}_2([B_P;B_H])-\operatorname{rank}_2(B_P)$, where the semicolon denotes vertical stacking. Consequently there are exactly $2^d$ compatible target responses for deficiency $d$. Any sequence of $k$ additional binary measurements has at most $2^k$ leaves, even if the addresses are selected from the preceding observations. Determining all target responses in the worst case requires $k\ge d$. Choosing $d$ independent rows of $B_H$ modulo the acquired row span attains equality. For the full grid, the kernel consists of the four choices of signs constant within blocks whose product is positive. Thus the full noiseless calibration rank is $3n-2$. The lower bound above concerns all target responses. A single threshold can identify many of these responses with the same decision.

\section{Constructing disjoint calibration settings}\label{app:distinct}
Index coordinates by $\mathbb Z_n$ and use centers $(u,v,u+v)$.
Each coordinate pair fixes a center, so a target triple forbids at most
three centers. After choosing a center, remove the at most $3n$ centers
sharing a coordinate. This ensures disjoint line addresses across rounds:
parallel lines are distinct, and a perpendicular intersection would force
a shared fixed coordinate. Within a round, the three lines meet only at
the center, which is read once.

Number centers $un+v$ and store flags for the first
$\min(n^2,3m+3nr+1)$ indices. Mark the at most $3m$ target exclusions,
ignoring indices outside the prefix. Scan one forward pointer for the next
allowed center, then mark its coordinate-sharing neighbors in $O(n)$ work.
Under the proposition's inequality, an allowed center remains before each
of the $r$ rounds. Initialization, marking and scanning cost $O(m+nr)$
operations and words. Selection uses only addresses.

In each round, the latent product for $(i,j,k)$ is $T_{ijk}$: every
center factor occurs twice. Reused sensor errors also cancel, leaving one
or three independent flips and bias at least $(1-2s)^3$. Across rounds
the products are independent for each reference sign. Hoeffding and a
union bound over the $3n$ references give the claimed error. Each recovered
reference is a majority of $r$ products of degree at most three, so the
score has degree at most $9r+1$; Appendix~\ref{app:upper} supplies the
polynomial comparison. With an independent donor, the mask fixes all
reading addresses and the record remains independent of target factors
and noise, as required in Appendix~\ref{app:lower}.

\FloatBarrier
\end{document}